\documentclass[12pt,a4paper,oneside]{thesis}
\usepackage{lmodern}
\usepackage{amssymb}
\usepackage{amsbsy}
\usepackage{amsmath}
\usepackage{amsthm} 
\usepackage{setspace}
\usepackage{ifpdf}
\usepackage{caption}
\usepackage{subcaption}
\usepackage{multirow}
\usepackage{hyperref}
\usepackage{lscape}
\usepackage{multicol}
\usepackage{eqnarray}
\usepackage{caption}
\usepackage{subcaption}
\usepackage{textcomp}
\usepackage{float}
\usepackage{bm}
\usepackage{color}
\usepackage{calligra} 
\usepackage{amssymb,amsmath}
\usepackage{multicol}
\usepackage{eqnarray}
\usepackage{caption}
\usepackage{subcaption}
\usepackage{textcomp}
\usepackage{multirow}
\usepackage{float}
\usepackage{bm}
\usepackage{color}
\usepackage{epstopdf}
\usepackage{url}
\usepackage[linesnumbered,ruled]{algorithm2e}
\ifpdf
\usepackage[pdftex]{graphicx}
\else
\usepackage{graphicx}
\fi
\usepackage{cite}

\title{*Template for MSc Thesis, Imperial College London*}
\author{*A. N. Other*}

\usepackage{fancyhdr}

\renewcommand{\chaptermark}[1]{\markright{\thechapter.\ #1}}

\fancypagestyle{plain}{
  \fancyhead[LE,RO]{\bfseries\thepage}
  \fancyhead[LO]{}
  \fancyhead[RE]{}
  
}

\def\mystretch{1.5}

\newlength{\tmplen}

\newlength{\matFigX}
\newlength{\matFigY}

\begin{document}

\thispagestyle{empty}

\begin{center}
\null \vspace{\stretch{0.2}}
\renewcommand{\baselinestretch}{2}
\vspace{\stretch{1}}
Department of Bioengineering \\
Imperial College London
\vspace{\stretch{1}}

\textsc{\huge{\textbf{Hardware Realisation of Nonlinear Dynamical Systems for and from Biology}}}

\vspace{\stretch{1}}
\textsc{\large{\textbf{Hamid Soleimani}}} \\
\textsc{Supervisor: Prof. Emmanuel M. Drakakis}\\
\today
\vspace{\stretch{1}}

\vspace{\stretch{1}}
PhD Thesis
\vspace{\stretch{1}}

\vspace{\stretch{0.1}}
Submitted in partial filfullment of the requirements for the degree of \\
Doctor of Philosophy of Imperial College London \\
and the Diploma of Imperial College London \\

\end{center}

\newpage
\frontmatter
\doublespace
\setlength{\tmplen}{\parskip}
\setlength{\parskip}{-1ex}
\renewcommand{\baselinestretch}{\mystretch}
\newenvironment{dedication}
  {
   \thispagestyle{empty}
   \vspace*{\stretch{1}}
   \itshape             
   \raggedleft          
  }
  {\par 
   \vspace{\stretch{3}} 
   \clearpage           
  }

 \frontmatter              
  \begin{dedication}
  \begin{center}
    \large{To my beloved parents,\\
    and\\
    To Loabat\\
    my collaborator\\
    my philosopher\\
    my tranquilliser\\
    my wife
    }
    \vspace{\baselineskip}
    \end{center}
  \end{dedication}
\mainmatter             
\renewcommand{\baselinestretch}{2}
\chapter*{Abstract}
\markright{Abstract}
The focus of this thesis is on the applications of nonlinear dynamical systems in bioengineering where they are mainly used in large scale and generally categorised into two groups: (1) dynamical systems from biology (those are inspired by operational, architectural and/or anatomical characteristics encountered in natural biology information processing systems) (2) dynamical systems for biology (those are inspired by the need for innovative instrumentation as dictated by a specific biological or medical need). The mathematical models describing the dynamical systems used in the above systems can be simulated with the use of powerful software such as MATLAB, however, for large--scale simulations software begins to collapse. Besides, computer--based simulations are not always suitable for interfacing with biological/physical systems where continuous monitoring with low power and area consumption might be required. To alleviate these issues, a few novel hardware techniques for the both groups are proposed and the hardware results are compared and validated by software simulations. In particular, a compact and fully reconfigurable digital hardware model capable of mimicking 1--D, 2--D and 3--D nonlinear dynamical systems in real--time and large scale is presented. The performance of the proposed hardware model is tested on intra and extracellular biological models and compared to other digital hardware approaches. The proposed cellular model is synthesized on a digital platform for a single unit and a network model. Hardware synthesis, physical implementation on FPGA, and theoretical analysis confirm that the proposed cellular model can mimic the biological model behaviour with considerably low hardware overhead. Various networks constructed by pipelining 10k to 40k cellular calcium units are compared with an equivalent simulation run on a standard PC workstation. Results show that the cellular hardware model is, on average, ~83 times faster than the CPU version. The proposed reconfigurable cellular model has been also fabricated in the commercially available AMS 0.35 $\mu m$ technology capable of emulating slow intracellular calcium dynamics. The fabricated chip occupies an area of 1.5 $mm^2$ (excluding the area of the pads) and consumes 18.93 $nW$ for each calcium unit from a power supply of 3.3 V. The presented cytomimetic topology follows closely the behaviour of its biological counterpart, exhibiting similar time--domain calcium ions dynamics. Results show that the implemented design has better performance compared to its analog counterpart to speed up large--scale simulations of slow intracellular dynamics by sharing cellular units in real--time. A novel analog circuit supporting a systematic synthesis procedure of log--domain and strong inversion circuits capable of computing bilateral dynamical systems at high and low speed with acceptable precision is also proposed. The application of the method is demonstrated by synthesizing four different case studies. The validity of our approach is verified by nominal and Monte Carlo simulated results with realistic process parameters from the commercially available AMS 0.35 $\mu m$ technology. The resulting continuous--time and continuous--value circuits exhibit various bifurcation phenomena, nominal time--domain responses in good agreement with their mathematical counterparts and fairly acceptable process variation results (less than $5\%$ STD). All the aforementioned hardware techniques were developed for the first group (dynamical systems from biology) and at the end a flexible and efficient hardware classifier for biomedical time series classification is proposed for the second group (dynamical systems for biology). In this classifier, throughput is traded off with hardware complexity and cost using resource sharing techniques. This compromise is only feasible in systems where the underying time series has slow dynamics, such as physiological systems. A Long-Short-Term-Memory (LSTM) based architecture with ternary weight precision is employed and synthesized on a Xilinx FPGA. Hardware synthesis and physical implementation confirm that the proposed hardware can accurately classify hand gestures using surface--electromyographical time series data with low area and power consumption. Most notably, our classifier reaches 1.46$\times$ higher GOPs/Slice than similar state of the art FPGA--based accelerators. Finally, an automatic feature extractor is added to the mentioned classifier which effectively detects important feature from input time--series. A Convolutional Neural Network (CNN) is employed to extract input features and then a Long-Short-Term-Memory (LSTM) architecture with ternary weight precision classifies the input signals according to the extracted features. Hardware implementation on a Xilinx FPGA confirm that the proposed hardware can accurately classify multiple complex heart related time series data with low area and power consumption and outperform all previously presented state--of--the--art records.

\setlength{\parskip}{\tmplen}
\frontmatter
\doublespace
\setlength{\tmplen}{\parskip}
\setlength{\parskip}{-1ex}
\tableofcontents
\listoffigures
\listoftables
\renewcommand{\baselinestretch}{\mystretch}
\setlength{\parskip}{\tmplen}
\mainmatter
\fancyhead[RE]{\emph{Chapter \thechapter}}
\renewcommand{\baselinestretch}{\mystretch}
\chapter{Introduction}
\renewcommand{\baselinestretch}{\mystretch}
\label{chap:Intro}
\section{Background and Problem Statement}
\par Dynamical systems are one of the basic mathematical objects capable of describing time--dependent activities in a geometrical space. Such systems include a set of variables and constants defining the state, and a functional law describing the evolution of the state variables through time. In other words, the dynamical laws establish a meaningful relation among the future state of the system, the inputs and its current state. General qualitative descriptions of dynamical systems can be observed by inspecting their phase portraits, demonstrating velocity and direction of motions in space. The applications of such systems are highly diverse in science and engineering. Here, we only focus on the bioengineering applications where they are mainly used in large scale and generally categorised into two groups: (1) dynamical systems for biology (2) dynamical systems from biology. 

\subsection{Dynamical Systems from Biology}
Such dynamical systems are inspired by operational, architectural and/or anatomical characteristics encountered in natural biology information processing systems. Fast and large scale simulation and modelling of such studies are important for three main reasons: 1) there is still vast missing knowledge in the biochemical signaling pathways; a fast and high speed simulator is needed in order to discover how these biological systems operate by exploring and validating such systems using experimental data; 2) such modelling systems facilitate the development of bio--inspired prostheses (i.e. replacement of a biological system by an electronic circuit), 3) the development of such platforms benefiting from new principles of bio--inspired massively parallel computation can be useful in engineering applications such as new devices capable of learning and independent decision making. Recent studies have demonstrated that intracellular and extracellular signalling malfunctions may be associated with many types of disease. According to the relevant literatures, here we first review four specific types of non-mental human pathology that are identified in this connection: heart disease, early or late puberty, cancer and obesity explained respectively in the following:

\begin{enumerate}
\item Heart disease embrace a wide range of pathological conditions due to abnormal activity of the heart in performing the role of contraction and pumping blood into the body. These abnormalities can potentially originate from other disorders in the body. For instance, the rise in blood pressure during hypertension results in a constant workload on the heart. In the long term this can consequently lead to negative consequences for the heart. Although, in some cases the heart is able to revert to the normal state, in other cases the dysfunctions are irreversible and lead to a constant enlargement of the heart (cardiac hypertrophy) demonstrating of congestive heart failure (CHF) \cite{Berridge2014}. $Ca^{2+}$-dependent cell signalling driven by action potentials from nervous system is one of the main signalling pathways that activates the contraction and also controls the stability of the heart. Any failure in this signalling pathway may cause remodelling of the cardiac functionality and consequently lead to heart disease. Building upon this implication, Berridge and his colleagues \cite{Berridge2014} \cite{Berridge2003} have developed an abstract model introducing signalling pathways in normal and abnormal cardiac hypertrophy, which is shown in Figure \ref{fig:Figure1}. According to this model, normal hypertrophy is activated by a rise in the concentration of insulin-like growth factor 1 (IGF-1), acting on the PtdIns 3-kinase signalling pathway to elevate the protein synthesis rate through messenger ribonucleic acids (mRNAs). This can be a normal reaction of the heart to regular exercise in athletes. However, when the heart is exposed to a high pressure workload, the $Ca^{2+}$ signal starts changing in shape (increasing in both amplitude and width) and driving foetal gene transcription that alters the cardiac signalling and consequently leads to pathological hypertrophy and heart disease.

\begin{figure}[t]
  \centering
  \includegraphics[scale=0.6]{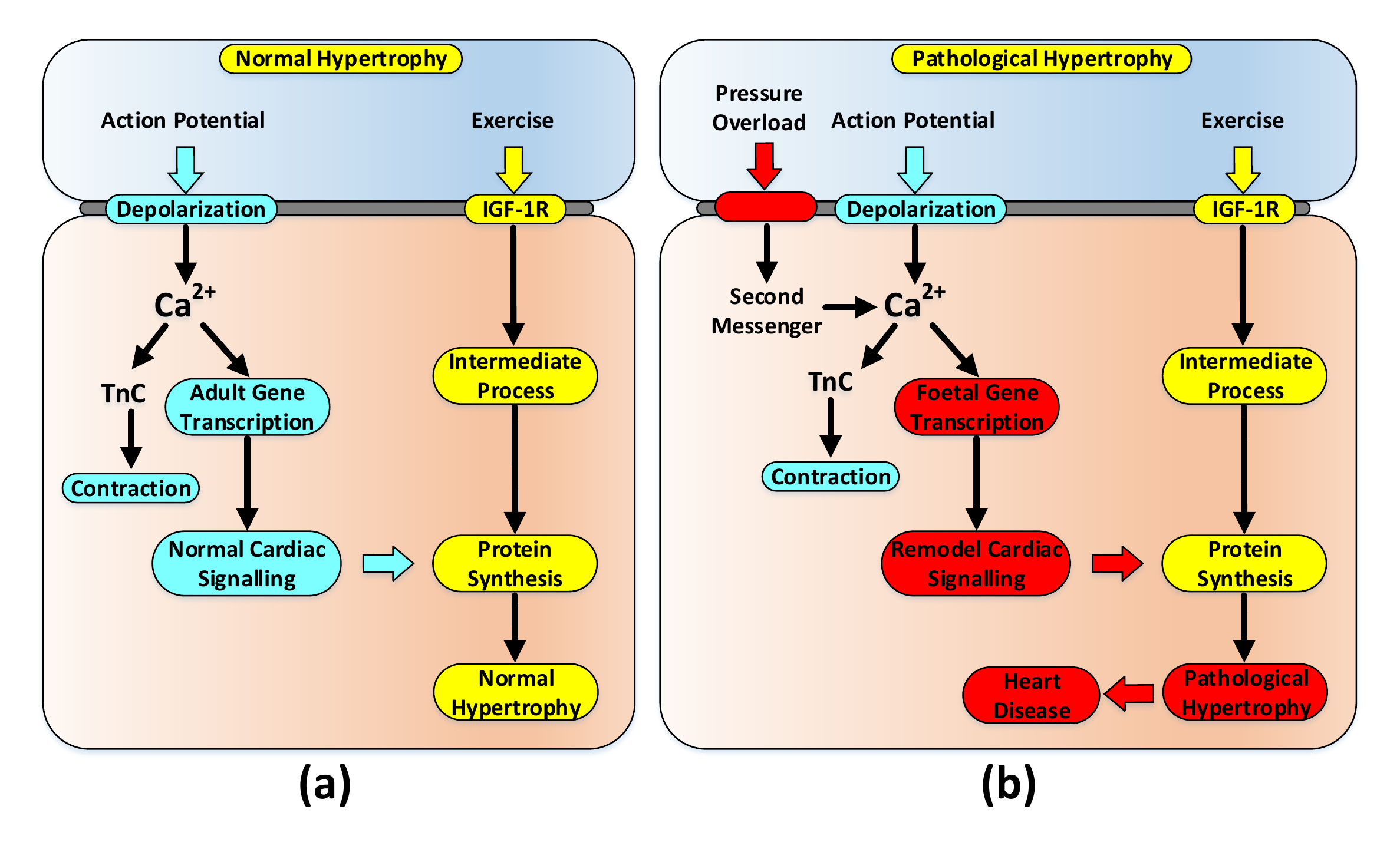}
  \vspace{-10pt}
  \caption{\footnotesize Signalling pathways in (a) normal and (b) abnormal cardiac hypertrophy. This figure is summarised and adopted from \cite{Berridge2014}.}
  \label{fig:Figure1}
\end{figure}

\item Gonadotropin-releasing hormone (GnRH) neurons are positioned in the hypothalamus and responsible for controlling the reproduction. This process is performed through the pituitary gland, which secrets GnRH into the blood in a very specific period of time during sexual development in adults. Failure in GnRH release by the hypothalamus or in stimulation of the pituitary gland implies that the puberty is delayed. In this section possible cell signalling pathways between GnRH neurons and other neural parts of hypothalamus are summarised. Besides, it is tried to highlight the defects in these pathways leading to the occurrence of early or late puberty. Early or late puberty disorder is an example of how a neural and an endocrine system collaborate together and how any abnormality in their interaction results in such a disorder. Various pathways from neural cells contributing in the excitatory and inhibitory regulation of GnRH neurons in the hypothalamus are shown in Figure \ref{fig:Figure2}(a). In this abstract model, Glu and KiSS neurons are excitatory and GABA neurons are inhibitory. Activation of excitatory neurotransmitters such as glutamate raises GnRH secretion and speeds up the sexual maturation process. In the same way, by releasing inhibitory neurotransmitters the GABA neurons are able to regulate GnRH secretion. These inhibitory neurotransmitters are bound to receptors positioned both on GnRH neurons and other nervous cells in the network and change the overall behaviour of the structure \cite{Ojeda2006}.

\begin{figure}[t]
  \centering
  \includegraphics[scale=0.55]{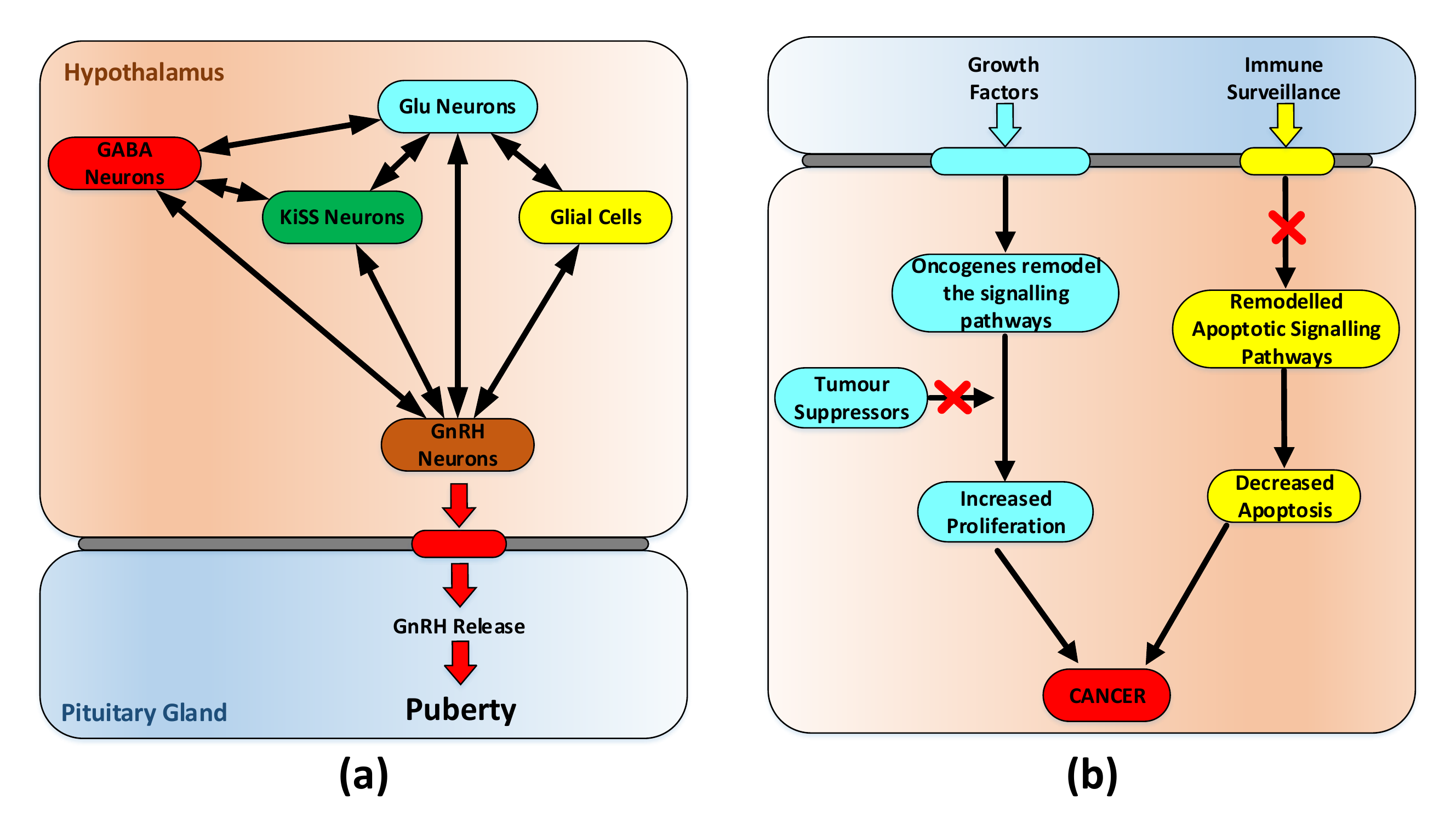}
  \vspace{-10pt}
  \caption{\footnotesize (a) An abstract model of various pathways applied from other neurons or glial cells contributing in the excitatory and inhibitory regulation of GnRH neurons in the hypothalamus. This figure is summarised and adopted from \cite{Ojeda2006}. (b) Summary of the main genotypic remodelling processes leading to the development of cancer. The suppressed or deactivated connections by means of cancer are marked in red cross. This figure is summarised and adopted from \cite{Berridge2014}.}
  \label{fig:Figure2}
\end{figure}

  \par As can be seen in the Figure, glial cells are also able to assist GnRH secretion through growth factor-dependent cell-cell signalling. The coordinated and time specified neuroglial interactions in the following structure depends on the dense control by a complex mechanism of excitatory and inhibitory neurotransmitters. Any defects in intracellular and extracellular signalling pathways may lead to early or late puberty. This has been confirmed by experiment in animal models. For instance, blocking of the receptors connected to the GnRH neurons of female rats leads to late sexual maturation \cite{Ojeda2008}.

\item Every day we see people who suffer from a type of cancer for which no medicine has been developed yet. Unfortunately, it can strike at any age without notification, and treatment is not attainable in many cases. The development of cancer is a multistep process associated with four to seven various genotypic mutations. Cancer is an example of intracellular remodelling and defects of the signalling pathways, in which the genotypic information is modified. An abstract model of the main genotypic remodelling has been introduced by \cite{Berridge2014} seen in Figure \ref{fig:Figure2}(b), in which the suppressed or deactivated connections by means of cancer are marked in red cross. These undesirable modifications occur in three main signalling systems: Firstly, through interaction of proteins and oncogenes, these genes are activated leading to proliferation of cancer cell. Secondly, the tumour suppressor genes that normally inhibit the proliferative signalling pathways, are switched off. Thirdly, there are mutations inner cell that decrease apoptosis process and result in the enhancement of cell proliferation. These abnormal processes have been observed in experiments on cancerous cells, but how and where they are initiated in many cases is still unclear.
\item Although obesity is not technically categorised as a severe disease, it is firmly associated with many debilitating disorders, such as diabetes and heart disease. Here, we discuss possible defects in cell signalling pathways that may lead to the development of obesity. In general, obesity is developed when there is an imbalance in the metabolic energy network. This can be caused by various cell signalling defects that are associated to the synaptic plasticity of hypothalamic neural system \cite{Kim2014}. Leptin and ghrelin are two hormones with a significant impact on energy balancing in the body. Leptin mediates long-term control of energy balance. Since it provides negative feedback in the brain to inhibit energy intake and to regulate energy homeostasis and body weight, it is called the “satiety hormone”. By contrast, ghrelin is a fast mediator, encouraging the digestive system to take a meal \cite{Klok2007}. Both hormones may be passed into the brain through the bloodstream or the vagal nerve, inducing the feeling of satiety or hunger. Clearly, any abnormality in the release of these hormones may lead to the development of obesity. Kim and his colleagues \cite{Kim2014} have investigated the role of astrocytes on leptin signalling and consequently obesity. A genetic mouse model was produced in which leptin receptors were not expressed in astrocytes located in the hypothalamus. The researchers report two important findings. First, the glial connections form and synaptic inputs onto hypothalamus neurons associated in feeding control was changed. Second, feeding after fasting and level of ghrelin hormone were raised. These results could demonstrate that, like heart disease, early or late puberty and cancer, obesity can also be associated with intracellular and extracellular signalling defects.
\end{enumerate}

\par The mathematical models describing the dynamical systems used in the above systems can be simulated with the use of powerful software such as MATLAB, however, for large--scale simulations software begins to collapse. Besides, computer--based simulations are not always suitable for interfacing with biological/physical systems where continuous monitoring with low power and area consumption might be required \cite{Papadimitriou2013}. This issue can be resolved by the means of specialized hardware tools capable of emulating dynamical behaviours in real--time \cite{Sarpeshkar2010}.

\subsection{Dynamical Systems for Biology}
Such dynamical systems are inspired by the need for innovative instrumentation as dictated by a specific biological or medical need. For example, recognizing internal activities of the human body based on biologically generated time series data is at the core of technologies used in wearable rehabilitation devices \cite{patel} and health support systems \cite{mazilu}. Some commercial examples include fitness trackers or fall detection devices. Wearable activity recognition systems are generally composed of sensors, such as accelerometers, gyroscopes or magnetic field/chemical sensors \cite{bulling} and a processor used to analyze the generated signals. Real--time and accurate interpretation of the recorded physiological data from these devices can be considerably helpful in preventing and treatment of a number of diseases  \cite{oscar}. For instance, patients with diabetes, obesity or heart disease are often required to be closely monitored and follow a specific exercise set as part of their treatments \cite{Jia}. Similarly, patients with mental pathologies such as epilepsy can be monitored to detect abnormal activities and therefore prevent negative consequences \cite{Yin}.

\par However, most current commercial products only offer relatively simple metrics, such as step count or heart beat and lack the complexity and computing power for many time series classification problems of interest in real time. The emergence of deep learning methodologies capable of learning multiple layers of feature hierarchies and temporal dependencies in time series problems and increased processing capabilities in wearable technologies lay the ground to perform more detailed data analysis on--node and in real time \cite{ravi}. The ability of performing more complex analysis, such as human activity classification on the wearable device could potentially filter data streaming from the device to host and save data bandwidth link. This data saving is more visible in the cases where the classification task should be continuously preformed on the patient such as in seizure detection for epileptic patients. The core of such deep learning classifiers includes a large number of nonlinear discrete dynamical systems. However, due to the high computational power and memory bandwidth required by deep learning algorithms, full realization of such systems on wearable and embedded medical devices is still challenging.

\section{Hardware Solutions for the Realisation of Nonlinear Dynamical Systems}
\par A number of solutions for realization of these systems have been devised yet, ranging from time-continuous low power analog circuits to time-discrete massively parallel digital ones \cite{Furber2014}--\cite{Soleimani2012}. Here, we summarize the main approaches:

\begin{enumerate}
\item Special purpose computing architectures have been developed to simulate complex biological networks via their special software tools \cite{Furber2014}--\cite{Minkovich2014}, \cite{GPU}, \cite{Intel}. Even though these systems are biologically plausible and flexible with remarkably high performance thanks to their massively parallel architecture, they run on bulky and power-hungry workstations with relatively high cost and long development time. Hence, this approach is often expensive and not widespread for public access.
\item Analog CMOS platforms are considered to be the main choice for direct implementation of intra-- and extracellular biological dynamics \cite{Indiveri2006}--\cite{Papadimitriou2013}. This approach is very power efficient, however, model adjustment is generally challenging in these circuits. Moreover, since the non--linear functions in the target models are directly synthesised by exploiting the inherent non--linearity of the circuit components, very good layout is imperative in order for the resulting topologies not to suffer from the variability and mismatch \cite{Houssein2015} and \cite{24}. Another issue has to do with the realization of slow biological dynamics (such as most intercellular dynamics in CytoMimetic circuits) and also when their time scale must be precisely implemented (e.g. for bio--inspired prosthesis that must be interfaced with real biological systems). In such cases, the size of capacitors may become very large (for example, in \cite{Woo2015} it is reported as 1 uF) for a monolithic realization.
\item Full custom analog/digital (mixed mode) implementations capable of sharing the features of both analog and digital world. This approach comprises low--power and fast analog circuits with programmable and mismatch immune digital circuits. Generally, in these systems, biological computation is performed in the analog domain and the communication of information between these units is implemented in the digital domain \cite{Schemmel2008}--\cite{Bavandpour2014}. However, analog part scaling down in technology is challenging due to mismatch and process variations. Furthermore, replication of slow biological dynamics suffers from similar issues as analog CMOS platforms.

\item Digital platforms are good candidates nowadays for implementing such biological and bio-inspired systems. Most digital approaches \cite{Truenorth2016}--\cite{Soleimani2012}, \cite{Google} use digital computational units to implement the mathematical equations codifying the behavior of biological intra/extracellular dynamics. Such a platform can be either implemented on FPGAs or custom ICs \cite{14} and \cite{15}, with FPGAs providing lower development time and more configurability, but at a higher cost in terms of power, area and speed compared to the digital custom ICs. Generally, a digital platform benefits from high reconfigurability, short development time, notable reliability and immunity to device mismatch. Although, the digital platform's silicon area and power consumption is comparatively high compared to its analog counterpart in typical academic technologies (such as 0.35 uM and 0.18 uM AMS), with remarkable progress in nowadays fabrication technologies it may be argued that realization in scaled down technology nodes (e.g. 28 nM \cite{Truenorth2016}) without scarifying the advantages is possible.
\end{enumerate}

\section{Contributions and Thesis Organisation}
The thesis is structured as follows:

\begin{itemize}
  \item \textbf{Chapter2:} This chapter presents a compact and fully reconfigurable digital hardware model capable of mimicking 1--D, 2--D and 3--D nonlinear dynamical systems in real--time and large scale. The performance of the proposed hardware model is tested on intra and extracellular biological models and compared to other digital hardware approaches. First, a nonlinear intracellular calcium model capable of mimicking Hopf bifurcation phenomenon and various nonlinear responses of the biological calcium dynamics is investigated. The proposed cellular model is synthesized on a digital platform for a single unit and a network model. Hardware synthesis, physical implementation on FPGA, and theoretical analysis confirm that the proposed cellular model can mimic the biological calcium behaviour with considerably low hardware overhead. The approach has the potential to speed up large--scale simulations of slow intracellular dynamics by sharing more cellular units in real--time. To this end, various networks constructed by pipelining 10k to 40k cellular calcium units are compared with an equivalent simulation run on a standard PC workstation. Results show that the cellular hardware model is, on average, ~83 times faster than the CPU version. Then the proposed hardware model is adopted and simplified for the realisation of (2--D) neuron models as well as other higher dimensional models. The model is applied to the Izhikevich and FitzHugh--Nagumo neuron models as 2--D case studies and to the Hindmarsh--Rose model as a 3--D case study. Hardware synthesis and physical implementations show that the resulting circuits can reproduce neural dynamics with acceptable precision and considerably low hardware overhead compared to previously published piecewise linear models.
  \item \textbf{Chapter3:} Low power/area cytomorphic chips may be interfaced and ultimately implanted in the human body for cell--sensing and cell--control applications of the future. In such electronic platforms, it is crucial to accurately mimic the biological time--scales and operate in real--time. This chapter proposes a methodology where slow nonlinear dynamical systems describing the behavior of naturally encountered biological systems can be efficiently realised in hardware. To this end, as a case study, a low power and efficient digital ASIC capable of emulating slow intracellular calcium dynamics with time--scales reaching to seconds has been fabricated in the commercially available AMS 0.35 $\mu m$ technology and compared with its analog counterpart. The fabricated chip occupies an area of 1.5 $mm^2$ (excluding the area of the pads) and consumes 18.93 $nW$ for each calcium unit from a power supply of 3.3 V. The presented cytomimetic topology follows closely the behavior of its biological counterpart, exhibiting similar time--domain calcium ions dynamics. Results show that the implemented design has the potential to speed up large--scale simulations of slow intracellular dynamics by sharing cellular units in real--time.
  \item \textbf{Chapter4:} Simulation of large--scale nonlinear dynamical systems on hardware with a high resemblance to their mathematical equivalents has been always a challenge in engineering. This chapter presents a novel current--input current--output circuit supporting a systematic synthesis procedure of log--domain circuits capable of computing bilateral dynamical systems with considerably low power consumption and acceptable precision. Here, the application of the method is demonstrated by synthesizing four different case studies: 1) a relatively complex two--dimensional (2--D) nonlinear neuron model, 2) a chaotic 3--D nonlinear dynamical system Lorenz attractor having arbitrary solutions for certain parameters, 3) a 2--D nonlinear Hopf oscillator including bistability phenomenon sensitive to initial values and 4) three small neurosynaptic networks comprising three FHN neuron models variously coupled with excitatory and inhibitory synapses. The validity of our approach is verified by nominal and Monte Carlo simulated results with realistic process parameters from the commercially available AMS 0.35 $\mu m$ technology. The resulting continuous--time, continuous--value and low--power circuits exhibit various bifurcation phenomena, nominal time--domain responses in good agreement with their mathematical counterparts and fairly acceptable process variation results (less than $5\%$ STD).
  \item \textbf{Chapter5:} Modern wearable rehabilitation devices and health support systems operate by sensing and analysing human body activities. The information produced by such systems requires efficient methods for classification and analysis. Deep learning algorithms have shown remarkable potential regarding such analyses, however, the use of these algorithms on low--power wearable devices is limited by resource constraints. Most of the available on--chip deep learning processors contain complex and dense hardware architectures in order to achieve the highest possible throughput. Such a trend in hardware design may not be efficient in applications where on--node computation is required and the focus is more on the area and power efficiency such as portable and embedded biomedical devices. The aim of this chapter is to overcome some of the limitations in a current typical deep learning framework and present a flexible and efficient platform for biomedical time series classification. Here, throughput is traded off with hardware complexity and cost using resource sharing techniques. This compromise is only feasible in systems where the underying time series has slow dynamics, such as physiological systems. A Long-Short-Term-Memory (LSTM) based architecture with ternary weight precision is employed and synthesized on a Xilinx FPGA. Hardware synthesis and physical implementation confirm that the proposed hardware can accurately classify hand gestures using surface--electromyographical time series data with low area and power consumption. Most notably, our classifier reaches 1.46$\times$ higher GOPs/Slice than similar state of the art FPGA--based accelerators.
    \item \textbf{Chapter6:} In the final Chapter of the thesis, a summary of this work is illustrated and the produced results are discussed in a critical way. The contribution of this work to the field of low-power BioElectronics is examined and is accompanied by some useful directions, regarding any potential future research in this field.
\end{itemize}
\chapter{Proposed Cellular Model}
\renewcommand{\baselinestretch}{\mystretch}
\label{chap:ANBCF}

\par This chapter presents a compact and fully reconfigurable digital hardware model \cite{Soleimani2017} capable of mimicking 1--D, 2--D and 3--D nonlinear dynamical systems in real--time and large scale. The performance of the proposed hardware model is tested on intra and extracellular biological models and compared to other digital hardware approaches. First, a nonlinear intracellular calcium model capable of mimicking Hopf bifurcation phenomenon and various nonlinear responses of the biological calcium dynamics is investigated. The proposed cellular model is synthesized on a digital platform for a single unit and a network model. Hardware synthesis, physical implementation on FPGA, and theoretical analysis confirm that the proposed cellular model can mimic the biological calcium behaviour with considerably low hardware overhead. The approach has the potential to speed up large--scale simulations of slow intracellular dynamics by sharing more cellular units in real--time. To this end, various networks constructed by pipelining 10k to 40k cellular calcium units are compared with an equivalent simulation run on a standard PC workstation. Results show that the cellular hardware model is, on average, ~83 times faster than the CPU version.
\par Then the proposed hardware model is adopted and simplified for the realisation of (2--D) neuron models as well as other higher dimensional models. The model is applied to the Izhikevich and FitzHugh--Nagumo neuron models as 2--D case studies and to the Hindmarsh--Rose model as a 3--D case study. Hardware synthesis and physical implementations show that the resulting circuits can reproduce neural dynamics with acceptable precision and considerably low hardware overhead compared to previously published piecewise linear models.

\par Digital platforms seem to be very promising especially for CytoMimetic circuit design in which the dynamics are slow (in certain cases, time scales reach minutes or even hours \cite{Elowitz2000}) and more cell units can be shared and synthesized in digital hardware. Our approach described here falls in this category and is a synchronous cellular--based system that discretizes intracellular calcium dynamics \cite{Dupont1993} into a cellular space and recreates the time domain signals with less computational effort compared to the other approaches. This system is compact, fully reconfigurable and applies no serious constraint on the hardware critical path. These features would make the approach appropriate for implementing other two dimensional neuromorphic and cytomorphic dynamical systems. 

\section{Calcium Released Calcium Induced (CICR) Model}
The CICR model introduced in \cite{Dupont1993}, describes accurately intracellular $Ca^{2+}$ oscillations. In this model the amount of $Ca^{2+}$ released is tuned by the level of the input stimulus modulated by the $IP_3$ level. The description of intracellular $Ca^{2+}$ oscillations in this model is given by the following two--dimensional (2--D) minimal model:

\begin{equation}\label{eq:2_1}
\frac{dx}{dt} =z_0+z_1\beta-z_2(x)+z_3(x,y)+k_fy-kx
\end{equation}
\begin{equation}\label{eq:2_2}
\frac{dy}{dt} =z_2(x)-z_3(x,y)-k_fy
\end{equation}
where
\begin{equation}\label{eq:2_3}
z_2(x) =V_{M_2}\frac{x^n}{K_2^n+x^n}
\end{equation}
\begin{equation}\label{eq:2_4}
z_3(x,y) =V_{M_3}\frac{y^m}{K_R^m+y^m}\frac{x^p}{K_A^p+x^p}
\end{equation}
with $y$ and $x$ representing the concentration of free $Ca^{2+}$ in the $IP_3$ insensitive pool and in the cytosol, respectively. Besides, $z_0$ represents the constant $Ca^{2+}$ input from the extracellular medium and $z_1$ is the $IP_3$ modulated release of $Ca^{2+}$ from the $IP_3$ sensitive pool. The parameter $\beta$ refers to the amount of $IP_3$ and measures the saturation of the $IP_3$ receptor. The biochemical rates $z_2$ and $z_3$ refer to the pumping of $Ca^{2+}$ into the $IP_3$ insensitive pool and to the release of $Ca^{2+}$ from that pool into the cytosol respectively. The parameters $V_{M_2}$ , $V_{M_3}$ , $K_2$, $K_R$, $K_A$, $k_f$ and $k$ are the maximum values of $z_2$ and $z_3$, threshold constants for pumping, release and activation and rate constants, respectively. Parameters $n$, $m$, and $p$ represent the Hill coefficients describing the pumping, release and activation processes, respectively. According to the values of the Hill coefficients, various degrees of cooperativity can be obtained and this allows us to simulate various intracellular calcium activity. The different values of the biological model parameters are shown in Table \ref{tab:Table}. In the next chapter, it is explained first how to derive the proposed cellular model and then all three different sets of Hill functions are implemented on the cellular model with the same structure in order to show the reconfigurability of the model.

\begin{table}[t]
\captionsetup{font=footnotesize}
\caption{Biological Values for the intracellular $Ca^{2+}$
Oscillations Model with Various Hill Functions.}   
\centering          
\begin{tabular}{c c c c}    
\hline\hline                        
Parameters &M=N=P=1&M=N=P=2&M=N=2, P=4\\ [0.5ex]  
\hline                      
$z_0~(\mu M/s)$ & 1 & 1  & 1  \\
$z_1\beta~(\mu M/s)$ & 2 & 6 & 3 \\
$V_{M_2}~(\mu M/s)$ & 250 & 100 & 65 \\
$V_{M_3}~(\mu M/s)$ & 2000 & 700 & 500 \\
$K_2~(\mu M)$ & 1 & 1 & 1 \\
$K_R~(\mu M)$ & 30 & 15 & 2 \\
$K_A~(\mu M)$ & 2.5 & 2.5 & 0.9 \\
$k_f~(s^{-1})$ & 0.1 & 0 & 1 \\
$k~(s^{-1})$ & 5 & 8 & 10 \\ [1ex]        
\hline          
\end{tabular}
\label{tab:Table}    
\end{table}

\section{Synchronous Cellular Calcium Model}
\par According to \cite{Bavandpour2014}, we convert the phase plane of the biological calcium models into a 2--D cellular space where $(x,y)\in Z^2$ represents the location of state points in the phase plane, and $(\frac{dx}{dt}, \frac{dy}{dt})$ determines the velocity and direction of the motion. The x--nullcline and y--nullcline are defined as the set of points where $\frac{dx}{dt}=0$ and $\frac{dy}{dt}=0$ respectively. Clearly the points of intersection between these two sets of arrays are defined as the equilibrium points. If we consider the phase plane as a cellular space, with velocity vectors in the cellular space, the time domain signals can be easily recreated. The cellular mapping of the proposed 2--D cellular system and the corresponding time--continuous nullclines are shown in Figure \ref{fig:Figure3}. In this figure, the minimum speed motion corresponding to the minimum velocity value is shown in blue, the maximum speed motion corresponding to the maximum velocity value is shown in red, and the equilibriums at which the velocity value is almost zero are shown in white. To explain the proposed cellular model we re--express equations (\ref{eq:2_1}) and (\ref{eq:2_2}) as the following:

\begin{figure*}[th]
\vspace{-10pt}
\centering
\includegraphics[scale=0.15]{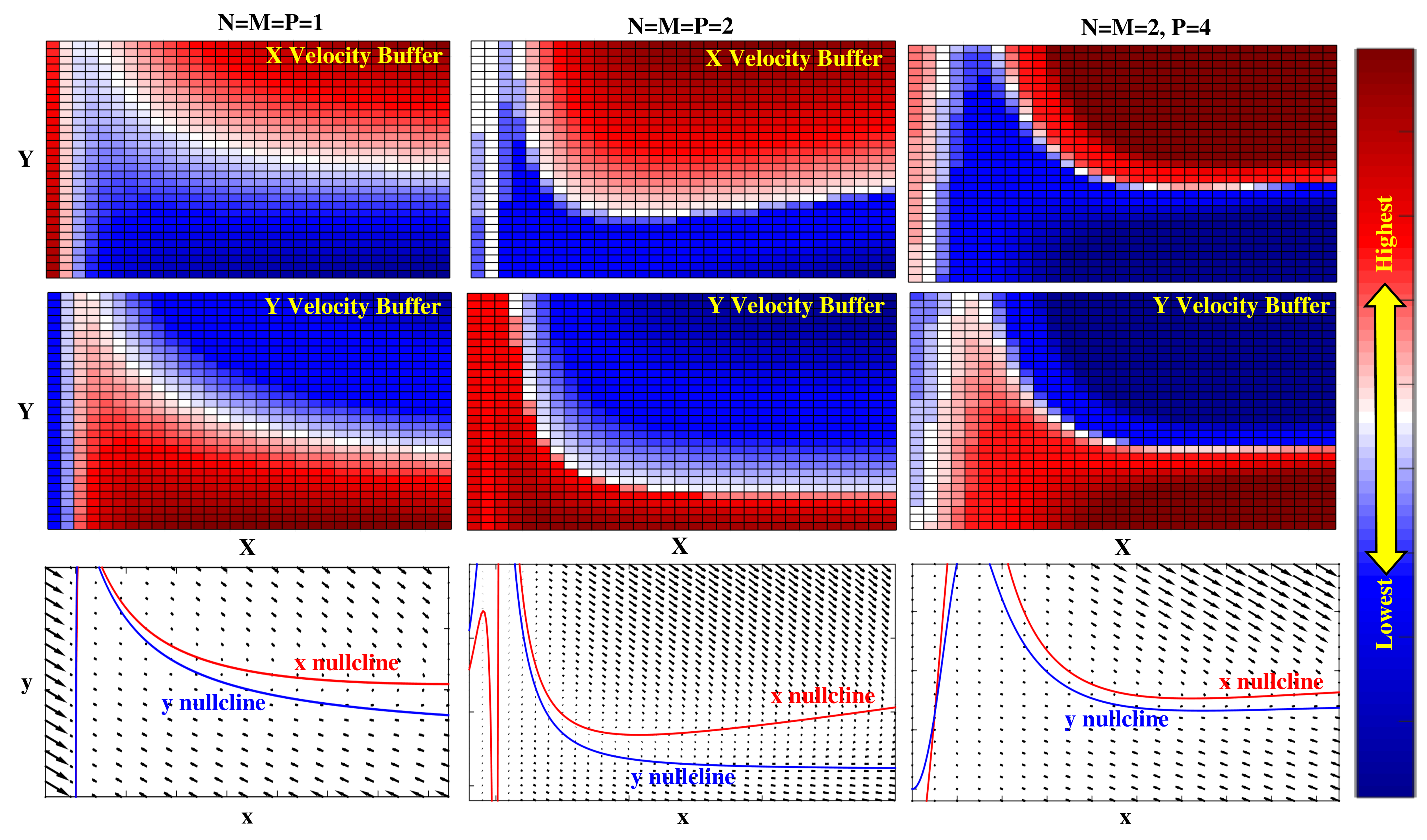}
\vspace{-10pt}
\captionsetup{font=footnotesize}
\caption{Cellular mapping of the X and Y velocity buffers and corresponding time-continuous nullclines for various set of Hill functions. The minimum speed motion corresponding to the minimum velocity value is shown in blue color, and the maximum speed motion corresponding to the maximum velocity value is shown in red, and the equilibriums in which the velocity value is almost zero are shown in white.}
\vspace{-5pt}
  \label{fig:Figure3}
\end{figure*}

\begin{equation}\label{eq:2_5}
\begin{cases}\frac{dx}{dt} = F(x,y)+IN_{ext}\\
\frac{dy}{dt}=G(x,y)
\end{cases}
\end{equation}
where
\begin{equation}\label{eq:2_6}
\begin{cases}F(x,y)=z_0-z_2(x)+z_3(x,y)+k_fy-kx\\
G(x,y)=z_2(x)-z_3(x,y)-k_fy\\
IN_{ext}=z_1\beta.
\end{cases}
\end{equation}
\par In the cellular space, we considered the phase plane as a discrete space and accordingly the state variables were points connecting the whole cellular phase plane together. One can show that $X(t)=i$ where $ i\in \textbf{N}\equiv {0,1,...,N-1}$ and $Y(t)=j$ where $ j\in \textbf{M}\equiv {0,1,...,M-1}$ are discrete variables corresponding to $X$ and $Y$ in the continuous space. The location of each state point in the $\textbf{N}\times \textbf{M}$ cellular space can be defined as:
\begin{equation}\label{eq:2_7}
\begin{cases}X(t)= \lfloor \frac{x(t)-x_{min}}{\Delta x} \rfloor ~~~;~~~~x(t) \in [x_{min},x_{max})\\
Y(t)= \lfloor \frac{y(t)-y_{min}}{\Delta y} \rfloor ~~~;~~~~y(t) \in [y_{min},y_{max})
\end{cases}
\end{equation}
where $\Delta x= \frac{x_{max}-x_{min}}{N}$ and $\Delta y= \frac{x_{max}-x_{min}}{M}$. In the next sections, it is explained how these values can change the truncation error in the proposed cellular model. The next state of each dynamical variables is obtained as:
\begin{equation}\label{eq:2_8}
\begin{cases}x^+-x^-=\Delta t\cdot F(x^-,y^-) \\
y^+-y^-=\Delta t\cdot G(x^-,y^-)
\end{cases}
\end{equation}
where
\begin{equation}\label{eq:2_9}
\begin{cases}F(x^-,y^-)=F(x_{min}+X^-.\Delta x,~y_{min}+Y^-.\Delta y) \\
G(x^-,y^-)=G(x_{min}+X^-.\Delta x,~y_{min}+Y^-.\Delta y).
\end{cases}
\end{equation}

\par Unlike the cellular model in \cite{Bavandpour2014}, the value of each state variable in the proposed model is changed synchronously. This implies that the timing of both state variables are the same, and the address of the next cellular velocity fetched from memory is indirectly related to the output variables. This feature allows the output register to work in any precision leading to more accuracy for the proposed approach with less hardware overhead compared to the previous cellular model \cite{Bavandpour2014}.

\par The direction of new motions on the cellular space can be formulated as:
\begin{equation}\label{eq:2_10}
\begin{cases}
X^-_{new}-X^-_{old}=0~\Rightarrow~\text{"no change"}\\
X^-_{new}-X^-_{old}\ge 1~\Rightarrow~\text{"upward motion"}\\
X^-_{new}-X^-_{old}\le -1~\Rightarrow~\text{"downward motion"}
\end{cases}
\end{equation}
\begin{equation}\label{eq:2_11}
\begin{cases}
Y^-_{new}-Y^-_{old}=0~\Rightarrow~\text{"no change"}\\
Y^-_{new}-Y^-_{old}\ge 1~\Rightarrow~\text{"upward motion"}\\
Y^-_{new}-Y^-_{old}\le -1~\Rightarrow~\text{"downward motion"}.
\end{cases}
\end{equation}

It should be noted that the resultant motion on the 2-D cellular phase planes is determined by a combination of motions in both the $X$ and $Y$ directions and unlike \cite{Bavandpour2014} the number of motions on the cellular space can correspond to more than one step in each clock cycle.
\section{Simulated Time Domain Analysis}
\subsection{Single Calcium Behavior}
\begin{figure*}[th]
\vspace{-10pt}
\centering
\includegraphics[scale=0.35]{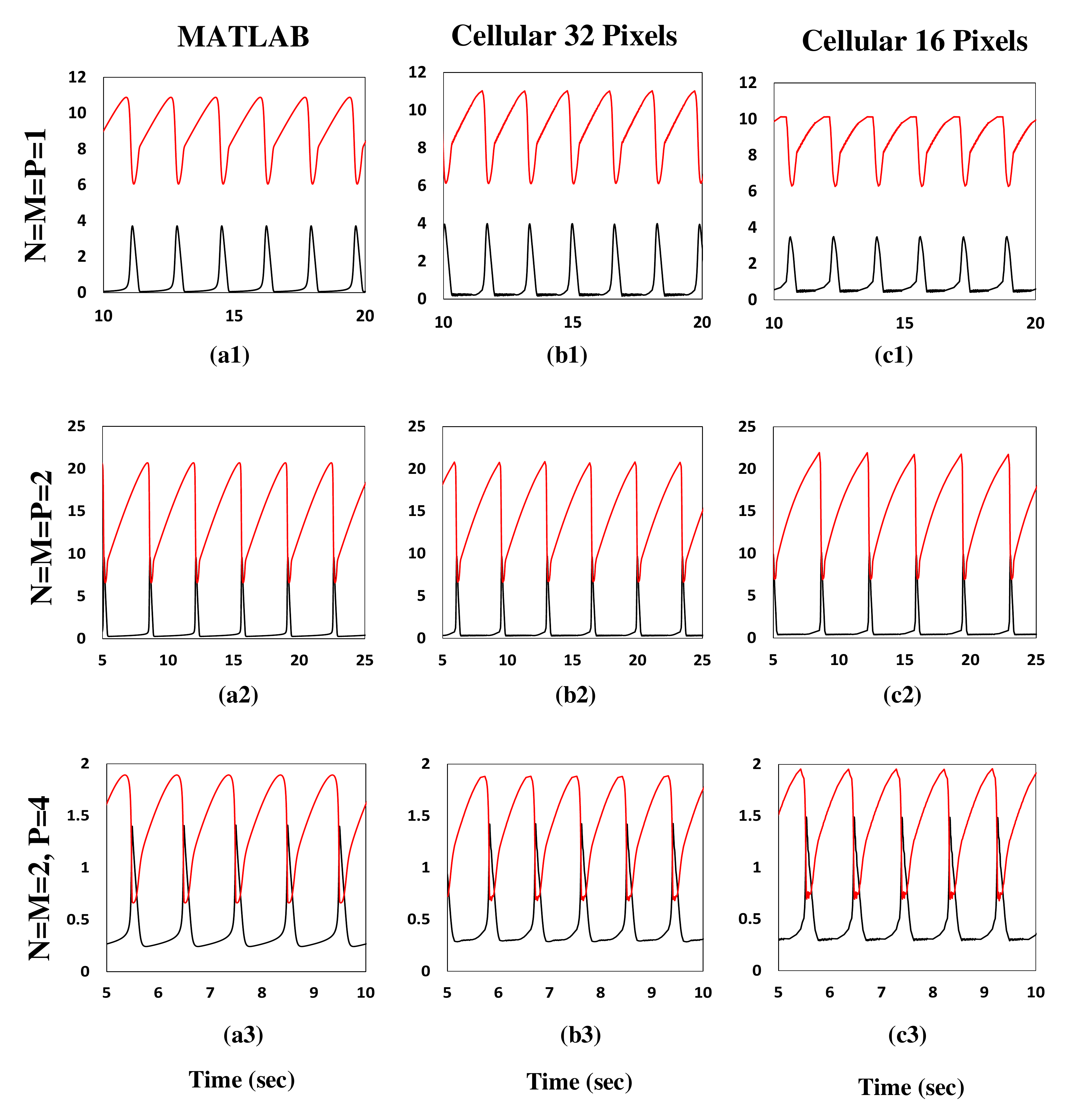}
\captionsetup{font=footnotesize}
\caption{Comparison of transient analysis results generated by the biological and the proposed cellular models for various sets of Hill functions. (a1-c3) The time domain comparison between the biological and cellular models with two different resolutions (32 and 16 pixels) for a single calcium unit.}
\vspace{-15pt}
  \label{fig:Figure4}
\end{figure*}

The simulation results for the biological model and cellular model with two different resolutions simulated by MATLAB are shown in Figure \ref{fig:Figure4} (a1-c3). According to these results, in this section we show that the proposed cellular model can exhibit various time domain responses with a remarkable compliance compared to the biological ones. The biological values extracted from \cite{Papadimitriou2013} and the cellular parameter correspond to each set of Hill functions are shown in Table \ref{tab:Table1} and \ref{tab:Table2} respectively.

\begin{table}[t]
\captionsetup{font=footnotesize}
\caption{Cellular Values for the intracellular $Ca^{2+}$
Oscillation Model with Various Hill Functions.}   
\centering          
\begin{tabular}{c c c c c}    
\hline\hline                        
Resolution& Parameters &M=N=P=1&M=N=P=2&M=N=2, P=4\\ [0.5ex]  
\hline                      
\multirow{6}{4em}{32-pixels} &$\Delta x$ & 0.25  & 0.5 & 0.0625 \\
&$\Delta y$ & 0.25 & 0.5 & 0.0625 \\
& $x_{min}$ & -0.6 & -1 & -0.1 \\
& $x_{max}$ & 7.4 & 15 & 1.9 \\
& $y_{min}$ & 2.5& 4 & -0.1 \\
& $y_{max}$ & 10.5& 20 & 1.9 \\
\hline
\multirow{6}{4em}{16-pixels}  &$\Delta x$ & 0.25 & 0.5 & 0.125 \\
&$\Delta y$ & 0.25 & 1 & 0.125 \\
& $x_{min}$ & -0.6 & -1& -0.1 \\
& $x_{max}$ & 7.4 &7 & 1.9  \\
& $y_{min}$ & 2.5 &4& -0.1 \\
& $y_{max}$ & 10.5 & 20&1.9  \\[1ex]
\hline          
\end{tabular}
\vspace{-10pt}
\label{tab:Table1}    
\end{table}

\par The first case of the CICR model ($m=n=p=1$) shows that the mechanisms of pumping, release and activation can be demonstrated by intrinsic Michaelian processes. The simulation results in Figure \ref{fig:Figure4} (a1-c1) show a good agreement between the biological model and the 32-pixels cellular model. However the 16-pixels cellular model cannot accurately mimic this behavior since the non-linearity degree of the dynamical system is relatively high.
\par The second case ($m=n=p=2$) is described by a Hill coefficient of 2 and demonstrates a less mild nonlinear system, in comparison with the previous case. The simulation results are shown in Figure \ref{fig:Figure4} (a2-c2). In this case as well, the time domain signals illustrate that the biological and cellular systems are adequately close while 16-pixels cellular model still shows a lower accuracy compared to the 32-pixels cellular model.
\par The third case ($m=n=2, p=4$) is the contains the highest-order of Hill coefficients equal to 4, corresponding to a stronger nonlinear response, in which small truncation errors can significantly disturb the targeted dynamics. The simulated results are shown in Figure \ref{fig:Figure4} (a3-c3), and the similarities between the two systems are satisfying even in the 16-pixels resolution.

\begin{figure*}[th]
\vspace{-10pt}
\centering
\includegraphics[scale=0.35]{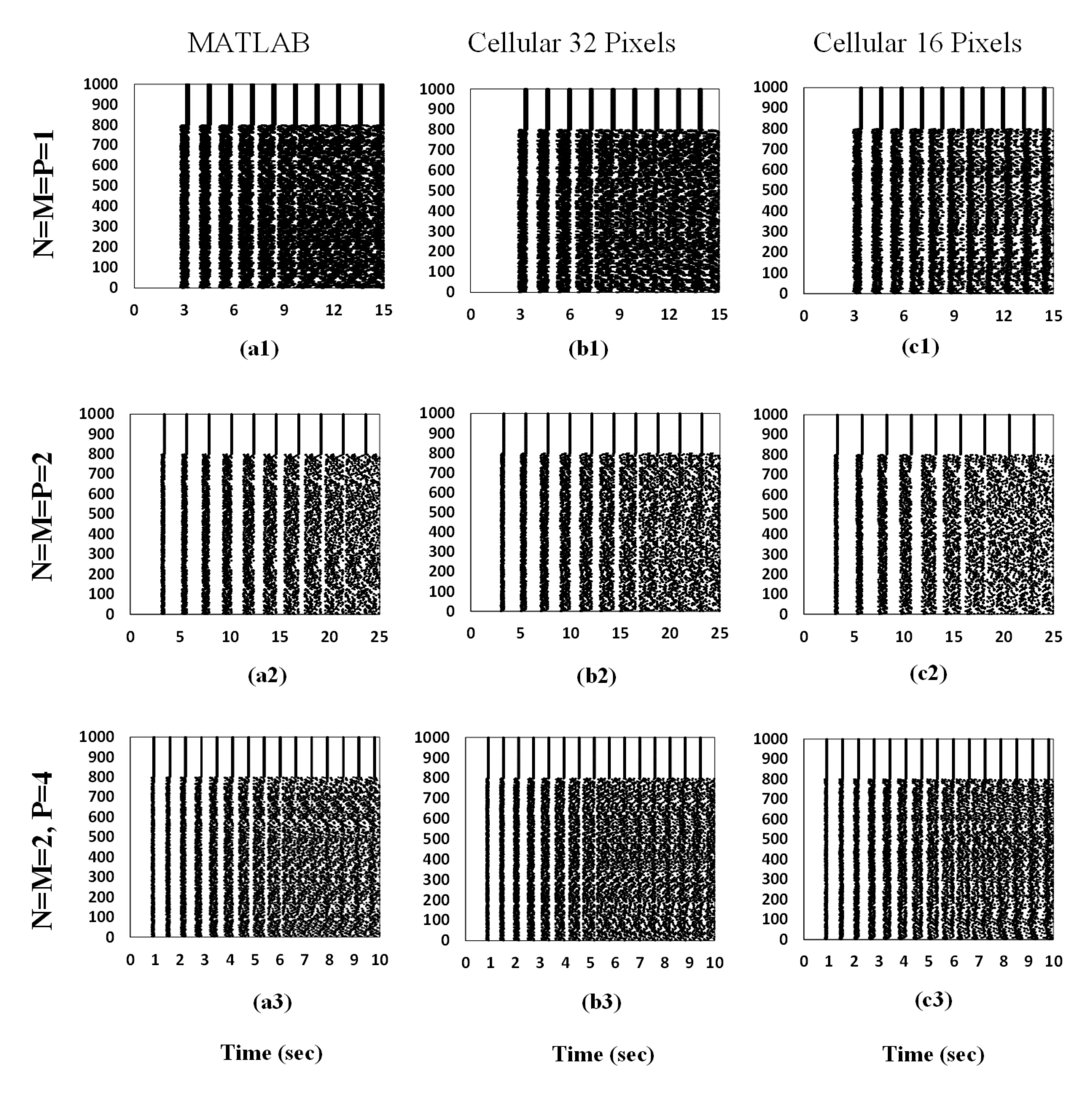}
\captionsetup{font=footnotesize}
\caption{Comparison of transient analysis results generated by the biological and proposed cellular models for various set of Hill functions. (a1--c3) Raster plots for a comparison of network activity of one thousands calcium units with inhomogeneous inputs between the biological and cellular models with different resolutions (32 and 16 pixels).}
\vspace{-15pt}
  \label{fig:Figure6}
\end{figure*}

\subsection{Networked Calcium Behavior}
To investigate the applicability of the proposed cellular model in a large scale simulation, a network model constructed by one thousands calcium units is simulated. To check the stability of the networks in both the proposed cellular and biological models, the network is partially excited with fixed and noisy inputs. The input function for the simulated network in both models is given by:
\begin{equation}\label{eq:2_12}
IN_{ext_i}=
\begin{cases}
\alpha.\eta+\gamma~~~\text{if}~~l_1\le i\le l_2\\
\gamma~~~~~~~~~~~~~\text{if}~~l_2<i\le l_3
\end{cases}
\end{equation}

\begin{table}[t]
\captionsetup{font=footnotesize}
\caption{Input Values for the Network Model Constructed by One Thousands of Calcium Units.}   
\centering          
\begin{tabular}{c c c c c}    
\hline\hline                        
Model& Parameters &M=N=P=1&M=N=P=2&M=N=2, P=4\\ [0.5ex]  
\hline                      
\multirow{2}{4em}{MATLAB} & $\alpha$ & 0.25 & 0.5 & 0.5 \\
& $\gamma$ & 4& 8 & 3 \\
\hline
\multirow{2}{4em}{32--pixels} & $\alpha$ & 0.25 & 0.5 & 0.6 \\
& $\gamma$ & 4& 8 & 3 \\
\hline
\multirow{2}{4em}{16--pixels} & $\alpha$ & 0.25 & 0.45 & 0.8 \\
& $\gamma$ & 4& 6.5 & 3 \\[1ex]
\hline          
\end{tabular}
\label{tab:Table2}    
\end{table}

where $l_1=0$, $l_2=800$, $l_3=1000$, $\eta$ is a random number between 0 to 1 generated by a uniform distribution and other parameters are presented in Table \ref{tab:Table2} . The raster plots of the simulations for the biological and cellular models with two different dimensions (32 and 16--pixels) are demonstrated in Figure \ref{fig:Figure6} (a1--c3).
\par In this figure, temporal evolution of the firing rate shows a satisfying agreement between the cellular and biological models. It can be seen that the calcium units excited by a noisy input are destabilised after a certain amount of time while other units remained stable and fire rhythmically. Since in large scale simulations the statistical nature of such activities is generally of interest, the trivial disagreement shown in the figure may not be significant. As we expected, the higher the dimension of the cellular model the higher accuracy is achieved. Thus, the 32--pixels cellular model is chosen to be implemented in the hardware synthesis section.
\begin{figure*}[!h]
    \centering
\includegraphics[scale=0.35]{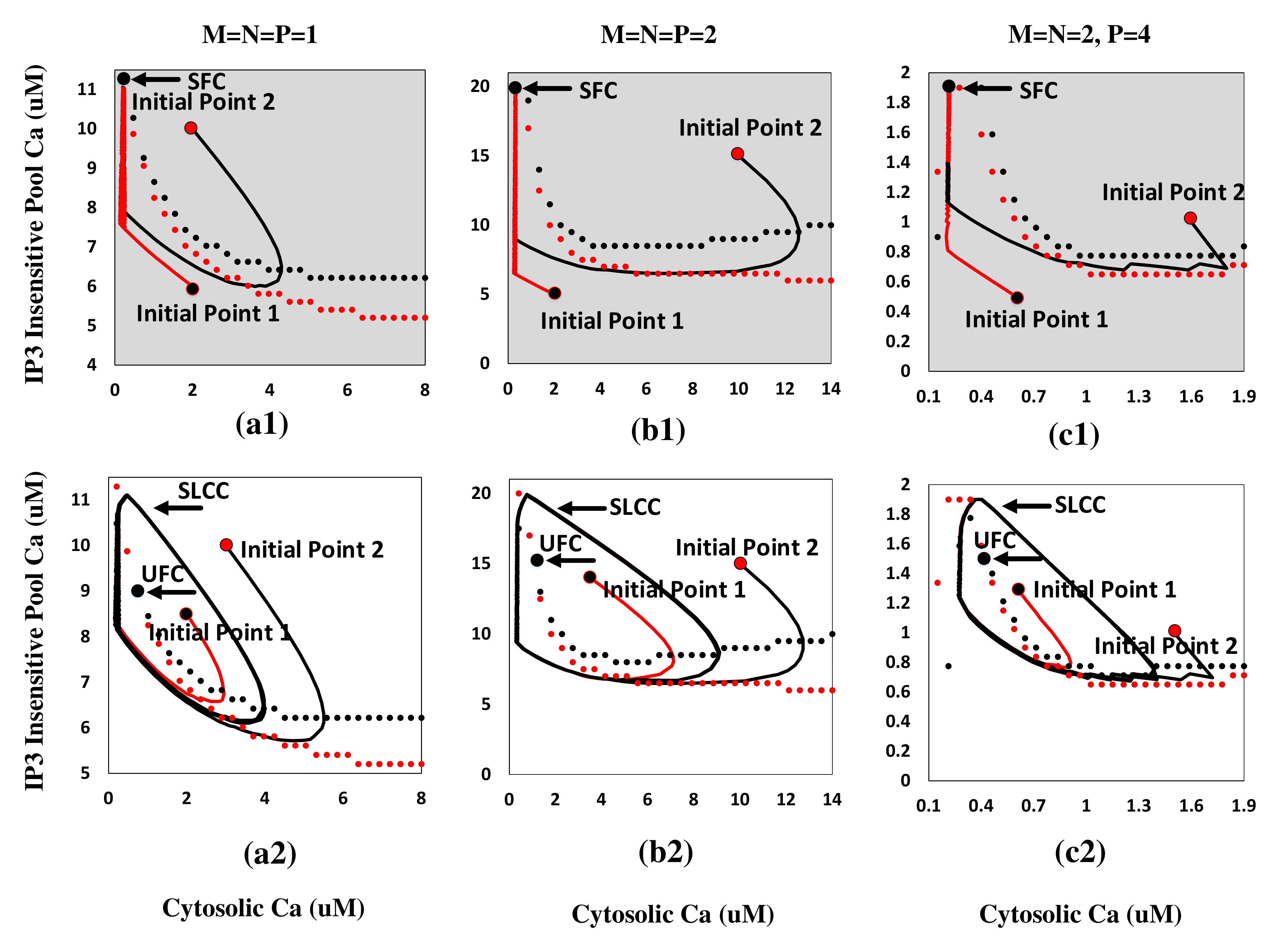}
\vspace{-15pt}
\captionsetup{font=footnotesize}
    \caption{Qualitative analysis of supercritical Hopf-Andronov bifurcation in two captured steps for each set of Hill functions in the calcium model mapped on the 32-pixels cellular phase plane.}
\vspace{-10pt}
    \label{fig:Figure5}
\end{figure*}

\section{Dynamical Behavior}
In this section, we qualitatively analyze the only supercritical Hopf-Andronov bifurcation of the proposed model in a discrete cellular space. From the dynamical systems viewpoint, the transition in the behavior of systems has a corresponding change in the phase portrait. This phenomenon determining the main features of the systems is called bifurcation \cite{Izhi2006}.

\par Since the time continuous bifurcation analysis rules cannot be directly applied to the proposed cellular model, here, we re-express a number of dynamical concepts according to the cellular phase plane \cite{Bavandpour2014}:
\par \textbf{Cellular Nullcline:} $X$ nullcline ($Xnull$), and $Y$ nullcline ($Ynull$) are two parts of the cellular phase plane $PP \equiv\{ (X,Y)|X\in \textbf{N},Y\in \textbf{M}\}$  and described as:
\begin{equation}\label{eq:2_13}
\begin{cases} Xnull\equiv \{(X,Y)|X\in \textbf{N} , Y\in \textbf{M},\\
 ~~~~~~~~~~~~~~~|F(X,Y)|\leq \textbf{$\delta$}\}\\
Ynull\equiv \{(X,Y)|X\in \textbf{N} , Y\in \textbf{M},\\
 ~~~~~~~~~~~~~~~|G(X,Y)|\leq \textbf{$\delta$}\}.
\end{cases}
\end{equation}
\par When the proper $\delta$ value is selected, the subsets $Xnull$ and $Ynull$ are the equivalents of $x$ and $y$ nullclines in the continuous phase plane.
\par \textbf{Equilibrium Cells (EC):} \textbf{EC} is a part of the \textbf{PP} and described as:
\begin{equation}\label{eq:2_14}
\textbf{EC}\equiv \{ (X,Y)|(X,Y)\in (Xnull\cap Ynull)\}
\end{equation}
where $\cap$ is an AND operator. \textbf{EC} is a subset of cellular phase plane corresponding to the equilibrium point in the continuous phase plane. To study of Hopf bifurcation, additional relevant dynamical concepts need to be defined.\\
\textbf{Stable Focus Cells (SFC)} defines as a part of the \textbf{EC} with the condition of:
\begin{equation}\label{eq:2_15}
\textbf{SFC}\equiv \{\textbf{EC}|\exists(X,Y)\in \textbf{EC}: (|x_{\rm min}+X\cdot\Delta x-x_{\rm sf}|<\frac{\Delta x}{2})\wedge (|y_{\rm min}+Y\cdot\Delta y-y_{\rm sf}|<\frac{\Delta y}{2})\}
\end{equation}
where $(x_{\rm sf},y_{\rm sf})$ refers to a stable equilibrium point of the cellular system based on the definition in the continuous time dynamics \cite{Izhi2006}.\\

\textbf{Unstable Focus Cells (UFC)} defines as a part of the \textbf{EC} with the condition of:
\begin{equation}\label{eq:2_16}
\textbf{UFC}\equiv \{\textbf{EC}|\exists(X,Y)\in \textbf{EC}: (|x_{\rm min}+X\cdot\Delta x-x_{\rm uf}|<\frac{\Delta x}{2})\wedge (|y_{\rm min}+Y\cdot\Delta y-y_{\rm uf}|<\frac{\Delta y}{2})\}
\end{equation}
where $(x_{\rm uf},y_{\rm uf})$ refers to an unstable equilibrium point of the cellular system based on the definition in the continuous time dynamics \cite{Izhi2006}.\\
\textbf{Cellular Attraction Domain (CAD)} defines as a part of \textbf{PP} with the condition of:
\begin{equation}\label{eq:2_17}
\textbf{CAD}\equiv \{(X,Y)|(X,Y)\in \textbf{PP}, \exists n<M\cdot N:~ (X^{(n)},Y^{(n)})\in \textbf{SFC}\}
\end{equation}
where $(X^{(n)},Y^{(n)})$ is the location of $(X,Y)$ after $n$ motions in the \textbf{PP}.\\
\textbf{Stable Limit Cycle Cells (SLCC)} is a subset of the \textbf{PP} with the following condition:
\begin{equation}\label{eq:2_18}
\textbf{SLCC}\equiv \{(X,Y)|\exists n<N\cdot M, n \neq 1 : (X^{(n)}, Y^{(n)})=(X,Y)\}.
\end{equation}
\par Figure \ref{fig:Figure5} (a1--c2) shows the supercritical Andronov-Hopf bifurcation in two captured steps of the process for each set of Hill functions. In the capture (a1), the first order calcium dynamical system (m=n=p=1) creates a \textbf{SFC} in the \textbf{PP} leading to a \textbf{CAD} subset all over the \textbf{PP}. As illustrated in the figure, all initial point in the \textbf{PP} subset is attracted to the \textbf{SFC} all over the cellular space. In the second capture (a2), the \textbf{CAD} subset is totally vanished, and the \textbf{SFC} turns into a \textbf{UFC} and a \textbf{SLCC} subset, where any random initial point in the cellular space attracts to the \textbf{SLCC}. These results for other set of Hill functions are also shown in Figure \ref{fig:Figure5} (b1--b2) and (c1--c2).

\section{Truncation Error}
As explained before, the approach converts the continuous phase plane into a cellular space with discrete cells locating any point with an address prepared by the outputs. This conversion allows our synchronous approach to track any trajectory in the phase plane with a sufficient number of pixels. However, depending on the cellular space dimension, a truncation error is observed in the system in each clock cycle. In this section, we formulate this error and discuss about its sources and how to find an optimum hardware architecture. First, let us assume that $dt$ is small enough and its corresponding error is negligible, so the values calculated from $F(x,y)$ and $G(x,y)$ in the (\ref{eq:2_8}) are called continuous. Now, if we fetch velocities from the memories, the next state variables are:
\begin{equation}\label{eq:2_19}
x^+=x^-+[x_{velocity}\pm \gamma_1]+IN_{ext}
\end{equation}
\begin{equation}
y^+=y^-+[y_{velocity}\pm \gamma_2]
\end{equation}
where $\gamma_1$ and $\gamma_2$ are the differences between the continuous state variables and the corresponding cellular values. Hence, we can rewrite the equations as the following:
\begin{equation}\label{eq:2_20}
x^+_{cellular}=x^+_{continous}\pm \gamma_1
\end{equation}
\begin{equation}\label{eq:2_21}
y^+_{cellular}=y^+_{continous}\pm \gamma_2
\end{equation}
where $x^+_{continous}$ and $y^+_{continous}$ are the continuous part of the iterative solution. Thus, the addresses of the next cellular values fetched from memory are given by:
\begin{equation}\label{eq:2_22}
X^-_{new}=\lfloor \frac{x^+_{continous}\pm \gamma_1-x_{min}}{\Delta x} \rfloor
\end{equation}
\begin{equation}\label{eq:2_23}
Y^-_{new}=\lfloor \frac{y^+_{continous}\pm \gamma_2-y_{min}}{\Delta y} \rfloor.
\end{equation}
\par This implies that the values of $\gamma_1$ and $\gamma_2$ are involved in the new address of cellular values and also the floor function leads to truncation error in the system since the $X^-_{new}$ and $Y^-_{new}$ must be integers. In other words, in the proposed cellular approach the velocity is defined for the cross point of the cells and considered as a constant all over intra-cell space. This truncation error appears in the form of momentary and permanent lag, lead and deviation in the time domain signals in which the velocity changes are more uneven and random. However, the virtual cellular trajectories can track fairly the continuous ones in the phase plane under certain conditions on the the number of pixels. Thus, the truncation error can be notably reduced by increasing the number of pixels as it is seen in the 32-pixels cellular model.

\par In this section, first for the sake of comparison, we present a regular digital implementation of the calcium model and then introduce the hardware implementation according to the mapping of the proposed cellular model for a single and a network of calcium units as explained in the previous sections.

\begin{figure*}[th]
\vspace{-10pt}
\centering
\includegraphics[scale=0.4]{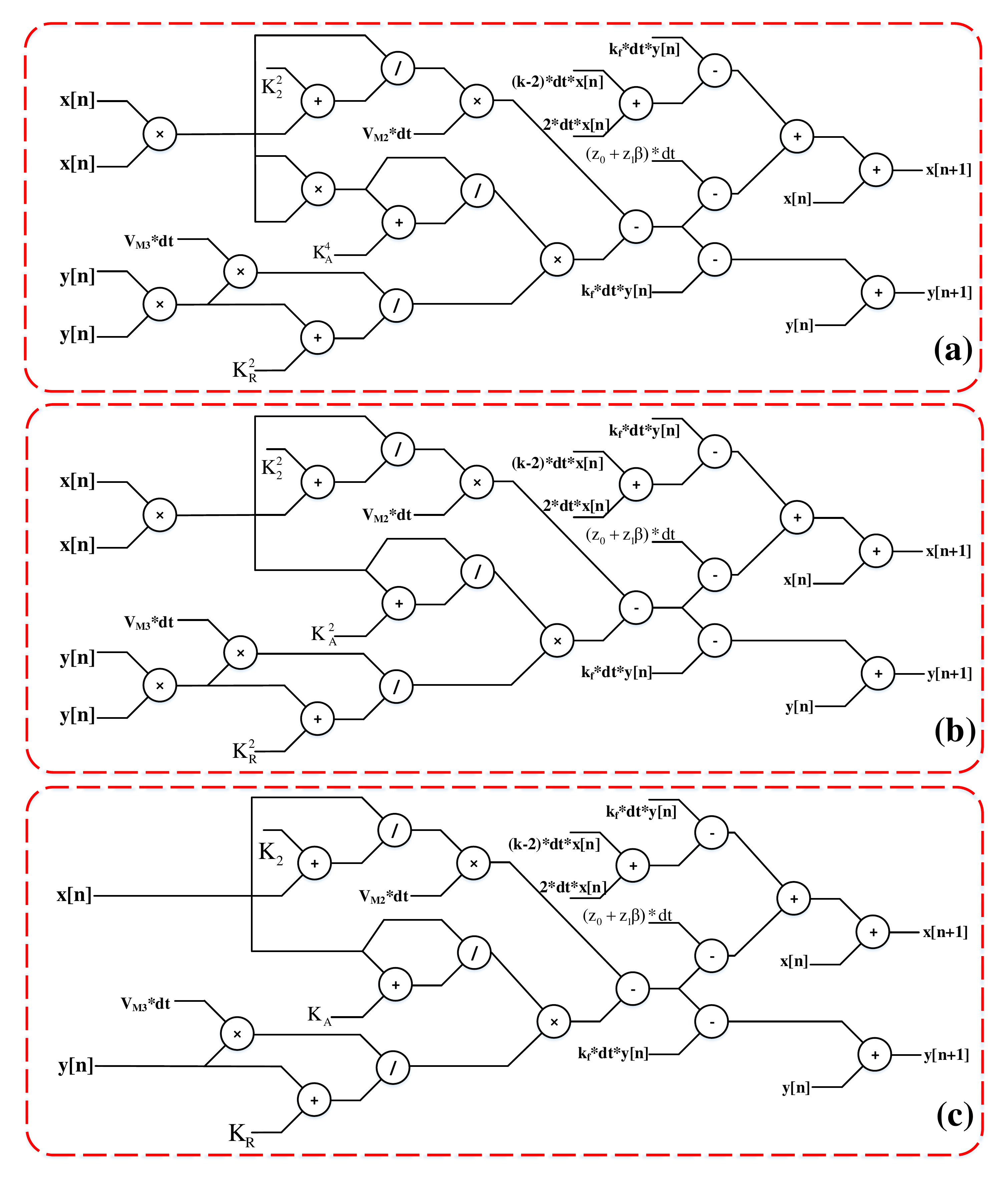}
\vspace{-10pt}
\captionsetup{font=footnotesize}
\caption{Arithmetic pipelines for regular implementation of the calcium model with various Hill functions. (a) Arithmetic pipeline for the case with Hill coefficients $m=n=2, p=4$. (b) Arithmetic pipeline for the case with Hill coefficients $m=n=p=2$. (c) Arithmetic pipeline for the case with Hill coefficients $m=n=p=1$.}
\vspace{-10pt}
    \label{fig:Figure7}
\end{figure*}

\begin{figure*}[th]
\vspace{-10pt}
\centering
\includegraphics[scale=0.5]{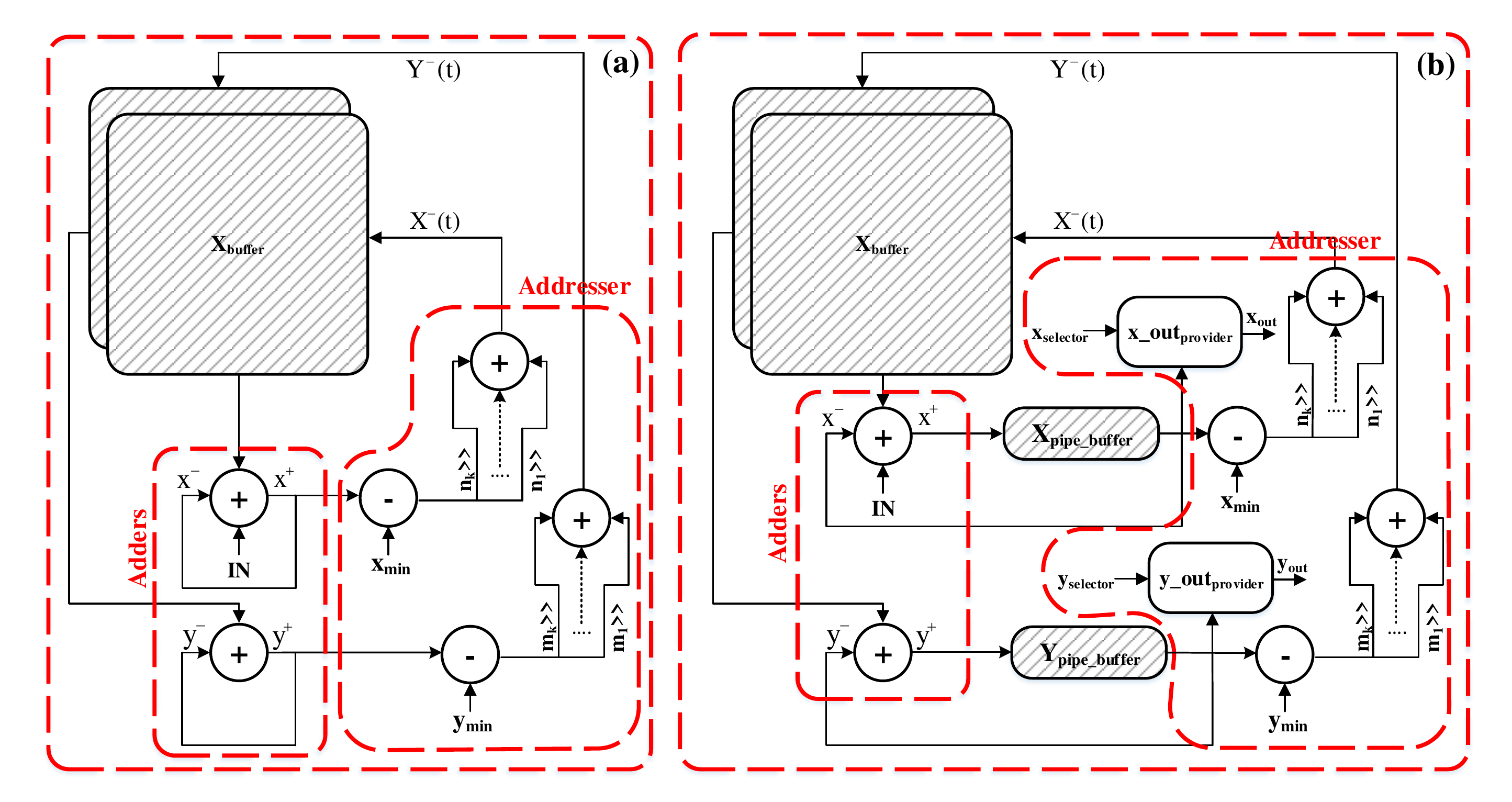}
\vspace{-10pt}
\captionsetup{font=footnotesize}
\caption{The detailed internal structure of the proposed cellular models. (a) The synchronous cellular hardware for a single calcium unit. (b) The synchronous cellular hardware for a network of calcium units. In this structure two buffers are added before output signal to share the whole hardware between all calcium units and an output provider prepares the output signal for each single calcium unit.}
\vspace{-10pt}
    \label{fig:Figure8}
\end{figure*}

\section{Regular Digital Implementation}
For regular digital implementation of the calcium model, the continuous time equations codified in (\ref{eq:2_1})--(\ref{eq:2_4}) are discretized using Euler method as the following:
\begin{multline}\label{eq:2_24}
x[k+1] =x[k]+dt[z_0+z_1\beta-z_2(x[k])+z_3(x[k],y[k])+k_fy[k]-kx[k]]
\end{multline}
\begin{equation}\label{eq:2_25}
y[k+1] =y[k]+dt(z_2(x[k])-z_3(x[k],y[k])-k_fy[k])
\end{equation}
where
\begin{equation}\label{eq:2_26}
z_2(x[k]) =V_{M_2}\frac{x[k]^n}{K_2^n+x[k]^n}
\end{equation}
\begin{equation}\label{eq:2_27}
z_3(x[k],y[k]) =V_{M_3}\frac{y[k]^m}{K_R^m+y[k]^m}\frac{x[k]^p}{K_A^p+x[k]^p}.
\end{equation}

\par These equations describe the same system behaviour as (\ref{eq:2_1}--\ref{eq:2_4}) if $dt$ is small enough and are implemented in 24 bit fixed--point (10 bits integer and 14 bits fraction) representation, so the constants $z_0+z_1\beta, k, k_f, V_{M_2}, V_{M_3}, K_R$ and $K_A$ must also be modified compared the biological model. The computational tree from the input to the output of the $x$ and $y$ pipelines for this regular digital implementation combined with the digitalized constants and $dt$ (equal to 1/128=0.0078125) for various Hill functions are shown in Figure \ref{fig:Figure7} (a--c). The arithmetic operations in (\ref{eq:2_24})--(\ref{eq:2_27}) are allocated to the arithmetic functional units so that the maximum efficiency and throughput are achieved. However, it is expected to obtain a large area consumption and low speed for the implemented digital hardware since the main building blocks of each regular model are: multiplier, divider, adder, and the critical path determining the operation frequency in each implementation model is limited by the time delay for division operation.

\section{Cellular Digital Implementation}
\subsection{Single Digital Cellular Model}
Unlike the regular implementation of the calcium models utilizing time consuming units to implement the (\ref{eq:2_24})--(\ref{eq:2_27}), the proposed cellular model has simple blocks and is multiplierless. This circuit contains three major parts and is demonstrated in \ref{fig:Figure8} (a):

\subsubsection{Storage Blocks}
\par As mentioned in section III, we calculate the cellularized velocity array for each cell and store this information in the storage blocks. The velocity vectors are calculated off--line, using (\ref{eq:2_7}) according to predetermined model parameters. These signed values are stored in two storage blocks with the size of $X\times Y$ (number of pixels) for each dimension. The size of velocity components is exactly the same with the bandwidth of the system and defines the length of each memory cell which is 24 bits. Corresponding velocity values are fetched from the storage blocks in accordance with the address $(X, Y)$ that shows the current cell.

\subsubsection{Adders}
\par This block contains an adder for each dimension which adds the velocity value fetched from the storage blocks with the previous state of the dynamical variable. According to (\ref{eq:2_5}) this value is also added by the input for $X$ cellular variable. The transferred values from the storage blocks and the input determine the motion direction on the cellular phase plane. For example when the value of velocity received from \textbf{$X_{buffer}$} plus the input is positive, we have an upward motion in the cellular space in the $X$ direction while if the value of velocity received from \textbf{$Y_{buffer}$} is negative, we have a downward motion in the $Y$ direction.
\par The absolute value of the velocities determine the amount of increase or decrease in the state variables in each step. Thus, the higher the velocity value, the bigger the increase of the state variable. Obviously, the reason why the cellular model is called "synchronous" is that in each clock cycle the dynamical variables evolve equally in time, unlike the previous cellular model \cite{Bavandpour2014} that was "asynchronous". This property of the model would allow us to implement and easily calibrate the velocity values even on analog memories such as floating gates or memristors.

\subsubsection{Addresser}
\par The location of each state variable in the cellular phase plane is calculated in this block. In other words, this block is responsible for converting the non--cellular output variables to the cellular addresses to fetch the next velocity values from the storage blocks. This conversion is based on the (\ref{eq:2_7}) where the cellular phase plane was introduced as a discrete mesh--like plane and built from the continuous space. As mentioned before in \cite{Bavandpour2014}, the velocity values change the frequency of the output Voltage Control Oscillator (VCO) leading to changes in the output register through a one--hot bit coding. This coding scheme limits the state variables to be changed in a few locations, and the more accuracy is needed the more velocity values should be stored in the memory leading to a higher area consumption in hardware. Besides, in \cite{Bavandpour2014} the output variables are represented by one--hot bit coding that is not appropriate for using as an input for other connected modules and should be again converted to non--cellular values. On the other hand, in the proposed approach the next address of the cellular phase plane is indirectly extracted from the output variables which is a non--cellular value. Besides, the address of the next state in cellular space can be easily implemented by one subtract unit and maximum two shifts if the values of $x_{min}$, $x_{max}$, $y_{min}$, $y_{max}$ and the number of pixels are properly chosen (see Table \ref{tab:Table1}).

\begin{figure}[th]
\vspace{-10pt}
\centering
\includegraphics[scale=0.4]{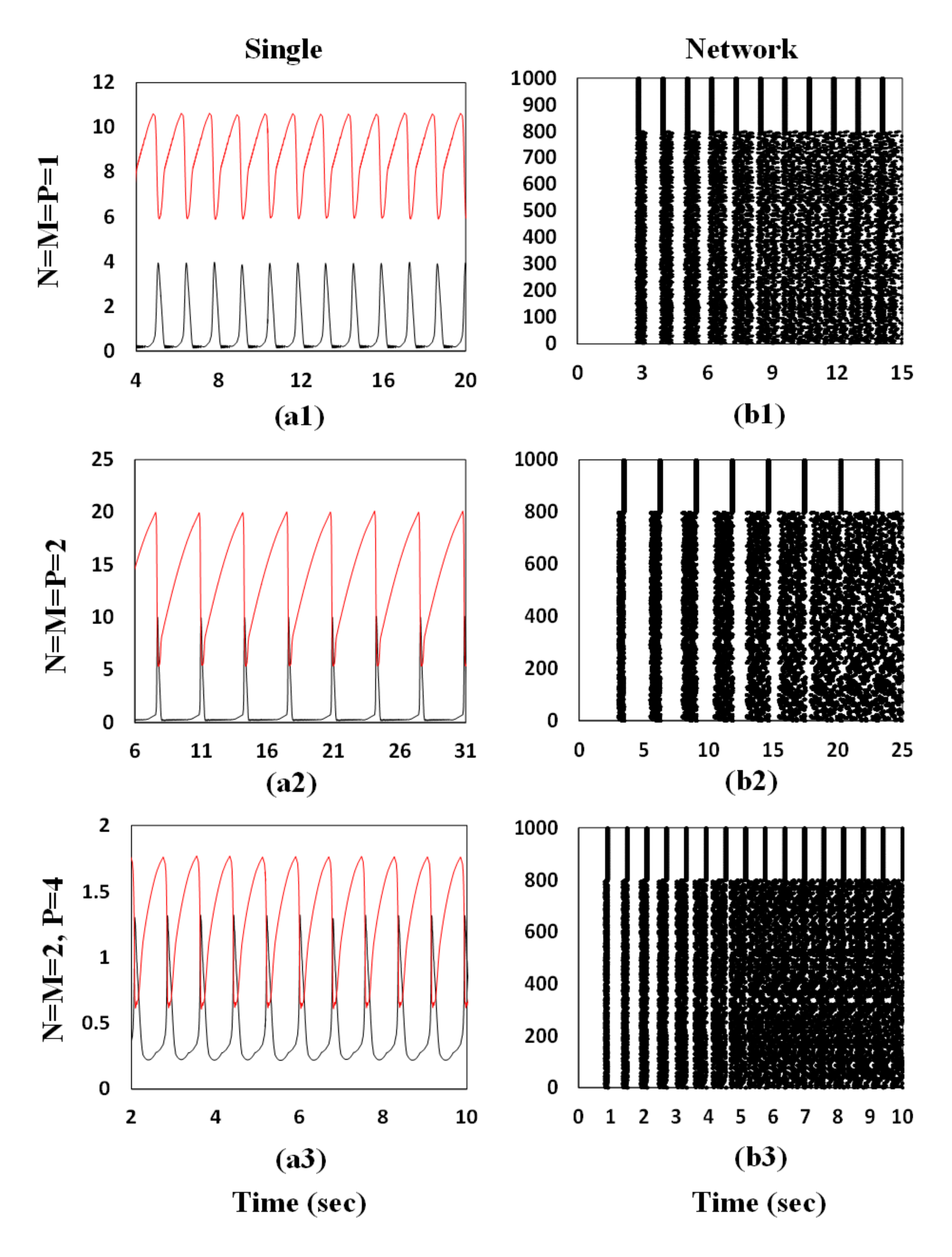}
\vspace{-10pt}
\captionsetup{font=footnotesize}
\caption{Output waveforms of a single and a network of calcium units implemented on the Kintex--7 (XC7K325T) FPGA for the proposed cellular model with 32--pixels introduced as the accurate cellular model in the previous chapter.}
\vspace{-10pt}
    \label{fig:Figure9}
\end{figure}

\subsection{Networked Digital Cellular Model}
Since the critical path in the proposed cellular model is determined only by one subtractor, we can easily share the digital hardware for large scale simulation in CytoMimetic circuit design. To this end, a network of pipelined calcium units is presented containing one thousand of CICR models that can be operated in real time. This number can be even bigger and equals (clock frequency)*(dt) which in this case dt=$\frac{1}{128}$, hence 1562500 calcium units can be simulated in real time. As illustrated in Figure \ref{fig:Figure8} (b), $X_{pipe-buffer}$ and $Y_{pipe-buffer}$ are added to the single cellular model to store the data of each calcium unit. These buffers are shifted in each clock pulse and the proper output is prepared by output provider. For example, in this case the output for each unit must be sampled after one thousand raising clock edges.

\begin{table}[t]
\captionsetup{font=footnotesize}
\caption{Device Utilization of the Kintex--7 (XC7K325T) FPGA for the Regular Digital Implementations and Synchronous Cellular Model with 32--pixels Resolution.}   
\centering          
\begin{tabular}{c c c c}    
\hline\hline                        
  \textbf{Parameter} & \textbf{LUT}& \textbf{FF} & \textbf{Freq. (MHz)}\\ [0.5ex]  
\hline                      
\textbf{Regular (N=M=P=1)}& 11404& 561& 49.44\\
\textbf{Regular (N=M=P=2)}& 12496& 585& 49.44\\
\textbf{Regular (N=M=2, P=4)}&13243& 703& 49.44\\
\textbf{Proposed Cellular Model}& 1534&48& 215.02\\
\textbf{Available}& 203800 & 407600& 200\\ [1ex]        
\hline          
\end{tabular}
\vspace{-10pt}
\label{tab:Table3}    
\end{table}

\section{Implementation Results}
\par To verify the validity of the proposed digital design for the cellular model, the circuit is implemented on a Genesys 2 development system, which provides a hardware platform containing a high performance Kintex--7 (XC7K325T) FPGA surrounded by a comprehensive collection of peripheral components. To compare the results, the regular digital models with various Hill functions are synthesized and implemented along with the proposed cellular hardware on the board. Figure \ref{fig:Figure9} shows digital measured outputs of the dynamical behavior for a single calcium unit and one thousand fully pipelined network implemented on the development platform using our cellular model. The device utilization for implementation of the proposed cellular model and the regular digital models is summarized in Table \ref{tab:Table3}.

\begin{table*}[t]
\captionsetup{font=footnotesize}
\caption{Speedup of the Cellular Hardware Model with Respect to a Single CPU for Four Networks Composed of 10k, 20k, 30k and 40k Units with Various Hill Functions.}   
\centering          
\resizebox{\columnwidth}{!}{%
\begin{tabular}{c c c c c c}    
\hline\hline                        
\textbf{Networks}& & & \textbf{Time (sec)}&& \\
                 & Proposed Cellular Model& Regular (N=M=P=1) & Regular (N=M=P=2)&Regular (N=M=2, P=4)& Average Speed up\\ [0.5ex]  
\hline                      
\textbf{10k}& 0.0064& 0.2380& 0.2646&1.1278& $84.95$\\
\textbf{20k}& 0.0128& 0.4533& 0.4541&2.3138& $83.88$\\
\textbf{30k}&0.0192& 0.6380& 0.6555&3.4147& $81.73$\\
\textbf{40k}& 0.0256&0.8018& 0.8714&4.5903& $81.55$\\
\hline          
\end{tabular}
\vspace{-10pt}
\label{tab:Table4}    
}
\end{table*}

\par The results of hardware implementation show that thanks to efficient design of the 32--pixels cellular model with the size of 24 bits velocity components, a smaller area and a higher clock frequency is achieved compared to the regular digital implementation. Since there is no high--cost operations with slow critical paths in the cellular structure, the reduction in area and increase in maximum operation frequency was expected. In particular, the proposed cellular model has reached 215 MHz clock frequency (almost 4.3 times faster compared to the regular digital implementation of various Hill functions) with an over 8 times less area for the cases $n=m=p=2$ and $n=m=2, p=4$ and 7 times for the case $n=m=p=1$ compared to the regular digital implementations. It should be noted that, the proposed cellular hardware is fully reconfigurable and only implemented once and reprogrammed for all cases with various Hill functions.

\par Moreover the performance of the pipelined cellular network is evaluated by scaling up the number of shared calcium units and comparing them with an equivalent simulation run on a standard PC workstation with Intel Core i7--4790 CPU, operating at 3.60 GHz and with 16 GB of RAM. The CPU version of the simulation is based on the CICR model represented by (\ref{eq:2_1})--(\ref{eq:2_4}) and written in Matlab. Similar to the structure introduced in the previous chapter, the networks in both the proposed cellular and the biological models are partially excited with fixed and noisy inputs. A snapshot of the comparison between hardware and software models for four different networks composed of 10k to 40k calcium units is shown in Table \ref{tab:Table4}. The results obtained for 1 sec simulation of the networks (i.e. simulation of 128 samples of each model with dt=1/128) reveal 84.95, 83.88, 81.73, 81.55 speed up for the networks containing 10k, 20k, 30k and 40k calcium units respectively. It should be stressed that due to the reconfigurable structure of the proposed hardware model, one run--time is reported for each cellular calcium unit codified by a set of Hill functions.

\section{Exceptional Cases}
In this section a simplified version of the previously proposed hardware cellular model is presented \cite{7} capable of mimicking various biological neurons' dynamics with good precision and low hardware cost. Unlike the "asynchronous" cellular neuron models introduced in \cite{Bavandpour2015}--\cite{Matsubara2013}, the synchronous property of the proposed model would make the approach appropriate for large--scale pipelined implementations since the effective operating frequency is not limited by the asynchronous time delay between each step. Another advantage of the proposed model compared to \cite{Bavandpour2015}--\cite{Matsubara2013} is the output coding of the system, in which the $membrane$ and $recovery$ registers change the output states through a relatively complex circuit with a one--hot bit coding. This coding scheme limits the state variables to be changed in a few locations and the higher the needed accuracy, the more complexity is applied to the hardware. In contrast, in the proposed approach, the output registers do not directly carry the address of the next velocity value that should be fetched from memory in the next clock. This implies that the output register can be set in any precision and not limited by the number of memory pixels. In other words, the proposed model is internally cellular and externally non–cellular, while the model introduced in \cite{Bavandpour2015}--\cite{Matsubara2013} is fully cellular which limits the output resolution leading to lower precision.

\section{Synchronous Cellular Neuron Model}
\par In this model, according to \cite{Bavandpour2015}, first we convert the phase plane of the biological neuron models into a 2--D cellular space where $(x,y)\in Z^2$ represents the location of state point in the phase plane, and $(\frac{dx}{dt}, \frac{dy}{dt} )$ determines the velocity and direction of the motion. The x--nullcline is the set of points where $\frac{dx}{dt}=0$. Similarly, y--nullcline is the set of points where $\frac{dy}{dt}=0$. Clearly the points of intersection between x--nullcline and y--nullcline are the equilibrium points. Most 2--D neuron models can be rewritten in a general form as follows:
\begin{equation}\label{eq:2_28}
\begin{cases}
\frac{dx}{dt} = F(x)+\alpha y+IN\\
\frac{dy}{dt}=G(x)+\beta y
\end{cases}.
\end{equation}
\par In the cellular space, $X[n]=i$ where $ i\in \textbf{N}\equiv {0,1,...,N-1}$ and $Y[n]=j$ where $ j\in \textbf{M}\equiv {0,1,...,M-1}$ are discrete variables corresponding to $x$ and $y$ in the continuous space. The location of each state point in the $\textbf{N}\times \textbf{M}$ cellular space can be defined as:
\begin{equation}\label{eq:2_29}
\begin{cases}X[n]= \lfloor \frac{x[n]-x_{min}}{\Delta x} \rfloor ~~~;~~~~x[n] \in [x_{min},x_{max})\\
Y[n]= \lfloor \frac{y[n]-y_{min}}{\Delta y} \rfloor ~~~;~~~~y[n] \in [y_{min},y_{max})
\end{cases}
\end{equation}
where $\Delta x= \frac{x_{max}-x_{min}}{N}$ and $\Delta y= \frac{x_{max}-x_{min}}{M}$. Since most 2--D neuron models can be expressed in the form of (\ref{eq:2_28}) and $F(x)$, $G(x)$ are functions of $x$, the $Y[n]$ introduced in (\ref{eq:2_29}) is not involved in the stored velocity functions and we have:
\begin{equation}\label{eq:2_30}
\begin{cases}X_{null}(X[n-1])=F(x_{min}+X[n-1]\cdot\Delta x)\\
Y_{null}(X[n-1])=G(x_{min}+X[n-1]\cdot\Delta x)
\end{cases}
\end{equation}
where $X_{null}$ and $Y_{null}$ are portions of time continuous nullclines and the next state of each dynamical variable is obtained by:
\begin{equation}\label{eq:2_31}
\begin{cases}x[n]-x[n-1]=\Delta t\cdot [X_{null}(X[n-1])+\alpha y[n-1]]\\
+\Delta t\cdot IN\\
y[n]-y[n-1]=\Delta t\cdot [Y_{null}(X[n-1])+\beta y[n-1]]
\end{cases}
\end{equation}
where $[n]$ and $[n-1]$ represent the current and previous states respectively. As can be seen, the proposed cellular model is independent of the neuron model's complexity since both $X_{null}$ and $Y_{null}$ are stored off--line. Moreover, the value of each state variable is synchronously changed, implying that the timing of both state variables is equal. Clearly, the address of the next cellular state is indirectly related to the output variables. This feature allows the output value to work in any precision leading to increased accuracy compared to the previous cellular models \cite{Bavandpour2015}--\cite{Matsubara2013}.

\begin{figure}[t]
\vspace{-10pt}
\centering
\includegraphics[trim = 0.35in 0.25in 0.15in 0.25in, clip, width=5in]{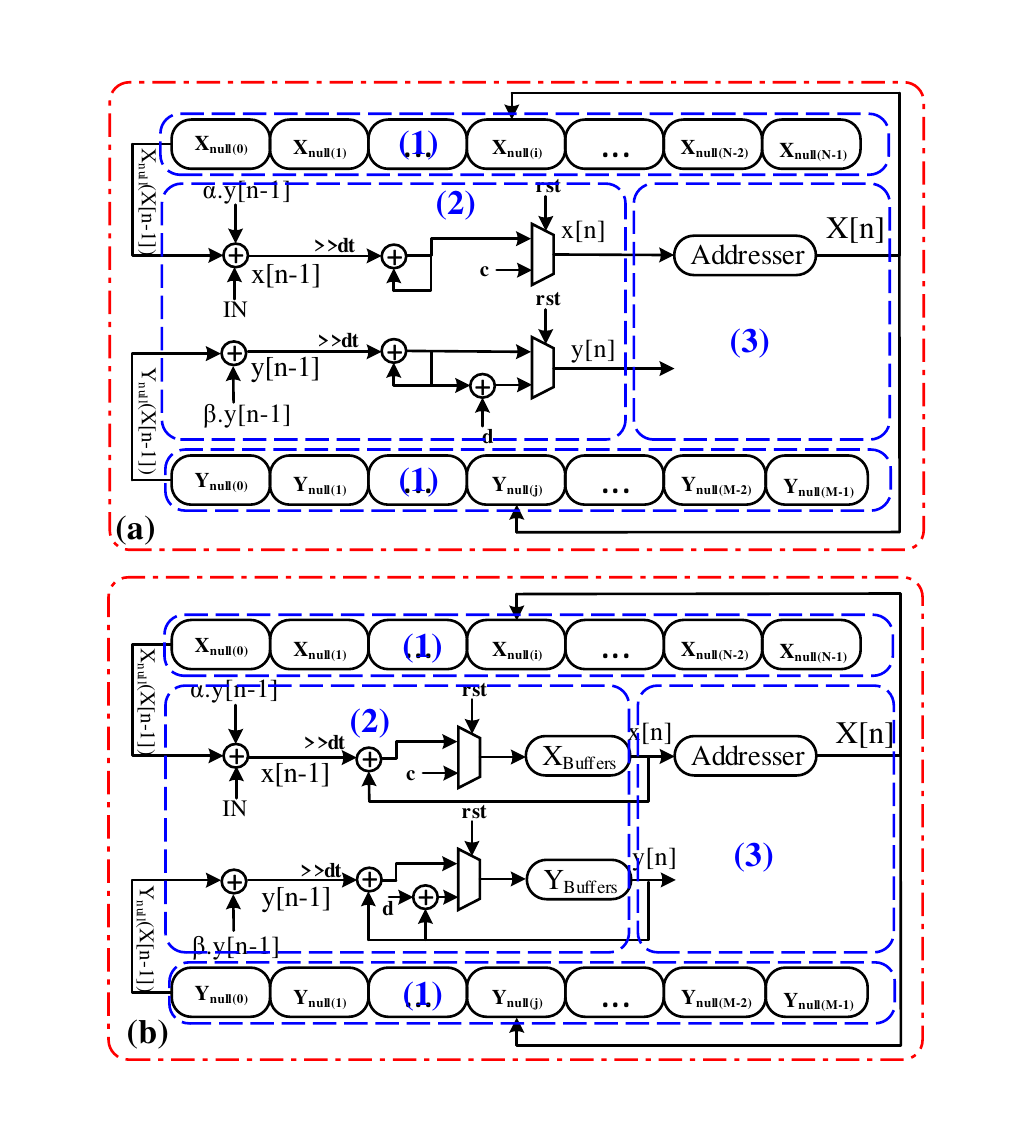}
\vspace{-5pt}
\captionsetup{font=footnotesize}
\caption{The detailed internal structure of (a) the synchronous cellular hardware for a single 2--D neuron unit and (b) the synchronous cellular hardware for a network of neuron units.}
\vspace{-10pt}
\label{fig:Figure10}
\end{figure}

\par The direction of new motions on the cellular space can be formulated as:
\begin{equation}\label{eq:2_32}
\begin{cases}
X[n]-X[n-1]=0~\Rightarrow~\text{"no change"}\\
X[n]-X[n-1]\ge 1~\Rightarrow~\text{"upward motion"}\\
X[n]-X[n-1]\le -1~\Rightarrow~\text{"downward motion"}
\end{cases}
\end{equation}

It should be stressed that the resultant motion on the 2--D cellular phase planes is determined only by the motion in $X$ direction and the number of motions on the cellular space unlike the one--hot bit coding in \cite{Bavandpour2015}--\cite{Matsubara2013} can be more than one step in each clock cycle.

\section{Hardware Architecture}
\par In this section, according to the cellular mapping introduced in the previous section, we present two digital architectures describing a single and a network of neuron units.

\subsection{Single Digital Cellular Model}
This circuit contains three major parts demonstrated in Figure \ref{fig:Figure10} (a):

\subsubsection{Storage Blocks}
\par As concluded from the previous section, we calculate the $X_{null}$ and $Y_{null}$ arrays and store them in a set of registers. The stored vectors are programmed off--line, using (\ref{eq:2_30}) in accordance to predetermined model parameters. For a 2--D neuron model, these signed values are stored in two sets of registers with the size of $N$ and $M$ for both $X$ and $Y$ respectively. The size of each nullcline's component is 18 bits and the corresponding values are fetched for both state variables according to the address of $X$.

\subsubsection{Next State Provider}
\par This block contains five adders, two multiplexers and two shifts to provide the next state of the dynamical variables. According to (\ref{eq:2_31}), the values fetched from the storage blocks (\textbf{$X_{null}$} or \textbf{$Y_{null}$}) are added to a portion of $y^-$, the time--dependent input (if applicable) and then shifted as $dt$. The result is added to the previous state of the dynamical variable. The obtained value is filtered according to the reset condition. The absolute value of the output for each dimension represents the amount of state--variable increase or decrease in each step. Thus, higher values lead to increasing state variable. Clearly, the reason why we term the cellular model "synchronous" is that in each clock cycle the dynamical variables evolve equally in time.

\subsubsection{Addresser}
\par The location of each state variable in the cellular phase plane is calculated in this block. In other words, this block is responsible for converting the non--cellular output variables to cellular addresses in order to fetch the next value from the storage blocks. Since the 2--D neuron models can be represented by (\ref{eq:2_28}), we need only one addresser block for both dimensions. This conversion can be easily implemented using (\ref{eq:2_29}) and by one subtract and one shift operation when the values $x_{min}$, $x_{max}$ and $N$ are properly chosen (these values are presented for each neuron model in Table \ref{tab:Table5}).

\subsection{Network of Digital Cellular Neuron}
Since the critical path in the proposed cellular model is determined only by one subtractor embedded in the $Addresser~Block$ (see Figure \ref{fig:Figure10}(b)), the digital hardware can be easily shared for large-scale simulation of spiking neural networks. To this end, a network of pipelined neuron units is presented, capable of real--time operation. The number of real-time neuron units that can be accommodated in this structure depends on the operating frequency and time step (i.e. clock frequency$\times$dt). As illustrated in Figure \ref{fig:Figure10} (b), $X_{buffers}$ and $Y_{buffers}$ are added to the single cellular model to store the data of each neuron unit. These buffers are shifted in each clock pulse and the proper output is fed to the $Addresser~Block$ to provide the next cellular address and to the $Next~State~Provider$ as the previous state of the dynamical system.

\begin{figure*}[t]
\vspace{-10pt}
\normalsize
\setcounter{equation}{5}
\centering
\includegraphics[trim = 0in 0in 0in 0in, clip, height=5in,width=5.5in]{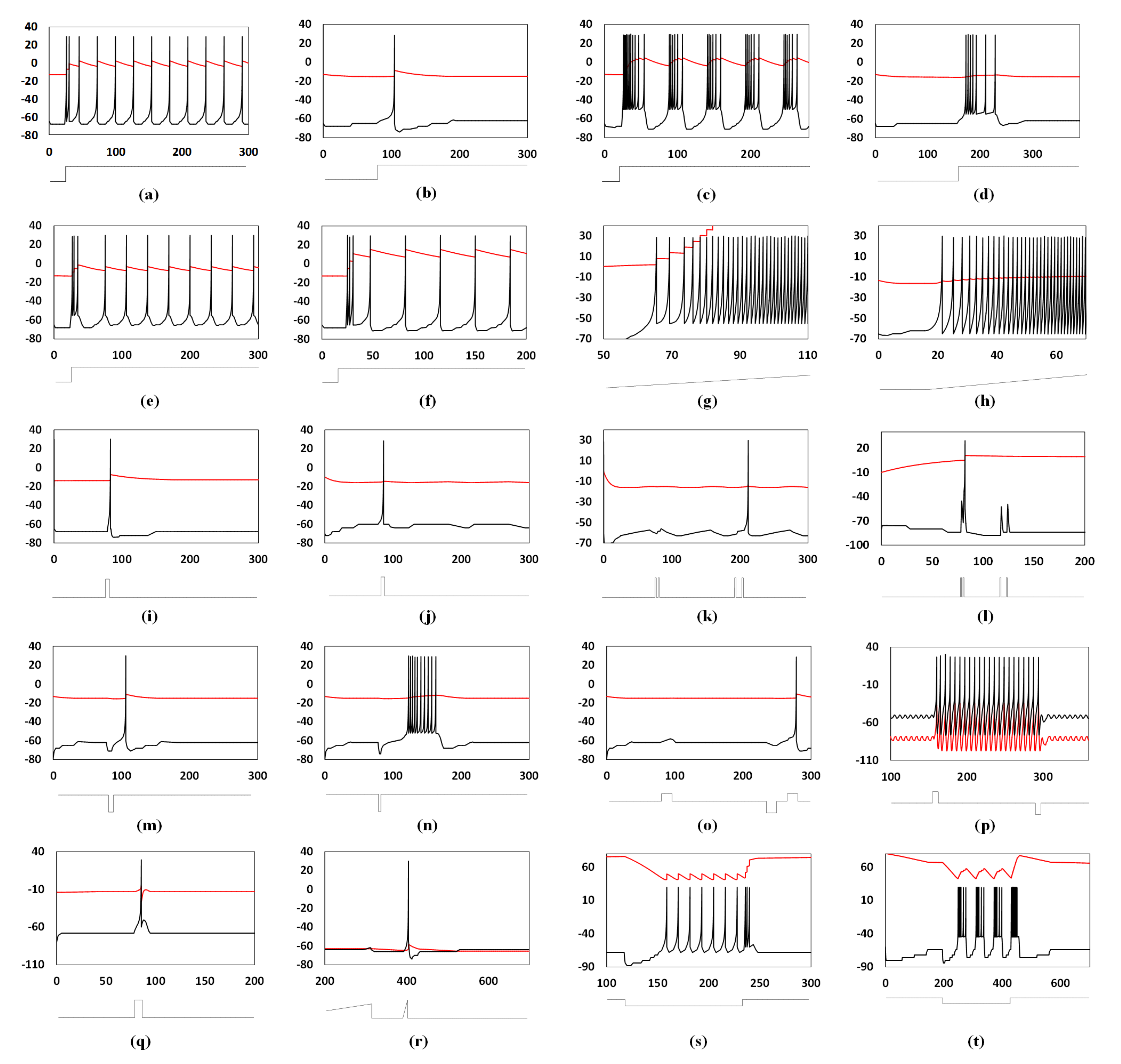}
\vspace{-10pt}
\captionsetup{font=footnotesize}
\caption{Time--domain representation of various dynamical responses of a cellular Izhikevich neuron circuit with 32 pixels. (a) Tonic spiking, (b) phasic
spiking, (c) tonic bursting, (d) phasic bursting, (e) mixed mode, (f) spike frequency adaption, (g) class I excitable, (h) class II excitable, (i) spike latency, (j) sub-threshold oscillation, (k) resonator, (l) integrator, (m) rebound spike, (n) rebound burst, (o) threshold variability, (p) bistability, (q) Depolarized After-Potential (DAP), (r) accommodation, (s) inhibitory induced spiking, and (t) inhibitory induced bursting.}
\vspace{-10pt}
\label{fig:Figure11}
\end{figure*}

\section{Hardware Time Domain Results}
\subsection{2-D Neuron Models}
In this section, we map the proposed cellular approach on two well--known 2--D neuron models and then synthesize them on a high--performance Kintex--7 (XC7K325T) FPGA according to the architecture introduced in the previous section.
\par The first test case is the Izhikevich neuron model \cite{Izhi2006} capable of mimicking a wide range of firing patterns and their underlying bifurcation
scenarios. The mathematical description of this model is denoted by:
\begin{equation}\label{eq:2_33}
  \left\{
  \begin{array}{l l}
  \dot{v}=0.04v^2+5v+140-u+I \\
  \dot{u}=a\cdot(b\cdot v-u) \\
  \end{array} \right.
\end{equation}

\begin{equation}\label{eq:2_34}
\text{if}~~v>30 \text{mV}~~\text{then} \left\{
  \begin{array}{l l}
  v\leftarrow c \\
  u \leftarrow u+d\\
  \end{array} \right.
  \end{equation}
where $v$ represents the membrane potential of the neuron, $u$ represents a membrane recovery variable and $a, b, c, d$ are dimensionless parameters. According to the cellular mapping introduced in section II, the model can be rewritten in the general form of (\ref{eq:2_28}) and then can be conveniently mapped on the proposed architecture as follows:
\begin{equation}\label{eq:2_34}
  \left\{
  \begin{array}{l l}
  x=v, y=u\\
  \alpha=1, \beta=a \\
  F(x)=0.04x^2+5x+140 \\
  G(x)=b.x \\
  \end{array} \right. .
\end{equation}

\par Figure \ref{fig:Figure11} illustrates the hardware results on a digital oscilloscope for various time domain waveforms produced by the synchronous cellular approach. As can be seen in the figure, there is remarkable compliance between the hardware results and the biological counterpart \cite{Izhi2006}.

\par The second neuron model test case is the FHN (FitzHugh--Nagumo) model \cite{FitzHugh1961} which is a two--dimensional simplification of the Hodgkin--Huxley model. Its equations are given by:
\begin{equation}\label{eq:2_35}
  \left\{
  \begin{array}{l l}
  \dot{v}=v-\frac{v^3}{3}-u+I \\
   \dot{u}=a.(v+0.7-0.8u) \\
  \end{array} \right.
\end{equation}
where $v$ is the membrane potential variable, $u$ is the recovery variable, $I$ is the input stimulus current, and $a$ is a dimensionless parameter. The FHN model can be also re--expressed in the form of (\ref{eq:2_28}) and then easily mapped on the proposed platform:
\begin{equation}\label{eq:2_36}
  \left\{
  \begin{array}{l l}
  x=v, y=u\\
  \alpha=1, \beta=0.8a \\
  F(x)=x-\frac{v^3}{3} \\
  G(x)=(x+0.7)/0.8 \\
  \end{array} \right. .
\end{equation}

\par Figure \ref{fig:Figure12} (a-h) illustrates the hardware results on a digital oscilloscope for four significant phenomena of the FHN model reproduced by the cellular model. As can be seen in the figure, the hardware results obtained from the proposed structure are practically identical compared to the biological ones \cite{FitzHugh1961}.

\subsection{3-D Neuron Models}
In order to show the applicability of the proposed cellular model in higher dimensions, in this section, we apply the approach on the Hindmarsh--Rose (HR) neuron model \cite{Hindmarsh1984} as a 3--D case study. The dynamical behavior of the HR neuron model is given by:
 \begin{equation}\label{eq:2_37}
\begin{cases}
  \dot{v}=u-v^3+3v^2-k+I \\
   \dot{u}=1-5v^2-u \\
   \dot{k}=r(4(v+\frac{8}{5})-k)
\end{cases}.
\end{equation}
\par In this model, $v$ is the membrane potential, $u$ is the spiking variable, $k$ is the bursting variable and $I$ is the applied neuron current. The equations can be re--expressed as:
\begin{equation}\label{eq:2_38}
\begin{cases}x[n]-x[n-1]=\Delta t\cdot [F(x[n-1])+\alpha y[n-1]+\gamma z[n-1]] \\
y[n]-y[n-1]=\Delta t\cdot [G(x[n-1])+\beta y[n-1]]\\
z[n]-z[n-1]=\Delta t\cdot [H(x[n-1])+\lambda z[n-1]]
\end{cases}
\end{equation}
where
\begin{equation}\label{eq:2_39}
\begin{cases}
  x=v, y=u, z=k,\\
  \alpha=1, \gamma=-1, \beta=-1, \lambda=-r \\
  F(x)=x^3-3x^2 \\
  G(x)=1-5x^2 \\
  H(x)=r(4(x+\frac{8}{5}))
\end{cases}.
\end{equation}

\begin{figure*}[t]
\vspace{-10pt}
\normalsize
\setcounter{equation}{5}
\centering
\includegraphics[trim = 0.1in 0.1in 0.1in 0.1in, clip, height=3in,width=5.5in]{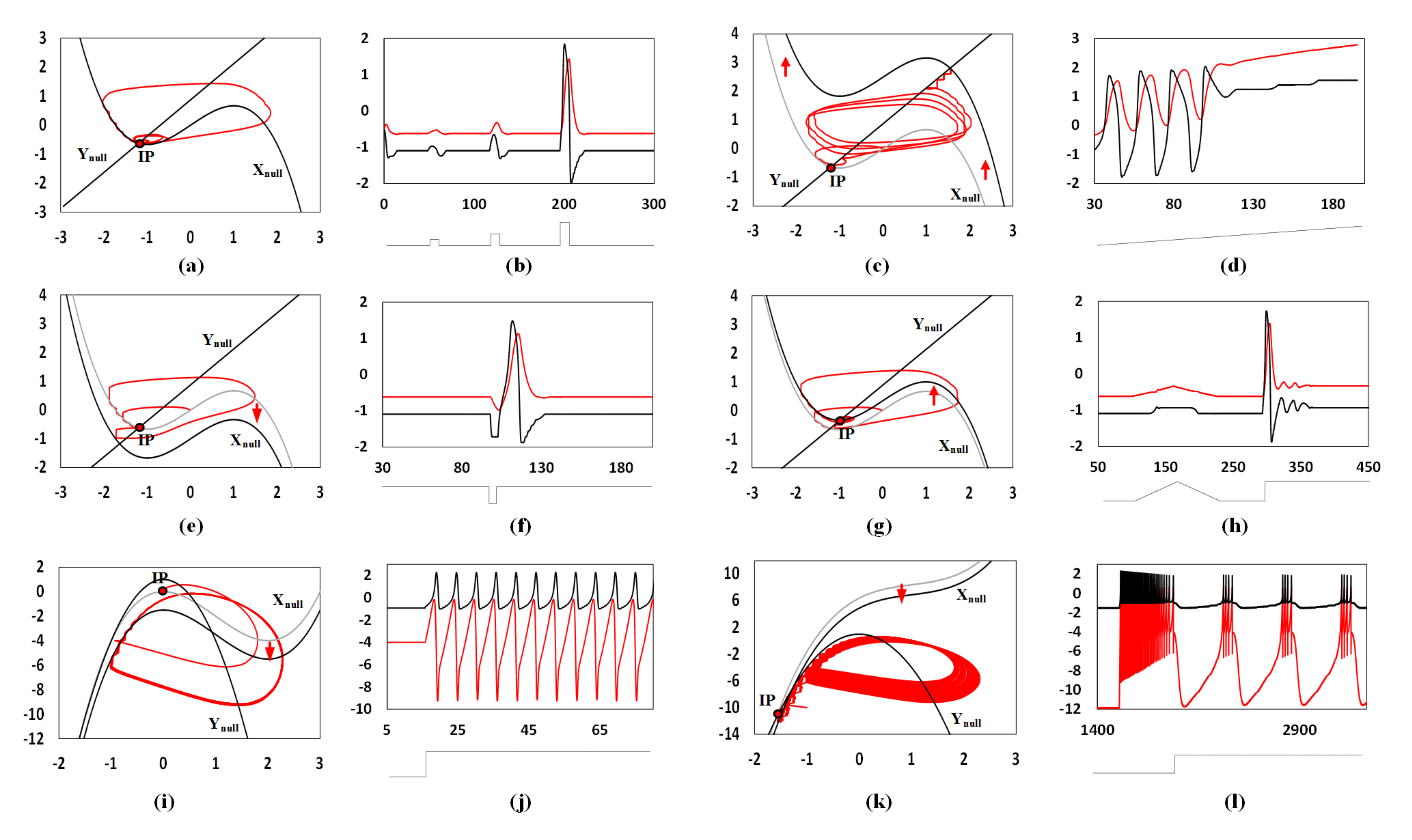}
\vspace{-10pt}
\captionsetup{font=footnotesize}
\caption{Phase plane (left side) and time-domain (right side) representations of (a-b) absence of all-or-none spikes phenomenon, (c-d) excitation block
phenomenon, (e-f) post-inhibitory rebound spike phenomenon, and (g-h) spike accommodation phenomenon for the cellular FHN neuron and (i-j) spiking mode, (k-l) bursting mode for the cellular HR neuron with 32 pixels. In the phase plane, $\textbf{IP}$ is the initial location of the state variable.}
\vspace{-10pt}
\label{fig:Figure12}
\end{figure*}

\par Figure \ref{fig:Figure12} (i-l) shows the hardware results on a digital oscilloscope for two spiking and bursting modes produced by the synchronous cellular approach. As can be seen in the figure, the cellular model is able to reproduce high precision time domain waveforms, very close to the biological model \cite{Hindmarsh1984}.
\par Note that, due to a lack of space, bifurcation analyses are not presented here, however it has been confirmed at least for two important bifurcations (saddle node on/off invariant circle) that the proposed cellular model can reproduce similar mechanisms in a fair manner.

\begin{table}[t]
\captionsetup{font=footnotesize}
\caption{Cellular Values for various neuron models.}   
\centering          
\begin{tabular}{c c c c}    
\hline\hline                        
Parameters &Izhikevich&FHN&HR\\ [0.5ex]  
\hline                      
$\Delta x$ & 2 &     0.1250 &  0.1250 \\
$x_{min}$ & -80 & -2 & -2 \\
$x_{max}$ & -16 & 2 & 2\\
$dt$ & $\frac{1}{32}$ & $\frac{1}{1024}$ & $\frac{1}{32}$\\
$N$ & 32 & 32 & 32\\
\hline          
\end{tabular}
\vspace{-10pt}
\label{tab:Table5}    
\end{table}

\section{Truncation Error}
As explained before, the approach detailed in this work converts the continuous nullclines into a cellular space with discrete cells locating any point with an address prepared by the dynamical system's outputs. This conversion allows our synchronous approach to track any trajectory in the phase plane when discretized by a sufficient number of pixels. However, depending on the cellular space dimension, a truncation error can be observed in the system after each clock cycle. In this section, we formulate this error for a 2--D dynamical system and discuss its sources and how to determine an optimum hardware architecture. First, let us assume that $dt$ is small enough and its corresponding error is negligible. Now, if we fetch the cellular nullclines values from the storage blocks, the next state variables are:
\begin{equation}\label{eq:2_40}
\begin{cases}
x[n]_{cellular}=\overbrace{x[n-1]+dt[F(x[n-1])+\alpha y[n-1]+IN]}^{x_{continuous}}\\
\pm\epsilon_1\\
y[n]_{cellular}=\overbrace{y[n-1]+dt[G(x[n-1])+\beta y[n-1]]}^{y_{continuous}}\pm\epsilon_2
\end{cases}
\end{equation}
where $\epsilon_1$ and $\epsilon_2$ are the differences between the continuous nullclines values and the corresponding cellular values. Thus, the address of the next cellular values fetched from the storage registers for both dimensions is given by:
\begin{equation}\label{eq:2_41}
X[n]=\lfloor \frac{x[n]_{continuous}\pm \epsilon_1-x_{min}}{\Delta x} \rfloor
\end{equation}
\par This implies that the value of $\epsilon_1$ and the floor function both contribute to the truncation error in the system since the $X[n]$ must be an integer for both addresses. This truncation error takes the form of momentary and permanent lag/lead deviations in the time domain signals. However, the cellular trajectories can track well the continuous ones in the phase plane under certain conditions on the number of pixels (i.e. the error is reduced by increasing the number of pixels).

\begin{table}[h]
\captionsetup{font=footnotesize}
\caption{RMSE and NRMSE Error for 1000 Points of Some Cases of Neuron Models with Various Resolutions.}   
\resizebox{\columnwidth}{!}{%
\centering          
\begin{tabular}{c c c c c c c}    
\hline\hline                        
Responses & \multicolumn{2}{c}{32}& \multicolumn{2}{c}{64} & \multicolumn{2}{c}{128}\\ [0.5ex]  
         & RMSE& NRMSE$\%$& RMSE& NRMSE$\%$&RMSE& NRMSE$\%$\\ [0.5ex]  
\hline                      
IZHI$_{TS}$& 2.482& 2.61& 1.893&1.98& 0.735& 0.77\\
FHN$_{EB}$& 0.107& 3.25& 0.080&2.13& 0.043& 1.15\\
HR$_{TS}$& 0.094& 2.87& 0.052&1.65& 0.031& 0.82\\
Mean Error& 0.894& 2.91& 0.675&1.92& 0.269& 0.91\\
\hline          
\end{tabular}
}
\label{tab:Table6}    
\end{table}

\section{Error Analysis}
As a consequence of the aforementioned truncation error, here we define a root mean square error (RMSE) to measure the time domain error of the proposed cellular system ($V_c$) compared to the original biological model ($V_o$). The error criterion is defined as follows:
\begin{equation}\label{eq:2_42}
\text{RMSE}(V_c,V_o)=\sqrt{\frac{\sum\limits_{i=1}^n (V_c-V_o)^2}{n}}
\end{equation}
where the normalized root mean square error (NRMSE) is described by:
\begin{equation}\label{eq:2_43}
\text{NRMSE}=\frac{\text{RMSE}}{V_{o_{max}}-V_{o_{min}}}.
\end{equation}
\par The measured time domain error for 1000 points of some cases of neuron models with various resolutions is shown in Table \ref{tab:Table6}.
The results show that by increasing the number of pixels, the error is decreased as predicted in the previous section.

\begin{table}[t]
\captionsetup{font=footnotesize}
\caption{Device Utilization of the Xilinx Virtex--II Pro FPGA for the Synchronous Cellular Izhikevich Model and Previous Published Piecewise Linear Model \cite{Soleimani2012}.}   
\centering          
\begin{tabular}{c c c c}    
\hline\hline                        
  Parameter & Cellular IZHI& IZHI (4pwl)\cite{Soleimani2012} & Performance\\ [0.5ex]  
\hline                      
FF& 265& 491& 1.85 times less\\
4--input LUT& 274& 602& 2.19 times less\\
Freq. (MHz)& 254.26& 204.31& 1.24 times more\\
\hline          
\end{tabular}
\label{tab:Table7}     
\end{table}

\begin{table}[t]
\captionsetup{font=footnotesize}
\caption{Device Utilization of the Xilinx Virtex--II Pro FPGA for the Synchronous Cellular FHN Model and Previous Published Piecewise Linear Model \cite{Nouri2015}.}   
\centering          
\begin{tabular}{c c c c}    
\hline\hline                        
  Parameter & Cellular FHN& FHN \cite{Nouri2015} & Performance\\ [0.5ex]  
\hline                      
FF& 370& 526& 1.42 times less\\
4--input LUT& 373& 1085& 2.9 times less\\
Freq. (MHz)& 248.14& Not reported& -\\
\hline          
\end{tabular}
\label{tab:Table8}     
\end{table}

\begin{table}[t]
\captionsetup{font=footnotesize}
\caption{Device Utilization of the Xilinx Virtex--II Pro FPGA for the Synchronous Cellular HR Model and Previous Published Piecewise Linear Model \cite{Hayati2016}.}   
\centering          
\begin{tabular}{c c c c}    
\hline\hline                        
  Parameter & Cellular HR& HR \cite{Hayati2016} & Performance\\ [0.5ex]  
\hline                      
FF& 425& 431& 1.01 times less\\
4--input LUT& 435& 659& 1.51 times less\\
Freq. (MHz)& 309.37& 81.2& 3.80 times more\\
\hline          
\end{tabular}
\label{tab:Table9}     
\end{table}

\begin{table}[t]
\captionsetup{font=footnotesize}
\caption{Device Utilization of the Kintex--7 (XC7K325T) FPGA for the All Synchronous Cellular Neuron Models.}   
\centering          
\begin{tabular}{c c c c}    
\hline\hline                        
  Parameter & Cellular IZHI& Cellular FHN & Cellular HR\\ [0.5ex]  
\hline                      
FF& 187& 256& 364\\
6--input LUT& 223& 414& 416 \\
Freq. (MHz)& 597.58& 563.88& 663.10\\
\hline          
\end{tabular}
\label{tab:Table10}     
\end{table}

\section{Hardware Synthesis Results}
\par The concept of employing digital piecewise linear models in neuromorphic engineering was first introduced in \cite{Soleimani2012}, and since then a number of valuable attempts were accomplished following the same general track \cite{Soleimani2012}--\cite{Hayati2016}. The basic idea behind these works is how to modify the critical path in the digital designs while preserving the time domain and phase domain properties of the biological model. For a fair comparison between the proposed cellular model and previously published piecewise linear models, the design is first synthesized on a Xilinx Virtex--II Pro (XC2VP30) FPGA. Device utilization details for the synthesis of the cellular and other piecewise linear models are summarized in Table \ref{tab:Table7}, \ref{tab:Table8} and \ref{tab:Table9}.
\par The results confirm that thanks to the efficient design of the cellular model (and as expected), a smaller area and a higher clock frequency are achieved compared to previously published piecewise linear neuron models. Since there are only 4--input LUTs (16-bits maximum) in the Xilinx Virtex--II Pro (XC2VP30) FPGA, each stored nullcline's component with 18 bits cannot be accommodated in one LUT, leading to a more complex routing process with lower operating frequency for the cellular model. Therefore, the cellular models are also synthesized on the Kintex--7 (XC7K325T) FPGA containing 6-input LUTs fabricated by more advanced technology. The synthesis results show a higher speed operating frequency and lower LUT usage as anticipated. The device utilization for this experiment is summarized in Table \ref{tab:Table10} .
\chapter{A Low Power Digital IC Emulating Intracellular Calcium Dynamics}
\renewcommand{\baselinestretch}{\mystretch}
\label{chap:Future}

\PARstart{L}{ow} Low power/area cytomorphic chips may be interfaced and ultimately implanted in the human body for cell--sensing and cell--control applications of the future. In such electronic platforms, it is crucial to accurately mimic the biological time--scales and operate in real--time. This chapter proposes a methodology where slow nonlinear dynamical systems describing the behavior of naturally encountered biological systems can be efficiently realised in hardware \cite{6}. To this end, as a case study, a low power and efficient digital ASIC capable of emulating slow intracellular calcium dynamics with time--scales reaching to seconds has been fabricated in the commercially available AMS 0.35 $\mu m$ technology and compared with its analog counterpart. The fabricated chip occupies an area of 1.5 $mm^2$ (excluding the area of the pads) and consumes 18.93 $nW$ for each calcium unit from a power supply of 3.3 V. The presented cytomimetic topology follows closely the behavior of its biological counterpart, exhibiting similar time--domain calcium ions dynamics. Results show that the implemented design has the potential to speed up large--scale simulations of slow intracellular dynamics by sharing cellular units in real--time.

\section{Introduction}
Depending on the specifications of the targeted application, one of these approaches or a mixture of both might be more efficient and useful. For example, in the case of large--scale implantable cytomorphic chips, where the biological dynamics is slow, the size of capacitors utilised in each analog processor may become very large (for example, in \cite{Woo2015} it is reported as 1 $\mu F$) and thus not practical. In such cases, digital platforms may benefit from the slow character of the targeted dynamics by emulating more units in a pipelined structure, although data conversion is necessary. It can be shown that as the number of emulating cells increases, the hardware cost of data conversion drops. In this paper, we show that digital hardware are efficient platforms in terms of area and power consumption for the implementation of slow biological dynamics in large--scale and real--time. The validity of this claim is confirmed by fabricating (in the commercially available AMS 0.35 $\mu m$ technology) and tested by means of a digital ASIC emulating the Calcium--Induced Calcium Release (CICR) model introduced in \cite{Dupont1993}.

\section{Hardware Implementation}
\par The proposed ASIC design presented in Figure \ref{fig:4_1}(a) contains five major blocks as follows:

\subsubsection{Input/Output Interfaces}
\par In this design, the output signals are provided in two parallel and serial forms. The parallel signals comprise 6 pins for each state variable ($X/Y$), while the serial output transmits the state variables on a single wire according to the UART protocol. The transmission bit rate for the parallel case is equal to $12\times CLK_{core}$ and for the serial one is determined by the UART clock ranging from 115200 to 110 bit per second. It should be noted that in the serial mode, since the UART packets carry 10 bits (one start, two stop and seven data bits), the $CLK_{core}$ is defined as $\frac{CLK_{serial1}}{10\times Num.~of~packets}$; in this case $Num.~of~packets$ is 4 since there are two packets per each state variable. $CLK_{serial2}$ must be at least 8 times faster than $CLK_{serial1}$ to make sure all input bits are captured properly by the UART receiver. Due to the limited number of pads, the input of the system only accepts values via the UART serial protocol and is calibrated by the $Controling~Signals$ pins.

\subsubsection{Storage Blocks}
\par The cellularized velocity values for all cells are stored in the storage blocks. The velocity vectors are calculated off--line according to predetermined model parameters explained in the previous chapter. These signed values are stored in two $R\times S$ (number of pixels) storage blocks. The size of velocity components is exactly the same as the bandwidth of the system and defines the length of each memory cell, which is 14 bits in this design.

\begin{figure*}[t]
\vspace{-10pt}
\centering
\includegraphics[trim = 0.2in 0in 0in 0in, clip, width=5.5in]{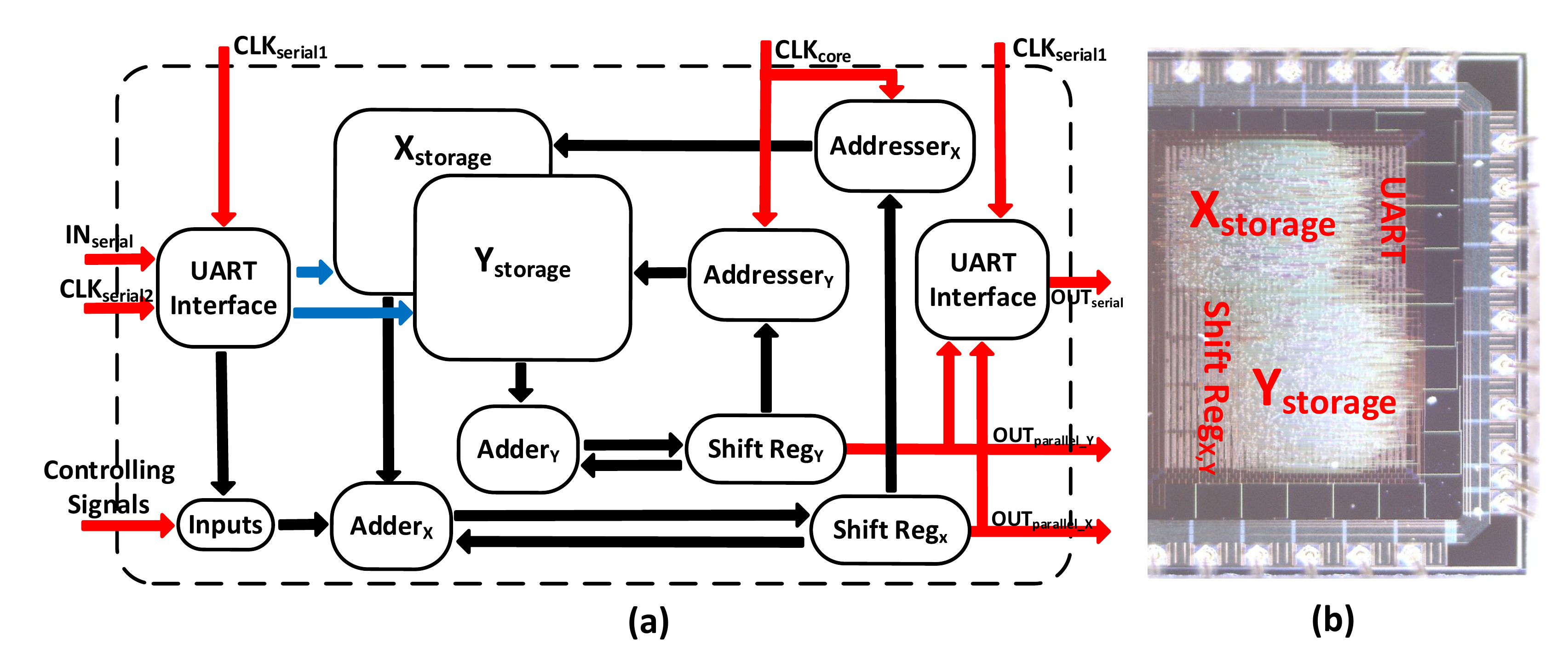}
\vspace{-10pt}
\captionsetup{font=footnotesize}
\caption{(a) The detailed internal structure of the ASIC design capable of emulating 16 calcium units in real--time. In this structure red, black and blue arrows respectively demonstrate in/out ports, internal buses and future possible connections leading to higher design flexibility (the blue connections are not implemented in the current version). (b) Microphotograph of the chip, fabricated in the commercially available 0.35 $\mu m$ AMS technology. The chip occupies an area of 1.5 $mm^2$ including the pipelined network and excluding the area of the pads.}
\vspace{-15pt}
\label{fig:4_1}
\end{figure*}

\subsubsection{Adders}
\par An adder is employed for each dimension, which adds the velocity value fetched from the storage blocks to the previous state of the dynamical variable. According to cellular model explained in the previous chapter, the input is also added to this value for the $X$ cellular variable. The transferred values from the storage blocks and the input determine the motion direction on the cellular phase plane. For example, when the value of velocity received from \textbf{$Buffer_X$} (see Figure \ref{fig:4_1}(a)) plus the input is positive, an upward motion in the $X$ direction occurs, while if the sum is negative the motion is downward. The same holds for \textbf{$Buffer_Y$} and the resultant direction on the 2D discrete phase planes is determined by motions in both $X$ and $Y$ directions. The absolute value of the velocities determines the amount of increase or decrease in the state variables in each step. Thus, the higher the absolute velocity value, the bigger the increase of the state variable.

\subsubsection{Addressers}
\par The location of each state variable in the cellular phase plane is calculated by this block. In other words, this block is responsible for converting the non--cellular output variables to the cellular addresses to fetch the next velocity values from the storage blocks. The address of the next state in the cellular space can be conveniently implemented by means of one subtract and one shift operation.

\subsubsection{Buffers}
Since the critical path in the proposed cellular model is determined only by one subtractor, we can easily share the digital hardware for large scale simulations in cytomimetic circuit design. To this end, a network of pipelined calcium units corresponding to 16 individual CICR models capable of operating in real--time is presented. The maximum value of this equals to $clock~core\cdot dt$ leading to a significant area reduction in such systems. As illustrated in Figure \ref{fig:4_1}(a), \textbf{$Buffer_X$} and \textbf{$Buffer_Y$} store and shift the data of each calcium unit in each clock pulse.

\begin{figure}[t]
\centering
\includegraphics[trim = 0in 0in 0in 0in, clip, width=4in]{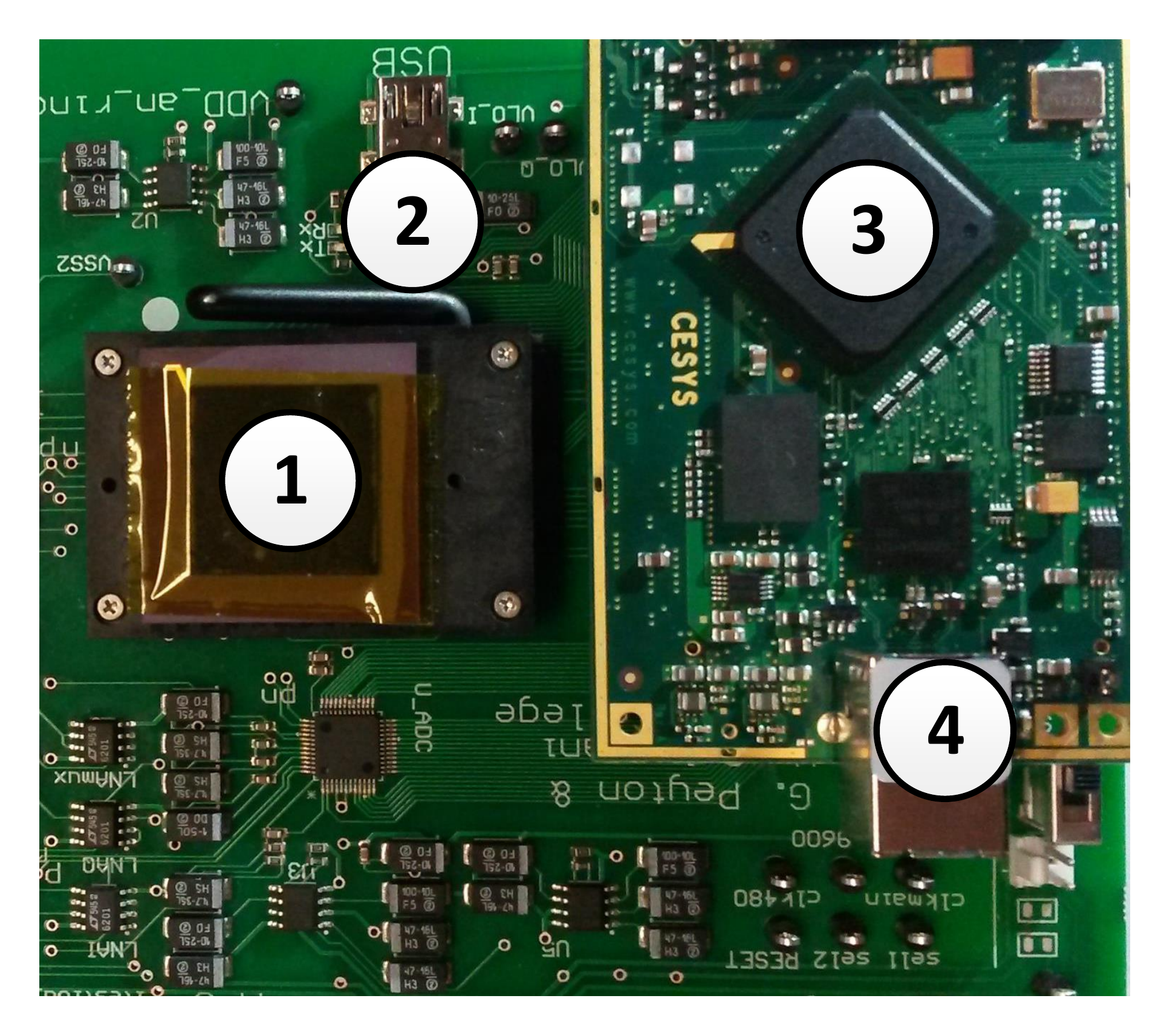}
\vspace{-5pt}
\captionsetup{font=footnotesize}
\caption{The printed circuit boards containing: (1) fabricated chip with PGA84 package mounted on a PGA ZIF header; (2) UART interface chip (FT232) and mini USB connector; (3) FPGA Spartan-6 XC6SLX150; (4) USB interface for programming the FPGA by the PC.}
\vspace{-15pt}
\label{fig:4_2}
\end{figure}

\subsection{Hardware Layout}
\par The resulting IC, including the serial UART interface and the pipelined network covers an area of 1.5 $mm^2$ (excluding the area of the pads), while consuming a power of 18.93 $nW$ for each calcium unit. It is powered from a 3.3 V supply and it contains 3788 basic gates. Figure \ref{fig:4_1}(b) displays a microphotograph of the chip layout in which different modules correspond to the structure shown in Figure \ref{fig:4_1}(a). The chip has one main module and four submodules. Thirteen outputs and six inputs are embedded in the design comprising the power supply pins to communicate with external devices.

\begin{figure*}[t]
\vspace{-10pt}
\centering
\includegraphics[trim = 0.2in 0in 0in 0in, clip, width=5.5in]{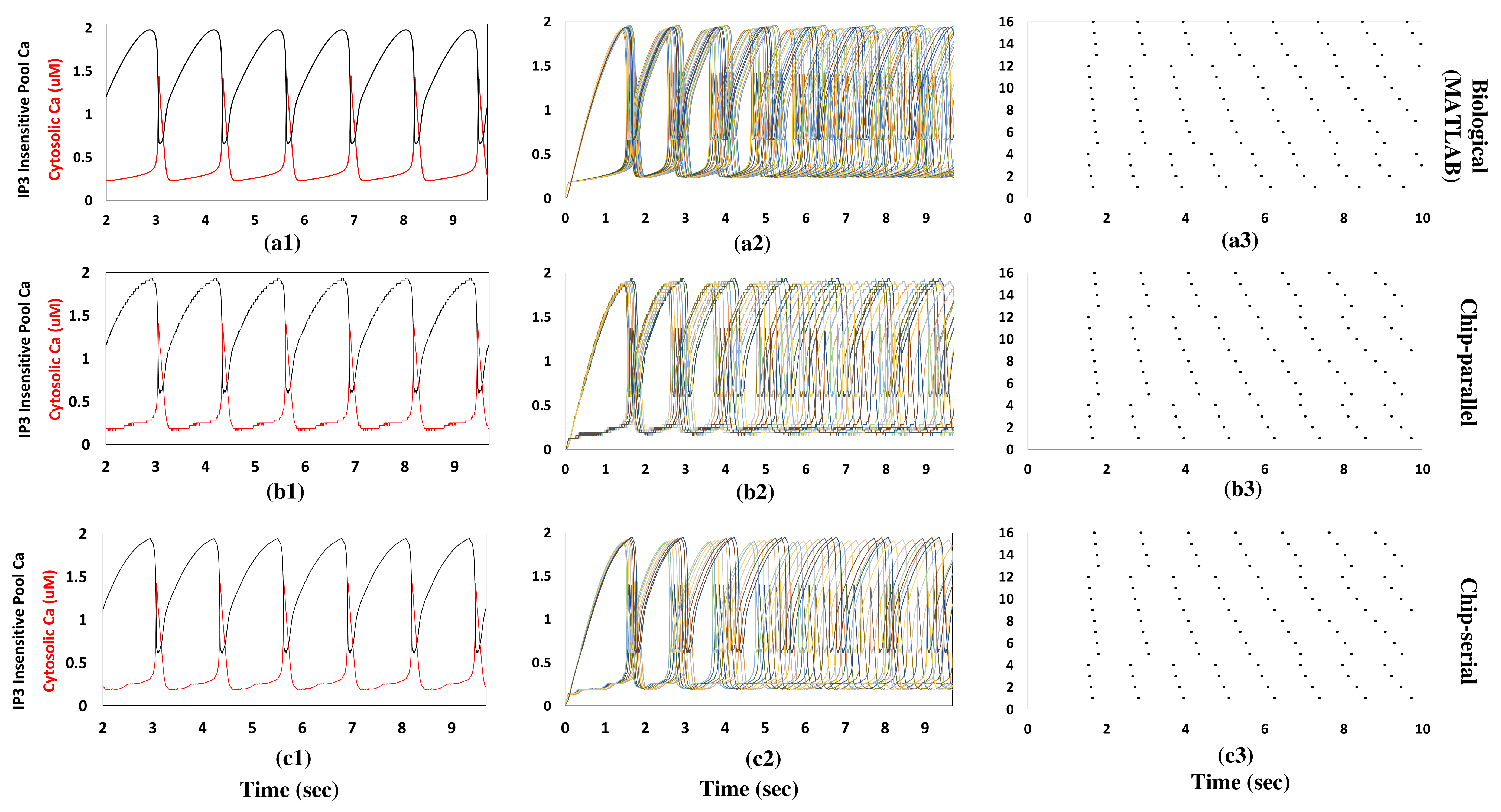}
\vspace{-5pt}
\captionsetup{font=footnotesize}
\caption{(a1--c1) Time domain comparison between the biological (MATLAB) and the cellular model with two different output protocols (chip--parallel and chip--serial) for a single calcium unit. Cytosolic $Ca^{2+}$ and $IP_3$ insensitive pool $Ca^{2+}$ waveforms are respectively shown in red and black color graphs. (a2--c2) and (a3--c3) are time domain and raster plot results respectively of a network activity comprised of 16 calcium units with inhomogeneous inputs.}
\vspace{-15pt}
\label{fig:4_3}
\end{figure*}

\section{Chip vs Simulated Time Domain Results}
\subsection{Experimental Setup}
The experimental setup consists of three main components: a PC, a generic FPGA interface and the cytomimetic digital ASIC. The PC records data and controls the cytomimetic system via the FPGA interface. Figure \ref{fig:4_2} shows the printed circuit boards that host two main hardware components of the system, including an FPGA, and the cytomimetic chip. The serial UART interface enables the chip to communicate bi-directionally with the PC. The parallel outputs of the chip are also connected to the FPGA and available to the PC through the UART serial interface implemented on the chip board.
\par Both FPGA and chip boards contain the circuitry needed to ensure the proper functionality and testing of the chip, FPGA and the serial transceiver, such as voltage regulators and connectors to measure digital input/output voltages from the chip. The chip is driven by three main clocks supplied by the FPGA board. In this setup, 115200 baud rate is used in the design to transmit data to PC, thus, the $CLK_{serial1}=115200$, $CLK_{core}=\frac{115200}{40}=2880$ (each output value is coded by two 10 bits and the system has two outputs at the time, therefore 40 bits are needed to be sent out for one meaningful set of the state variables) and the $CLK_{serial2}=115200*8$.

\subsection{Single Calcium Behavior}
Time domain waveforms for the biological and cellular model with two different output forms (parallel and serial) simulated respectively by MATLAB and ASIC are shown in Figure \ref{fig:4_3}(a1--c1). The cellular parameters corresponding to the implemented 32--pixel cellular model are $\Delta x=0.0625$, $\Delta y=0.0625$, $x_{min}=-0.1$, $x_{max}=1.9$, $y_{min}=-0.1$ and $y_{max}=1.9$. Results show acceptable agreement between the MATLAB and chip results, however due to the low resolution of the fixed point parallel outputs, Figure \ref{fig:4_3}(b1) shows poorer results compared to the serial case.

\subsection{Network Behavior}
To investigate the applicability of the proposed cellular model in a large scale design, the results simulated by both MATLAB and ASIC of a network model constructed with 16 calcium units are compared. The input function for the simulated network in both cases is given by:
\begin{equation}\label{eq:4_1}
IN_{ext_i}=\alpha.\eta+\gamma
\end{equation}
where $\eta$ is a random number between 0 to 1 generated by a uniform distribution and other parameters are presented in Table \ref{table:4_1}. The time domain signals and the corresponding raster plots for the biological and the chip models with two different output protocols (parallel and serial) are demonstrated in Figure \ref{fig:4_3}(a2--c2) and (a3--c3) respectively.
\par In this figure, temporal evolution of the firing rate shows good agreement between the cellular and biological models. It can be seen that the calcium units excited by a noisy input are destabilised after a certain amount of time. The chip with parallel outputs still shows a bit poorer time domain result (Figure \ref{fig:4_3}(b2)). However, since in large scale simulations/emulations, the statistical nature of such activities is of interest in general, the trivial disagreement is not significant in the raster plot shown in Figure \ref{fig:4_3}(b3).

\begin{table}[t]
\captionsetup{font=footnotesize}
\caption{Input Values for the Network Model Constructed by 16 of Calcium Units.}   
\centering          
\begin{tabular}{c c c}    
\hline\hline                        
Parameters &Biological Model&Cellular Model\\ [0.5ex]  
\hline                      
$\alpha$ ($\mu M/s$)&  0.2 & 0.27\\
\hline
$\gamma$ ($\mu M/s$)& 2.7& 2.7 \\
\hline
\end{tabular}
\label{table:4_1}    
\end{table}

\begin{table}[b]
\captionsetup{font=footnotesize}
\caption{Area and Average Power Comparison Between the Proposed ASIC Design and the Equivalent Analog Counterpart Emulating the Same Case Study.}   
\centering          
\begin{tabular}{c c c c}    
\hline\hline                        
  \textbf{Quantitative Parameters} & Digital (per calcium unit)+ ADC+ DAC& Analog \cite{Papadimitriou2013}\\ [0.5ex]  
  & measured & simulated \\
\hline                      
\textbf{Area ($mm^2$)}& $\frac{1.50+0.86+0.35}{16}=0.169$& $0.661$\\
\textbf{Average Power ($\mu W$)}& $\frac{0.30+0.64+0.43}{16}=0.08$& $1.53$\\ [1ex]        
\hline          
\end{tabular}
\vspace{-10pt}
\label{table:4_2}  
\end{table}

\section{Analog vs Digital}
\par To verify the suitability of digital platforms for slow biological dynamics, here we qualitatively and quantitatively compare both analog and digital designs on the same case study. In analog designs, computations are performed continuously and based on the physics of the devices (continuous--time continuous--value designs). In contrast, in digital designs, computations are performed upon discrete values of physical variables (discrete--time discrete--value designs). This would require to use data converters in the digital design in order to interact with biological systems.
\par In general it can be argued that analog designs consume less area compared to their digital equivalents, but when emulating slow biological systems in large--scale and real--time, the size of capacitors in analog designs may become large and impractical. In such cases, digital designs become more area efficient as they can benefit from the slow character of biological dynamics by emulating more units in real--time. Quantitative measurements of the fabricated chip are compared with simulated results reported in \cite{Papadimitriou2013}. For the sake of comparison, the area and power consumptions of a 10--bit CMOS DAC and a 6--bit ADC are extracted from \cite{Van2001}--\cite{Sandner2005} and adapted according to the fabrication technology, operating frequency and power supply used in this paper. It should be noted that one ADC and one DAC converter can be shared between all 16 units. Table \ref{table:4_2} shows almost 4 and 18 times area and power reduction respectively including ADC and DAC modules. Note that the simulated power consumption reported from \cite{Papadimitriou2013} is static, thus the total average value may be even higher leading to further power reduction for the digital design. Such a reduction in area and power consumptions is only based on 16 pipelined calcium units, while further reductions can be achieved by sharing more units. The maximum number of shared units is limited by the operating frequency and integration time step and can be calculated as $clock~core\times dt$, which in this case is ~$700k$ shared units. In large--scale designs the hardware cost consumed by data converters becomes small compared to the processing part. The efficiency of digital designs for the emulation of slow biological dynamics increases with scaling down of the feature size.
\par On the other hand, the operation of the digital design is characterised by current spikes whose typical duration is very short. Such current spikes observed during measurements of the fabricated chip were managed at layout stage by sizing the width of the power wires and adding decoupling capacitors. Large--scale analog designs (e.g. very long cochlear cascades) are prone to noise, mostly due to thermal fluctuations in physical devices, while in digital designs noise is due to round--off error which can be alleviated significantly at the expense of increased datapath bit length. Moreover, in large--scale analog designs computation is also offset prone due to mismatches in the parameters of the physical devices leading to lower accuracy. While in digital designs, computation is not offset prone since it is insensitive to mismatches in the parameters of the physical devices. Generally, it is also accepted that digital designs have a better scalability property in comparison with their analog counterparts \cite{Sarpeshkar1998}. It should be noted however that in the case of cytomorphic chips in which a few implantable cell--sensing and cell--control units are needed and scaling--up is not critical, analog ultra--low--power designs \cite{Houssein2015} may offer more practical interfacing (and better performance) with the biological systems.

\begin{table}[t]
\captionsetup{font=footnotesize}
\caption{Average Power Consumption for the Fabricated Digital ASIC Consisting 16 Pipelined Calcium Units with the Area of 1.5 $mm^2$.}   
\centering          
\begin{tabular}{c c c c}    
\hline\hline                        
  Bio--timescale (sec) & dt (sec) & Freq. (Hz) & Ave. Power ($\mu W$) \\ [0.5ex]  
\hline                      
1& $\frac{1}{64}$& $64\times16$&0.200\\
0.5&$\frac{1}{128}$& $128\times16$&0.303\\
0.25&$ \frac{1}{256}$&$256\times16$&0.486\\
0.125& $\frac{1}{512}$&$512\times16$&0.849\\
0.0625& $\frac{1}{1024}$&$1024\times16$&1.586\\ [1ex]
\hline          
\end{tabular}
\vspace{-10pt}
\label{table:4_3}   
\end{table}

\section{Area and Power Tradeoffs for Large Scale Designs}
\par Nowadays, many biological processes are codified in the form of mathematical dynamical systems. According to the nature of these processes, various time-domain evolution speeds are observed ranging from milliseconds to hours. For example, in bioelectrical systems (e.g. spiking neural networks in the brain) time scales are in the order of milliseconds while in pure biochemical systems (e.g. the expression of proteins in the cell) time scales can be in the order of hours. In digital synchronous designs, the frequency of the system is defined as follows:
\begin{equation}\label{eq:4_2}
Operating~Frequency=\frac{Pipelined~Units}{dt}
\end{equation}
where $Pipelined~Units$ is the number of embedded pipelined units in the digital design and $dt$ is the Euler time step. Table \ref{table:4_3} illustrates the relation between the speed of the biological dynamical system (CICR model) and the digital ASIC power consumption. As expected, the lower (the slower) the biological timescale, the lower the power consumption. The reason for this decrease is that the dominant part of the power consumption in such designs is the dynamic switching part demonstrated by \cite{Sarpeshkar1998}:
\begin{equation}\label{eq:4_3}
P_{switching}\propto V_{DD}^2\times C_{load}\times f
\end{equation}
where $V_{DD}$ is the power supply voltage, $C_{load}$ and $f$ are the capacitance load and the number of toggles in the output node. This means that faster biological dynamical systems with higher operating frequency demand more power consumption. As shown in Table \ref{table:4_3}, when the biological timescale decreases, $dt$ should also decrease leading to higher operating frequency and consequently higher power consumption. However, such an issue can be alleviated by using more silicon area and less sharing of hardware resources especially with the modern fabrication technology capable of being scaled down even to 16nm. For example, IBM has implemented the most dense neuromorphic chip (28 nm technology) comprising 1 and 256 million individual neurons and synapses respectively with only 73 $m W$ power consumption \cite{Truenorth2016}.
\par By scaling down the fabricated chip to 28 nm, the current area would shrink to 0.0096 $mm^2$ (almost 156 times smaller). Such a reduction in area permits the designer to implement more physical calcium units in silicon. Therefore optimum design would share maximum pipelined designs so that the power constraints are met and the implemented design can properly operate in real--time.

\section{Reconfigurability}
\begin{figure}[h]
\centering
\includegraphics[trim = 0.2in 0in 0.2in 0.2in, clip, width=5in]{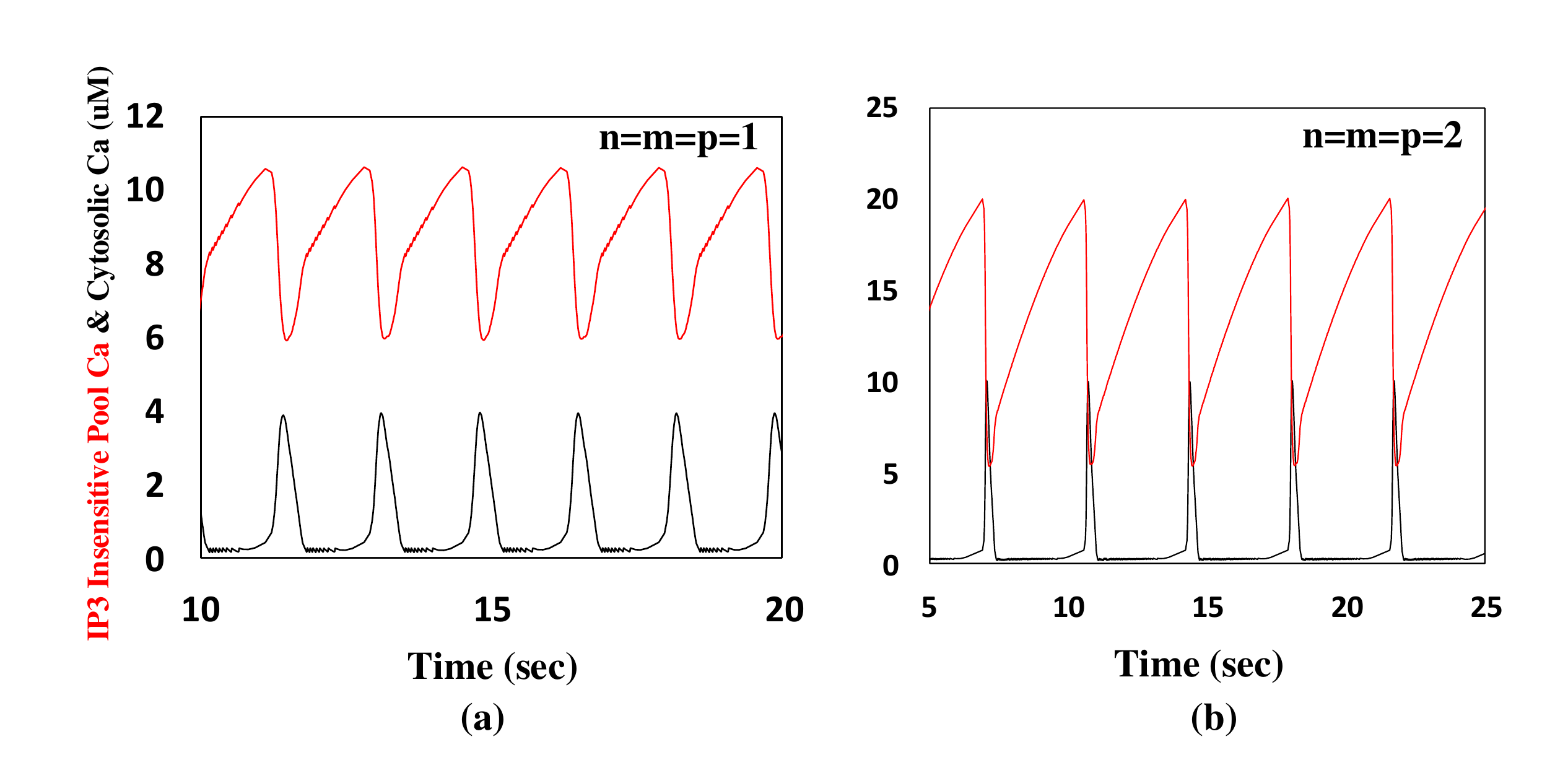}
\vspace{-10pt}
\captionsetup{font=footnotesize}
\caption{Simulated hardware waveforms of a single calcium unit for the proposed cellular model with two different Hill coefficient value sets of (a) m=n=1, (b) m=n=2. Cytosolic $Ca^{2+}$ and $IP_3$ insensitive pool $Ca^{2+}$ waveforms are respectively shown in black and red color.}
\label{fig:4_4}
\vspace{-10pt}
\end{figure}

The proposed ASIC has the ability to be fully reconfigurable and reprogrammable for all possible CICR dynamics with various Hill coefficient values. This reconfigurability stems from the memory--based architecture of the proposed digital hardware. Such a flexibility is limited in practice when it comes to the implementation of dynamical systems in analog CMOS--based design. Such reconfigurability could be implemented in hardware as depicted by two blue arrows in Figure \ref{fig:4_1}(a). The simulated hardware results are shown in Figure \ref{fig:4_4} for two other calcium dynamics. Such a reconfigurability comes with a negligibly higher hardware cost (about $7\%$ and $9\%$ higher area and power consumption respectively) and calls for properly tuned cellular parameters for a certain output dynamic range.

\section{Truncation Error}
As explained in \cite{Jokar2017}, depending on the cellular space dimension, a truncation error can be observed in the system in each clock cycle. By assuming a small enough $dt$ so that its corresponding error is negligible, two error criteria can be defined \cite{Jokar2017}: I) The quantity
\begin{equation}\label{eq:4_4}
\epsilon=\sum_{i=0}^{R-1}\sum_{j=0}^{S-1} \frac{\epsilon_{x_{(i,j)}}+\epsilon_{y_{(i,j)}}}{2R\cdot S}.
\end{equation}
where $\epsilon_x$ and $\epsilon_y$ are the difference between the continuous state variables and the corresponding cellular values and $R$ and $S$ are the dimensions of the cellular space. Clearly, if the memory bit--length increases, the corresponding $\epsilon$ decreases, leading to smaller time domain errors. II) The quantity
\begin{equation}\label{eq:4_5}
\zeta=\sum_{i=0}^{\tau-1}\frac{x^+_{(i)}-X^+_{(i)}+y^+_{(i)}-Y^+_{(i)}}{2\tau}
\end{equation}
where $x^+=\frac{x^+_{continuous}\pm \epsilon_x-x_{min}}{\Delta x}$, $y^+=\frac{y^+_{continuous}\pm \epsilon_y-y_{min}}{\Delta y}$ and $\tau$ is the error measurement time, a multiple of $dt$. By increasing the resolution of the stored velocities (number of pixels), the corresponding $\zeta$ decreases, leading to smaller time domain errors. As a consequence of the aforementioned truncation errors, we define a root mean square error (RMSE) to measure the time domain error of the proposed cellular system compared to the biological model. The measured time and phase domain errors for a 32--pixel cellular model are shown in Table \ref{table:4_4}. It should be stressed that the truncation errors appear in the form of momentary and permanent lag/lead and deviation in the time domain signals where the velocity changes are more erratic and uneven. However, under certain conditions applied upon the memory bit-length and the number of pixels, the cellular trajectories can track fairly the continuous ones in the phase plane as seen in the 32--pixel cellular model with 14--bit (4.10) memory bit--length.

\begin{table}[t]
\captionsetup{font=footnotesize}
\caption{The Measured Time and Phase Domain Errors for a 32--pixel Hardware Cellular Model with Various Hill Functions.}   
\centering          
\begin{tabular}{c c c c}    
\hline\hline                        
  Cellular model & $\epsilon$ & $\zeta$ & RSME \\ [0.5ex]  
\hline                      
m=n=p=1& 4.1425e-06& 0.1157&0.3126\\
m=n=p=2& 5.3421e-06&0.1793&0.2932\\
m=n=2, p=4&8.9856e-06&0.0293&0.1031\\
Average&6.1567e-06&0.1081&0.2363\\ [1ex]
\hline          
\end{tabular}
\vspace{-10pt}
\label{table:4_4}    
\end{table}
\chapter{Systematic Computation of Nonlinear Bilateral Dynamical Systems (NBDS) with a Novel CMOS Circuit}
\renewcommand{\baselinestretch}{\mystretch}
\label{chap:Future}

\par Simulations of large--scale nonlinear dynamical systems on hardware with a high resemblance to their mathematical equivalents has been always a challenge in engineering. This paper presents a novel current--input current--output circuit supporting a systematic synthesis procedure of log--domain circuits capable of computing bilateral dynamical systems with considerably low power consumption and acceptable precision. Here, the application of the method is demonstrated by synthesizing four different case studies: 1) a relatively complex two--dimensional (2--D) nonlinear neuron model, 2) a chaotic 3--D nonlinear dynamical system Lorenz attractor having arbitrary solutions for certain parameters, 3) a 2--D nonlinear Hopf oscillator including bistability phenomenon sensitive to initial values and 4) three small neurosynaptic networks comprising three FHN neuron models variously coupled with excitatory and inhibitory synapses. The validity of our approach is verified by nominal and Monte Carlo simulated results with realistic process parameters from the commercially available AMS 0.35 $\mu m$ technology. The resulting continuous--time, continuous--value and low--power circuits exhibit various bifurcation phenomena, nominal time--domain responses in good agreement with their mathematical counterparts and fairly acceptable process variation results (less than $5\%$ STD).

\section{Introduction}
\par Dynamical systems are one of the basic mathematical objects capable of describing time--dependent activities in a geometrical space. Such systems include a set of variables and constants defining the state, and a functional law describing the evolution of the state variables through time. In other words, the dynamical laws establish a meaningful relation among the future state of the system, the inputs and its current state. General qualitative descriptions of dynamical systems can be observed by inspecting their phase portraits, demonstrating velocity and direction of motions in space.

\par The applications of such systems are highly diverse in science and engineering. The mathematical models describing the dynamical systems can be simulated with the use of powerful software such as MATLAB, however, for large--scale simulations software begins to collapse. Besides, computer--based simulations are not always suitable for interfacing with biological/physical systems where continuous monitoring with low power and area consumption might be required \cite{Papadimitriou2013}.
\par This issue can be resolved by the means of specialized hardware tools capable of emulating dynamical behaviours in real--time. The remarkable resemblance between the mathematical description of dynamical systems and the equations governing the current--voltage relations between interconnected log--domain MOS transistors and capacitors lays a groundwork to emulate real--time dynamics with the use of ultra--low power electrical circuits \cite{Sarpeshkar2010}.

\par To this end, a number of valuable attempts have been accomplished ranging from continuous--time low power analog circuits to discrete--time massively parallel digital ones. Exceptional examples can be found in $neuromorphic~electronics$ where brain's neural dynamical systems are mimicked by the use of very-large-scale integration (VLSI) systems containing electronic analog/digital circuits \cite{Furber2014}, \cite{Soleimani2015}, \cite{Indiveri2006}, \cite{Erdener2016}, \cite{Soleimani2012}, \cite{Schlottmann2012}, \cite{Basu2010_1}, \cite{Basu2010_2}, \cite{Brink2013} and \cite{Truenorth2016}. On the other hand, other research efforts have focused on the synthesis and study of intra/extracellular chemical dynamics demonstrating a bold shift of emphasis from the neural system. For example, in \cite{Papadimitriou2013}, \cite{Sarpeshkar2010}, \cite{Papadimitriou2014_1} and \cite{Houssein2015} $cytomorphic/cytomimetic~electronics$ are introduced. The logarithmic behaviour \cite{Frey1993}, \cite{Frey1996}, \cite{Tsividis1997}, \cite{Drakakis1997}, \cite{Drakakis1999_1}, \cite{Drakakis1999_2}, \cite{Frey2009} and \cite{Toumazou1998} of weakly inverted MOS devices is exploited. Log--domain circuits have been shown able to produce a variety of nonlinear dynamics \cite{Serdijn2013} and \cite{Papadimitriou2014_2}. However, the systematic synthesis tool (NBCF) presented in \cite{Papadimitriou2013} and \cite{Papadimitriou2014_1}, though directly applicable for cellular dynamics with strictly positive variables, does not level itself naturally to the realization of bilateral dynamical systems such as neuronal dynamics \cite{Papadimitriou2014_2}.

\par As a solution, this paper presents a novel current--input current--output circuit leading to a systematic synthesis methodology of bilateral dynamical systems onto low--power log--domain circuits. To the best of our knowledge, this is the first systematic log-domain circuit capable of emulating such nonlinear bilateral dynamical systems. The application of the method is verified by synthesizing four different case studies and transistor--level simulations confirm that the resulting circuits are in good agreement with their mathematical counterparts.

\section{The NBDS Circuit}
As mentioned before, the previous attempts in designing log--domain circuits capable of emulating nonlinear dynamics entail state variables that are only strictly positive such as intracellular concentrations of substances, genes and etc. Nevertheless, there exist numerous biological systems that are not limited to such dynamics with the prime example being the area of neuronal dynamics or various biological rhythms. The state variables in such systems could represent for example membrane potentials, a quantity that can possess positive and negative values. In the FHN neuron model \cite{FitzHugh1961} with the following representation: $\dot{v}=v-\frac{v^3}{3}-w+I_{ext}$ and $\dot{w}=0.18(v+0.7-0.8w)$ describing the membrane potential's and the recovery variable's velocity, the state variables in the absence of input stimulations remain at $(v, w)\approx(-1.2, -0.6)$, while these values go up to $(v, w)\approx(2, 1.7)$ in the presence of input stimulations.

\begin{figure*}[t]
\normalsize
\centering
\includegraphics[trim = 0in 0in 0in 0in, clip, height=2.4in]{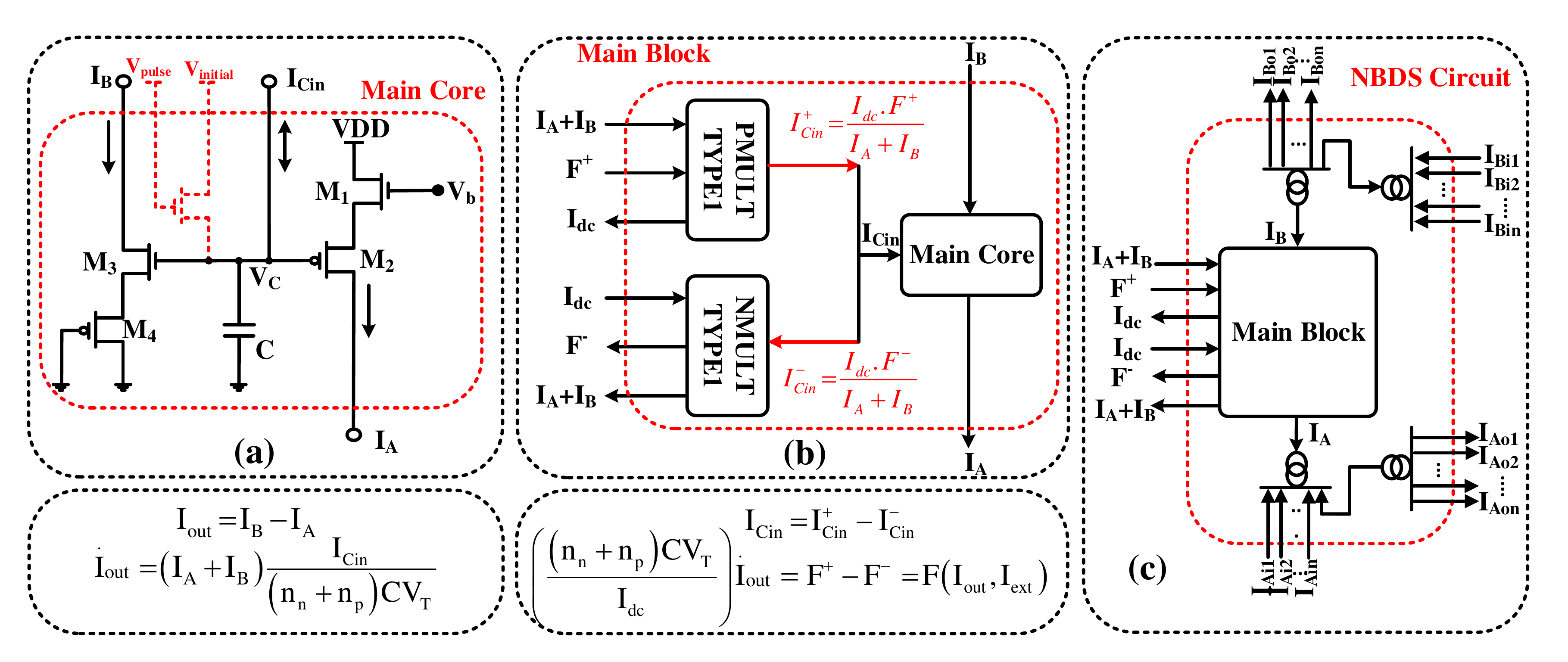}
\vspace{-5pt}
\captionsetup{font=footnotesize}
\caption{Hierarchical representations of the proposed low--power log--domain circuit along with corresponding input/output equations. (a) The ``main core" including the initialization circuit highlighted with red color. (b) The ``main block" including the main core and two current--mode PMOS and NMOS multipliers. The PMULT (NMULT) sources (sinks) the output current and has one sink (source) and two source (sink) inputs (see Section III for further explanations). (c) The NBDS circuit including the ``main block" with several copied currents (the current mirrors are represented with double circle symbols). $F^+$ and $F^-$ respectively refer to the positive and negative parts of $F(I_{out},I_{ext})$ explained in (\ref{eq:5_15}).}
\vspace{-15pt}
\label{fig:5_1}
\end{figure*}

\par Although bilateral dynamical systems may be mapped to strictly positive systems by shifting up the state variables, such an approach leads to inefficient hardware realizations. For example, consider the case of the system $\dot x=x^3$. By shifting up the state variable to $X=x+\alpha$, (where $\alpha$ is positive and constant in time), the mapped dynamical system becomes $\dot X=(X-\alpha)^3=X^3-\alpha^3-3X^2\alpha+3X\alpha^2$ leading to a quite complicated system. Besides, to realize such mappings, in the case of topologies where the stated variables are represented by means of currents, constant currents need to be injected to shift the state variables up and eventually drawn to map them back, resulting in higher power consumption.

\par The reason why the previous electrical realizations have not been yet able to emulate bilateral dynamics without such mappings might stem from the inherent nature of the transistor (BJT or MOSFET). For example, the MOSFET transistors, due to their structures, are only able to conduct current towards a single direction, from drain to source. This structure is helpful for implementing class-AB linear dynamical systems by processing positive and negative components of the input/state variables separately resulting in two capacitors per each state variable of the prototype system \cite{Frey1994}, \cite{Katsiamis2009}, \cite{Yang2015}, \cite{Houssein2016}. However according to the superposition principle, such a procedure, in general, cannot be applied to nonlinear systems, therefore, it has been challenging to realize nonlinear systems that contain bidirectional state variables.

\par This limitation is resolved in the proposed NBDS circuit comprising a single capacitor per each state variable. In the following, the structure and corresponding mathematical formulations of the proposed NBDS circuit are explained in detail.

\subsection{Mathematical Framework}
Figure \ref{fig:5_1} shows hierarchical representations of the proposed nonlinear bilateral dynamical system. Here, we first introduce the main core of the proposed circuit in Figure \ref{fig:5_1}(a) as a fundamental transistor-level element in computing the nonlinear dynamics. It is formed by connecting a grounded capacitor of value $C$ at the gate of two PMOS and NMOS transistors ($M_2$ and $M_3$ in Figure \ref{fig:5_1}(a)) each followed by an NMOS ($M_1$) and PMOS ($M_4$) transistor, respectively. The initialization circuit is shown in dotted red color. It is used for dynamical systems whose initial values affect the resulting dynamics. The realization of such dynamical systems will be discussed further in Section IV.
\par As we know, the current relation of an NMOS and PMOS device operating in subthreshold saturation when $\lvert V_{DS}\rvert>4V_T$ is described by the following equations \cite{Andreou1996}:
\begin{equation}\label{eq:5_1}
I_{D_n}=(\frac{W}{L})_nI_{D_{0_n}}exp(\frac{V_{GS}}{n_nV_T})
\end{equation}

\begin{equation}\label{eq:5_2}
I_{D_p}=(\frac{W}{L})_pI_{D_{0_p}}exp(\frac{V_{SG}}{n_pV_T})
\end{equation}
where $n_n$ and $n_p$ are the subthreshold process-dependent slope factors for NMOS and PMOS transistors, respectively; $V_T$ denotes the thermal voltage ($\approx 26mV$ at $300K$), $I_{D_{0_n}}$ and $I_{D_{0_p}}$ are the leakage currents of the NMOS and PMOS transistors, respectively and $W$, $L$ are the width and length of the devices, respectively.
\par Setting $I_{S_n}=(\frac{W}{L})_nI_{D_{0_n}}$ and $I_{S_p}=(\frac{W}{L})_pI_{D_{0_p}}$ in (\ref{eq:5_1}) and (\ref{eq:5_2}) and differentiating with respect to time, the current expression for $I_A$ (see Figure \ref{fig:5_1}(a)) yields:
\begin{equation}\label{eq:5_3}
\dot{I}_A=\overbrace{(I_{S_n}\cdot exp(\frac{V_{GS_1}}{n_nV_T}))}^{I_A}(\frac{\dot{V}_{GS_1}}{n_nV_T})
\end{equation}

\begin{equation}\label{eq:5_4}
\dot{I}_A=\overbrace{(I_{S_p}\cdot exp(\frac{V_{SG_2}}{n_pV_T}))}^{I_A}(\frac{\dot{V}_{SG_2}}{n_pV_T})
\end{equation}
obviously (\ref{eq:5_3}) and (\ref{eq:5_4}) are equal, therefore:
\begin{equation}\label{eq:5_5}
\dot{V}_{SG_2}=\frac{n_p}{n_n}\dot{V}_{GS_1}=\alpha \dot{V}_{GS_1}
\end{equation}
where $\alpha$ is the ratio of the subthreshold slope factors. Similarly, we can derive the following equation for transistors $M_3$ and $M_4$:
\begin{equation}\label{eq:5_6}
\dot{V}_{SG_4}=\frac{n_p}{n_n}\dot{V}_{GS_3}=\alpha \dot{V}_{GS_3}.
\end{equation}

 On the other hand, the application of Kirchhoff's Voltage Law (KVL) and the derivative function describe the following relations for the capacitor voltage $V_C$ when the voltage $V_b$ is constant (see Figure \ref{fig:5_1}(a)):
 \begin{equation}\label{eq:5_7}
\dot{V}_{C}=-(\dot{V}_{GS_1}+\dot{V}_{SG_2})
\end{equation}

  \begin{equation}\label{eq:5_8}
\dot{V}_{C}=+(\dot{V}_{GS_3}+\dot{V}_{SG_4})
\end{equation}
and substituting (\ref{eq:5_5}) and (\ref{eq:5_6}) into (\ref{eq:5_7}) and (\ref{eq:5_8}) respectively leads to:
 \begin{equation}\label{eq:5_9}
\dot{V}_{C}=-\dot{V}_{GS_1}\cdot(1+\alpha)
\end{equation}

\begin{equation}\label{eq:5_10}
\dot{V}_{C}=+\dot{V}_{GS_3}\cdot(1+\alpha).
\end{equation}

By setting the current $I_{out}=I_B-I_A$ in Figure \ref{fig:5_1}(a) as the state variable of our system and using (\ref{eq:5_3}) and the corresponding equation for $I_B$, the following relation is derived:
\begin{equation}\label{eq:5_11}
\dot{I}_{out}=\dot{I}_B-\dot{I}_A=I_B(\frac{\dot{V}_{GS_3}}{n_nV_T})-I_A(\frac{\dot{V}_{GS_1}}{n_nV_T})
\end{equation}
by substituting (\ref{eq:5_9}) and (\ref{eq:5_10}) in (\ref{eq:5_11}):
\begin{equation}\label{eq:5_12}
\dot{I}_{out}=(I_A+I_B)\cdot \frac{\dot{V}_C}{(1+\alpha)n_nV_T}.
\end{equation}
\par Bearing in mind that the capacitor current $I_{Cin}$ can be expressed as $C\dot{V}_C$, relation (\ref{eq:5_12}) yields:
\begin{equation}\label{eq:5_13}
\dot{I}_{out}=(I_A+I_B)\cdot \frac{I_{Cin}}{(1+\alpha)n_nCV_T}.
\end{equation}
\par One can show that:
\begin{equation}\label{eq:5_14}
\frac{(1+\alpha)n_nCV_T}{I_{dc}}\dot{I}_{out}=\frac{(I_A+I_B)}{I_{dc}}\cdot I_{Cin}.
\end{equation}
\par Equation (\ref{eq:5_14}) is the main core's constitutive relation where $\alpha=\frac{n_p}{n_n}$. In order for a mathematical dynamical system with the following general form to be mapped to (\ref{eq:5_14}):
\begin{equation}\label{eq:5_15}
\tau\dot{I}_{out} =F(I_{out}, I_{ext})
\end{equation}
where $I_{ext}$ and $I_{out}$ are the external and state variable currents, the quantities $\frac{C}{I_{dc}}$ and $I_{Cin}$ must be respectively equal to $\frac{\tau}{(n_n+n_p)V_T}$ and $\frac{F(I_{out}, I_{ext})I_{dc}}{(I_A+I_B)}$. Note that the ratio value $\frac{C}{I_{dc}}$ can be satisfied with different individual values for $C$ and $I_{dc}$. These values should be chosen appropriately according to practical considerations (see Section V.G). Since $F$ is a bilateral function, in general, it will hold:
\begin{equation}\label{eq:5_16}
I_{Cin}=\overbrace{\frac{F^+(I_A,I_B,I_{ext}^+,I_{ext}^-)I_{dc}}{(I_A+I_B)}}^{I_{Cin}^+}-\overbrace{\frac{F^-(I_A,I_B,I_{ext}^+,I_{ext}^-)I_{dc}}{(I_A+I_B)}}^{I_{Cin}^-}
\end{equation}
where $I_{Cin}^+$ and $I_{Cin}^-$ are calculated respectively by a PMOS and NMOS multiplier (see Figure \ref{fig:5_1}(b)) and $I_{ext}$ is separated to + and -- signals by means of splitter blocks. Note that $I_{dc}$ is a scaling dc current and $\tau$ has dimensions of $second(s)$. Since $I_{Cin}$ can be a complicated nonlinear function in dynamical systems, we need to provide copies of $I_{out}$ or equivalently of $I_A$ and $I_B$ to simplify the systematic computation at the circuit level. Therefore, the higher hierarchical block shown in Figure \ref{fig:5_1}(c) is defined as the NBDS circuit (see Figure \ref{fig:5_1}(c)) including the main block and associated current mirrors. The form of (\ref{eq:5_15}) is extracted for a 1--D dynamical system and can be extended to $N$ dimensions in a straightforward manner as follows:
\begin{equation}\label{eq:5_17}
\tau_N\dot{I}_{out_N} =F_N(\bar{I}_{out},\bar{I}_{ext})
\end{equation}
where $\frac{C_N}{I_{dc_N}}=\frac{\tau_N}{(n_n+n_p)V_T}$ and $I_{Cin_N}=\frac{F_N(\bar{I}_{out},~\bar{I}_{ext})I_{dc_N}}{(I_{A_N}+I_{B_N})}$.

\begin{figure}[t]
\centering
\includegraphics[trim = 0.25in 0.25in 0.15in 0.25in, clip, width=4in]{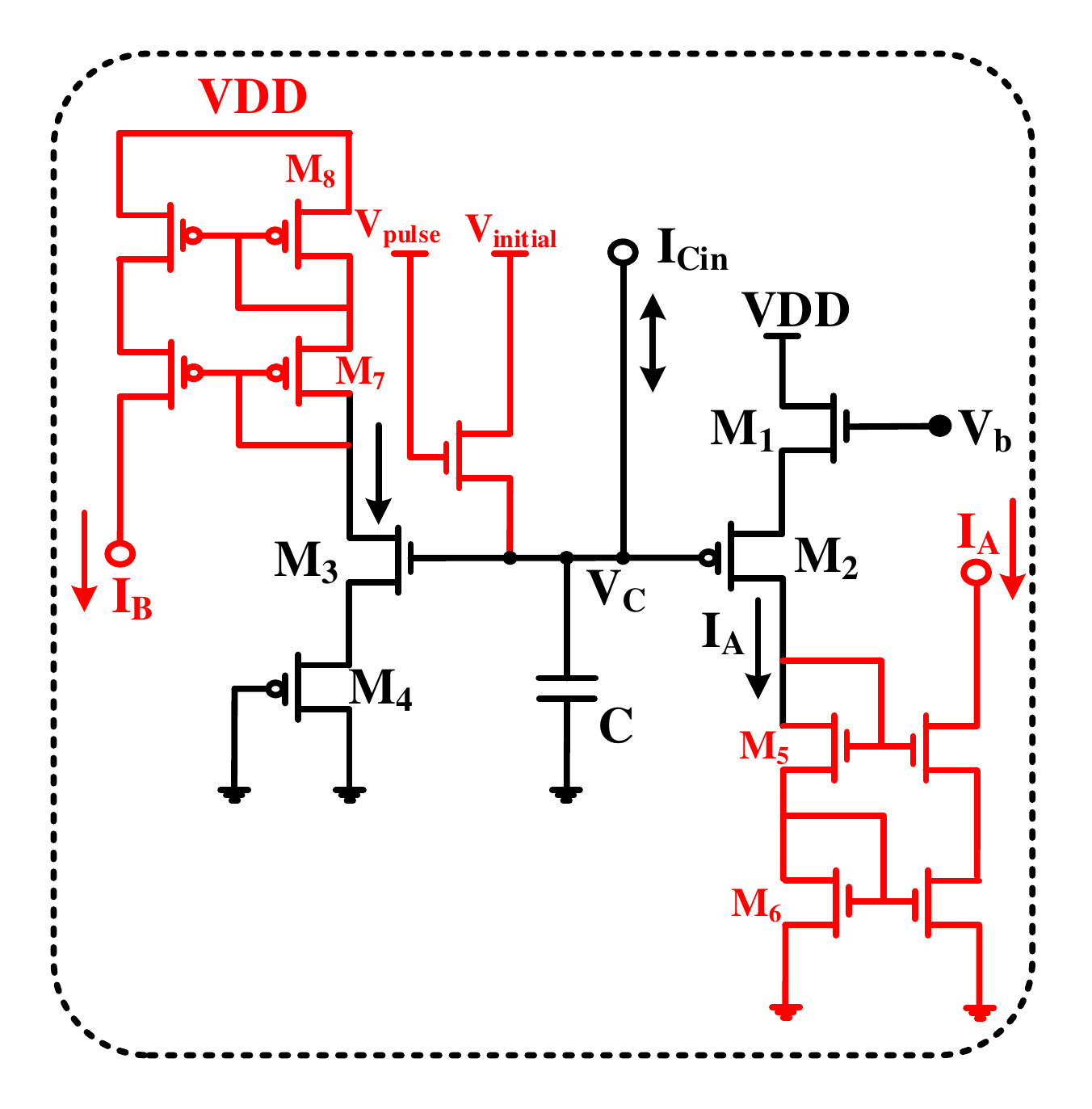}
\captionsetup{font=footnotesize}
\caption{The main core circuit including additional current mirrors.}
\vspace{-10pt}
\label{fig:5_2}
\end{figure}

 \subsection{The Bias Voltage $V_b$}
In the proposed circuit, the bias voltage ($V_b$) regulates the dynamic range of the output signal ($I_{out}$) as well as the circuit's power. By setting a proper value of $V_b$ in the circuit for a certain output dynamic range, an optimum design can be delivered. In the following, we explain how $V_b$ can affect the output dynamic range and consequently the value of power consumption. To this end, we show that the larger $V_b$, the higher negative dynamic range and the power consumption. However, according to the desirable dynamic range, $V_b$ is saturated after a certain value and further increase just leads to consuming extra power.
\par One can show that when the transistors $M_1$--$M_4$ operate in subthreshold saturation, $I_{out}$ is derived as (see Appendix A):
\begin{equation}\label{eq:5_18}
I_{out}=I_{B}-I_{A}=\beta [exp(\frac{V_C}{(n_n+n_p)V_T})-exp(\frac{V_b-V_C}{(n_n+n_p)V_T})]
\end{equation}
where $\beta=\sqrt{I_{S_n}I_{S_p}}exp(\frac{n_n-n_p}{2(n_n+n_p)}ln\frac{I_{S_n}}{I_{S_p}})$. Clearly, by setting $V_C=\frac{V_b}{2}$, $I_{out}$ will be equal to zero and by decreasing $V_C$, $I_{out}$ decreases and vice versa.
\par As shown in Figure \ref{fig:5_2}, by considering the cascoded current mirrors, the minimum $V_C$ that holds the transistors in saturation is determined by transistors $M_2$ and $M_4$. The minimum $V_C$ that holds $M_2$ in saturation is equal to (see Appendix B):
\begin{equation}\label{eq:5_19}
V_{C_{min_{M_2}}}=\frac{1+\alpha}{3}4V_T+\frac{2-\alpha}{3}V_b+n_pV_Tln\frac{I_{S_p}}{I_{S_n}}
\end{equation}
where $\alpha$ is equal to $\frac{n_p}{n_n}$. Similarly, one can show that the minimum $V_C$ that holds $M_4$ in saturation is equal to:
\begin{equation}\label{eq:5_20}
V_{C_{min_{M_4}}}=\frac{1+\alpha}{\alpha}4V_T+n_nV_Tln\frac{I_{S_p}}{I_{S_n}}.
\end{equation}
\par By regulating $(\frac{W}{L})_n$ and $(\frac{W}{L})_p$ so that $I_{S_n}\approx I_{S_p}$ and by assuming $n_n\approx n_p$, (\ref{eq:5_19}) and (\ref{eq:5_20}) are respectively simplified as:
\begin{equation}\label{eq:5_21}
V_{C_{min_{M_2}}}=\frac{8}{3}V_T+\frac{1}{3}V_b
\end{equation}
\begin{equation}\label{eq:5_22}
V_{C_{min_{M_4}}}=8V_T.
\end{equation}
\par As can be seen in the above equations if $V_b>16V_T$, by decreasing $V_C$, $M_2$ enters to triode before $M_4$. Since in the case studies that we will synthesize, the scale of circuit's current is $nA$, it is most likely $V_b>16V_T$. Therefore $M_2$ is dominant and (\ref{eq:5_19}) is considered as the minimum $V_C$ in the rest of this subsection. The corresponding $I_{out}$ can be obtain as:
\begin{equation}\label{eq:5_23}
I_{out_{min}}=\beta[\gamma exp(\frac{2-\alpha}{3(n_n+n_p)}V_b)-\frac{1}{\gamma}exp(\frac{1+\alpha}{3(n_n+n_p)}V_b)]
\end{equation}
where $\gamma=exp(\frac{4}{3}(1+\alpha)+n_pln\frac{I_{S_p}}{I_{S_n}})$. Since the second term in (\ref{eq:5_23}) carries more weight compared to the left one, by increasing $V_b$, the lower bound (the negative bound) of the output dynamic range increases. Thus, for the dynamical systems whose negative side of the output dynamic range is wider, we must increase $V_b$ to extract more current from $M_1$ and $M_2$ while holding the transistors in saturation.

\begin{table}[t]
\captionsetup{font=footnotesize}
\caption{Main Electrical Parameter Values.}   
\centering          
\begin{tabular}{c c}    
\hline\hline                        
Specifications & Value\\ [0.5ex]  
\hline                      
$\alpha$& $\frac{n_p}{n_n}$\\
$\beta$& $\sqrt{I_{S_n}I_{S_p}}exp(\frac{n_n-n_p}{2(n_n+n_p)}ln\frac{I_{S_n}}{I_{S_p}})$\\
$\zeta$& $\frac{1}{2}(\frac{I_{out_{init}}}{\beta}+\sqrt{\frac{I_{out_{init}}^2}{\beta^2}+4exp(\frac{V_b}{(n_n+n_p)V_T})})$\\
$\theta$&$\frac{V_C}{(n_n+n_p)V_T}$\\
$\gamma$& $exp(\frac{4}{3}(1+\alpha)+n_pln\frac{I_{S_p}}{I_{S_n}})$\\
$V_{C_{max}}$& $\frac{n_n+n_p}{3n_p}(V_{DD}-4V_T)+n_nV_Tln\frac{I_{S_p}}{I_{S_n}}$\\
$V_{C_{min}}$& $\frac{1+\alpha}{3}4V_T+\frac{2-\alpha}{3}V_b+n_pV_Tln\frac{I_{S_p}}{I_{S_n}}$\\
\hline          
\end{tabular}
\label{table:5_1}    
\end{table}

\begin{figure*}[t]
\normalsize
\centering
\includegraphics[trim = 0in 0in 0in 0in, clip, width=5.5in]{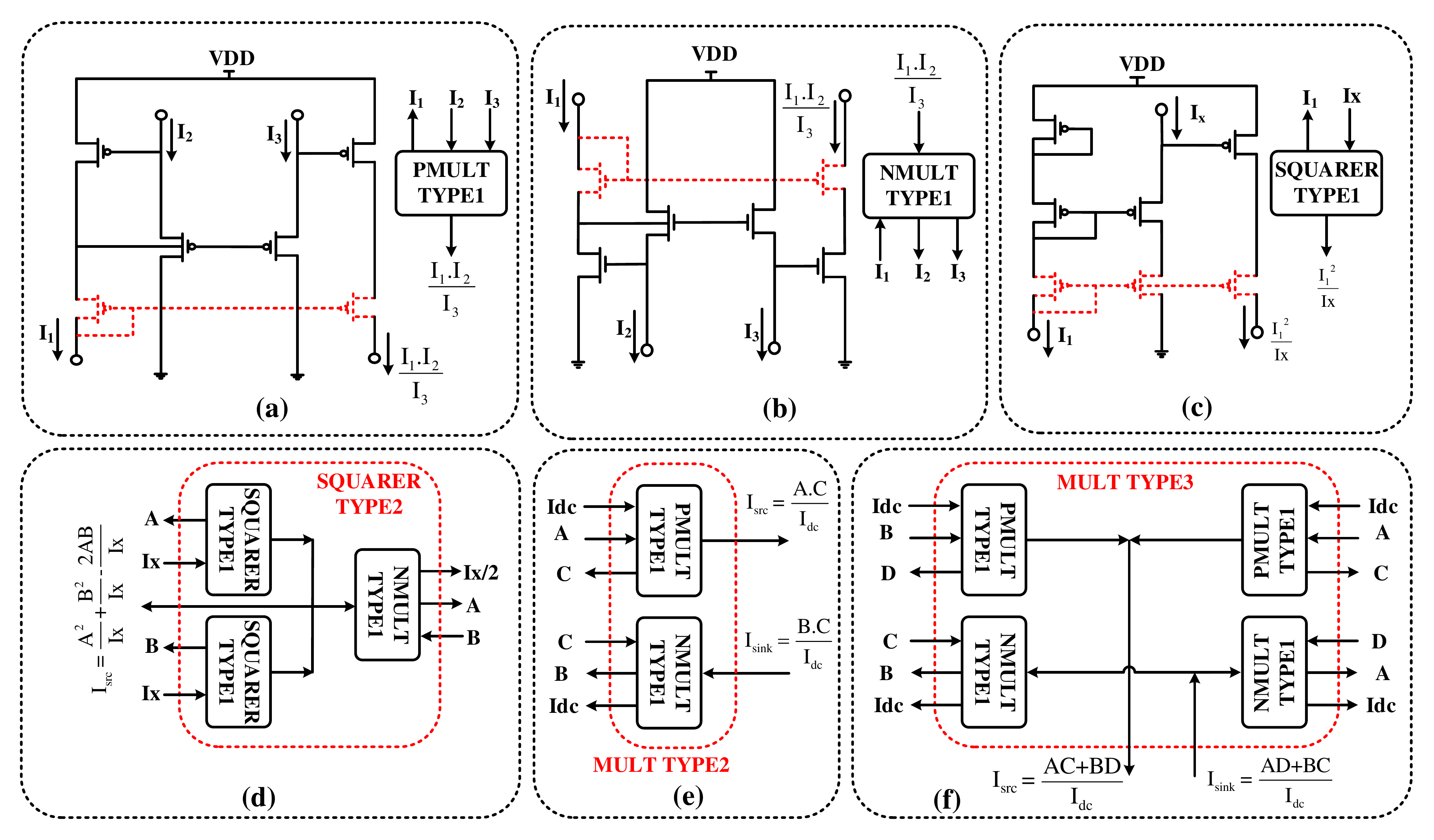}
\vspace{-5pt}
\captionsetup{font=footnotesize}
\caption{Schematic and symbolic representation of the basic TL blocks. (a) The PMULT TYPE1. (b) The NMULT TYPE1. (c) The SQUARER TYPE1. (d) The SQUARER TYPE2.
 (e) The MULT TYPE2. (f) The MULT TYPE3.}
\vspace{-15pt}
\label{fig:5_3}
\end{figure*}

\par As mentioned before, by increasing $V_C$ towards $V_b$, $I_{out}$ increases. When it reaches $V_b$, $I_A$ will be zero leading to:
\begin{equation}\label{eq:5_24}
I_{out}=I_B=\beta exp(\frac{V_C}{(n_n+n_p)V_T}).
\end{equation}
\par On the other hand, the output current can increase up to a certain point in which the $M_3$ and $M_4$ still operate in subthreshold saturation. The capacitor voltage at this point is referred to as $V_{C_{max}}$ and similar to the $V_{C_{min}}$ case, it can be calculated as:
\begin{equation}\label{eq:5_25}
V_{C_{max}}=\frac{n_n+n_p}{3n_p}(V_{DD}-4V_T)+n_nV_Tln\frac{I_{S_p}}{I_{S_n}}.
\end{equation}

\par In those designs where $V_b<V_{C_{max}}$, $V_b$ cannot affect the upper bound of the dynamic range, however if $V_b>V_{C_{max}}$, by increasing $V_b$, the positive dynamic range decreases.
\par Besides, the increase of $V_b$ results in higher power consumption for the circuit. From (\ref{eq:5_46}) and (\ref{eq:5_47}) (see Appendix A), one can show that:
\begin{equation}\label{eq:5_26}
I_A\cdot I_B=\beta^2 exp(\frac{V_b}{(n_n+n_p)V_T}).
\end{equation}
\par It is clear that by increasing $V_b$, the product $I_A\cdot I_B$ increases. On the other hand, the circuit operates properly for a range of $V_b$ implying this fact that by increasing $V_b$, $I_{out}=I_B-I_A$ remains almost unchanged. Therefore, by increasing $V_b$, both $I_A$ and $I_B$ must almost equally increase leading to higher power consumption. Thus in conclusion, for an optimum design, the minimum $V_b$ value, needed to cover the output dynamic range must be selected. The main electrical parameter values are shown in Table \ref{table:5_1}.

 \subsection{Initial Values}
Here, we explain the importance of initial values in a specific group of dynamical systems and how it can be mapped onto the NBDS circuit. As a common example of such dynamical systems, we refer to Hopf oscillator including the Hopf bifurcation in which a limit cycle is given birth from an equilibrium point \cite{Strogatz2014}. The bifurcation can be supercritical or subcritical resulting in a stable or unstable limit cycle, respectively. Here, we focus on the subcritical case in which two coexisting attractors separated by an unstable limit cycle cause bistablilty in the system. The evolution of the state variable in such systems depends on which attraction domain the initial condition is placed in initially. Moreover, sufficiently strong perturbations can change it from one state to another with the unstable limit cycle playing the role of the threshold \cite{Izhi2006}. However, in order to cope with the bistability phenomenon, an additional NMOS transistor is employed within the main core of the proposed circuit shown in Figure \ref{fig:5_1}(a). During the initialization process, the transistor highlighted with red is triggered by a short external pulse to pull up the capacitor voltage instantaneously to an acceptable initial value.
\begin{equation}\label{eq:5_27}
I_{out_{init}}=\beta [exp(\theta)-exp(\frac{V_b}{(n_n+n_p)V_T}) exp(-\theta)]
\end{equation}
where $\theta=\frac{V_C}{(n_n+n_p)V_T}$. By substituting $\zeta=exp(\theta)$ in (\ref{eq:5_27}), the following second order polynomial equation is derived:
\begin{equation}\label{eq:5_28}
\zeta^2-\frac{I_{out_{init}}}{\beta}\zeta-exp(\frac{V_b}{(n_n+n_p)V_T})=0.
\end{equation}
\par Solving (\ref{eq:5_28}) and bearing in mind that only the positive solution is acceptable we have:
\begin{equation}\label{eq:5_29}
\zeta=\frac{1}{2}(\frac{I_{out_{init}}}{\beta}+\sqrt{\frac{I_{out_{init}}^2}{\beta^2}+4exp(\frac{V_b}{(n_n+n_p)V_T})})
\end{equation}
given that $\theta=ln(\zeta)$ and (\ref{eq:5_29}), $V_C=\theta(n_n+n_p)V_T$ yields:
\begin{equation}\label{eq:5_30}
V_C=(n_n+n_p)V_Tln[\frac{1}{2}(\frac{I_{out_{init}}}{\beta}+\sqrt{\frac{I_{out_{init}}^2}{\beta^2}+4exp(\frac{V_b}{(n_n+n_p)V_T})})].
\end{equation}
\par Therefore, according to the CMOS process parameters, the bias voltage ($V_b$) and the desirable initial output current ($I_{out_{init}}$), $V_{initial}$ (see Figure \ref{fig:5_1}(a)) can be set to $V_C$ in (\ref{eq:5_30}).

\begin{figure*}[t]
\vspace{-20pt}
\normalsize
\centering
\includegraphics[trim = 0.1in 0.1in 0.1in 0.1in, clip, width=5in]{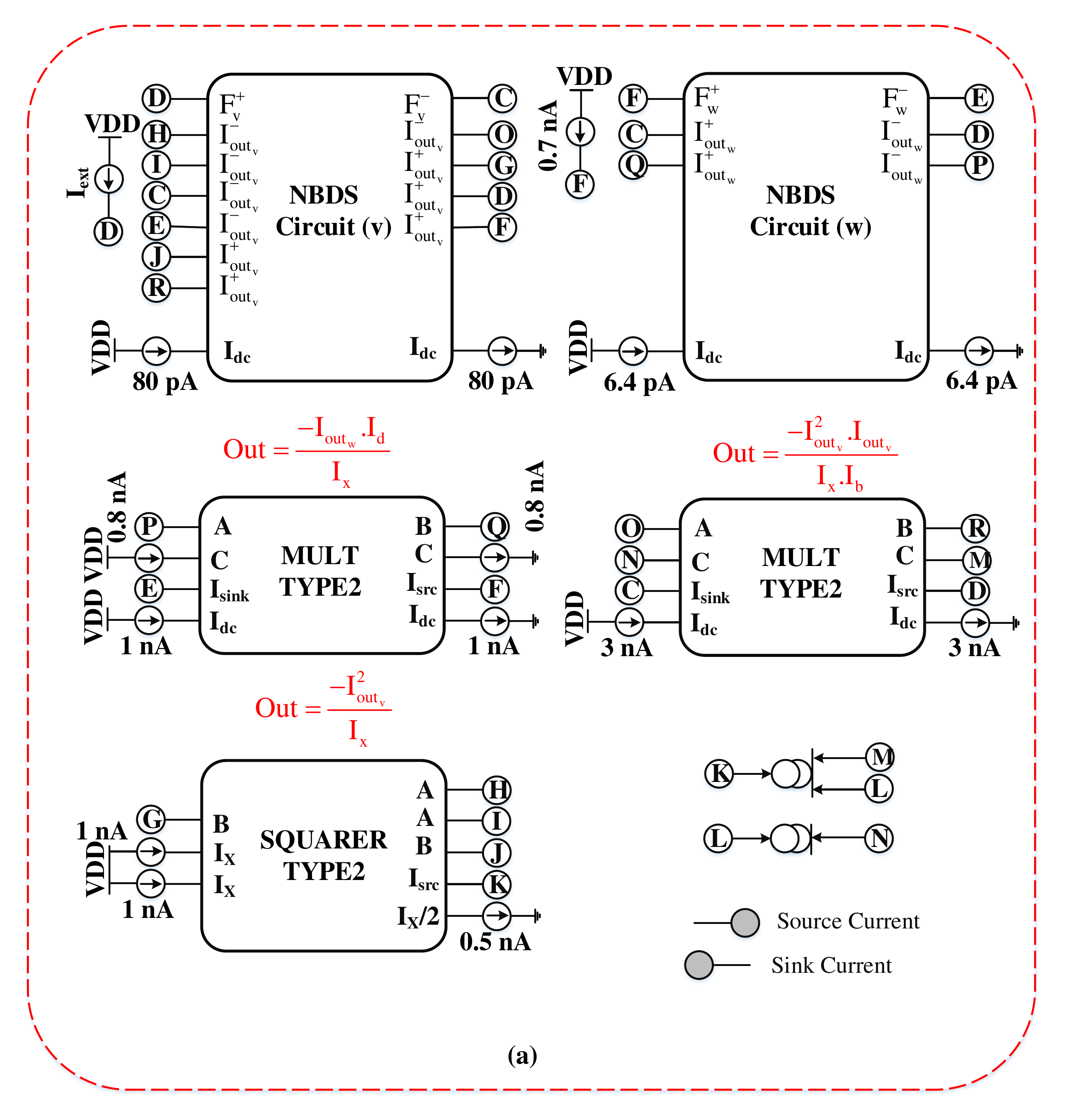}
\vspace{-5pt}
\captionsetup{font=footnotesize}
\caption{A block representation of the total circuit implementing the 2--D FHN neuron model where $I_{out_v}=I_{out_v}^+-I_{out_v}^-$, $I_{out_w}=I_{out_w}^+-I_{out_w}^-$. To preserve consistency compared to Fig.~1(c), it should be noted that for example, $I_{out_v}^+(source)=I_{Bo}$, $I_{out_v}^+(sink)=I_{Bi}$, $I_{out_v}^-(source)=I_{Ao}$ and $I_{out_v}^-(sink)=I_{Ai}$.}
\vspace{-15pt}
\label{fig:5_4}
\end{figure*}

\begin{figure*}[t]
\normalsize
\centering
\includegraphics[trim = 0.1in 0.1in 0.1in 0.1in, clip, width=5in]{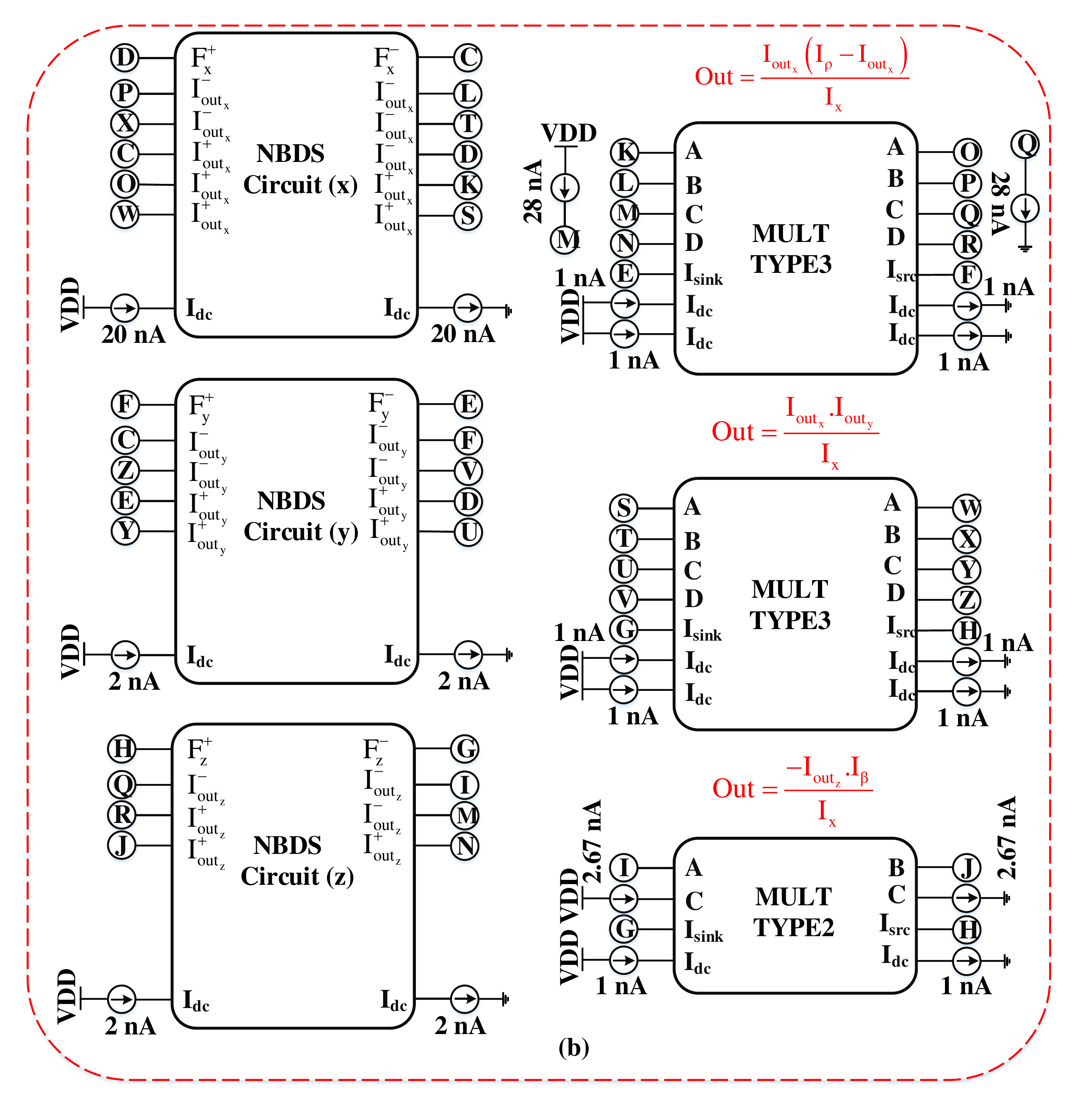}
\vspace{-5pt}
\captionsetup{font=footnotesize}
\caption{A block representation of the total circuit implementing the 3--D Lorenz attractor where $I_{out_x}=I_{out_x}^+-I_{out_x}^-$, $I_{out_y}=I_{out_y}^+-I_{out_y}^-$ and $I_{out_z}=I_{out_z}^+-I_{out_z}^-$.}
\vspace{-15pt}
\label{fig:5_5}
\end{figure*}

\section{Electrical Circuit Blocks}
To explore the applicability of the proposed NBDS circuit in real--world case studies, we need to implement static operations leading to meaningful time--domain dynamics. Although there are various mathematical models for nonlinear dynamical systems including different static functions, most of them can be viewed as a combination of simple basic blocks such as multipliers, dividers and squarers. These mathematical operations can be also implemented using different TL network realizations. For the sake of simplicity, in this work we use the $stacked~loop~topology$, however, it should be stressed that regardless of the TL structure chosen to generate the mathematical operations, the NBDS circuit will be held. To preserve the systematic nature of the proposed framework in this paper, the following TL blocks are used along with the NBDS circuit for the implementation of the case studies.

$\textbf{The multiplier block (type1)}$: This block is able to perform current mode multiplication or division operations on single--sided input signals. By setting the same $\frac{W}{L}$ aspect ratio, the governing TL principle for this block becomes: $I_{OUT}=\frac{I_1\cdot I_2}{I_3}$. To satisfy the conditions demanding sink or source currents from the output of such blocks, here we present two NMOS and PMOS based multipliers demonstrated in Figure \ref{fig:5_3}(a-b). The cascoded topology is highlighted in the figure with red color and employed to minimize output current errors.

$\textbf{The squarer block (type1)}$: This block produces the squaring operation of a single--sided input current over a scaling current $I_X$ (see Figure \ref{fig:5_3}(c)). Again the cascoded topology is highlighted in the figure with red colour.

$\textbf{The squarer block (type2)}$: This block contains two SQUARER TYPE1 and one NMULT TYPE1 realizing the square of bilateral input signals (see Figure \ref{fig:5_3}(d)). Obviously, if we consider the input signal as $I_{in}=A-B$, the output signal is strictly positive and described by $I_{OUT}=\frac{A^2}{I_X}+\frac{B^2}{I_X}-\frac{2AB}{I_X}$.

$\textbf{The multiplier block (type2)}$: This block is made of two MULT TYPE1 producing the multiplication operation of a single--sided ($C$) and a bilateral ($A-B$) input signal (see Figure \ref{fig:5_3}(e)). The output signal is consequently bilateral and represented by $I_{OUT}=\frac{A\cdot C}{I_X}-\frac{B\cdot C}{I_X}$ difference between two strictly positive currents.

$\textbf{The multiplier block (type3)}$: This block contains four MULT TYPE1 performing multiplication operation on bilateral input signals (see Figure \ref{fig:5_3}(f)). The output signal is bilateral and difference between two strictly positive currents expressed by $I_{OUT}=\frac{AC+BD}{I_{dc}}-\frac{AD+BC}{I_{dc}}$.

 \section{Systematic Synthesis}
The systematic synthesis procedure provides the flexibility and convenience required for the realization of nonlinear dynamical systems by computing their time-dependent dynamical behaviour. In this section, we showcase the methodology through which we systematically map the mathematical dynamical models onto the proposed electrical circuit. To this end, we apply the proposed systematic synthesis on four nonlinear and fairly complex bilateral dynamical systems.
 \subsection{FitzHugh-Nagumo (FHN)}
The FHN is a 2-D neuron model originally from the Hodgkin-Huxley (HH) \cite{FitzHugh1961}. This model is considered to be a relatively complicated dynamical system due to the third power factor (significantly non-linear) in its equations and codified as follows:
\begin{equation}\label{eq:5_31}
\begin{cases}
\dot{v}=v-\frac{v^3}{3}-w+I_{ext}\\
\dot{w}=a\cdot(v+0.7-0.8w)
\end{cases}
\end{equation}
where $v$ and $w$ are the membrane potential and the recovery variables, respectively; $I_{ext}$ is the injected stimulus current, and $a$ is an adjustable parameter. The FHN model employs no auxiliary resetting function to reproduce spiking behaviors and consequently such a resetting mechanism is not needed to be considered in the circuit implementation.
\par According to this biological dynamical system, we can start forming the electrical equivalent using (\ref{eq:5_17}):
\begin{equation}\label{eq:5_32}
\begin{cases}
\frac{(n_n+n_p)C V_T\dot{I}_{out_{v}}}{I_{dc_v}}=F_v(I_{out_v},I_{out_w},I_{ext})\\
\frac{(n_n+n_p)C V_T\dot{I}_{out_{w}}}{I_{dc_w}}=F_w(I_{out_v},I_{out_w})
\end{cases}
\end{equation}
where $I_{dc_v}=80pA$, $I_{dc_w}=a\cdot I_{dc_v}=6.4pA$, $F_v$ and $F_w$ are functions given by:
\begin{equation}\label{eq:5_33}
\begin{cases}
F_v(I_{out_v},I_{out_w},I_{ext})=I_{out_v}-\frac{I_{out_v}^3}{I_bI_x}-I_{out_w}+I_{ext}\\
F_w(I_{out_v},I_{out_w})=(I_{out_v}+I_{c}-\frac{I_dI_{out_w}}{I_x})
\end{cases}
\end{equation}
where $I_b=3 nA$, $I_c=0.7 nA$, $I_d=0.8 nA$ and $I_x=1 nA$.
\par Schematic diagrams for the FHN neuron model are seen in Figure \ref{fig:5_4}, including the symbolic representation of the basic TL blocks
introduced in section III. According to these diagrams, it is observed how the mathematical model described in (\ref{eq:5_31}) is mapped onto the proposed electrical circuit. The schematic contains two NBDS circuits implementing the two dynamical variables, followed by two MULT TYPE2, one SQUARER TYPE2 blocks and current mirrors realizing the dynamical functions. As shown in the figure, according to (\ref{eq:5_31}), proper bias currents are selected and the correspondence between the biological voltage and electrical current is $V\iff nA$.

\begin{figure*}[t]
\normalsize
\centering
\includegraphics[trim = 0in 0in 0in 0in, clip, width=6in]{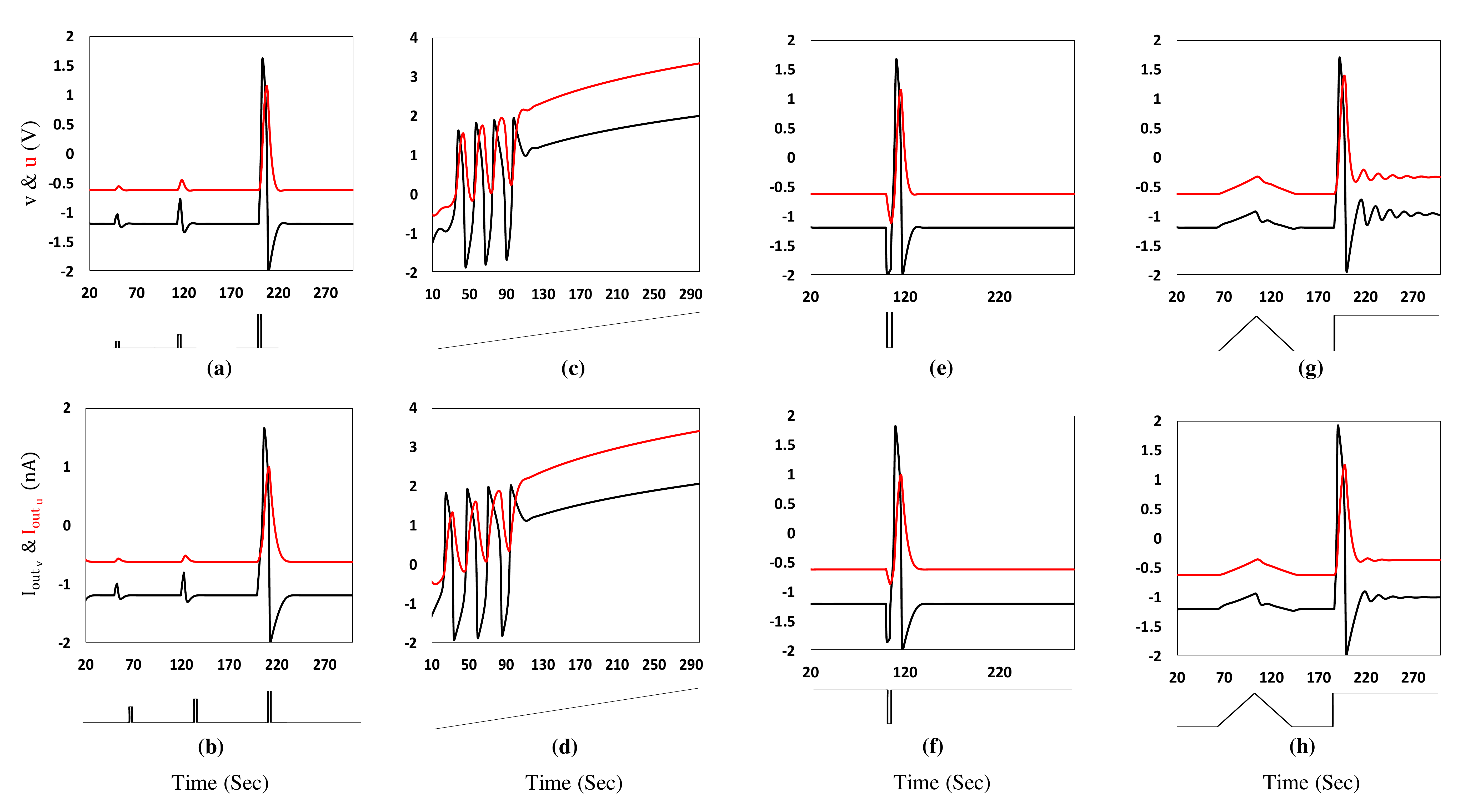}
\vspace{-5pt}
\captionsetup{font=footnotesize}
\caption{Time-domain representations of (a-b) absence of all-or-none spikes phenomenon, (c-d) excitation block phenomenon, (e-f) post-inhibitory rebound spike phenomenon, and (g-h) spike accommodation phenomenon for MATLAB and Cadence respectively.}
\vspace{-15pt}
\label{fig:5_6}
\end{figure*}

 \subsection{Lorenz Attractor}
The Lorenz attractor is an arbitrary dynamical system having chaotic solutions for certain parameter values \cite{lorenz1963}. In particular, the Lorenz attractor is a set of chaotic solutions of the Lorenz system forming a butterfly in phase portrait plots. The system is 3--D and its time--domain signals have a wide dynamic range, which is an important feature of biological signals. The mathematical description of this model is presented as follows:
\begin{equation}\label{eq:5_34}
\begin{cases}
\dot{x}=\sigma(y-x)\\
\dot{y}=x(\rho-z)-y\\
\dot{z}=xy-\beta z
\end{cases}
\end{equation}
where $x$, $y$ and $z$ are state variables and $\sigma$, $\rho$, $\beta$ are parameters.
\par According to this chaotic system, we can represent the electrical equivalent using (\ref{eq:5_16}):
\begin{equation}\label{eq:5_35}
\begin{cases}
\frac{(n_n+n_p)CV_T\dot{I}_{out_{x}}}{I_{dc_x}}=F_x(I_{out_x},I_{out_y},I_{out_z})\\
\frac{(n_n+n_p)CV_T\dot{I}_{out_{y}}}{I_{dc_y}}=F_y(I_{out_x},I_{out_y},I_{out_y})\\
\frac{(n_n+n_p)CV_T\dot{I}_{out_{z}}}{I_{dc_z}}=F_z(I_{out_x},I_{out_y},I_{out_y})
\end{cases}
\end{equation}
where $I_{dc_x}=\sigma\cdot I_{dc_y}=20nA$, $I_{dc_y}=2nA$, $I_{dc_z}=2nA$, $F_x$, $F_y$ and $F_z$ functions are given by:
\begin{equation}\label{eq:5_36}
\begin{cases}
F_x(I_{out_x},I_{out_y},I_{out_z})=I_{out_y}-I_{out_x}\\
F_y(I_{out_x},I_{out_y},I_{out_y})=\frac{I_{out_x}(I_{\rho}-I_{out_z})}{I_x}-I_{out_y}\\
F_z(I_{out_x},I_{out_y},I_{out_y})=\frac{I_{out_x}I_{out_y}}{I_x}-\frac{I_{\beta}I_{out_z}}{I_x}
\end{cases}
\end{equation}
where $I_\rho=28 nA$, $I_\beta=\frac{8}{3} nA$ and $I_x=1 nA$.
\par Schematic diagrams for the Lorenz attractor model are seen in Figure \ref{fig:5_5}. The schematic contains three NBDS circuits implementing the three dynamical variables, followed by two MULT TYPE3 and one MULT TYPE2 blocks realizing the dynamical functions. As shown in the figure, in accordance with (\ref{eq:5_34}), proper bias currents are selected.

\subsection{Hopf Oscillator}
In this case study, by means of the proposed systematic synthesis method we first explain how to realize the Hopf oscillator dynamical system categorized as a complex system and then in the next section, we show how to cope with its bistability phenomenon. The mathematical description of the Hopf oscillator is presented as follows:
\begin{equation}\label{eq:5_37}
\begin{cases}
\dot{x}=-y+x(x^2+y^2+\mu)(1-x^2-y^2)\\
\dot{y}=x+y(x^2+y^2+\mu)(1-x^2-y^2)
\end{cases}
\end{equation}
where $\mu$ is called as the bifurcation parameter. When setting $\mu<0$, the subcritical Hopf bifurcation is observed in (\ref{eq:5_37}).

\begin{figure*}[t]
\normalsize
\centering
\includegraphics[trim = 0in 0in 0in 0in, clip, width=6in]{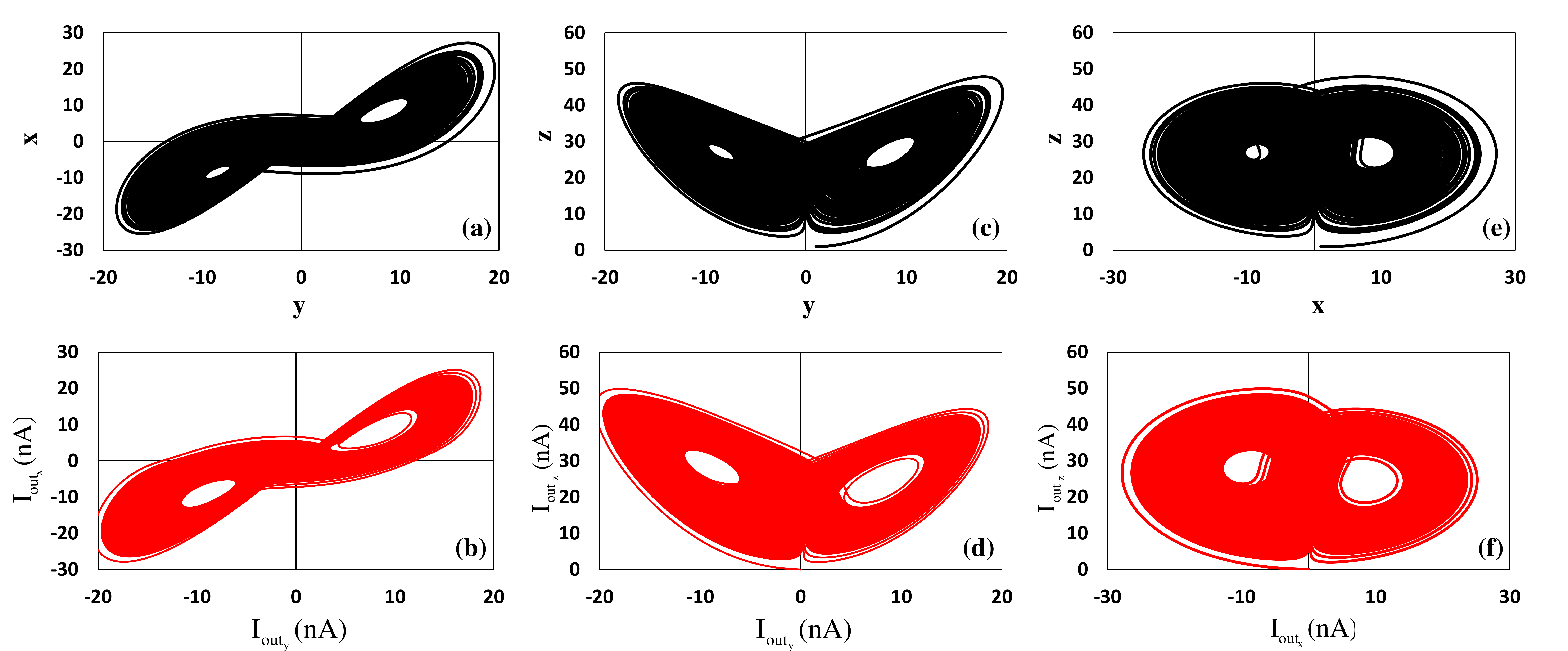}
\captionsetup{font=footnotesize}
\caption{The projections of (a-b) x--y plane, (c-d) z--y plane (butterfly--wings) and (e-f) z--x plane for MATLAB and Cadence, respectively.}
\vspace{-10pt}
\label{fig:5_7}
\end{figure*}

According to the mathematical description, we can demonstrate the electrical equivalent using (\ref{eq:5_17}):
\begin{equation}\label{eq:5_38}
\begin{cases}
\frac{(n_n+n_p)CV_T\dot{I}_{out_{x}}}{I_{dc_x}}=F_x(I_{out_x},I_{out_y})\\
\frac{(n_n+n_p)CV_T\dot{I}_{out_{y}}}{I_{dc_y}}=F_y(I_{out_x},I_{out_y})
\end{cases}
\end{equation}
where $I_{dc_x}=I_{dc_y}=0.5nA$, $F_x$ and $F_y$ functions are presented as follows:
\begin{equation}\label{eq:5_39}
\begin{cases}
F_x(I_{out_x},I_{out_y})=-I_{out_y}+\frac{I_{out_x}\cdot(\frac{I_{out_x}^2}{I_x}+\frac{I_{out_y}^2}{I_x}+I_{\mu})}{I_x}\\
\cdot(1-\frac{I_{out_x}^2}{I_x}-\frac{I_{out_y}^2}{I_x})\\
F_y(I_{out_x},I_{out_y})=I_{out_x}+\frac{I_{out_y}\cdot(\frac{I_{out_x}^2}{I_x}+\frac{I_{out_y}^2}{I_x}+I_{\mu})}{I_x}\\
\cdot(1-\frac{I_{out_x}^2}{I_x}-\frac{I_{out_y}^2}{I_x})
\end{cases}
\end{equation}
where $I_\mu=-0.5 nA$ and $I_x=1 nA$.
\par Schematic diagrams for this system are not presented here due to lack of space and left to the interested readers.

\subsection{Neurosynaptic Network Structure}
In this case study, the feasibility of the proposed NBDS circuit is investigated by means of three small neurosynaptic network structures. The networks comprise three FHN neuron models (see Section IV. A) variously coupled with excitatory and inhibitory synapses. The strength of the excitatory and inhibitory connections are 0.01 and -0.01 respectively. The dynamics of each synapse is modeled by means of a simple low pass filter with cut--off frequency 20 Hz:
\begin{equation}\label{eq:5_40}
\tau \dot{x}=-x+I
\end{equation}
where $\tau=0.05$ sec and $I$ represents the weighted input vectors connected from other neurons. It should be noted that, similar to the FHN neuron model, the synapse circuits are also implemented by the NBDS circuit demonstrating the applicability of the proposed topology in the realization of linear dynamical systems such as filters.

\section{Simulated Results}
This section illustrates the simulation-based results of the aforementioned nonlinear bilateral dynamical systems in the previous section. The hardware results for various case studies simulated by the Cadence Design Framework (CDF) using the process parameters of the commercially available AMS 0.35 $\mu m$ CMOS technology are validated by means of MATLAB simulations. The mathematical parameters have been extracted from the literature, while the electrical ones have been calculated from the scaled relation between the two systems.

\subsection {FHN Neuron Model}
The proposed electrical FHN circuit is able to reproduce all significant qualitative phenomena of the mathematical equivalent and underlying
bifurcations. Here, we focus on the main responses of the FHN model and individually investigate them in detail along with the MATLAB simulations. Generally, results confirm an acceptable compliance between the MATLAB and Cadence simulations. Table \ref{table:5_2} summarizes the specifications of the proposed circuit applied to this case study. It should be stressed that in order to mimic the biological time--scale, relatively large capacitors must be employed in the design while in the pure simulation studies we can reduce the value of capacitors even to $10 pF$.
\subsubsection {Absence of All-or-none Spike}
In this behaviour, the amplitude of the output signal is directly related to the amplitude of the injected input current. Therefore, weak stimulations lead to small changes in the output termed as the subthreshold response. Stronger stimulations result in intermediate changes, and strong stimulations produce large changes in the output termed as the supra--threshold firing response. Figure \ref{fig:5_6}(a--b) show the time--domain and representation of this phenomenon produced by MATLAB and Cadence, respectively.

\begin{table}[t]
\captionsetup{font=footnotesize}
\caption{Electrical Parameter Values for the Simulated FHN Neuron Model.}   
\centering          
\begin{tabular}{c c}    
\hline\hline                        
Specifications & Value\\ [0.5ex]  
\hline                      
Power Supply (Volts)& 3.3\\
Bias Voltage (Volts)& 1.2\\
Capacitances (pF)& 800\\
$\frac{W}{L}$ ratio of PMOS and NMOS Devices ($\frac{\mu m}{\mu m}$)& $\frac{30}{9}$ and $\frac{10}{2}$\\
Static Power Consumption ($n W$)& 93.86\\
\hline          
\end{tabular}
\label{table:5_2}    
\end{table}

\subsubsection {Excitation Block}
In this response, the neuron ceases periodic firing and leans to a stable resting state as the amplitude of the input current increases. The transition from the resting state to the periodic firing and then the blocking state from MATLAB and Cadence, using a ramp input current, are respectively demonstrated in Figure \ref{fig:5_6}(c--d).

\subsubsection {Post--inhibitory Rebound Spike}
This behaviour is produced in response to a short negative pulse to the model. As the negative pulse is applied, hyperpolarization occurs. This transient state results in a single spike in time--domain. This response is illustrated for MATLAB and Cadence, respectively in Figure \ref{fig:5_6}(e--f).

\subsubsection {Spike Accommodation}
In this response, slow increase in the input current up to a certain amplitude cannot cause a firing response, while a quick increase of the input current to the same (even smaller) amplitude leads to a spike in the output. The time--domain waveforms from MATLAB and Cadence are respectively seen in Figure \ref{fig:5_6}(c--d)(g--h).

\subsection {Lorenz Attractor}
As shown in Figure \ref{fig:5_7}, the proposed Lorenz attractor's circuit is able to project the trajectory onto the x--y, z--y (butterfly--wings) and z--x planes similar to the mathematical model. The results illustrate a good agreement between MATLAB and Cadence simulations and also confirm that the proposed electrical circuit can mimic the nonlinear bilateral dynamical systems with such a high dynamic range (in this case $\approx 50 nA$). Table \ref{table:5_3} summarizes the specifications of the proposed circuit applied to this case study.

\begin{figure*}[t]
\normalsize
\centering
\includegraphics[trim = 0in 0in 0in 0in, clip, width=6in]{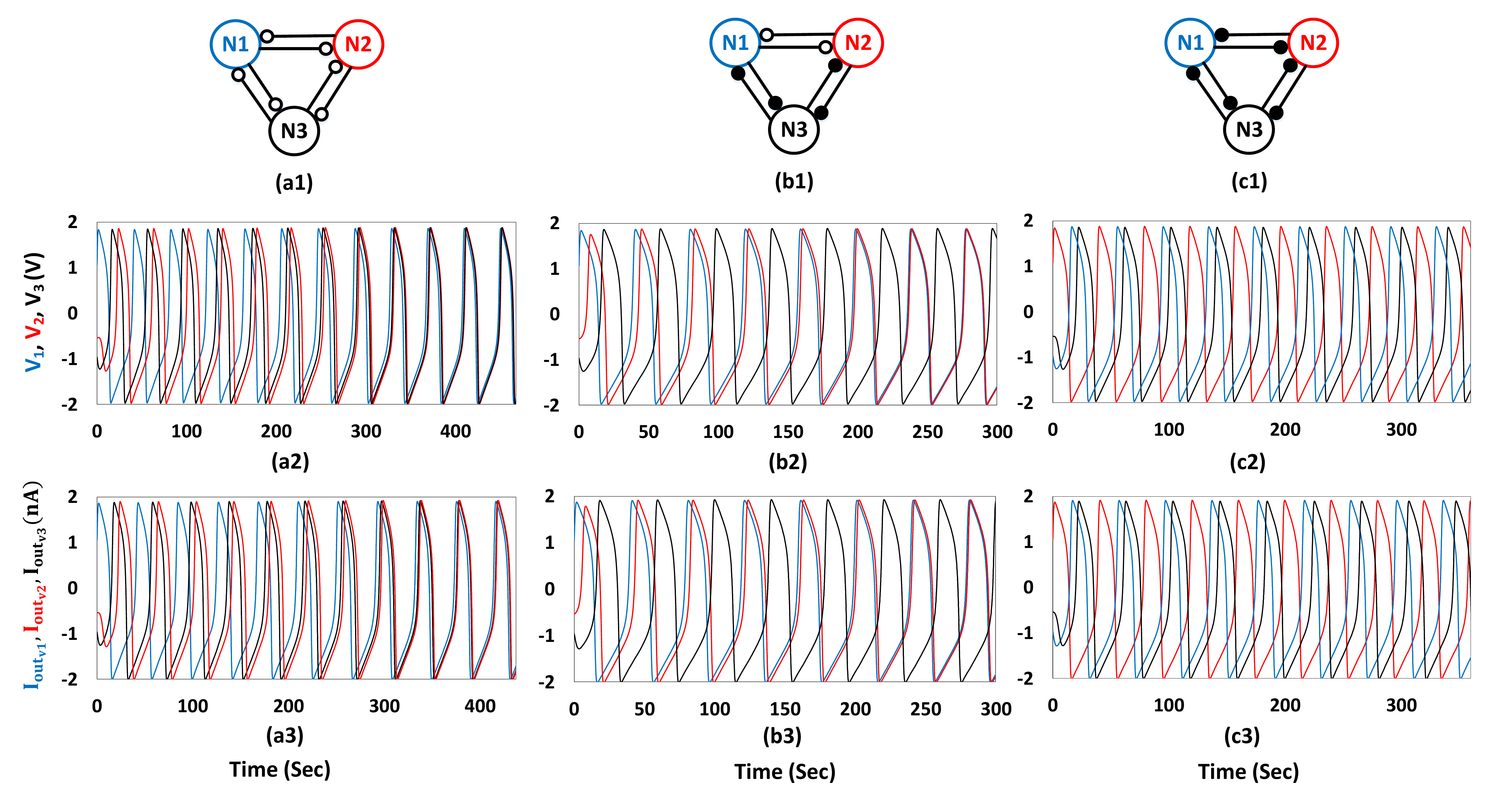}
\vspace{-10pt}
\captionsetup{font=footnotesize}
\caption{(a1--c1) represent three different combinations of the neurosynaptic structures. The excitatory and inhibitory synapses are respectively shown with white and black circles. (a2--c2) and (a3--c3) are the corresponding time--domain responses of (a1--c1), extracted from MATLAB and Cadence respectively.}
\vspace{-10pt}
\label{fig:5_8}
\end{figure*}

\begin{table}[t]
\captionsetup{font=footnotesize}
\caption{Electrical Parameter Values for the Simulated Lorenz Attractor.}   
\centering          
\begin{tabular}{c c}    
\hline\hline                        
Specifications & Value\\ [0.5ex]  
\hline                      
Power Supply (Volts)& 3.3\\
Bias Voltage (Volts)& 1.5\\
Capacitances (pF)& 400\\
$\frac{W}{L}$ ratio of PMOS and NMOS Devices ($\frac{\mu m}{\mu m}$)& $\frac{30}{9}$ and $\frac{10}{2}$\\
Static Power Consumption ($u W$)& 10.46\\
\hline          
\end{tabular}
\label{table:5_3}    
\end{table}

\subsection {Hopf Oscillator}
\begin{figure}[t]
\vspace{-10pt}
\centering
\includegraphics[trim = 0.25in 0.25in 0.15in 0.25in, clip, width=5in]{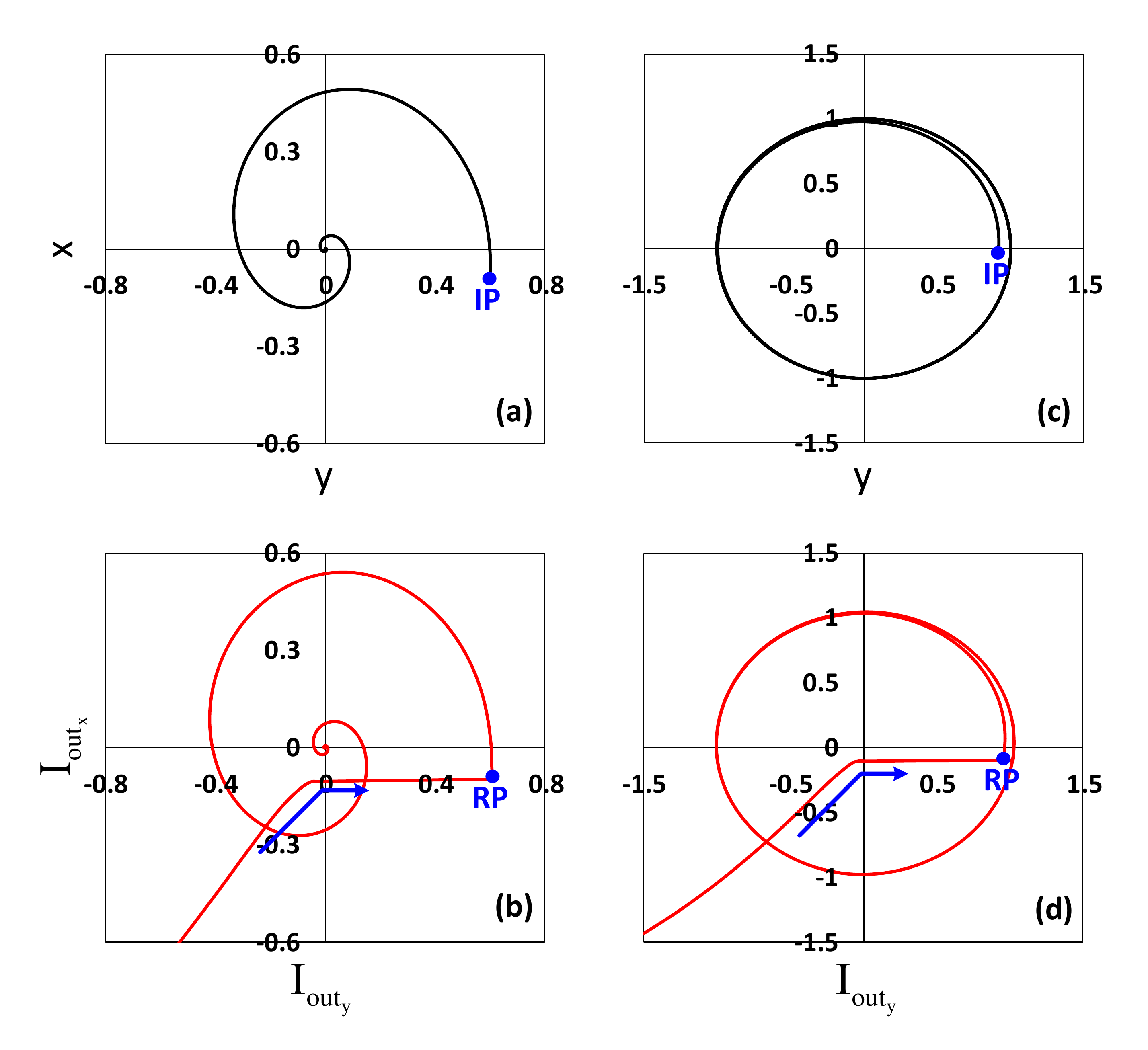}
\vspace{-5pt}
\captionsetup{font=footnotesize}
\caption{The projections of (a-b) x--y plane demonstrating damping behaviour and (c-d) x--y plane demonstrating oscillatory behaviour for MATLAB and Cadence respectively. IP stands for initial point and RP refers to reset point.}
\vspace{-10pt}
\label{fig:5_9}
\end{figure}

As explained before, initial values in a specific group of dynamical systems can affect the time--domain evolution of the whole system. Figure \ref{fig:5_9} illustrates the damping and oscillatory behaviours for MATLAB and Cadence simulations. As shown in the figure, different initial points (IPs) result in changing the behaviour of the system from damping to oscillatory response. Here, we use the circuit introduced in Section II.C in order to reset the initial output current to a specific value and show bistability. However, it should be stressed that before applying a short pulse ($V_{pulse}$), the capacitor voltage is zero volt leading to a negative current in the output signal. Then by applying the pulse, the NMOS transistor tries to pull up the capacitor voltage so that the output current can be set to the initial value. Therefore, we expect to see the output current reaching the reset point (RP) through a highlighted path by a blue arrow in Figure \ref{fig:5_9}(b) and (d). It can be seen that the peak-to-peak oscillation amplitude error between Cadence and MATLAB simulations is around 1.4 $\%$ ($\frac{2.035-2.007}{2.007}*100$) which is a tolerable precision. It also confirms that the proposed electrical circuit is able to mimic bistability in the nonlinear bilateral dynamical systems. Table \ref{table:5_4} summarizes the specifications of the proposed circuit applied to this case study.

\begin{table}[t]
\captionsetup{font=footnotesize}
\caption{Electrical Parameter Values for the Simulated Hopf Oscillator.}   
\centering          
\begin{tabular}{c c}    
\hline\hline                        
Specifications & Value\\ [0.5ex]  
\hline                      
Power Supply (Volts)& 3.3\\
Bias Voltage (Volts)& 1.2\\
Capacitances (pF)& 500\\
$\frac{W}{L}$ ratio of PMOS and NMOS Devices ($\frac{\mu m}{\mu m}$)& $\frac{30}{9}$ and $\frac{10}{2}$\\
Static Power Consumption ($n W$)& 118.14\\
\hline          
\end{tabular}
\label{table:5_4}    
\end{table}

\subsection{Neurosynaptic Network Structure}
\par As mentioned before, the networks shown in Figure \ref{fig:5_8}(a1--c1) contain three FHN neuron models variously coupled with excitatory and inhibitory synapses. The MATLAB time--domain simulations of each network along with the corresponding Candence responses are demonstrated in Figure \ref{fig:5_8}(a2--c3). The results reveal a remarkable agreement between the MATLAB and Cadence simulations. In the first case (a1), the neurons are coupled with excitatory connections, leading to synchronization in the network in spite of the different initial values. Such an activity is due to the positive interaction between the neurons pushing them into an identical time--domain response. In the second case (a2), the connections between neurons 1 and 2 are excitatory while the others are inhibitory. It is expected that the neurons (1 and 2) with positive cross--coupling strength are pushed into synchronization but neuron 3; due to its negative couplings, is repelled form the others and pushed to be asynchronized. In the third case (a3), all neurons are connected to each other through inhibitory synapses and fire with almost $120^{\circ}$  phase shift, illustrating the repelling interaction among them all.

\subsection{Power Analysis for Different Bias Voltages}
\begin{figure}[t]
\vspace{-10pt}
\centering
\includegraphics[trim = 0.25in 0.25in 0.15in 0.25in, clip, width=5in]{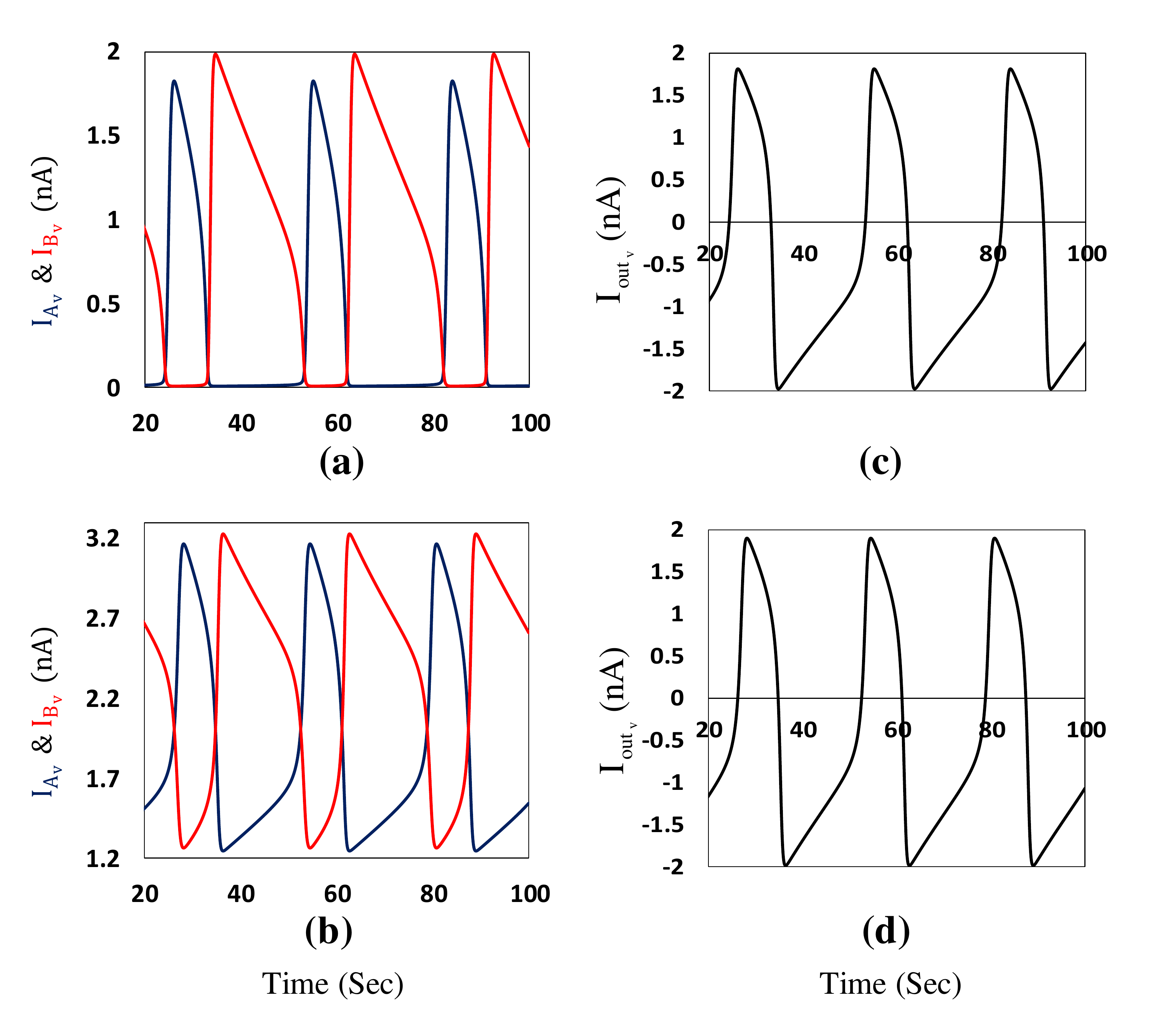}
\vspace{-5pt}
\captionsetup{font=footnotesize}
\caption{Cadence time-domain representations of (a-b) $I_A$ and $I_B$ waveforms corresponding to the oscillatory behaviour of the FHN neuron model for $V_b=1.2V$ and $V_b=1.6V$ respectively, (c-d) $I_{out}$ waveform corresponding to the oscillatory behaviour of the FHN neuron model for $V_b=1.2V$ and $V_b=1.6V$ respectively.}
\vspace{-10pt}
\label{fig:5_10}
\end{figure}

As stressed before, the bias voltage ($V_b$) is able to regulate the dynamic range of the output signal ($I_{out}$) as well as the circuit's power consumption. By setting a proper value of $V_b$ in the circuit for a certain output dynamic range, an optimum design can be delivered. We analytically showed that by increasing $V_b$, the negative dynamic range increases and consequently since stronger currents propagate through the NBDS circuit and basic blocks in the system, the power consumption also increases. However, it was stressed that for an optimum design, the minimum $V_b$ must be selected so that the output dynamic range is covered and minimum power consumed. In this subsection, we set up two experiments on the FHN neuron model with various bias voltages in order to show the effect of $V_b$ on the power consumption. In the first case, we set $V_b=1.2V$ so that the circuit can properly operate and the second case, $V_b=1.6V$ where the circuit still operates properly in terms of functionality and accuracy. Figure \ref{fig:5_10}(a-b) shows time-domain waveforms of $I_A$ and $I_B$ corresponding to the oscillatory behaviour of the FHN neuron model for $V_b=1.2V$ and $V_b=1.6V$ leading to identical output currents (see Figure \ref{fig:5_10}(c-d) where $I_{out}=I_B-I_A$). However, the average power (AP) consumptions are different in two experiments ($AP_{1.2}=93.86$ $nW$ and $AP_{1.6}=237.98$ $nW$) confirming this fact that for having an optimum design $V_b$ must be set to the minimum voltage so that the circuit operates properly and for higher voltages just the AP increases.

\subsection{Robustness and Process Variations}
\begin{figure}[t]
\vspace{-10pt}
\centering
\includegraphics[trim = 0.25in 0.25in 0.15in 0.25in, clip, width=5in]{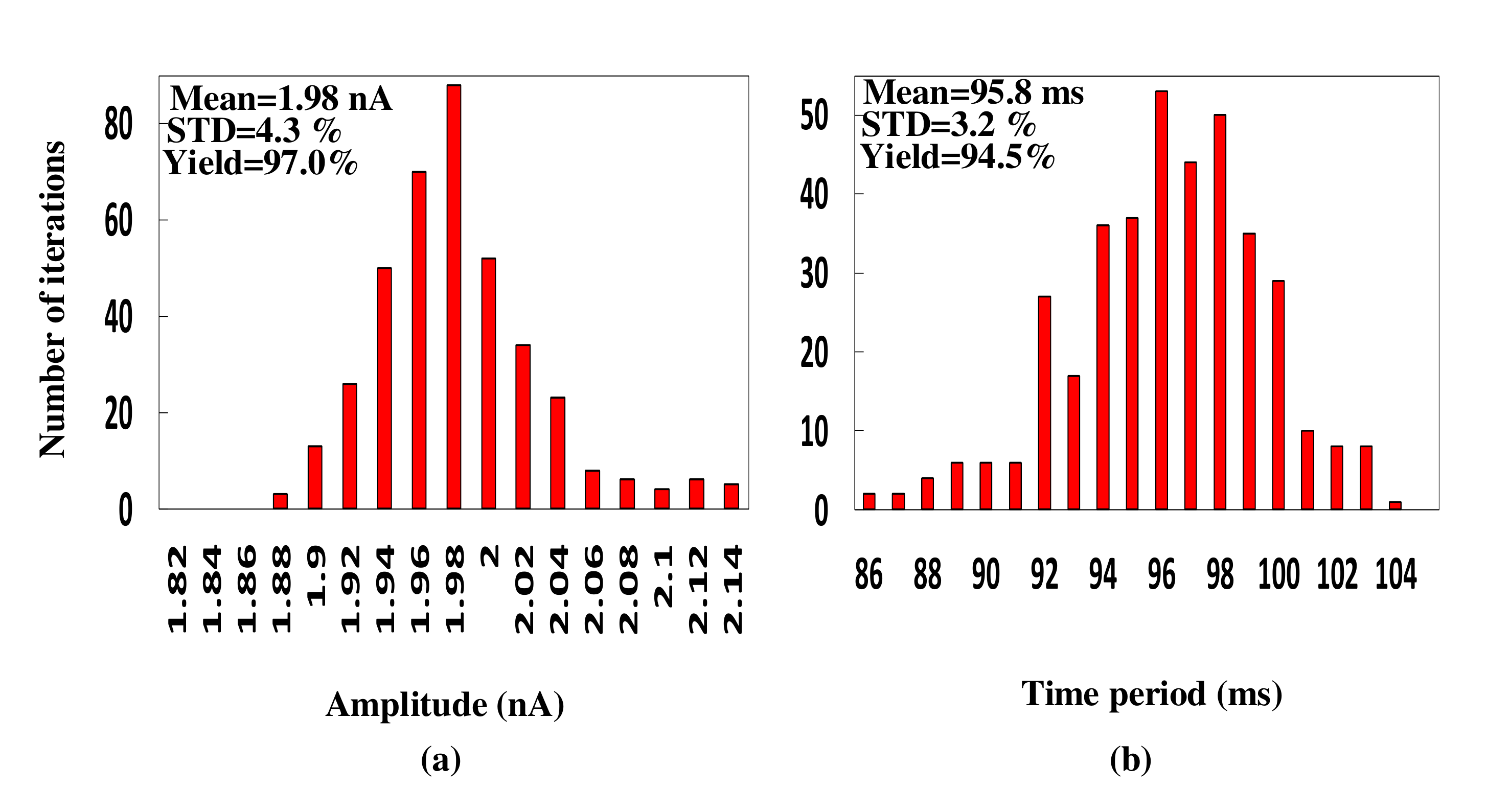}
\vspace{-5pt}
\captionsetup{font=footnotesize}
\caption{Monte Carlo analysis for the FHN neuron model targeting (a) amplitude, (b) time period (1/frequency) of oscillations.}
\vspace{-10pt}
\label{fig:5_11}
\end{figure}

In this section, the susceptibility of the FHN neuron model against process variations is investigated by means of extensive Monte Carlo analyses. To deliver an acceptable performance for such analyses, larger sizes ($\frac{W}{L}$ aspect ratios of PMOS and NMOS devices are $\frac{90}{30}$ and $\frac{80}{10}$ ($\frac{\mu m}{\mu m}$) respectively) are used for the basic building blocks and current mirrors. Besides, the aspect ratios of the core part (see Figure \ref{fig:5_1}(a), M1, M2, M3 and M4) of the design are adjusted using the embedded $Yield~Optimizer$ in Cadence software resulting in $\frac{41}{1}$, $\frac{11}{36}$, $\frac{31}{31}$ and $\frac{51}{6}$ for M1, M2, M3 and M4 respectively. The capacitor value and input amplitude are also chosen 2 nF and 0.6 nA. The amplitude and frequency of oscillations are selected as the two quantities for the proposed circuit. Results demonstrate that the mean MC amplitude and time period are 1.98 nA and 95.8 ms with standard deviation of $4.3\%$ and $3.2\%$ respectively (shown in Figure \ref{fig:5_11}). The total percentage of successful oscillations are also $97.0\%$ and $94.5\%$ which are reassuringly high for such nonlinear log--domain circuits. It should be stressed that by applying no optimization and selecting identical ratios for PMOS and NMOS devices used in the basic building blocks, the resulting MC amplitude and time period values change to 1.91 nA and 92.8 ms with standard deviations of $8.1\%$ and $10.6\%$ respectively. The total percentage of successful oscillations for 400 points in this case are $93.2\%$ and $91.7\%$. Although in the optimised experiment the core sizes for PMOS and NMOS devices are not matched, the equations (\ref{eq:5_15} and \ref{eq:5_17}) proven before are still valid with small changes which are left to the interested readers.

\subsection {Noise and Area Tradeoff}
$Signal*noise$ intermodulation characterizes the noise behaviour of the proposed logarithmic topologies. This is a general feature of Externally--Linear--Internally--Nonlinear (ELIN) topologies \cite{Tsividis1997} stemming from their internal nonlinear behaviour: when the input signal power increases beyond the Class--A limit of operation, then the noise power also increases with the input and the SNR saturates, i.e. the SNR remains constant for increasing inputs. This behaviour is detectable in Class-AB log--domain and hyperbolic--sine (Sinh) structures via both simulations and measurements \cite{Kardoulaki2014}, \cite{Kardoulaki2013}. In such structures the ratio of the input current over the input bias current value, termed modulation index in international literature, can take very high instantaneous values (e.g. 3000 or more).
\par Transient noise simulations of the presented topologies in this paper have confirmed the presence of $signal*noise$ intermodulation. However, this is not pronounced because the modulation index of the proposed circuits is not high. Simulations further confirmed that the input and output referred noise decreases when the capacitor values increases (all the $I_{dc}$ currents scale appropriately) to achieve the same targeted dynamics. Recall that, as explained in Section II, for a certain value of $\tau$ various individual $C$ and $I_{dc}$ value combinations can be chosen. For example, it has been confirmed that for the FHN neuron circuit (see Table \ref{table:5_2}), setting the capacitor $C$ value from 5.5nF to a reduced value of 275pF (area reduction of $95\%$) and scaling appropriately the bias currents $I_{dc}$, leads to increased (by 1.4 to 2 times) input and output referred noise levels. Given that increased $C$ values lead to increased area, an optimal design targeting a specific set of dynamics should consider both noise and area constraints. An additional approach to reduce area consumption at the expense of less optimal noise level is the exploitation of MOS-capacitors \cite{Houssein2016}.

\section{Proof of (\ref{eq:5_18})}
According to Figure \ref{fig:5_2} and using (\ref{eq:5_1})--(\ref{eq:5_2}), it can be shown that:
\begin{equation}\label{eq:5_41}
I_{A}^2=I_{S_n}I_{S_p}\cdot exp(\frac{V_b-V_{S_1}}{n_nV_T})\cdot exp(\frac{V_{S_1}-V_{C}}{n_pV_T})
\end{equation}
the square root on (\ref{eq:5_41}) leads to:
\begin{equation}\label{eq:5_42}
I_{A}=\sqrt{I_{S_n}I_{S_p}} \cdot exp(\frac{V_b}{2n_nV_T}-\frac{V_{C}}{2n_pV_T}+\frac{V_{S_1}}{2V_T}\cdot (\frac{1}{n_p}-\frac{1}{n_n})).
\end{equation}
\par On the other hand $I_{M_1}=I_{M_2}$, thus:
\begin{equation}\label{eq:5_43}
I_{S_n}\cdot exp(\frac{V_b-V_{S_1}}{n_nV_T})=I_{S_p}\cdot exp(\frac{V_{S_1}-V_{C}}{n_pV_T})
\end{equation}
and applying the $ln$ function on (\ref{eq:5_43}) yields:
\begin{equation}\label{eq:5_44}
\frac{V_b-V_{S_1}}{n_nV_T}=\frac{V_{S_1}-V_{C}}{n_pV_T}+ln\frac{I_{S_p}}{I_{S_n}}
\end{equation}
which leads to:
\begin{equation}\label{eq:5_45}
V_{S_1}=\frac{n_nn_pV_T}{n_n+n_p}(\frac{V_{b_1}}{n_nV_T}+\frac{V_{C_2}}{n_pV_T}+ln\frac{I_{S_n}}{I_{S_p}}).
\end{equation}
\par By substituting (\ref{eq:5_45}) in (\ref{eq:5_42}):
\begin{equation}\label{eq:5_46}
I_{A}=\sqrt{I_{S_n}I_{S_p}}\cdot exp(\frac{V_b}{(n_n+n_p)V_T}-\frac{V_{C}}{(n_n+n_p)V_T}+\frac{n_n-n_p}{2(n_n+n_p)}ln\frac{I_{S_n}}{I_{S_p}}).
\end{equation}
\par Setting $V_C=0$ and $V_b=V_C$ in (\ref{eq:5_46}), we can derive a similar relation for $I_B$:
\begin{equation}\label{eq:5_47}
I_{B}=\sqrt{I_{S_n}I_{S_p}}\cdot exp(\frac{V_C}{(n_n+n_p)V_T}+\frac{n_n-n_p}{2(n_n+n_p)}ln\frac{I_{S_n}}{I_{S_p}}).
\end{equation}
\par Therefore, $I_{out}$ is equal to:
\begin{equation}\label{eq:5_48}
I_{out}=I_{B}-I_{A}=\sqrt{I_{S_n}I_{S_p}}exp(\frac{n_n-n_p}{2(n_n+n_p)}ln\frac{I_{S_n}}{I_{S_p}})[exp(\frac{V_C}{(n_n+n_p)V_T})-exp(\frac{V_b-V_C}{(n_n+n_p)V_T})].
\end{equation}

\section{Proof of (\ref{eq:5_19})}
According to Figure \ref{fig:5_2} when $M_2$ operates in subthreshold saturation:
\begin{equation}\label{eq:5_49}
V_C>4V_T+2V_{GS_5}-V_{SG_2}
\end{equation}
and assuming that $M_1$ and $M_5$ are matched $I_{M_1}=I_{M_5}$; then $V_{GS_1}=V_{GS_5}$, therefore:
\begin{equation}\label{eq:5_50}
V_{C_{min}}=4V_T+2V_{GS_1}-V_{SG_2}
\end{equation}
where $V_{C_{min}}$ is considered as the minimum voltage dropping over the capacitor. Obviously, if $M_1$ and $M_2$ are matched $I_{M_1}=I_{M_2}$, one can show that:
\begin{equation}\label{eq:5_51}
\frac {V_{GS_1}}{n_nV_T}=\frac {V_{SG_2}}{n_pV_T}+ln\frac{I_{S_{p}}}{I_{S_{n}}}
\end{equation}
and consequently:
\begin{equation}\label{eq:5_52}
V_{SG_2}=-n_pV_Tln\frac{I_{S_{p}}}{I_{S_{n}}}+\frac{n_p}{n_n}V_{GS_1}.
\end{equation}
\par By substituting (\ref{eq:5_52}) in (\ref{eq:5_50}):
\begin{equation}\label{eq:5_53}
V_{C_{min}}=4V_T+(2-\frac{n_p}{n_n})V_{GS_1}+n_pV_Tln\frac{I_{S_{p}}}{I_{S_{n}}}.
\end{equation}
\par On the other hand, if $V_C=V_{C_{min}}$, one can show that:
\begin{equation}\label{eq:5_54}
V_{GS_1}+V_{SG_2}=V_b-V_{C_{min}}.
\end{equation}
\par From (\ref{eq:5_52}) and (\ref{eq:5_54}):
\begin{equation}\label{eq:5_55}
V_{GS_1}=\frac{1}{1+\alpha}(V_b-V_{C_{min}}+n_pV_Tln\frac{I_{S_{p}}}{I_{S_{n}}})
\end{equation}
where $\alpha=\frac{n_p}{n_n}$. By substituting (\ref{eq:5_55}) in (\ref{eq:5_53}):
\begin{equation}\label{eq:5_56}
V_{C_{min}}=\frac{1+\alpha}{3}4V_T+\frac{2-\alpha}{3}V_b+n_pV_Tln\frac{I_{S_p}}{I_{S_n}}.
\end{equation}

\section{A High Speed Systematic Computation of Nonlinear Bilateral Dynamical Systems}
In this section, a novel current--input current--output circuit is proposed \cite{25} that supports a systematic realisation procedure of strong--inversion circuits capable of computing bilateral dynamical systems at higher speed compared to the previously proposed log--domain circuit. Here, the application of the method is only demonstrated by synthesizing the 2--D nonlinear FitzHugh–Nagumo neuron model. The validity of our approach is verified by nominal simulated results with realistic process parameters from the commercially available AMS 0.35 $\mu m$ technology. The resulting high speed circuits exhibit time--domain responses in good agreement with their mathematical counterparts. Further simulation results on different case studies similar to those investigated for the log--domain in the previous section circuit is left to the interested readers.
\renewcommand{\baselinestretch}{\mystretch}
\label{chap:Future}
\begin{figure*}[ht]
\vspace{-20pt}
\normalsize
\centering
\includegraphics[trim = 0in 0in 0in 0in, clip, height=2.8in]{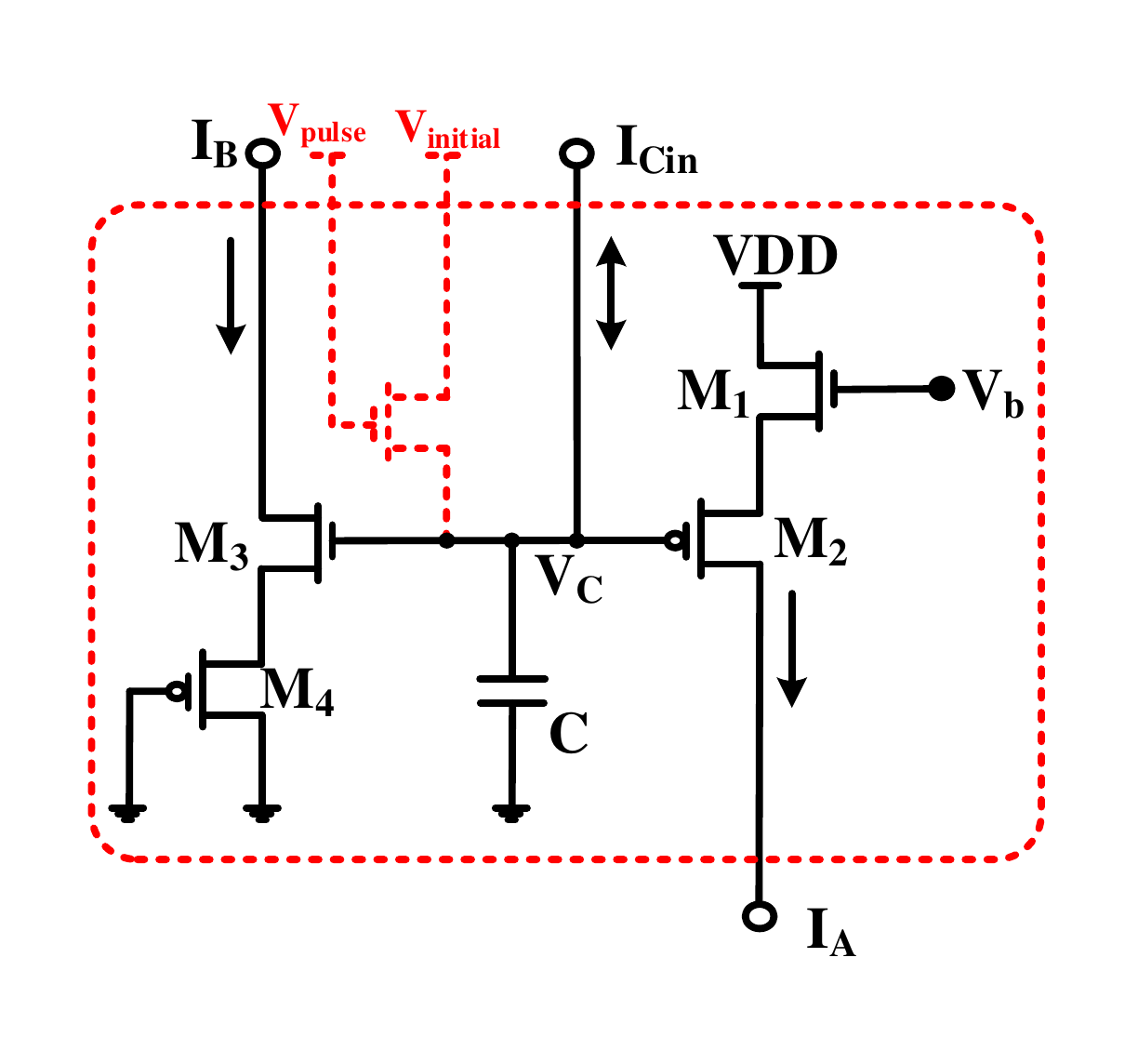}
\vspace{-5pt}
\captionsetup{font=footnotesize}
\caption{The ``main core" including the initialization circuit highlighted with red color.}
\vspace{-15pt}
\label{fig:51_1}
\end{figure*}

\section{Circuit Realization}
The current relationship of an NMOS and PMOS transistor operating in strong--inversion saturation when $\lvert V_{DS}\rvert>\lvert V_{GS}\rvert-\lvert V_{th}\rvert$ can be expressed as follows:
\begin{equation}\label{eq:51_1}
I_{D_n}=\frac{1}{2}\mu_n C_{ox}(\frac{W}{L})_n(V_{GS}-V_{th})^2
\end{equation}

\begin{equation}\label{eq:51_2}
I_{D_p}=\frac{1}{2}\mu_p C_{ox}(\frac{W}{L})_p(V_{SG}-V_{th})^2
\end{equation}

where $\mu_n$ and $\mu_p$ are the charge--carrier effective mobility for NMOS and PMOS transistors, respectively; $W$ is the gate width, $L$ is the gate length, $C_{ox}$ is the gate oxide capacitance per unit area and $V_{th}$ is the threshold voltage of the device.

\par Setting $k_{n}=\frac{1}{2}\mu_n C_{ox}(\frac{W}{L})_n$ and $k_{p}=\frac{1}{2}\mu_p C_{ox}(\frac{W}{L})_p$ in (\ref{eq:51_1}) and (\ref{eq:51_2}) and differentiating with respect to time, the current expression for $I_A$ (see Figure \ref{fig:51_1}) yields:

\begin{equation}\label{eq:51_3}
\dot{I}_A=\overbrace{2k_n(V_{GS}-V_{th})}^{\sqrt {k_nI_A}}\dot{V}_{GS_1}
\end{equation}

\begin{equation}\label{eq:51_4}
\dot{I}_A=\overbrace{2k_p(V_{SG}-V_{th})}^{\sqrt{k_pI_A}}\dot{V}_{SG_2}
\end{equation}

(\ref{eq:51_3}) and (\ref{eq:51_4}) are equal, therefore:
\begin{equation}\label{eq:51_5}
\dot{V}_{SG_2}=\sqrt{\frac{k_n}{k_p}}\dot{V}_{GS_1}=\beta \dot{V}_{GS_1}
\end{equation}
where $\beta=\sqrt{\frac{k_n}{k_p}}$. Similarly, we can derive the following equation for transistors $M_3$ and $M_4$:
\begin{equation}\label{eq:51_6}
\dot{V}_{SG_4}=\sqrt{\frac{k_n}{k_p}}\dot{V}_{GS_3}=\beta \dot{V}_{GS_3}.
\end{equation}
\par The application of Kirchhoff's Voltage Law (KVL) and applying the derivative function show the following relations:
 \begin{equation}\label{eq:51_7}
\dot{V}_{C}=-(\dot{V}_{GS_1}+\dot{V}_{SG_2})
\end{equation}

  \begin{equation}\label{eq:51_8}
\dot{V}_{C}=+(\dot{V}_{GS_3}+\dot{V}_{SG_4})
\end{equation}
 where $V_C$ is the capacitor voltage and $V_b$ the bias voltage which is constant (see Figure \ref{fig:51_1}). Substituting (\ref{eq:51_5}) and (\ref{eq:51_6}) into (\ref{eq:51_7}) and (\ref{eq:51_8}) respectively yields:

 \begin{equation}\label{eq:51_9}
\dot{V}_{C}=-\dot{V}_{GS_1}\cdot(1+\beta)
\end{equation}

\begin{equation}\label{eq:51_10}
\dot{V}_{C}=+\dot{V}_{GS_3}\cdot(1+\beta).
\end{equation}

Setting the current $I_{out}=I_B-I_A$ in Figure \ref{fig:51_1} as the state variable of our system and using (\ref{eq:51_3}) and the corresponding equation for $I_B$, the following relation is derived:

\begin{figure*}[t]
\vspace{-20pt}
\normalsize
\centering
\includegraphics[trim = 0in 0in 0in 0in, clip, height=3in]{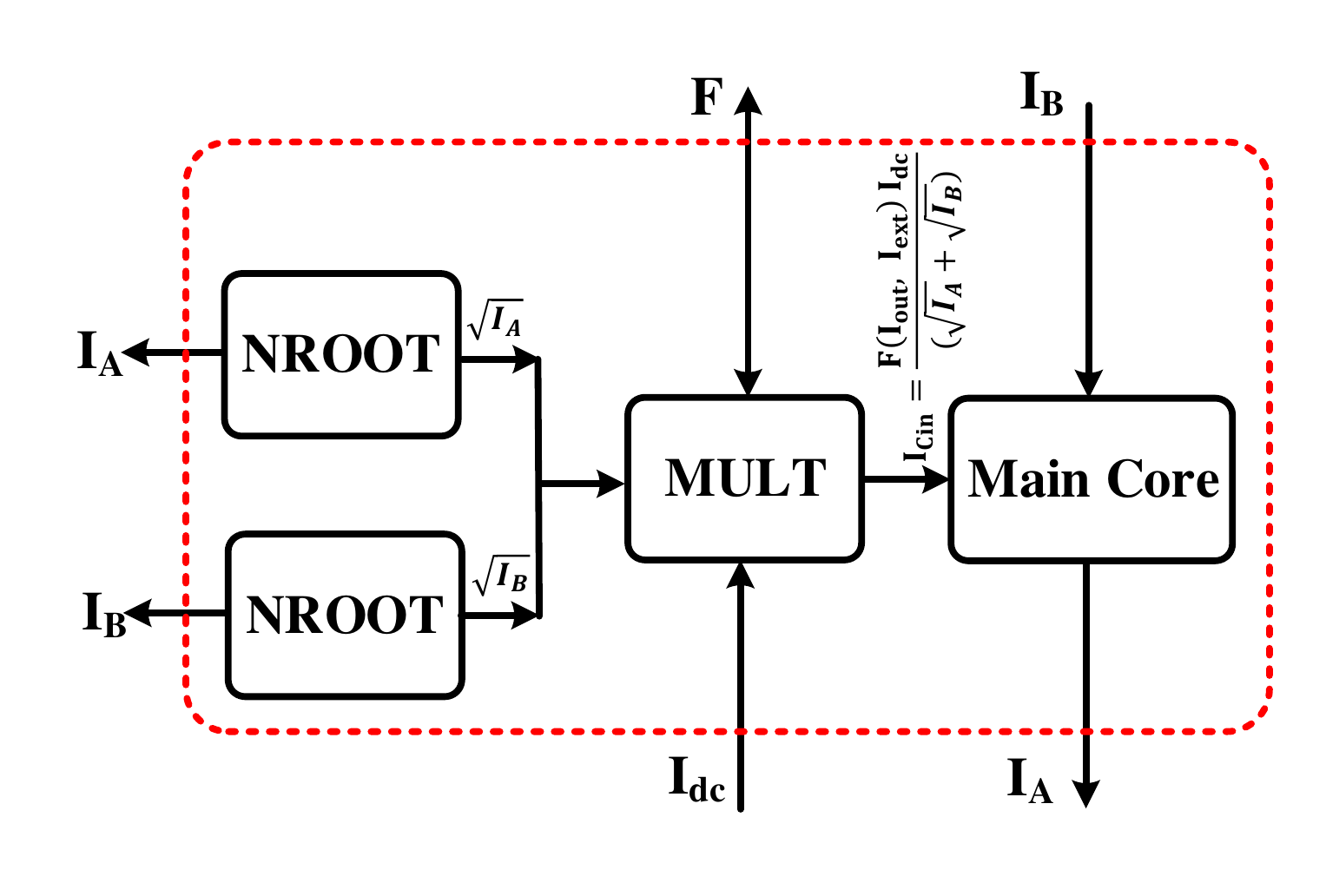}
\vspace{-5pt}
\captionsetup{font=footnotesize}
\caption{The ``main block" including the main core and two current--mode root square blocks and a bilateral multiplier.}
\vspace{-15pt}
\label{fig:51_2}
\end{figure*}

\begin{equation}\label{eq:51_11}
\dot{I}_{out}=\dot{I}_B-\dot{I}_A=2\sqrt{k_nI_B}\dot{V}_{GS_3}-2\sqrt{k_pI_A}\dot{V}_{GS_1}
\end{equation}

by substituting (\ref{eq:51_9}) and (\ref{eq:51_10}) in (\ref{eq:51_11}):
\begin{equation}\label{eq:51_12}
\dot{I}_{out}=(\sqrt{I_A}+\sqrt{I_B})\cdot \frac{2\sqrt{k_n}\dot{V}_C}{2+\beta}.
\end{equation}
\par Bearing in mind that the capacitor current $I_{Cin}$ can be expressed as $C\dot{V}_C$, relation (\ref{eq:51_12}) yields:
\begin{equation}\label{eq:51_13}
\dot{I}_{out}=(\sqrt{I_A}+\sqrt{I_B})\cdot \frac{2\sqrt{k_n}I_{Cin}}{(2+\beta)C}.
\end{equation}

\begin{figure*}[t]
\vspace{-20pt}
\normalsize
\centering
\includegraphics[trim = 0in 0in 0in 0in, clip, height=3.2in]{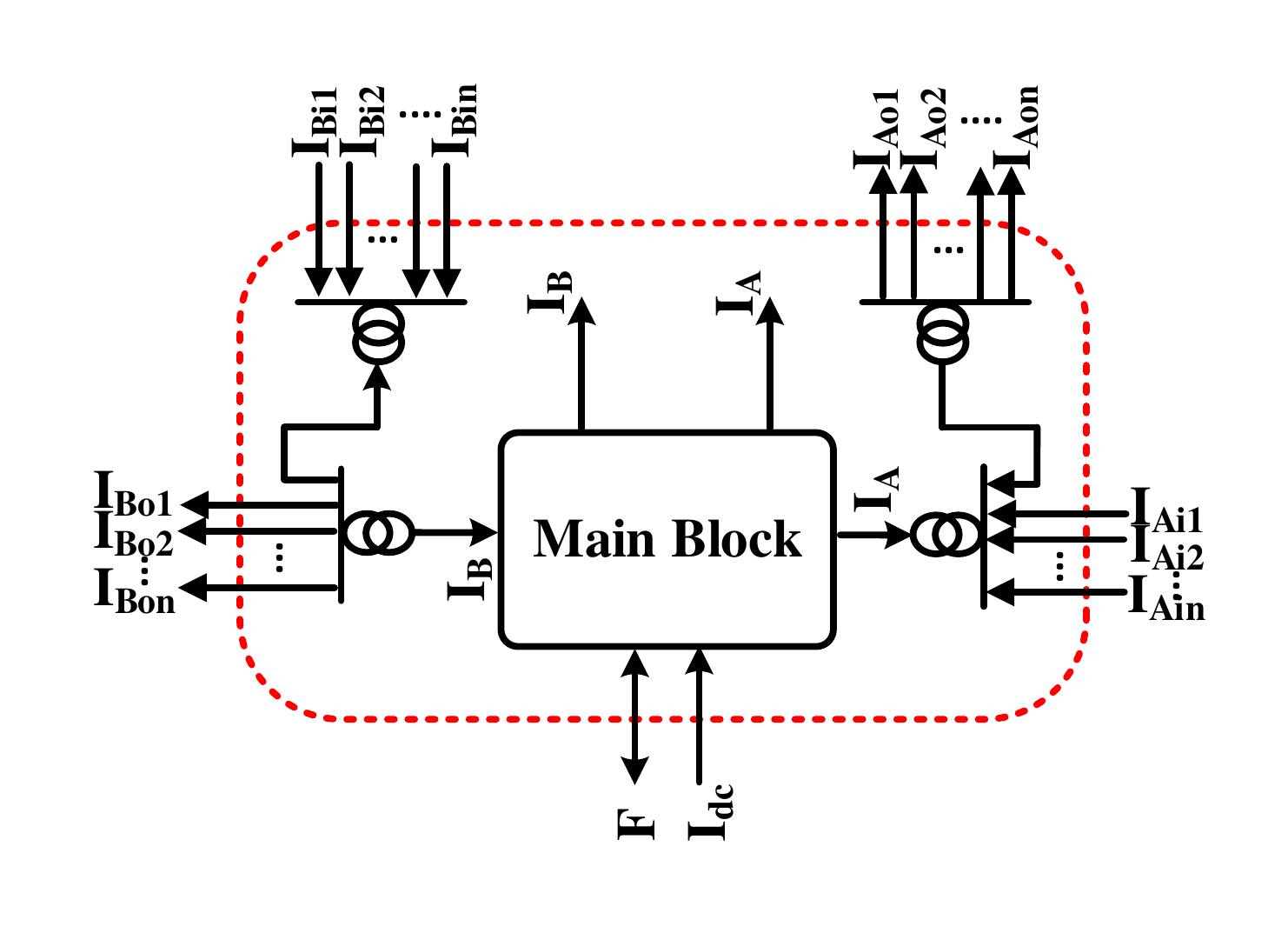}
\vspace{-5pt}
\captionsetup{font=footnotesize}
\caption{The final high speed circuit including the ``main block" with several copied currents (the current mirrors are represented with double circle symbols).}
\label{fig:51_3}
\end{figure*}

\par One can show that:
\begin{equation}\label{eq:51_14}
\frac{(2+\beta)C}{2\sqrt{k_n}\cdot I_{dc}}\dot{I}_{out}=\frac{(\sqrt{I_A}+\sqrt{I_B})}{I_{dc}}\cdot I_{Cin}.
\end{equation}
\par Equation (\ref{eq:51_14}) is the main core's relation. In order for a high speed mathematical dynamical system with the following general form to be mapped to (\ref{eq:51_14}):
\begin{equation}\label{eq:51_15}
\tau\dot{I}_{out} =F(I_{out}, I_{ext})
\end{equation}
where $I_{ext}$ and $I_{out}$ are the external and state variable currents, the quantities $\frac{C}{I_{dc}}$ and $I_{Cin}$ must be respectively equal to $\frac{2\tau\sqrt{k_n}}{(2+\beta)}$ and $\frac{F(I_{out}, I_{ext})I_{dc}}{(\sqrt{I_A}+\sqrt{I_B})}$. Note that the ratio value $\frac{C}{I_{dc}}$ can be satisfied with different individual values for $C$ and $I_{dc}$. These values should be chosen appropriately according to practical considerations (see Section V.G). Since $F$ is a bilateral function, in general, it will hold:
\begin{equation}\label{eq:51_16}
I_{Cin}=\overbrace{\frac{F^+(I_A,I_B,I_{ext}^+,I_{ext}^-)I_{dc}}{(\sqrt{I_A}+\sqrt{I_B})}}^{I_{Cin}^+}-\overbrace{\frac{F^-(I_A,I_B,I_{ext}^+,I_{ext}^-)I_{dc}}{(\sqrt{I_A}+\sqrt{I_B})}}^{I_{Cin}^-}
\end{equation}
where $I_{Cin}^+$ and $I_{Cin}^-$ are calculated respectively by a root square block (see Figure \ref{fig:51_2}) and $I_{ext}$ is separated to + and -- signals by means of splitter blocks. Note that $I_{dc}$ is a scaling dc current and $\tau$ has dimensions of $second(s)$. Since $I_{Cin}$ can be a complicated nonlinear function in dynamical systems, we need to provide copies of $I_{out}$ or equivalently of $I_A$ and $I_B$ to simplify the systematic computation at the circuit level. Therefore, the higher hierarchical block shown in Figure \ref{fig:51_3} is defined as the NBDS circuit (see Figure \ref{fig:51_3}) including the main block and associated current mirrors. The form of (\ref{eq:51_15}) is extracted for a 1--D dynamical system and can be extended to $N$ dimensions in a straightforward manner as follows:
\begin{equation}\label{eq:51_17}
\tau_N\dot{I}_{out_N} =F_N(\bar{I}_{out},\bar{I}_{ext})
\end{equation}
where $\frac{C_N}{I_{dc_N}}=\frac{2\tau_N\sqrt{k_n}}{(2+\beta)}$ and $I_{Cin_N}=\frac{F_N(\bar{I}_{out},~\bar{I}_{ext})I_{dc_N}}{(\sqrt{I_{A_N}}+\sqrt{I_{B_N}})}$.

\begin{figure*}[t]
\vspace{-20pt}
\normalsize
\centering
\includegraphics[trim = 0in 0in 0in 0in, clip, height=3.2in]{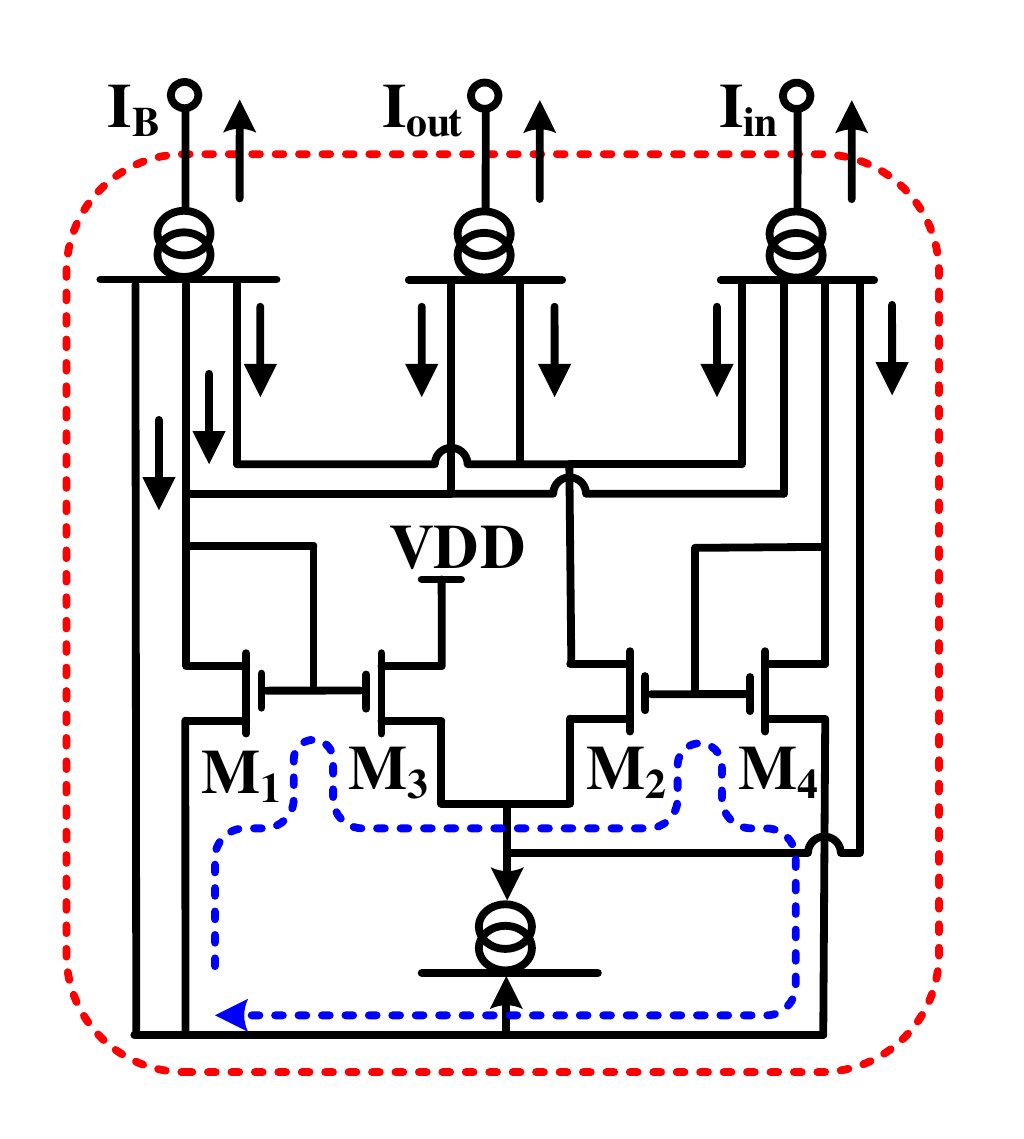}
\vspace{-5pt}
\captionsetup{font=footnotesize}
\caption{Transistor level representation of the basic Root Square block. The current mirrors are represented with double circle symbols.}
\vspace{-15pt}
\label{fig:51_4}
\end{figure*}

\section{Basic Electrical Blocks}
\subsection{Root Square Block}
This block performs current mode root square function on single–sided input signals. By setting $(\frac{W}{L})_{1\&2}=4\times(\frac{W}{L})_{3\&4}$, considering $I_1, I_2, I_3$ and $I_4$ as the currents flowing respectively into $M_1, M_2, M_3$ and $M_4$ and all transistors operate in strong--inversion saturation, the governing TL principle for this block becomes (highlighted with dotted blue arrow):
\begin{equation}\label{eq:51_18}
\frac{1}{2}(\sqrt{I_1}+\sqrt{I_2})=\sqrt{I_3}+\sqrt{I_4}
\end{equation}
\par By pushing specific currents (copied by current mirrors) according to Figure \ref{fig:51_4} into the TL's transistors we have:
\begin{equation}\label{eq:51_19}
\begin{cases}
I_1=I_2=I_{in}+I_{out}+I_b\\
I_3=I_b,~I_4=I_{in}
\end{cases}
\end{equation}
Substituting (\ref{eq:51_19}) into (\ref{eq:51_18}) yields:
\begin{equation}\label{eq:51_20}
\frac{1}{2}\times(\sqrt{I_{in}+I_{out}+I_b}+\sqrt{I_{in}+I_{out}+I_b}=\sqrt{I_{in}}+\sqrt{I_b}
\end{equation}
By squaring both sides of (\ref{eq:51_20}):
\begin{equation}\label{eq:51_21}
I_{in}+I_{out}+I_b=\sqrt{I_{in}}+\sqrt{I_{in}\cdot I_b}
\end{equation}
and finally:
\begin{equation}\label{eq:51_22}
I_{out}=2\sqrt{I_{in}\cdot I_b}
\end{equation}

\begin{figure*}[t]
\vspace{-20pt}
\normalsize
\centering
\includegraphics[trim = 0in 0in 0in 0in, clip, height=3.2in]{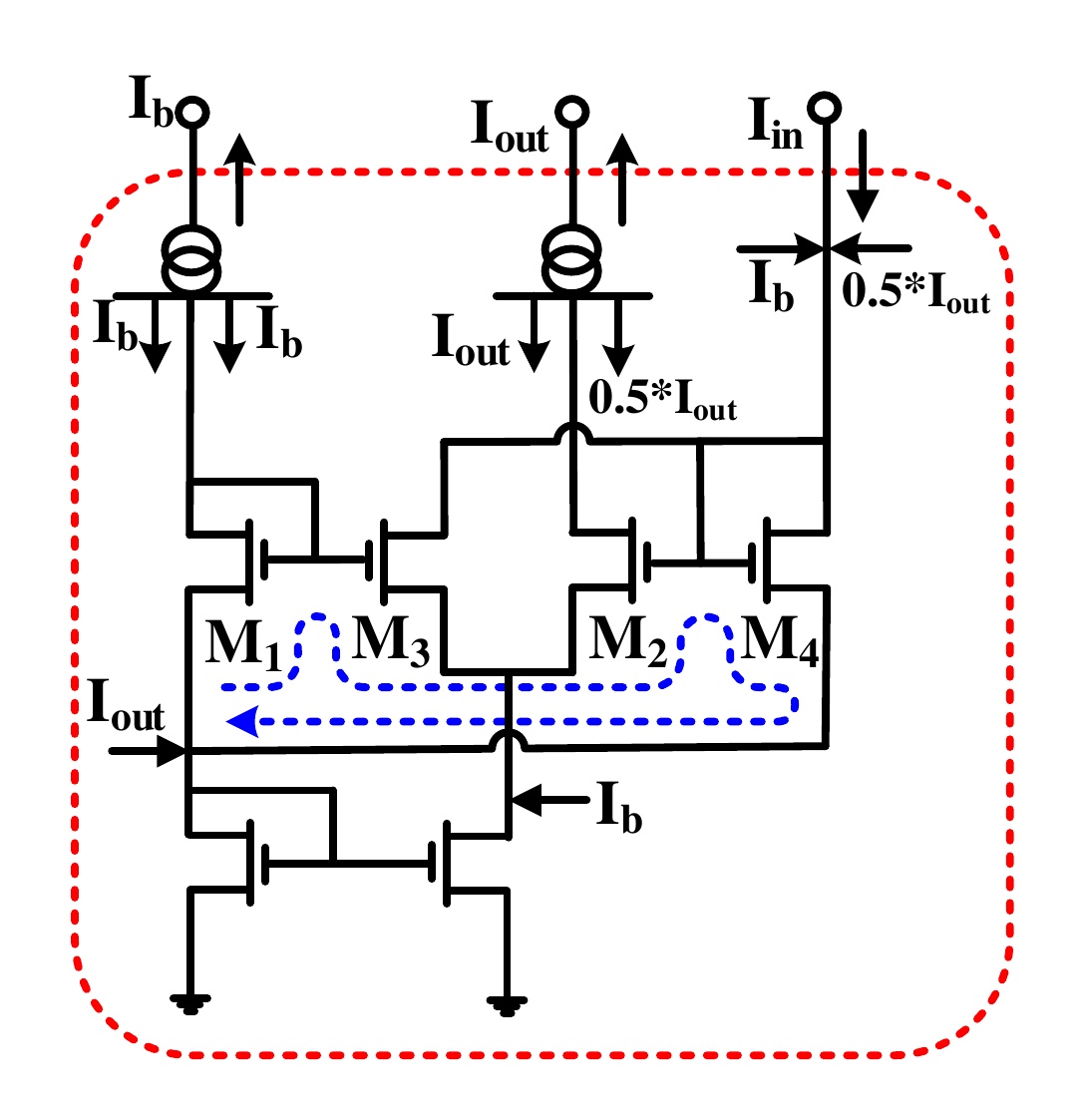}
\vspace{-5pt}
\captionsetup{font=footnotesize}
\caption{Transistor level representation of the MULT core block. The current mirrors are represented with double circle symbols.}
\vspace{-15pt}
\label{fig:51_5}
\end{figure*}
\subsection{MULT Core Block}
This block is the main core forming the final bilateral multiplier which is introduced in the next subsection. The block contains six transistors as well as two current mirrors. By assuming $I_1, I_2, I_3$ and $I_4$ as the currents flowing respectively into $M_1, M_2, M_3$ and $M_4$ and the same $\frac{W}{L}$ aspect ratio for all transistors operating in strong--inversion saturation, the KVL at the highlighted TL with dotted blue arrow yields:
\begin{equation}\label{eq:51_23}
\sqrt{I_1}+\sqrt{I_2}=\sqrt{I_3}+\sqrt{I_4}
\end{equation}
\par By forcing specific currents (copied by current mirrors) according to Figure \ref{fig:51_5} into the TL's transistors we have:
\begin{equation}\label{eq:51_24}
\begin{cases}
I_1=I_b,~I_2=I_{out}\\
I_3=I_4=\frac{1}{2}(I_{in}+\frac{I_{out}}{2}+I_b)
\end{cases}
\end{equation}
Substituting (\ref{eq:51_24}) into (\ref{eq:51_23}) yields:
\begin{equation}\label{eq:51_25}
\sqrt{I_{out}}+\sqrt{I_b}=2\sqrt{\frac{1}{2}(I_{in}+\frac{I_{out}}{2}+I_b)}
\end{equation}
By squaring both sides of (\ref{eq:51_25}):
\begin{equation}\label{eq:51_26}
\sqrt{I_{out}\cdot I_b}=I_{in}+\frac{1}{2}I_b
\end{equation}
and:
\begin{equation}\label{eq:51_27}
I_{out}=\frac{(I_{in}+\frac{1}{2}I_b)^2}{I_b}
\end{equation}

\begin{figure*}[t]
\vspace{-20pt}
\normalsize
\centering
\includegraphics[trim = 0in 0in 0in 0in, clip, height=3in]{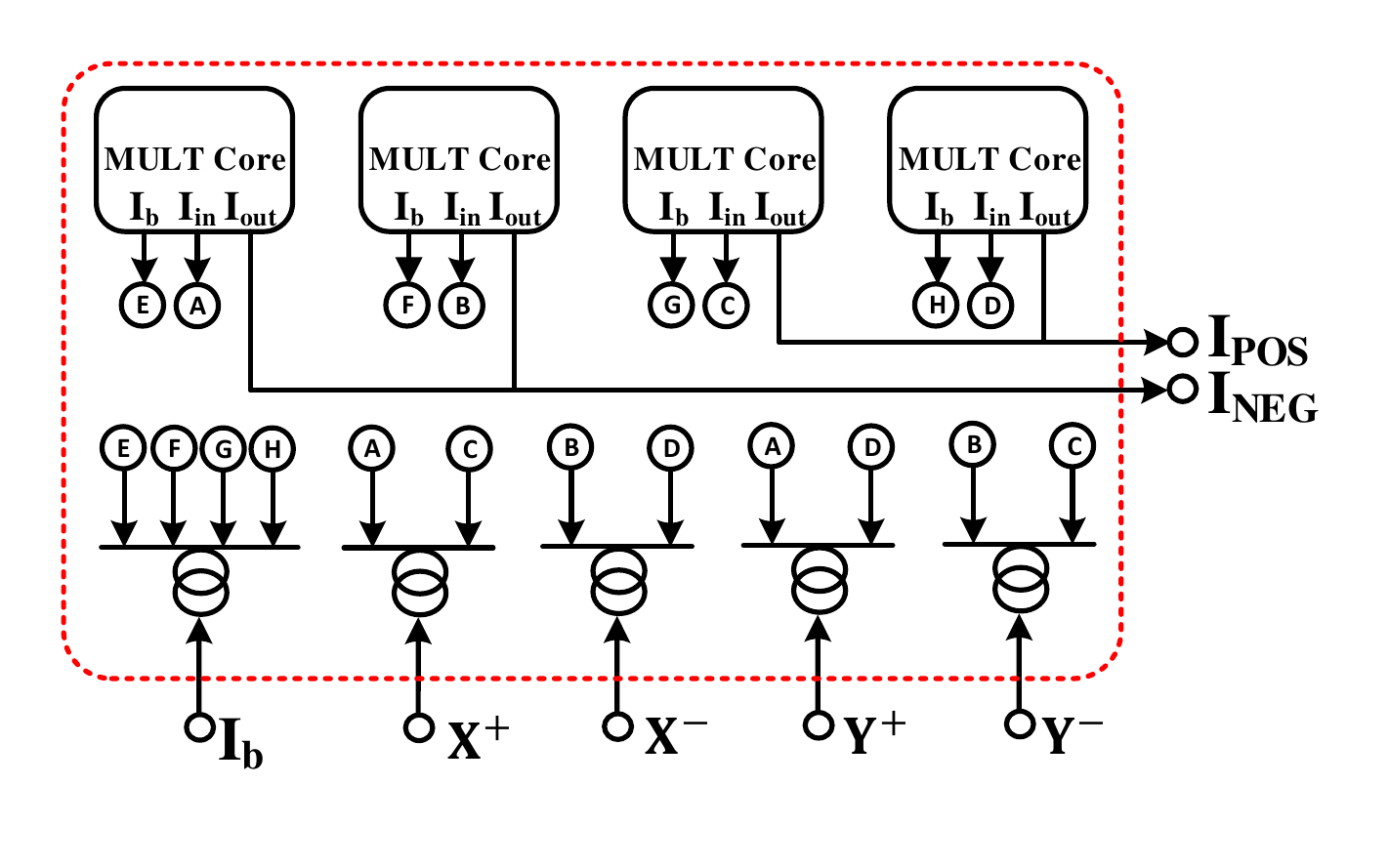}
\vspace{-5pt}
\captionsetup{font=footnotesize}
\caption{Schematic and symbolic representation of the bilateral MULT block comprising current mirrors and MULT Core block.}
\vspace{-15pt}
\label{fig:51_6}
\end{figure*}
\subsection{Bilateral MULT Block}
This block is able to perform current mode multiplication operation on bilateral input signals. If inputs are split to positive and negative sideS we have:
\begin{equation}\label{eq:51_28}
\begin{cases}
X=X^+-X^-\\
Y=Y^+-Y^-.
\end{cases}
\end{equation}
\par The multiplication result can be expressed as $XY=X^+Y^++X^-Y^--(X^-Y^++Y^-X^+)$. By extending equation (\ref{eq:51_27}) to $\frac{I_{in}^2}{I_b}+\frac{I_b}{4}+I_{in}$ for every basic MULT core block, the output signal constructed by a positive and negative side can be written as:
\begin{multline}
I_{out}=\frac{(X^++Y^+)^2}{I_b}+(X^++Y^+)+\frac{I_b}{4}+\frac{(X^-+Y^-)^2}{I_b}+(X^-+Y^-)+\frac{I_b}{4}\\
-\frac{(X^-+Y^+)^2}{I_b}-(X^-+Y^+)-\frac{I_b}{4}-\frac{(X^++Y^-)^2}{I_b}-(X^++Y^-)-\frac{I_b}{4}
\end{multline}
and by further simplifications:
\begin{equation}\label{eq:51_29}
I_{out}=\overbrace{\frac{2(X^+Y^++X^-Y^-)}{I_b}}^{I_{out}^+}-\overbrace{\frac{2(X^-Y^++X^+Y^-)}{I_b}}^{I_{out}^-}=\frac{2XY}{I_b}
\end{equation}

\subsection{Circuit Realisation of FHN neuron model}
The systematic synthesis procedure provides the flexibility and convenience required for the realization of nonlinear dynamical systems by computing their time-dependent dynamical behaviour. In this subsection, we showcase the methodology through which we systematically map the mathematical dynamical models onto the proposed electrical circuit. To this end, we apply the proposed systematic synthesis on the bilateral FHN neoruon model. The mathematical expression of model is shown in (\ref{eq:5_31}). According to this biological dynamical system, we can start forming the electrical equivalent using (\ref{eq:51_17}):
\begin{equation}\label{eq:51_30}
\begin{cases}
\frac{(2+\beta)C}{2\sqrt{k_n}\cdot I_{dc_v}}\dot{I}_{out_v}=F_v(I_{out_v},I_{out_w},I_{ext})\\
\frac{(2+\beta)C}{2\sqrt{k_n}\cdot I_{dc_w}}\dot{I}_{out_w}=F_w(I_{out_v},I_{out_w})
\end{cases}
\end{equation}
where $I_{dc_v}=80nA$, $I_{dc_w}=a\cdot I_{dc_v}=6.4nA$, $F_v$ and $F_w$ are functions given by:
\begin{equation}\label{eq:5_33}
\begin{cases}
F_v(I_{out_v},I_{out_w},I_{ext})=I_{out_v}-\frac{I_{out_v}^3}{I_bI_x}-I_{out_w}+I_{ext}\\
F_w(I_{out_v},I_{out_w})=(I_{out_v}+I_{c}-\frac{I_dI_{out_w}}{I_x})
\end{cases}
\end{equation}
where $I_b=3 uA$, $I_c=0.7 uA$, $I_d=0.8 uA$ and $I_x=1 uA$.

\begin{figure*}[t]
\vspace{-20pt}
\normalsize
\centering
\includegraphics[trim = 0.1in 0.1in 0.1in 0.1in, clip, width=5in]{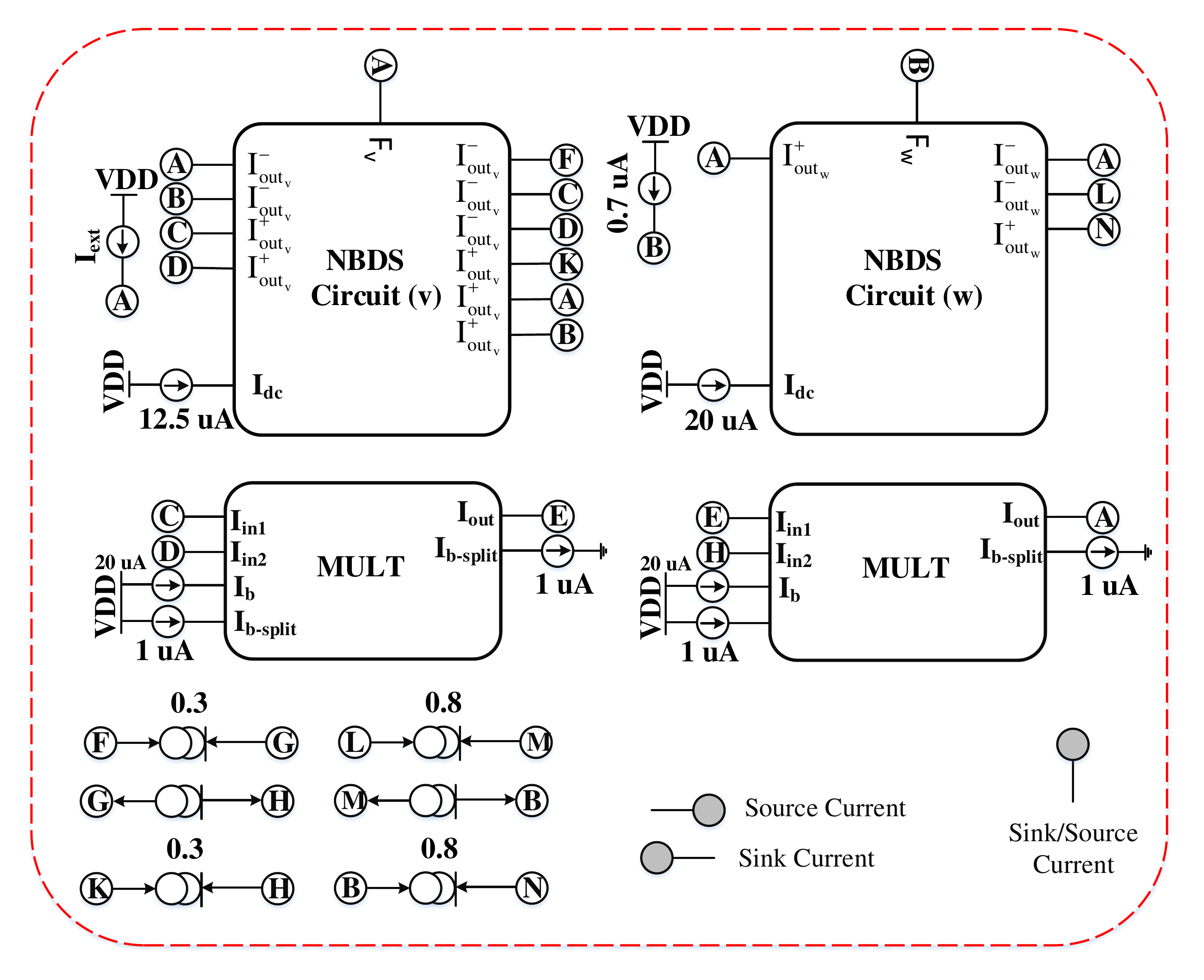}
\vspace{-5pt}
\captionsetup{font=footnotesize}
\caption{A block representation of the total circuit implementing the 2--D FHN neuron model.}
\vspace{-15pt}
\label{fig:51_7}
\end{figure*}

\par Schematic diagrams for the FHN neuron model is seen in Figure \ref{fig:51_7}, including the symbolic representation of the basic TL blocks
introduced in the previous sections. According to these diagrams, it is observed how the mathematical model described in (\ref{eq:5_31}) is mapped onto the proposed electrical circuit. The schematic contains two NBDS circuits implementing the two dynamical variables, followed by two MULT and current mirrors realizing the dynamical functions. As shown in the figure, according to (\ref{eq:5_31}), proper bias currents are selected and the correspondence between the biological voltage and electrical current is $V\iff uA$.

\begin{figure*}[t]
\vspace{-20pt}
\normalsize
\centering
\includegraphics[trim = 0.1in 0.1in 0.1in 0.1in, clip, width=5.5in]{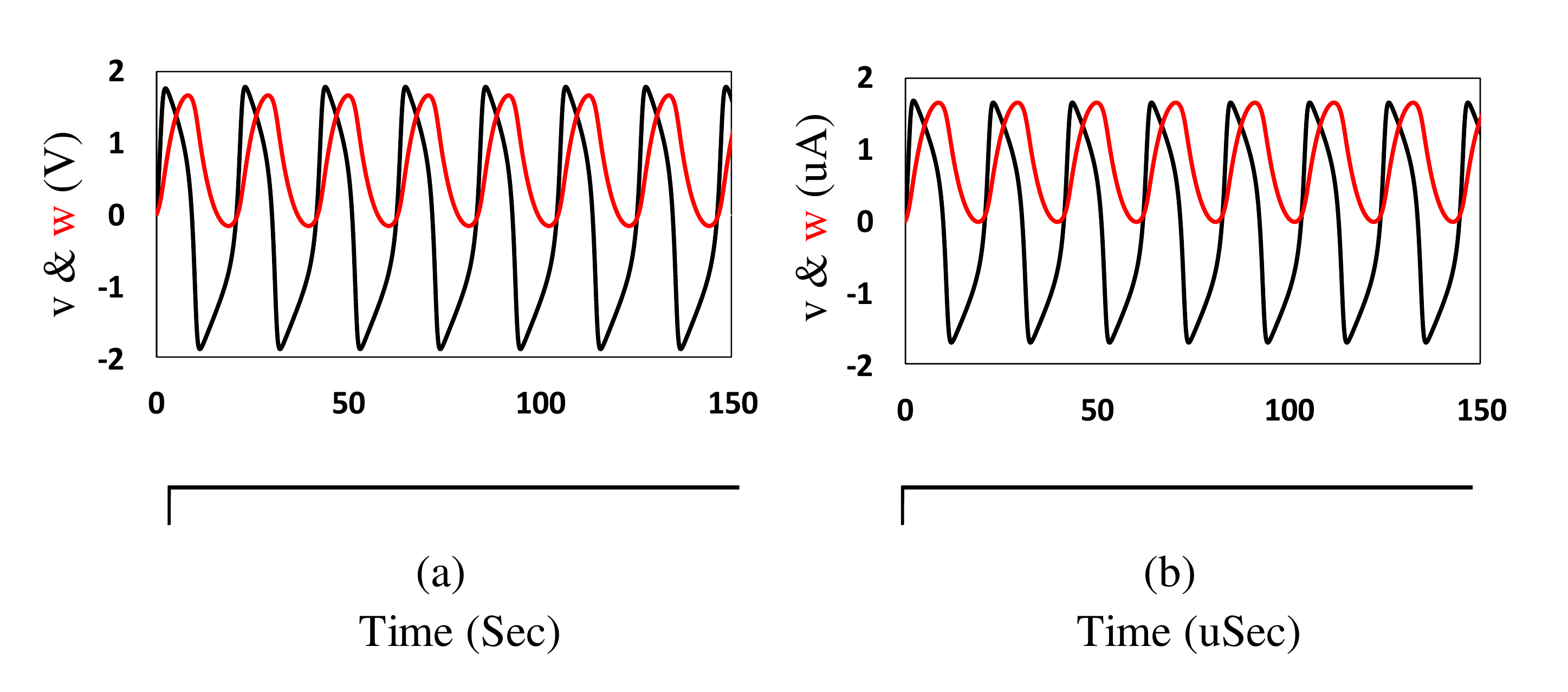}
\vspace{-5pt}
\captionsetup{font=footnotesize}
\caption{Time-domain representations of regular spiking for (a) for MATLAB and (b) Cadence respectively.}
\vspace{-15pt}
\label{fig:51_8}
\end{figure*}

\par Here, we demonstrate the simulation--based results of the high speed circuit realisation of the FHN neuron model. The hardware results simulated by the Cadence Design Framework (CDF) using the process parameters of the commercially available AMS 0.35 $\mu m$ CMOS technology are validated by means of MATLAB simulations as shown in Figure \ref{fig:51_8}. For the sake of frequency comparison, a regular spiking mode is chosen. Generally, results confirm an acceptable compliance between the MATLAB and Cadence simulations while the hardware model operates at higher speed (almost 1 million times faster than real--time). Table \ref{table:51_1} summarizes the specifications of the proposed circuit applied to this case study. As shown in the table, the circuit uses a higher $V_b$ compared to the subthreshold version to force the circuit to operate in strong inversion region. This comes at the expense of higher power consumption (95000 times higher than the subthreshold version). 

\begin{table}[t]
\captionsetup{font=footnotesize}
\caption{Electrical Parameter Values for the Simulated FHN Neuron Model operating in strong inversion.}   
\centering          
\begin{tabular}{c c}    
\hline\hline                        
Specifications & Value\\ [0.5ex]  
\hline                      
Power Supply (Volts)& 3.3\\
Bias Voltage (Volts)& 3.3\\
Capacitances (pF)& 800\\
$\frac{W}{L}$ ratio of PMOS and NMOS Devices ($\frac{\mu m}{\mu m}$)& $\frac{12}{1}$ and $\frac{10}{1}$\\
Static Power Consumption ($m W$)& 8.94\\
\hline          
\end{tabular}
\label{table:51_1}    
\end{table}
\chapter{A High GOPs/Slice Time Series Classifier for Portable and Embedded Biomedical Applications}
\renewcommand{\baselinestretch}{\mystretch}
\par Modern wearable rehabilitation devices and health support systems operate by sensing and analysing human body activities. The information produced by such systems requires efficient methods for classification and analysis. Deep learning algorithms have shown remarkable potential regarding such analyses, however, the use of such algorithms on low--power wearable devices is challenged by resource constraints. Most of the available on--chip deep learning processors contain complex and dense hardware architectures in order to achieve the highest possible throughput. Such a trend in hardware design may not be efficient in applications where on--node computation is required and the focus is more on the area and power efficiency as in the case of portable and embedded biomedical devices. The aim of this paper is to overcome some of the limitations in a current typical deep learning framework and present a flexible and efficient platform for biomedical time series classification. Here, throughput is traded off with hardware complexity and cost exploiting resource sharing techniques. This compromise is only feasible in systems where the underlying time series is characterised by slow dynamics as in the case of physiological systems. A Long-Short-Term-Memory (LSTM) based architecture with ternary weight precision is employed and synthesized on a Xilinx FPGA. Hardware synthesis and physical implementation confirm that the proposed hardware can accurately classify hand gestures using surface--electromyographical time series data with low area and power consumption. Most notably, our classifier reaches 1.46$\times$ higher GOPs/Slice than similar state of the art FPGA--based accelerators.

\section{Introduction}
Recognizing internal activities of the human body based on biologically generated time series data is at the core of technologies used in wearable rehabilitation devices \cite{patel} and health support systems \cite{mazilu}. Some commercial examples include fitness trackers or fall detection devices. Wearable activity recognition systems are generally composed of sensors, such as accelerometers, gyroscopes or magnetic field/chemical sensors \cite{bulling} and a processor used to analyze the generated signals. Real--time and accurate interpretation of the recorded physiological data from these devices can be considerably helpful in prevention and treatment of a number of diseases  \cite{oscar}. For instance, patients with diabetes, obesity or heart disease are often required to be closely monitored and follow a specific exercise set as part of their treatments \cite{Jia}. Similarly, patients with mental pathologies such as epilepsy can be monitored to detect abnormal activities and therefore prevent negative consequences \cite{Yin}.

\par However, most current commercial products only offer relatively simple metrics, such as step count or heart beat.  Further, they lack the complexity and computing power for many time series classification problems of interest in real time. The emergence of deep learning methodologies capable of learning multiple layers of feature hierarchies and temporal dependencies in time series problems and increased processing capabilities in wearable technologies lay the ground work to perform more detailed data analysis on--node and in real time. The ability to perform more complex analysis, such as human activity classification, on the wearable device/node could potentially relax the burden of data streaming from the device to host and thus save data bandwidth link. This bandwidth saving is more visible in the cases where the classification task should be continuously preformed on the patient such as in seizure detection for epileptic patients or continuous imaging and classification of bladder, liver and heart using ultrasound machines. However, due to the high computational power and memory bandwidth required by deep learning algorithms, full realization of such systems on wearable and embedded medical devices is still challenging.

\par On the other hand, the recent development focus of deep learning hardware accelerators has been mainly on achieving the highest possible throughput. This is to keep up with the real--time requirements of complex and embedded machine learning algorithms. In such systems, a large number of Multiply Accumulate (MAC) processors are embedded into the architecture in order to achieve the highest possible throughput. Such strategies in hardware design would lead to large area ($\sim$600~mm$^2$) and power ($\sim$500~W) requirements \cite{GPU} \cite{Google} \cite{Intel}. In this paper we propose an alternative approach based on the observation that most physiological time series are slow, ranging between 0--500 Hz. In such systems, since the real--time classification rate is low, the throughput can be traded off with hardware complexity using resource sharing techniques. Therefore, we introduce a low cost time series classifier based on Long-Short-Term-Memory (LSTM) \cite{hochreiter} networks which is ideally suited for portable and embedded biomedical applications. As a case study, we examine the performance of the proposed hardware on a hand gesture database \cite{geng} recorded by instantaneous surface electromyography (sEMG). Furthermore, as in the case of deep--networks, our hardware classifier can be also applied to post--feature extraction.
\par The rest of the paper is organized as follows: in Section 2, the proposed hardware--oriented classifier is described in detail, while in Section 3, the performance of the classifier when sweeping free parameters is measured on a synthetic database. Sections 4 and 5 investigate high level and detailed structure of the proposed hardware classifier respectively. FPGA implementation results are presented in Section 6 and the proposed classifier is applied to a biomedical case study in Section 7.

\section{Hardware--oriented Time--series Classifier}
LSTM networks are very powerful Recurrent Neural Networks (RNN) that explicitly add memory gates \cite{hochreiter}. This makes the training procedure more stable and allows the model to conveniently learn both long and short--term dependencies. There are some variations on the LSTM architecture, however in this paper we use the following model \cite{schmid}:
\begin{equation}
\begin{cases}
\bm h^f_{n+1}=\sigma( \bm{W}_f^T \cdot\bm x_{n}+\bm b_f) \\
\bm h^i_{n+1}=\sigma( \bm{W}_i^T\cdot\bm x_n+\bm b_i)\\
\bm h^o_{n+1}=\sigma( \bm{W}_o^T\cdot \bm x_n+\bm b_o)\\
\bm h^c_{n+1}=\tanh( \bm{W}_c^T\cdot \bm x_n+\bm b_c)\\
\bm c_{n+1} = \bm{h}^f_n \circ \bm{c}_n+\bm{h}_n^c \circ \bm h\bm_n^i;\\
\bm h_{n+1}=\bm h^o_n \circ \tanh(\bm c_n);\\
\bm y_{n+1}= \bm W^T \bm h_n +\bm b_y
\end{cases}
\end{equation}
where $\bm{x}_n=[\bm{h}_n,\bm u_n]$ and $\bm u_n$, $\bm{h}_n$ and $\bm{c}_n$ are the input, output and cell state vectors respectively at discrete time index, $n$.  The operator $\circ$ denotes the Hadamard element by element product. The variables $\bm h^f_n$, $\bm h^i_n$, $\bm h^o_n$ represent the forgetting, input and output gating vectors. The parameter $\bm y_n$ is a fully connected layer following the LSTM block and serves as the network output and its size is determined by the number of output classes and as well as the number of hidden neurons. Finally, $\bm W_f$, $\bm W_i$, $\bm W_o$, $\bm W_c$ ,$\bm W_y$ and $\bm b_f$, $\bm b_i$, $\bm b_o$, $\bm b_c$, $\bm b_y$ are the weights and biases for the different layers, respectively.

\par One of the main bottlenecks for the hardware realization of RNNs and convolutional neural networks (CNNs) is the large memory size and bandwidth required to fetch weights in each operation. To alleviate the need for such high bandwidth memory access, we investigate two quantization methods (binary and ternary) introduced in \cite{hubara} \cite{li} to quantize weights embedded in the network architecture. As the changes during gradient descent are small, it is important to maintain sufficient resolution otherwise no change is seen during the training process, therefore the real--valued gradients of weights are accumulated in real--valued variables. We also set the bias values to zero to achieve further efficiency in hardware realization while delivering an acceptable classification accuracy for the experimented biomedical case study. Such quantization methods can be considered as a form of regularization that can help the network to generalize.  In particular, the binary and ternary quantization are a variant of Dropout, in which weights are binarized/ternarized instead of randomly setting part of the activations to zero when computing the parameter gradients \cite{hubara}.

\par The quantization of weights in the forward path must be also reflected in the calculation of the gradient descent. Here, we use the version of the straight--through estimator introduced in \cite{hubara} that takes into account the saturation effect. Consider the sign and round functions for binary and ternary quantization respectively as follows:
\begin{equation}
\begin{cases}
q_b=\text{sign}(r)\\
q_t=\text{round}(r)
\end{cases}
\end{equation}
where $r=W_x$ and assume that estimators $g_{q_b}$ and $g_{q_t}$ of the gradients $\frac{\delta C}{\delta q_b}$ and $\frac{\delta C}{\delta q_t}$ ($C$ is the cost function) are derived. Then, the straight--through estimators of $\frac{\delta C}{\delta r}$ are:
\begin{equation}
\begin{cases}
g_{r_b}=g_{q_b}1_{\mid r\mid\leq 1}\\
g_{r_t}=g_{q_t}1_{\mid r\mid\leq 1}
\end{cases}
\end{equation}
\par This implies that the gradient is applied to the weights if their real values are between 1 and -1 otherwise the gradient is cancelled when $r$ is outside the range. The overall architecture of the proposed LSTM--based classifier is shown in Fig.~\ref{fig:arch}. In this approach, the sequential target replication technique inspired from \cite{lipton} is used during the training phase. However our approach is slightly different. In the proposed architecture, during forward path and for all $steps$, the same output label is used to calculate the output error. The error is stored $s$ times in memory in order to calculate the gradient descent during the backward path. In the backward pass, the RNN is unrolled back in time and the weights are updated. This technique forces the network to better memorize the previous sequences of the input windows \cite{andrej}. It should be also noted that during the training phase loss values are averaged over all steps, while at the inference (test) time, the output at the final step is chosen as the actual classification value.

\begin{figure}[t]
\centering
\includegraphics[trim = 0.3in 0.1in 0.1in 0.1in, clip, width=4.2in]{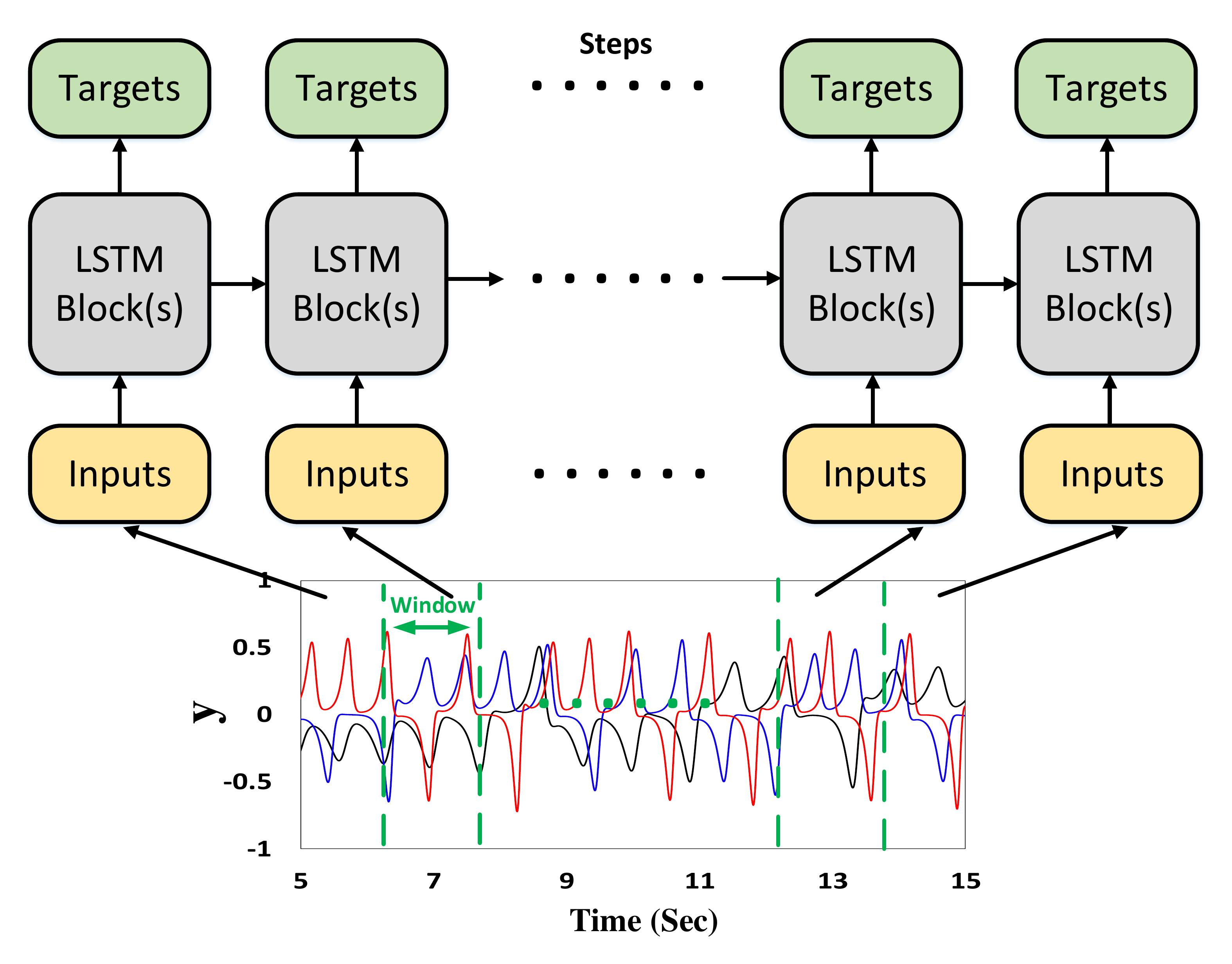}
\captionsetup{font=footnotesize}
\caption{The overall architecture of the proposed LSTM--based classifier along with three input synthetic time series reproduced by the Lorenz attractor dynamical systems. A portion of the input signal termed as $window$ is highlighted in green color, sequentially fed to the systems after $steps$ times.}
\vspace{-10pt}
\label{fig:arch}
\end{figure}

\begin{figure}[t]
\vspace{-20pt}
\centering
\includegraphics[trim = 0.0in 0.0in 0.0in 0.0in, clip, width=4.0in]{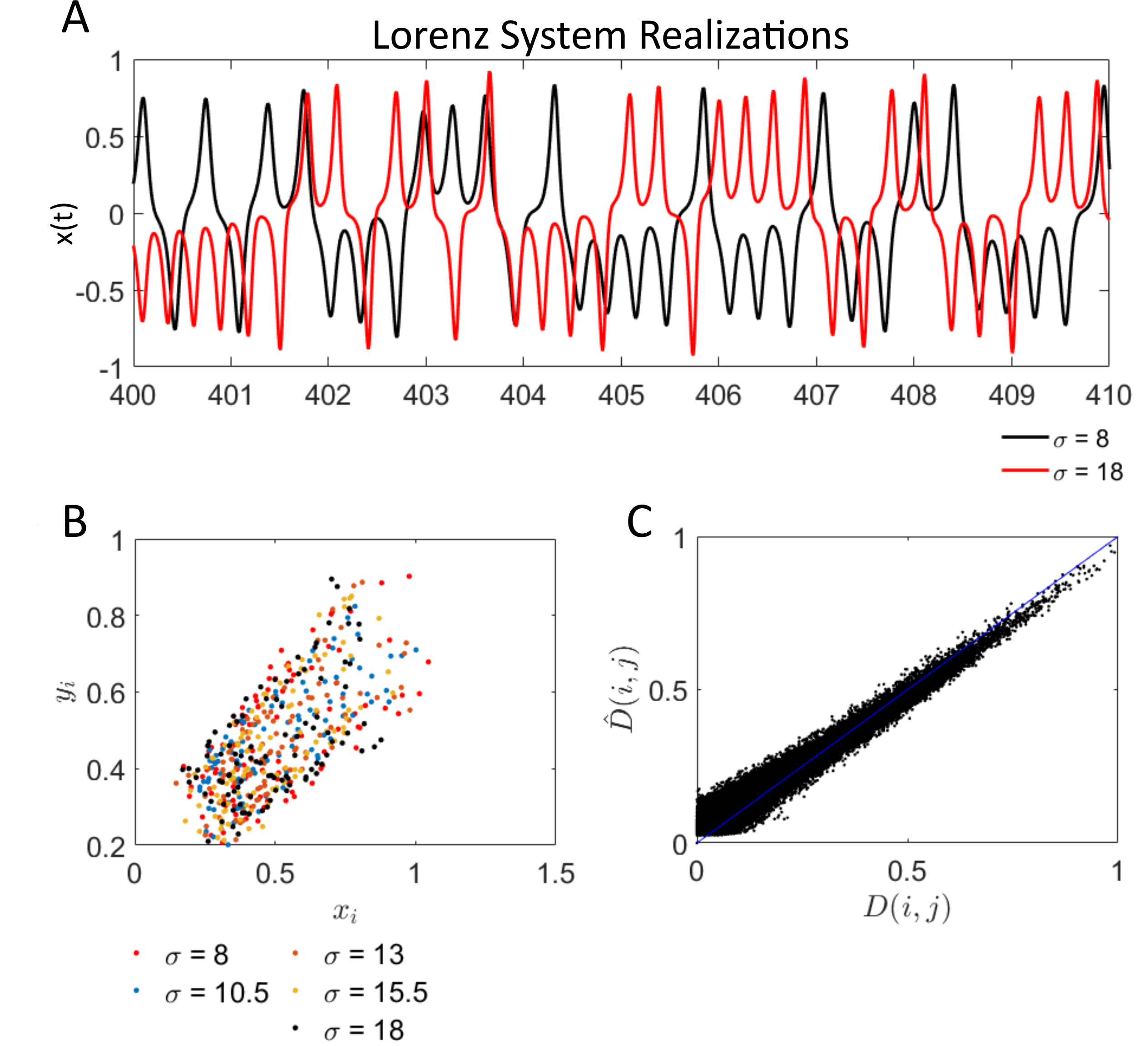}
\caption{(A) A realization of the lorenz system for $\sigma = 8$ (red) and $\sigma = 18$ (black).  The other parameters were taken to be $\rho=28$ and $\beta =\frac{5}{3}$. The $x(t)$ variable is plotted, downscaled by a factor of 40 such that it is bounded in the interval $[-1,1]$ for subsequent network training. (B)  The final $(x_i,y_i)$ generated by the Fourier Spectrum and the visualization procedure outlined in the text.  The data displays no clear clustering in the Fourier domain with a 2-dimensional projection for the 5 classes considered.  (C)  A plot of $\hat{D}(i,j)$ vs $D(i,j)$.  The computed correlation coefficient for the Lorenz data was $ p = 0.9826$, indicating that the Fourier domain data is well described as lying on a 2-D manifold.}
\label{fig:visual}
\end{figure}

\section{Hardware--oriented Simulations}
In order to measure the performance of each quantization method, we synthetically produce and classify a number of time series as a preliminary proof of principle. In this example, we generate synthetic time series data from well known chaotic dynamical systems in different parameter regimes. In all parameter regimes, these systems have chaotic attractors. The unique and isolated parameter regime corresponds to discrete classes. The LSTM network is tasked to classify the resulting time series as belonging to a unique class. The considered dynamical systems are the logistic (discrete time) and Lorenz systems (continuous time). The logistic map is given by:
\begin{eqnarray*}
x_{n+1}=r\cdot x_n(1-x_n)
\end{eqnarray*}
where $x$ is the state variable and $r$ is a parameter \cite{may}.  The Lorenz system is given by:
\begin{eqnarray*}
\dot{x}&=&\sigma(y-x)\\
\dot{y}&=&x(\rho-z)-y\\
\dot{z}&=&xy-\beta z
\end{eqnarray*}
where $x$, $y$ and $z$ are state variables and $\sigma$,~$\rho$,~$\beta$ are parameters \cite{lorenz}.
\par By sweeping $r$ in the Logistic map and $\sigma$ in the Lorenz attractor, various responses can be observed (Figure 2A). The time series data generated by the chaotic dynamical systems must be similar between the classes in terms of both time and frequency features so that the signals are not easily distinguishable by the classifier. To determine how similar the synthetic data classes are and visualize our synthetic data set in a simple way, we computed a distance matrix in the Fourier domain as
\begin{eqnarray}
D(i,j) = \int_{\omega} \left(|P_i(\omega)|-|P_j(\omega)|\right) ^2\,d\omega
\end{eqnarray}
where $P(\omega)$ is the logarithm of the power of the Fourier transforms of $\frac{x_i(t)}{20}$, for $i=1,2,\ldots N_S$ realizations for a sample of $N_S=100$ realizations per discrete parameter class.  The factor of $1/20$ multiplying $x_i(t)$ bounds the trajectories in the unit interval for subsequent learning in the LSTM network. Then, we looked for a set of points in $\mathbb{R}^2$, $(x_i,y_i)$ (Figure 2B) such that $$\hat{D}(i,j) = (x_i-x_j)^2 - (y_i-y_j)^2$$ by stochastically minimizing the sum:
$$ E = \sum_{i\neq j } \left(\hat{D}(i,j) - D(i,j) \right)^2$$
This approach finds low dimensional (in this case 2D) manifolds that the data may lie on.  Alternatively, one may also use the singular value decomposition, however we do not take that approach here.  The stochastic minimization occurs by initializing the $(x_i,y_i)$ from a joint uniform distribution on $[0,1]^2$, randomly perturbing every point to compute $E$.  The perturbations were drawn from a normal distribution with mean 0 and standard deviation $\eta = 10^{-3}$.  At each time step, the network computes $E$ after $(x_i,y_i)$ have been perturbed and compares $E$ to the smallest value of $E$ so far, $E^*$.  If $E<E^*$, then we set the new $E^* = E$ and keep the perturbed $(x_i,y_i)$.  If $E^*<E$, we disregard the perturbation and iterate.  The results of this process are shown in Fig.~\ref{fig:visual}A-C for the Lorenz system without noise. The results demonstrate that the data has no readily visible clusters in a 2-dimensional projection, however clustering may appear in a higher dimensional projection. In order for the network to generalize the input features better, a uniform random noise is added to the training and test data. Since binarization is a form of regularization \cite{li}, we do not use other regularization methods such as Dropout. All weights are initialized by random numbers with normal distribution. The same analysis applied to the Logistic map showed that the generated time series are separable in the Fourier domain, however still difficult to visually classify in the time domain.

\begin{figure}[t]
\centering
\includegraphics[trim = 0.3in 0.1in 0.3in 0.1in, clip, width=4.0in]{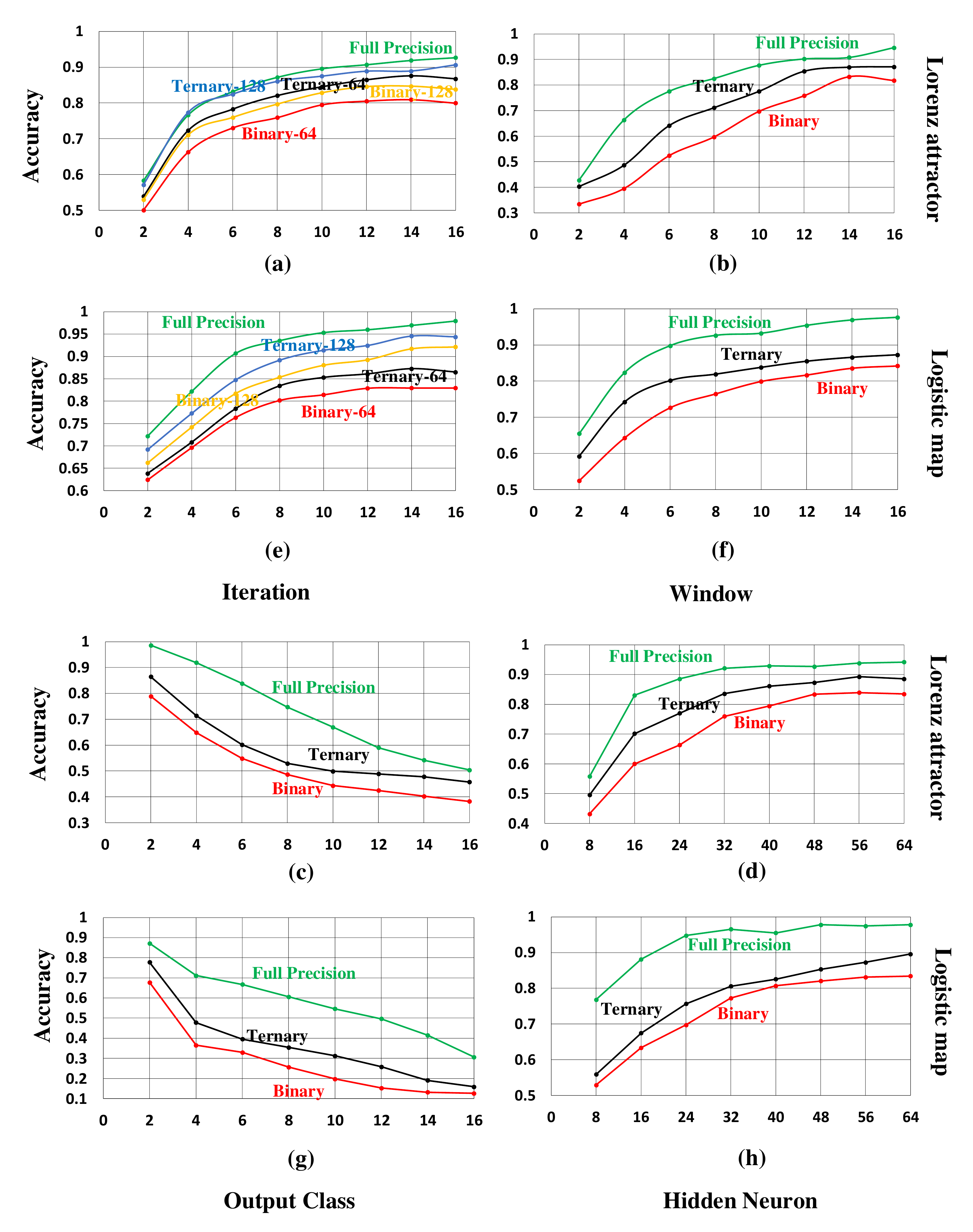}
\captionsetup{font=footnotesize}
\caption{Free parameter sweeping for three networks with various weight precision on two synthetic database extracted from Lorenz (a--d) and logistic map (e--h). The networks' accuracies are changed by varying $iteration$ in (a) and (e), $window$ size in (b) and (f) , $output~class$ in (c) and (g) and $hidden~neuron$ in (d) and (h). In addition, two sets of binary and ternary networks are compared in (a) and (e).}
\vspace{-10pt}
\label{fig:sweep}
\end{figure}

\par To compare the performance of the network with different weight precision, experiments with different free parameters were performed on the database. The free parameters are defined as follows:
\begin{itemize}
  \item Window~($\omega_s$):  the length of each part of the input time series fed to the network to be classified in the output. The length of input signal ($\bm u$) is equal to $M \cdot \omega_s$ where $M$ is the dimension of the input signal. For example, $M$ for data collected from a three dimensional gyroscope is 3 as the input signal is presented to the system by 3 independent time series. A highlighted window sample is shown in Fig.~\ref{fig:arch}.
  \item Iteration ($s$): the number of successive windows that must be introduced to the network sequentially so that the network can classify the input signals properly. The partition of the input signals into window sizes and then introducing them to the network sequentially would allow the RNNs to use recurrent feedback and internal memories to make decisions, leading to a significant reduction in hardware area consumption.
  \item Output~Class~($N_y$): the number of classes that the network must classify based on the input signals.  We can specify this by considering more discrete parameter sets in our chaotic systems.
  \item Hidden~neurons~($N_h$): the number of neurons embedded in the network. Accuracy increases with the number of hidden neurons at the expense of higher hardware cost.  After a certain point, there are diminishing returns in increasing the number of hidden neurons.
\end{itemize}
\par In these experiments, the LSTM network takes a sequence of continuous/discrete arrays defined by the $\omega_s$ size as input, and after $s$ steps classifies it into one of the output classes. The training objective (loss function) is the cross--entropy loss over all target sequences as follows:
\begin{equation}
L_i=-\log(p_{y_i});~p_k=\frac{e^{y_{i}}}{\sum_j e^{y_{j}}}
\end{equation}
where $k$ is the array of class scores for a single example and $p$ is a vector of the output normalised probabilities. In order to backpropagate the output error to all layers, $\frac{\delta L_i}{\delta f_{k}}$ can be derived using chain rule as follows:
\begin{equation}
\frac{\delta L_i}{\delta f_{k}}=p_k-1(y_i=k).
\end{equation}
\par Adagrad is used as the learning algorithm with learning rate of 5e-2 \cite{Duchi}. The weights are randomly drawn from a uniform distribution in [-0.01,0.01]. After each iteration, the gradients are clipped to the range [-5,5]. The results of sweeping on the free parameters of the test synthetic database for the binary, ternary and full precision networks are shown in Fig.~\ref{fig:sweep}. As can be seen in Fig.~\ref{fig:sweep} (a) and (e), by increasing $s$, the accuracy of the classifier increases at the expense of longer latency and higher power consumption for the hardware realization. It can be also observed that the quantized networks with the same number of neurons (64) can classify the input signals with a lower accuracy rate. However, this reduction in the accuracy can be compensated by increasing the number of hidden neurons. For example, 128 neurons in the quantized network have similar performance compared to 64 neurons in a full precision network. Although requiring more neurons in a quantized network,  a significant hardware efficiency improvement can be still seen. It is also observed that the ternary network possesses better accuracy performance compared to its binarized counterpart thereby confirming the results in \cite{li}.

\par The effect of varying window size $\omega_s$ on the accuracy for all networks are shown in Fig.~\ref{fig:sweep} (b) and (f). Increasing $\omega_s$, increases the accuracy for all networks but this imposes a higher hardware cost in terms of area and power. Thus, by increasing the length of the scanned input signals, the number of operations and the memory bandwidth per input increase. Moreover, these plots show that the accuracy drops as quantization is applied to the weights. Again, this reduction in accuracy can be compensated by increasing the number of hidden neurons while still achieving better hardware area performance compared to the full precision.
\par Fig.~\ref{fig:sweep} (c) and (g) show that the accuracy of the classifier drops in all networks if the number of output classes increases. However, similar to the previous experiments this reduction in accuracy can be compensated up to a certain point by increasing the number of hidden neurons as observed in Fig.~\ref{fig:sweep}~(d) and (h).

\section{Finite State Machine}
\begin{figure}[t]
\vspace{-10pt}
\centering
\includegraphics[trim = 0.1in 0.1in 0.1in 0.1in, clip, width=3.2in]{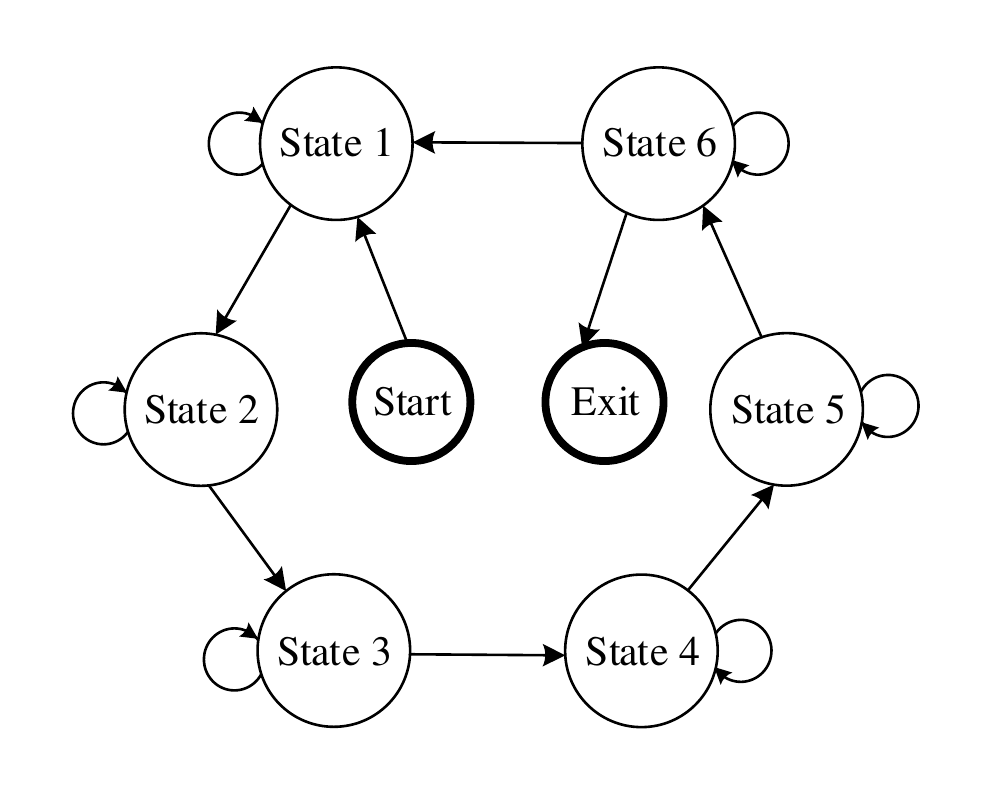}
\vspace{-10pt}
\captionsetup{font=footnotesize}
\caption{Finite state machine block diagram for the proposed hardware architecture. The system iterates through six states and
only one state is active at a time.}
\vspace{-10pt}
\label{fig:FSM}
\end{figure}

As illustrated in Fig.~\ref{fig:FSM}, the proposed hardware classifier functions as a finite state machine that iterates through six states and only one state is active at a time. This structure can also be implemented in a pipelined form with multiple active states and higher throughput at the expense of increased power and area consumption. The general functionality of each state is briefly described as follows:
\par $\textbf{State~1:}$ After initialization at the start state, the first input transmission according to the defined $\omega_s$ is carried out and the system enters the first state where $\bm W_f^T \bm x_1$, $\bm W_i^T \bm x_1$, $\bm W_o^T \bm x_1$ and $\bm W_c^T \bm x_1$ are independently calculated each with eight additions per clock cycle. A total of  $[N_h+\omega_s ]\times \frac{N_h}{8}$ clock pulses are needed for this state to finish the calculations.
\par $\textbf{State~2:}$ At the end of the calculations, the system enters the second state. In this state, according to (1), nonlinear functions ($\sigma(.)$ and $\tanh(.)$) are applied to the previous values fetched from memories. After $N_h$ clock pulses, the system enters the next state.
\par $\textbf{State~3:}$ In this state, using two embedded multipliers, the value of variable $\bm c$ is calculated in $N_h$ clock pulses. The critical path of the proposed architecture is limited by this state which can be alleviated by using pipelined or serial multipliers at the expense of increased latency and hardware cost.
\par $\textbf{State~4:}$ As the variable $c$ is provided, the system enters this state where the $\tanh(.)$ function is applied to the previous values, taking $N_h$ clock pulses, then the system enters the next state.
\par $\textbf{State~5:}$ In this state, using one of the two embedded multipliers, the value of variable $h$ is calculated in $N_h$ clock pulses and the system enters the next state.
\par $\textbf{State~6:}$ Finally, if the number of scan times is equal to $s$, by calculating $\bm W_y^T \bm h$ in $N_h\times N_y$ clock pulses, the system determines the classified output and exits, otherwise enters State 1. This process is repeated for each window of the input signal(s) and managed by a master controller circuit, embedded into the system.

\section{Hardware Architecture}
\par As stated before, the aim of this paper is to present a flexible and efficient hardware--based time series classifier that exploits the slow nature of physiological signals for reducing hardware complexity and cost. In such systems, given that the classification rate is low, a few high speed processor modules are enough to share the computational burden in different layers. This would also allow us to actively and efficiently reconfigure the system according to the user's specifications for the set number of neurons, window size, iterations and input/output classes. The architecture of the proposed system is shown in Fig.~\ref{fig:detail} in which the hardware modules (maximum 64 operations per clock cycle) are shared through a 96--bit bus. It should be stressed that since the accuracy of the ternary network is generally higher than its binary counterpart as shown in Fig.~\ref{fig:sweep}, in the hardware implementation, the ternary quantization is used and two bits are allocated for storing each weight value. In the following, the architecture of each hardware module is explained in detail:
\par $\textbf{WBs~(Weight~Banks):}$ This block contains five sets of buffers to store the truncated 2--bit weights ($\omega_f,~\omega_i,~\omega_o,~\omega_c$, and $\omega_y$), trained off-line. The $WBs$ module is able to read maximum 64 bits in each clock cycle. The utilised volume of each buffer is defined by the user which is equal to $2N_h(N_h+\omega_s) $ bits. However, the maximum volume of these buffers must be selected based on the available resources on the chip. The greater the volume size, the wider the range of flexibilities for the network/input size. The reading process of this block is controlled by the $MC$ unit and the block is only activated upon its use. It should be noted that, as the proposed architecture is implemented on a Xilinx FPGA in this work, the buffers are realised using block RAMs and the address of each reading operation is provided on the negative clock edge by the $MC$ module. This module is used only in states~1 and 6 of the FSM.

\begin{figure}[t]
\vspace{-10pt}
\centering
\includegraphics[trim = 0.3in 0.1in 0.1in 0.1in, clip, width=3.5in]{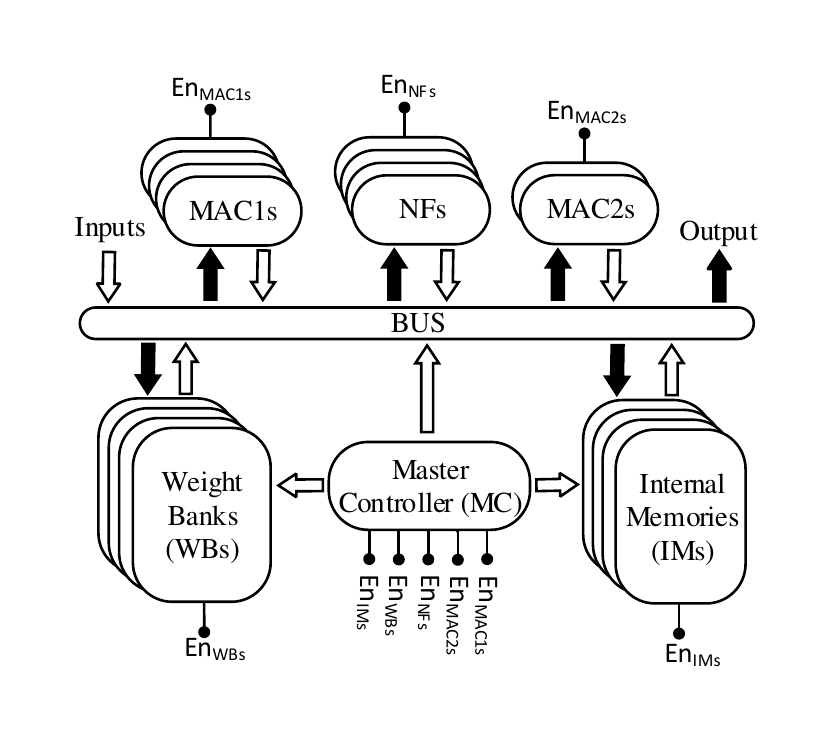}
\vspace{-10pt}
\captionsetup{font=footnotesize}
\caption{The proposed architecture of the system comprising six blocks: MAC1s, NFs, MAC2s, WBs, IMs and MC. Black and white arrows represent output and input signals respectively in accordance with bus connections}
\vspace{-10pt}
\label{fig:detail}
\end{figure}

\par $\textbf{IMs~(Internal~Memories):}$ This block contains seven sets of buffers to store 12--bit values produced by the intermediate stages (${\bm h_f,~\bm h_i,~\bm h_o,~\bm h_c,~\bm h,~\bm c,}$ and $\bm y$). The maximum and utilised volume of the buffers are again determined by the available on--chip memory and the user's specifications respectively. The writing and reading of this block is also controlled by the $MC$ unit and the reading address is provided on the negative clock edge as they are implemented using block RAMs. The maximum data bandwidth of this module is utilised in states~1, 2 and 3 which is 48 bits and the module is active in all states.
\par $\textbf{MAC1s~(Multply--Acummulate~[limitted~precision]):}$ This blocks contains 32 parallel MAC units with limited precision. Each unit computes the product of 12 and 2--bit signed numbers and adds the product to an accumulator. The number of iterations that this unit needs to operate is defined by the user and assigned by the $MC$ block. In state~1, the $k$th value of each intermediate stage is calculated as follows:
\begin{equation}
\begin{cases}
h_{f_k}=\sum_{j=1}^{(N_h + \omega_s)/z} \sum_{i=1}^{z}\bm x _{ss_{(j-1)\times z+i}}\times \omega_{f_{k,(j-1)\times z+i}}\\
h_{i_k}=\sum_{j=1}^{(N_h + \omega_s)/z} \sum_{i=1}^{z}\bm x_{ss_{(j-1)\times z+i}}\times \omega_{i_{k,(j-1)\times z+i}}\\
h_{o_k}=\sum_{j=1}^{(N_h + \omega_s)/z} \sum_{i=1}^{z}\bm x_{ss_{(j-1)\times z+i}}\times \omega_{o_{k,(j-1)\times z+i}}\\
h_{c_k}=\sum_{j=1}^{(N_h + \omega_s)/z} \sum_{i=1}^{z}\bm x_{ss_{(j-1)\times z+i}}\times \omega_{c_{k,(j-1)\times z+i}}.
\end{cases}
\end{equation}
\par The 32 MAC units calculate the internal sum of $z=8$ for each intermediate stage in parallel.
\par Moreover, the $p$th output value is calculated by means of this block in state~6 as follows:
\begin{equation}
y_{s_p}=\sum_{j=1}^{N_h} h_j\times \omega_{y_{p,j}}.
\end{equation}
\par It should be stressed that since the weight values are truncated to 2--bit precision (-1,0,1), the multiplications in this block are simply implemented by multiplexers.
\par $\textbf{MAC2s~(Multply--Acummulate~[full~precision])}:$ Two signed multipliers and a 12--bit adder are embedded in this block and used in states~3 and 5. This module acts as a MAC processor in state~3 and is reset in each cycle. This means that, the two multipliers and the adder are used in a normal operation not necessarily as MACs. In this state, the $k$th value of variable $c$ is sequentially calculated as follows:
\begin{equation}
c_{k}=(h_{f})_k \cdot c_{k}+(h_{i})_k\cdot (h_{c})_k.
\end{equation}
\par In state~5, the multipliers are only used and the $k$th value of the variable $c$ is sequentially calculated as follows:
\begin{equation}
h_{k}=(h_{o})_k\cdot \tanh(c_k).
\end{equation}

\begin{figure}[t]
\centering
\includegraphics[trim = 0.3in 0.1in 0.5in 0.1in, clip, width=3.5in]{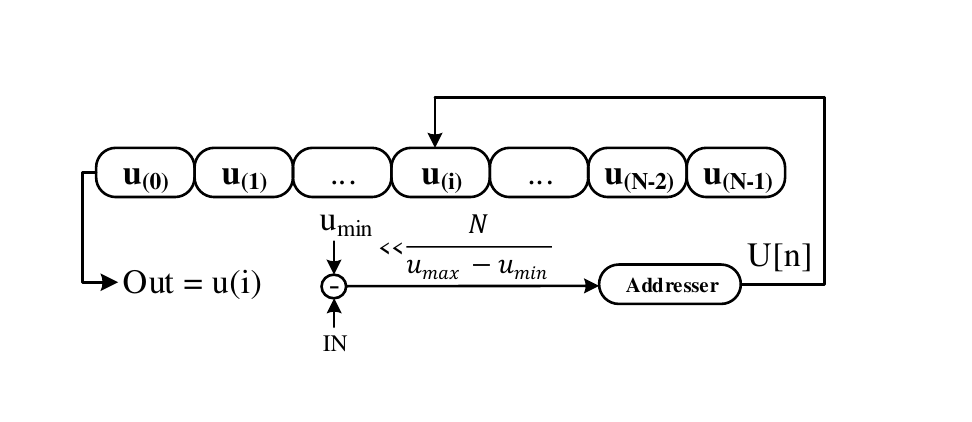}
\vspace{-10pt}
\captionsetup{font=footnotesize}
\caption{The look-up table based architecture of NFs. In each use, according to the input value, the corresponding address ($U[n]$) is generated and output value ($u[i]$) is fetched from the memory.}
\vspace{-15pt}
\label{fig:NF}
\end{figure}

\par $\textbf{NFs~(Nonlinear~Functions)}:$ This block is responsible for the calculation of nonlinear functions ($\sigma(u)$ and $\tanh(u)$) employed in states~2 and 4 where $u$ is the input value of these functions. As shown in Fig.~\ref{fig:NF}, we store $N$ discrete values of each nonlinear function in look-up tables with 10-bit length. The quantity $u[i]$ is the $i$th stored value where $ i\in \textbf{N}\equiv {0,1,...,N-1}$. The address of each stored value is defined as:
\begin{equation}
U[n]= \lfloor \frac{u-u_{min}}{\Delta u} \rfloor ~~~;~~~~u \in [u_{min},u_{max})
\end{equation}
where $\Delta u= \frac{u_{max}-u_{min}}{N}$. If the $u_{max}$ and $u_{min}$ parameters are orders of 2, the division in (12) can be easily performed by arithmetic shifts. Therefore, the values of these parameters are respectively chosen to be 8 and -8 for $\sigma(.)$ function and 4 and -4 for $\tanh(.)$ function. According to (12), by preparing the address, the corresponding output value can be fetched in one clock pulse from the look-up table. In our experimental setup, the value of $N$ is considered to be 64, providing enough accuracy for the calculation of nonlinear functions.
\par $\textbf{MC~(Master~Controller)}:$ According to the defined parameters by the user, this block manages and controls all resources used in the architecture through a shared bus and controlling signals ($En_{MAC1s}$, $En_{MAC2s}$, $En_{NFs}$, $En_{IMs}$ and $En_{WBs}$). In other words, this block actively changes the state of the FSM by assigning proper tasks to the hardware modules and actively turning off the unused modules. For example, in state~1, the $MAC1s$ and $WBs$ modules are only active and others are deactivated. This block also controls the input and output transmissions as required. For example, the input/output ports do not update the new values unless the number of iterations is equal to $steps$.

\begin{table}[t]
\captionsetup{font=footnotesize}
\caption{Performance comparison between the proposed system, implemented the Kintex--7 (XC7K325T) FPGA and other state of the art hardware.}   
\centering          
\resizebox{\columnwidth}{!}{%
\begin{tabular}{c c c c c c}    
\hline\hline                        
  & \textbf{2015\cite{sankaradas}} & \textbf{2016\cite{motamedi}} &\textbf{2016\cite{qiu}}&\textbf{2017\cite{yonekawa}}& \textbf{This Work}\\ [0.5ex]  
\hline                      
\textbf{Precision}& 32bits float&32bits float&16bits float&binary&12bits fixed\\
\textbf{Frequency}& 100 MHz & 100 MHz&150 MHz& 150 MHz&100 MHz\\
\textbf{FPGA Chip}& VX485T& VX485T& XC7Z045&ZU9EG&XC7K325T\\
\textbf{GOPs}& 61.6& 84.2& 187.8&460.8&6.3\\
\textbf{Slice}& 75123& 75924& 52458&47950&447\\
\textbf{GOPs/Slice}& 8.12E-04& 11.09E-04& 35.8E-04&96.1E-04&141.0E-4\\[1ex]        
\hline          
\end{tabular}
}
\label{table:hardware}    
\end{table}

\section{Hardware Results}
\par To verify the validity of the proposed hardware classifier, the architecture designed in the previous section is implemented on a Genesys 2 development system, which provides a high performance Kintex--7 (XC7K325T) FPGA surrounded by a comprehensive collection of peripheral components. The device utilization for the implementation of the proposed hardware is summarized in Table~\ref{table:hardware} along with other state of the art implementations. The focus of all other implementations is mainly on the hardware realization of CNNs; however, as the nature of computations in all deep learning algorithms is the same, for the sake of comparison the implementation results of those studies are included here. The results of hardware implementations show that the proposed classifier reaches 1.46$\times$ higher GOPs/Slice than similar state of the art FPGA--based accelerators. Obviously, less power consumption is also achieved as the number of FPGA slices used in the proposed system is lower than in other state of art hardware. Such a trade off constrains the GOPs factor, which is not critical for most slow biomedical applications. It should be also stressed that the proposed hardware is fully reconfigurable in terms of defined free parameters introduced in Section 2 and only implemented once and reprogrammed for all other cases. The proposed hardware architecture can only implement the networks with single layer LSTM, however, it can be conveniently extended for multilayer LSTM structures by modifying the \textbf{MC} module and increasing the memory volume. The required response time of the system must be seriously considered upon such modifications. For example, by adding another layer to the current design, the amount of calculations is almost doubled, therefore, the number of parallel MAC processors in the \textbf{MAC1s} module must be doubled to keep the response time of the systems constant.

\begin{table}[t]
\captionsetup{font=footnotesize}
\caption{Classification test error rates of the LSTM networks with different weight resolutions/structure and the hardware results trained on the DB-a with 8 output classes.}   
\centering          
\resizebox{\columnwidth}{!}{%
\begin{tabular}{c c c c c c c}    
\hline\hline                        
  \textbf{Model} & \textbf{Learning~Rate}& \textbf{$N_h$} & \textbf{Input~Window}& \textbf{Steps}& \textbf{Loss} &\textbf{Accuracy $\%$}\\ [0.5ex]  
\hline                      
\textbf{Full precision}&0.05& 150 & 10& 15& 0.01 &97.63\\
\textbf{Full precision}&\textbf{0.05}& \textbf{150} & \textbf{5}& \textbf{30}& \textbf{0.01} &\textbf{97.73}\\
\textbf{Ternary}& 0.1& 250 & 5 & 30& 0.08 & 96.41\\        
\textbf{Hardware}& 0.1& 250 & 5 & 30& 0.09& 95.71\\ [1ex]        
\hline          
\end{tabular}
}
\label{table:accuracy_DB_a}    
\end{table}

\begin{table}[t]
\captionsetup{font=footnotesize}
\caption{Classification test error rates of the LSTM networks with different weight resolutions/structure and the hardware results trained on the DB-c with 12 output classes.}   
\centering          
\resizebox{\columnwidth}{!}{%
\begin{tabular}{c c c c c c c}    
\hline\hline                        
  \textbf{Model} & \textbf{Learning~Rate}& \textbf{$N_h$} & \textbf{Input~Window}& \textbf{Steps}& \textbf{Loss} &\textbf{Accuracy $\%$}\\ [0.5ex]  
\hline                      
\textbf{Full precision}&0.05& 250 & 15& 10& 0.06 &95.47\\
\textbf{Full precision}&\textbf{0.05}& \textbf{250} & \textbf{10} & \textbf{15}& \textbf{0.04} &\textbf{95.85}\\
\textbf{Ternary}& 0.1& 350 & 10 & 15& 0.12 & 94.52\\        
\textbf{Hardware}& 0.1& 350 & 10 & 15& 0.14& 93.83\\ [1ex]        
\hline          
\end{tabular}
}
\label{table:accuracy_DB_c}    
\end{table}

\begin{figure}[t]
\centering
\includegraphics[trim = 0.4in 0.1in 0.4in 0.1in, clip, width=5.4in]{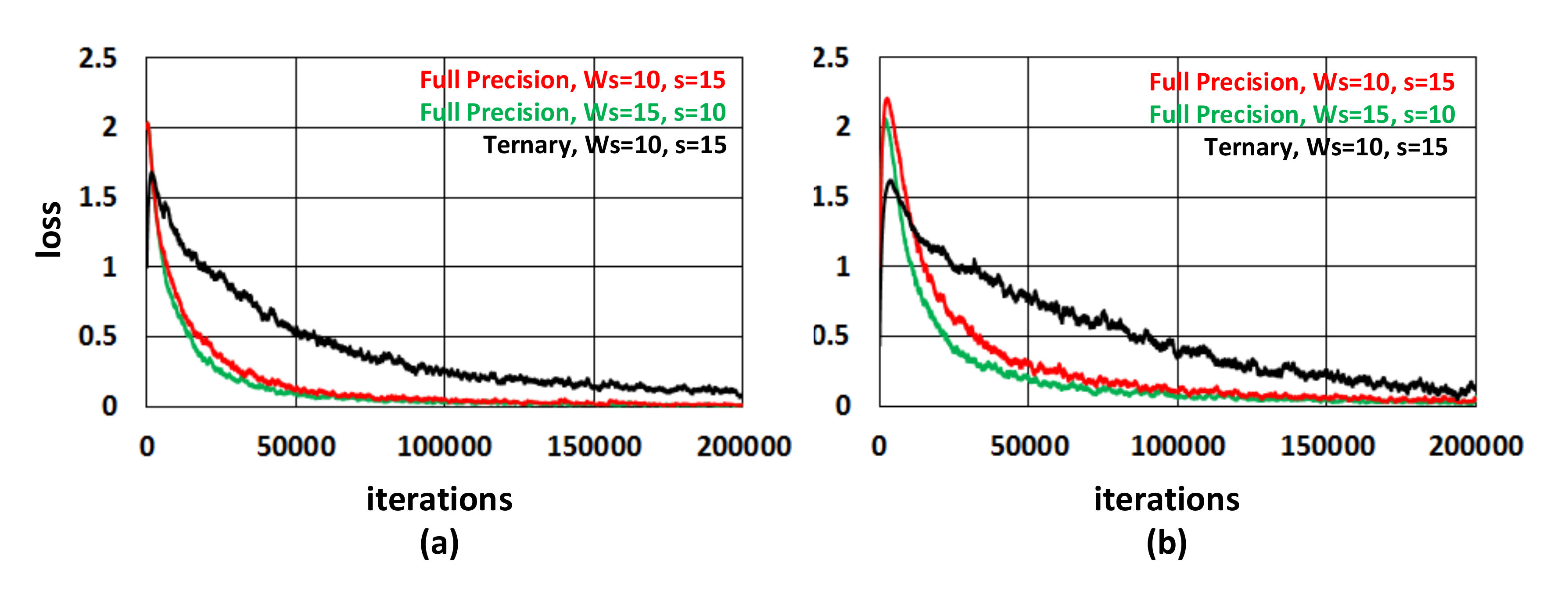}
\captionsetup{font=footnotesize}
\caption{Training loss traces for various network structures on (a) DB--a and (b) DB--c experiment. It is evident that the ternary network converges to its final value slower that full precision networks.}
\label{fig:loss}
\end{figure}

\begin{table}[t]
\captionsetup{font=footnotesize}
\caption{Confusion Matrix for DB-a database with 128 input time series and 8 output classes.}   
\centering          
\resizebox{\columnwidth}{!}{%
\begin{tabular}{c c c c c c c c c}    
\hline\hline                        
  \textbf{Output Class} & \multicolumn{8}{c}{\textbf{Target Class}}\\ [0.5ex]  
  \hline                      
 &\textbf{1}& \textbf{2} & \textbf{3}& \textbf{4}& \textbf{5} &\textbf{6}& \textbf{7} &\textbf{8} \\
\hline                      
\textbf{1}&\textbf{98.91}& 0.03 & 0.02 & 0.01& 0 &0.43& 0.01 &0.07\\
\hline                      
\textbf{2}&0.15& \textbf{96.47} & 1.02& 2.05& 0.01 &0 & 0.01 &0.27\\
\hline                      
\textbf{3}&0.14 & 0.79 & \textbf{97.80}& 0.34 & 0.53 & 0 &0.02&0.34\\
\hline                      
\textbf{4}&0& 0.57 & 0.04& \textbf{95.60} & 0.52 &0&0.26&2.98\\
\hline                      
\textbf{5}&0&0.39&0.30&2.68&\textbf{94.71} &0&0.33&1.57\\
\hline                      
\textbf{6}&0.34&0&0.02&0&0.04&\textbf{99.32} &0.26&0\\
\hline                      
\textbf{7}&0.01&0.10&0.09&0&0.02&0.36&\textbf{99.31}&0.08\\
\hline                      
\textbf{8}&0&0.07& 0.06&0.79&0.02&0.14&0.15&\textbf{98.73}\\
[1ex]        
\hline          
\end{tabular}
}
\label{table:confusion_db_a}    
\end{table}

\begin{table*}[t]
\captionsetup{font=footnotesize}
\caption{Confusion Matrix for DB-c database with 128 input time series and 12 output classes.}   
\centering          
\resizebox{\columnwidth}{!}{%
\begin{tabular}{c c c c c c c c c c c c c}    
\hline\hline                        
  \textbf{Output Class} & \multicolumn{12}{c}{\textbf{Target Class}}\\ [0.5ex]  
  \hline                      
 &\textbf{1}& \textbf{2} & \textbf{3}& \textbf{4}& \textbf{5} &\textbf{6}& \textbf{7} &\textbf{8} & \textbf{9} &\textbf{10}&\textbf{11}&\textbf{12}\\
\hline                      
\textbf{1}&\textbf{95.56}& 1.09 & 0.37& 0.96& 0.30&0.24& 0.23 &0 &0.91&0.14&0.02&0.14\\
\hline                      
\textbf{2}&0.68& \textbf{95.15} & 0.49& 0.53& 0.03&0&0.11&0.16&1.36 &0.96& 0.18 &0.24\\
\hline                      
\textbf{3}&0.66& 0.08 &\textbf{97.38}& 0& 0.41&0.85&0.30&0.18&0.08 &0&0.01&0\\
\hline                      
\textbf{4}&0.09&0 &0&\textbf{99.01}&0.26&0.09&0.21&0&0.27&0.01&0.03&0\\
\hline                      
\textbf{5}&0.01&0.26&0.31&0.44&\textbf{98.26}&0.04&0.05&0&0.02&0.36&0.19&0\\
\hline                      
\textbf{6}&0.02&0.01&0&0.39& 0 &\textbf{96.17}& 3.04 &0.01 &0.20 &0&0.11 &0.01\\
\hline                      
\textbf{7}&0&0.29&0&0.54& 0 &1.89&\textbf{96.37}&0&0.76&0.12&0.01 &0\\
\hline                      
\textbf{8}&0.08&0& 0.01& 0& 0 &0.35& 0.01 &\textbf{99.50}&0.03 &0.01&0 &0\\
\hline                      
\textbf{9}&0.03&0.07&0.01&0.01& 0 &0.07& 0.15 &0.32 &\textbf{97.65}&0.25& 0.53&0.14\\
\hline                      
\textbf{10}&0.09&0.18 &0&0.02& 0 &0.18&0.05 &0.01 &7.07&\textbf{89.94}& 1.21 &1.21\\
\hline                      
\textbf{11}&0.20&0.19&0.01&0.39& 0.11 &2.81& 0.79 &0.02& 2.95&0.60&\textbf{91.48}&0.41\\
\hline                      
\textbf{12}&0.37&0.51 &0.01& 0& .01 &0.07& 0.34&0.11 &0.88 &1.71& 2.47&\textbf{93.49}\\
[1ex]        
\hline          
\end{tabular}
}
\label{table:confusion_db_c}    
\end{table*}

\section{Biomedical Case Study}
To test the proposed architecture, we use CapgMyo \cite{geng}, a hand gesture time--series database recorded by instantaneous surface electromyography (sEMG). The data was collected by a non--invasive wearable device consisting of 8 acquisition modules. Each module contained a matrix--type (2 $\times$ 8) electrode array with an inter--electrode horizontal distance of 7.5 mm and a vertical distance of 10.05 mm. The 128 sEMG time--series were band--pass filtered at 20-380 Hz and sampled at 1,000 Hz with a 16--bit ADC conversion. Two different experiments were tested.  In experiment 1, each one of 18 subjects performed 8 basic isometric and isotonic hand gestures including \textit{thumb up}, \textit{extension of index and middle, flexion of the others}, \textit{flexion of ring and little finger, extension of the others}, \textit{thumb opposing base of little finger}, \textit{abduction of all fingers}, \textit{fingers flexed together in fist}, \textit{pointing index} and \textit{adduction of extended fingers}. The result of this experiment is termed as DB--a in the database. In experiment 2, each of the 10 subjects performed 12 gestures performed the maximal voluntary contraction (MVC) force hand gestures including \textit{index flexion}, \textit{index extension}, \textit{middle flexion}, \textit{middle extension}, \textit{ring flexion}, \textit{ring extension}, \textit{little finger flexion}, \textit{little finger extension}, \textit{thumb adduction}, \textit{thumb abduction}, \textit{thumb flexion}, \textit{thumb extension}. The result of this experiment is termed as DB--c in the database.
\par First, we aim at evaluating the system by classifying the DB--a actions. Therefore, the output classes are separated into 8 different actions, and the envelope of the EMG signals (using a Hilbert Transform) are extracted and applied to the networks as inputs. The classification training loss and test accuracy rates of various networks with different sizes along with hardware results are shown in Table~\ref{table:accuracy_DB_a}. The table shows that the hardware classification rate obtained from the proposed structure is similar to the performance in \cite{geng}. It should be noted that 150 frames, (equivalent to 150 ms) is the window size suggested by several studies of pattern recognition based prosthetic control \cite{geng}. Therefore various options for $s$ and $\omega_s$ can be considered while the multiplication of both these parameters should be no more than 150. As $q\cdot\omega_s$ is the latency of the system to make the final classification decision, the hardware can be efficiently used if $\omega_s$ takes the lowest possible value while keeping $q\cdot\omega_s $ fixed by increasing $s$. In this case, the network with $\omega_s$ of 10 is chosen to be implemented on hardware. Similar experiments are performed on DB--c actions and results are reported in Table~\ref{table:accuracy_DB_c}, however the classification rate of this network is not reported in \cite{geng}. Here, again the network architecture with narrower length of input is chosen to be implemented on hardware. Results from both tables show that the proposed hardware can achieve an accuracy comparable with a full precision network with about 40$\%$ and 30$\%$ more neurons respectively for DB--a and DB--c experiments. Although the number of neurons in the quantized ternary networks is higher than in the full precision ones, a significant hardware efficiency improvement is still seen in the quantized networks. Training loss traces for both experiments with different network structures are shown in Fig.\ref{fig:loss}. Results show that the loss function in the ternary network reaches to the required minimum value, albeit slower than the full precision networks in both DB--a and DB--c experiments. Note that, this would only create delay in the training phase which is not critical as the network is trained once for every application.

\par Considering the sampling rate of 1000 Hz and according to $\omega_s$, $\frac{1}{1000}\times10=10~ms$ per input window is the required response time for the system in order to operate in real--time. According to Table. II, the required operations per input window for the DB--a experiment is $(5\times128+250)\times250=\sim220$ K operations which can be delivered in $\sim35~\mu s$ by the hardware classifier and is negligible compared to the required response time ($10~ms$). These operations may take longer for the DB--c experiment as more neurons are embedded in the network. According to Table. III, the required operations per input window for the DB--c experiment is $(10\times128+350)\times350=\sim570$ K operations which can be delivered in $\sim90~\mu s$ by the hardware classifier and again is negligible compared to the required response time ($10~ms$).

\par
\begin{figure}[t]
\centering
\includegraphics[trim = 0.4in 0.1in 0.3in 0.1in, clip, width=4.2in]{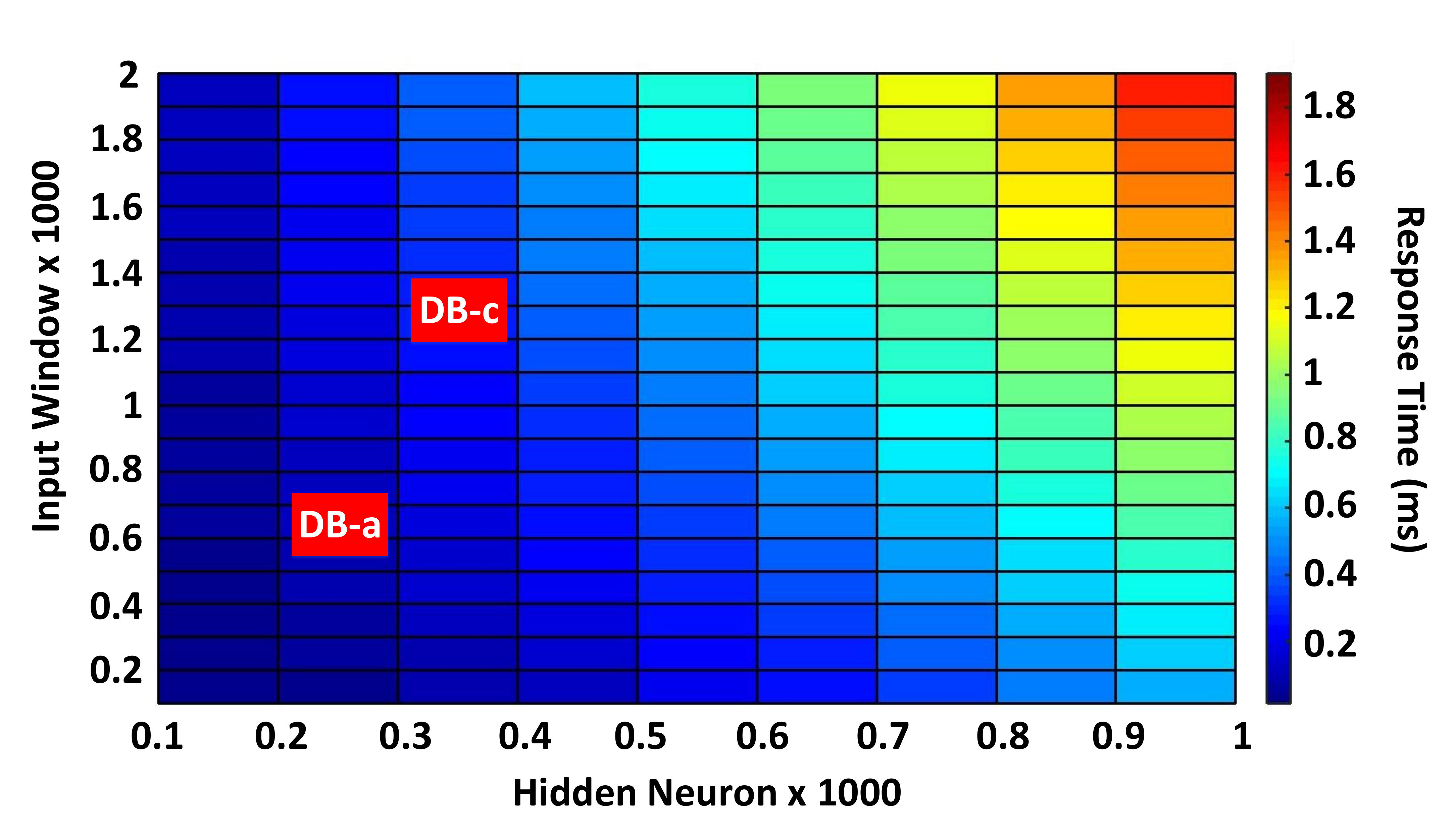}
\captionsetup{font=footnotesize}
\caption{Response time of the proposed hardware classifier for various input window size and hidden neuron ($N_h$). The response time for the employed datasets (DB-a and DB-c) is shown with red square box.}
\label{fig:tradeoffs}
\end{figure}

 The confusion matrices extracted from CapgMyo dataset for DB-a and DB-c are respectively illustrated in Tables~\ref{table:confusion_db_a} and \ref{table:confusion_db_c}. In these experiments, the trained classifier is run 200000 times on DB-a and DB-c database. The confusion matrices compare target and predicted hand gesture classes during the test stage to identify the nature of the classification errors, as well as their quantities. The correct predictions for each output class are bolded in the tables. According to the similarities of the hand gestures, the tables highlight the occurring misclassifications accordingly. For example, in Table~\ref{table:confusion_db_a}, class 1 (\textit{Thumb up}) is misclassified 0.43 $\%$ as class 6 (\textit{Fingers flexed together in fist}) which is the closest gesture in the dataset compared to class 1. The same applies in Table V, where class 6 (\textit{Ring extension}) is misclassified 3.04 $\%$ as class 7 (\textit{Little finger flexion}). Fig.\ref{fig:tradeoffs} illustrates the response time of the proposed hardware classifier for various input window size and hidden neuron ($N_h$). The response time for the employed datasets (DB-a and DB-c) is shown with red square boxes. It should be noted that the proposed architecture can be conveniently modified for larger networks while delivering enough response time.

\section{Discussion}
In this work, we have demonstrated how complex deep learning algorithms can be modified for portable and embedded biomedical applications. By using hardware resource sharing, a flexible and efficient classifier can be delivered where throughput of the system is traded off with hardware complexity and cost. It was also highlighted that such a compromise is only feasible in systems where the underlying time series has slow dynamics, as in the case of physiological systems. Otherwise, the faster the dynamics, the less resource sharing can be exploited. Moreover, for the purpose of performance evaluation, two case studies were investigated and results illustrated acceptable accuracy for both cases. Hardware synthesis and physical implementation also confirmed that the proposed hardware achieved 1.46$\times$ higher GOPs/Slice than similar state of the art FPGA--based accelerators. The proposed hardware architecture may lay the groundwork towards developing a generalized on--chip classifier for wearable biomedical applications where continuous classification on the patient is required such as in seizure detection for epileptic patients or continuous imaging and classification of bladder, liver and heart using ultrasound machines.

\section{Support Materials}
A python code and the corresponding Lorenz training/test data for the results provided in Section~3 are available online at: \url{https://github.com/hs15/LSTM_CLASSIFIER}

\chapter{Heart Signal Classifications Using a Low-cost Deep Hardware Time Series Classifier}
\renewcommand{\baselinestretch}{\mystretch}
\label{chap:Future}

\par Nowadays a diverse range of physiological data can be captured continuously for various applications in particular wellbeing and healthcare. Such data require efficient methods for classification and analysis. Deep learning algorithms have shown remarkable potential regarding such analyses, however, the use of these algorithms on low--power wearable devices is challenged by resource constraints such as area and power consumption. Most of the available on--chip deep learning processors contain complex and dense hardware architectures in order to achieve the highest possible throughput. Such a trend in hardware design may not be efficient in applications where on--node computation is required and the focus is more on the area and power efficiency as in the case of portable and embedded biomedical devices. This paper presents an efficient time-series classifier capable of automatically detecting effective features and classifying the input signals in real--time. In the proposed classifier, throughput is traded off with hardware complexity and cost using resource sharing techniques. A Convolutional Neural Network (CNN) is employed to extract input features and then a Long-Short-Term-Memory (LSTM) architecture with ternary weight precision classifies the input signals according to the extracted features. Hardware implementation on a Xilinx FPGA confirm that the proposed hardware can accurately classify multiple complex heart related time series data with low area and power consumption and outperform all previously presented state--of--the--art records.

\section{Introduction}
Advent of technologies such as wearable sensor systems could be an answer to the rising issues such as increasing individuals with critical medical conditions, providing quality care for remote areas and methods to maximize the participation of disable patients \cite{patel} that healthcare system struggle with. Chronicle electronic health data that can reformed to the time series in machine learning tasks are prominent information should be sensed and analyzed using human biologically activities \cite{mazilu}. The interest for wearable systems originates from the need for monitoring patients over extensive periods of time \cite{Subhas}. Wearable activity systems mainly include sensors such as accelerometers, gyroscopes or magnetic field communication/chemical sensors \cite{bulling}, communication systems and process systems for analyzing generated signals. Smart wearable sensors are effective and reliable for preventative methods in many different facets of medicine such as, cardiopulmonary, vascular \cite{Geoff}. Further, the use of wearable sensors has made it possible to have the necessary treatment at home for patients after heart-attacks and diagnosis of some heart diseases such as cardiac \cite{Subhas}.

\par Regardless of these achievements, most contemporary commercial products only can measure simple metrics such as heart beats or steps. In addition, high computational requirement to classify high dimensional, ordered attributes time series of interest makes it practically impossible in real-time. Compare to traditional time series classifiers deep learning algorithms, armed with multiple layer of feature hierarchical, capable of extracting temporal dependencies in time series and more powerful processing capacities in wearable systems pave the way for performing more data analysis on-node and in real--time. This capability to perform more complex data analysis on the wearable device/node provides the opportunity to decrease transition data from device to host, or on the other word save data bandwidth link. The bandwidth saving is more exposes itself in the heart disease patients who should continuously monitored and classified using ultrasound machines or the victims such as cardiovascular disease does not have access to health care service , even if the doctor, relatives are not near the patient and also during the non--availability of the cellular network \cite{Kala}. However, full hardware implementation of deep neural networks still challenging for designers on wearable sensors and embedded platform due to memory bandwidth and energy inefficiency of high computational units.

\par Recent studies on the development of deep learning hardware accelerators mainly have tried to achieve highest throughput, keeping up with real-time demands of complicated and embedded machine learning algorithms, led to intricate systems with a large number of Multiply Accumulate (MAC) processors. As a result, when it comes on the hardware realization regarded systems consume large silicon area ($\sim$600~mm$^2$) and power ($\sim$500~W) \cite{GPU}\cite{Google}\cite{Intel}. Based on the observation that most biological time series signals have small rates of frequency (0--500 Hz), an alternative approach, by trading off throughput and hardware complexity using sharing resources is proposed in \cite{Hamid}. This method utilized LSTM \cite{hochreiter} to capture motor signals of arm muscles. This architecture was shown to be very effective in sequence learning, however, feature extraction of the input time series still remains an issue. In this paper we propose a generalized time series classifier that implements both feature extraction and sequence learning respectively through an CNN and an LSTM network. The architecture first is compared to \cite{Hamid} and then applied to multiple heart disease database. It will be also shown that the proposed system is compact, portable as well as accurate. In contrast to muscle signals which can be classified using features such as amplitude of signals, heart signals classification needs to highlight more subtle features. These feature could be extracted through training an CNN. The advantage of doing so compared to other classical feature extraction methods is the ability of learning new features. The proposed generalized architecture can be reconfigurable and trained for different applications.


\begin{figure}[t]
\centering
\includegraphics[trim = 0.3in 0.1in 0.1in 0.1in, clip, width=4.0in]{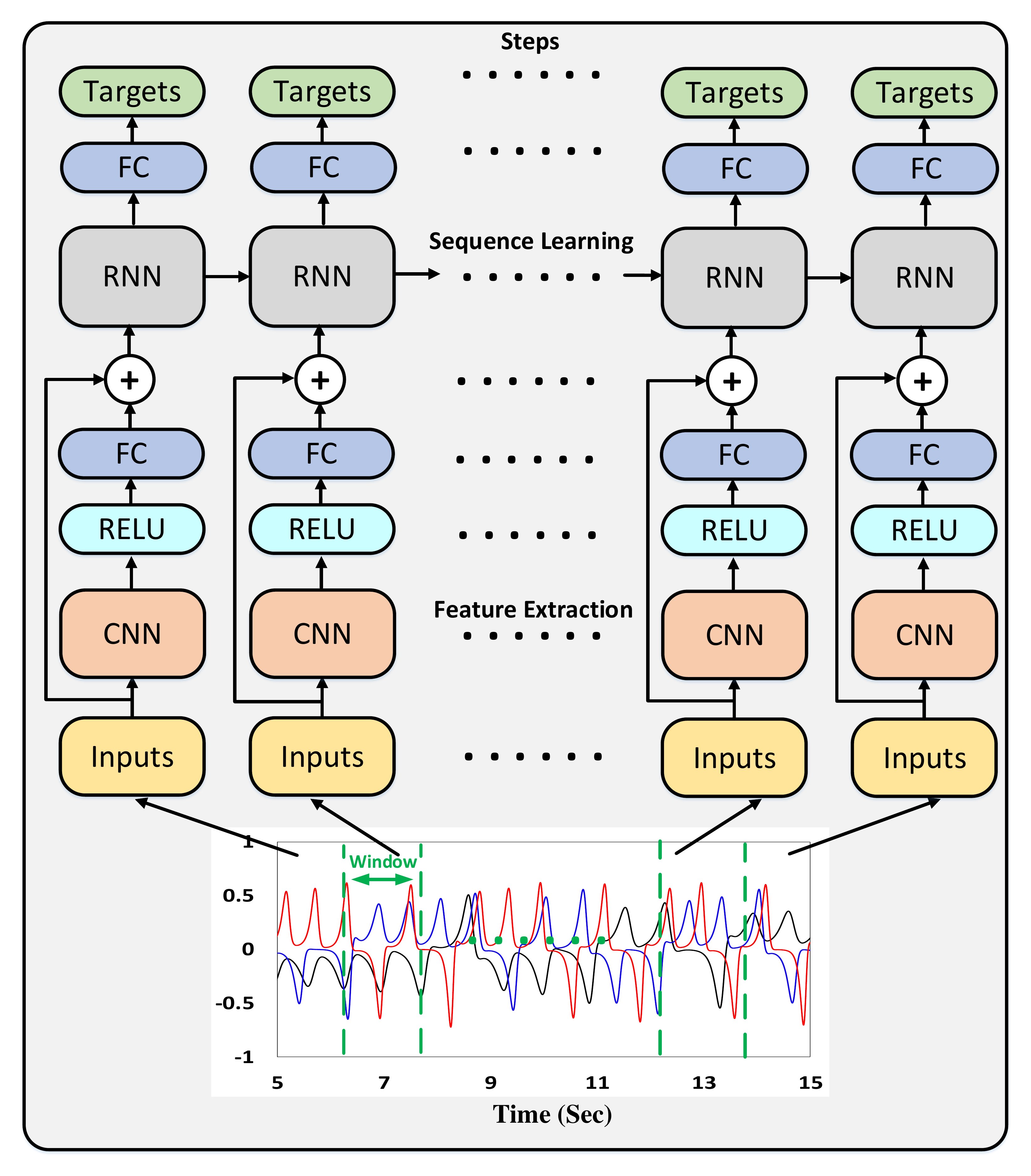}
\captionsetup{font=footnotesize}
\caption{The overall architecture of the proposed classifier along with three input time series. A portion of the input signal termed as $window$ is highlighted in green color, sequentially fed to the systems after $steps$ times.}
\label{fig:arch}
\end{figure}

\section{Hardware--oriented Time--series Classifier}
The overall architecture of the proposed time series classifier is shown in Fig.~\ref{fig:arch}. In this approach, feature extraction is carried out using CNN blocks and then data is entered to the RNN blocks for sequence learning. The sequential target replication technique inspired from \cite{lipton} is used during the training phase. However our approach is slightly different. In the proposed architecture, during forward path and for all $steps$, the same output label is used to calculate the output error. The error is stored $q$ times in memory in order to calculate the gradient descent during the backward path. In the backward pass, the RNN is unrolled back in time and the weights are updated. This technique forces the network to better memorize the previous sequences of the input windows \cite{andrej}. It should be also noted that during the training phase loss values are averaged over all steps, while at the inference (test) time, the output at the final step is chosen as the actual classification value. Here is a detailed explanation of the two main blocks in the system:

\subsection{CNN Blocks}
CNNs have shown remarkable performance in image processing tasks such as object detection \cite{Ren}, face recognition \cite{Florian} and are normally composed of two types of layers: pooling layers and convolutional layers, where in this work just later layer is utilized for developing proposed architecture. Each convolutional layer is responsible for three dimension calculation of inner product of input window and weights, which referred as kernels. In contrast to regular convolution which determines output using whole input, in machine learning applications, this is done through regional products of input with a single filter. Each filter is responsible for extracting a feature from input signal. In our case, the input is a 1D time--series array, therefor the CNN filters are also 1D. The resulted output is denoted by feature map. The 1D convolution operation can be represented as follow:
\begin{equation}
\bm{z}_{(i)}^\ell = \bm{b}_{(i)}^\ell+ \sum_{a=0}^{m-1} \bm{W}_{cnn(a)}\cdot\bm{x}_{(i+a)}^{\ell - 1}
\end{equation}
where $\bm{W}_{cnn}$ and $\bm{b}$ are weight and bias of each channel, $l$ represents index of layers, $m$ is the length of each 1D filter, $\bm{x}$ and $\bm{z}$ are respectively the input and output of the network.
\par The output of each filter in an CNN layer is rectified using an activation function called ReLU which is mathematically described as follows:
\begin{equation}
\bm{r}_{(i)}^\ell = max(0,\bm{z}_{(i)}^\ell)
\end{equation}
\par The output of the activation function in the last layer is fed to a fully connected network as given here:
\begin{equation}
\bm{P}_{(i)} = \sum_{c=1}^{f-1} \sum_{d=0}^{n-1} \bm{W}_{f(c)}\cdot\bm{r}_{(d)}
\end{equation}
where $\bm{W}_{f}$ is weight of the full connection layer, $f$ is the number of filters per layer and $n$ is the length of the output feature map.
\par Current state of the art CNN networks such as ResNets \cite{Kaiming} or GoogLeNet \cite{Szegedy} have privilege of utilizing depth layers for achieving higher accuracy in the image related tasks. With increasing number of layers process of convergence becoming harder again due to the exploding/vanishing gradients. Techniques such as normalization layers \cite{Ioffe}\cite{Glorot} enabled designing networks with depth layers. In addition, these networks are vulnerable to the problem of accuracy saturations when with decreasing energy of system, accuracy does not improve \cite{Kaiming}. For dealing with this issue ResNet or GoogLeNet exploit structures such as inception module or residual learning. In our simulations we faced both problems of hampering convergence and accuracy saturations. To address these problems we used residual learning technique which is more hardware friendly and straightforward compared to the other structures. In Fig.~\ref{fig:arch} typical structure of residual module have been shown, where CNN is chosen to be two or three layers.

\subsection{RNN Blocks}
LSTM networks are very powerful Recurrent Neural Networks (RNN) that explicitly add memory gates \cite{hochreiter}. This makes the training procedure more stable and allows the model to conveniently learn both long and short--term dependencies. There are some variations on the LSTM architecture, however in this paper we use the following model \cite{schmid}:
\begin{equation}
\begin{cases}
\bm h^f_{n+1}=\sigma( \bm{W}_f^T \cdot\bm{xx}_n+\bm b_f) \\
\bm h^i_{n+1}=\sigma( \bm{W}_i^T\cdot\bm{xx}_n+\bm b_i)\\
\bm h^o_{n+1}=\sigma( \bm{W}_o^T\cdot \bm{xx}_n+\bm b_o)\\
\bm h^c_{n+1}=\tanh( \bm{W}_c^T\cdot \bm{xx}_n+\bm b_c)\\
\bm c_{n+1} = \bm{h}^f_n \circ \bm{c}_n+\bm{h}_n^c \circ \bm h\bm_n^i;\\
\bm h_{n+1}=\bm h^o_n \circ \tanh(\bm c_n);\\
\bm y_{n+1}= \bm W^T_y \bm h_n +\bm b_y
\end{cases}
\end{equation}
where $\bm{xx}_n=[\bm{h}_n,(\bm x_n+P_n)]$ and $\bm{h}_n$ and $\bm{c}_n$ are the output and cell state vectors respectively at discrete time index, $n$.  The operator $\circ$ denotes the Hadamard element by element product.  The variables $\bm h^f_n$, $\bm h^i_n$, $\bm h^o_n$ represent the forgetting, input and output gating vectors. The parameter $\bm y_n$ is a fully connected layer following the LSTM block and serves as the network output and its size is determined by the number of output classes and as well as the number of hidden neurons. Finally, $\bm W_f$, $\bm W_i$, $\bm W_o$, $\bm W_c$ ,$\bm W_y$ and $\bm b_f$, $\bm b_i$, $\bm b_o$, $\bm b_c$, $\bm b_y$ are the weights and biases for the different layers, respectively.

\par One of the main bottlenecks for the hardware realization of RNNs and convolutional neural networks (CNNs) is the large memory size and bandwidth required to fetch weights in each operation. To alleviate the need for such high bandwidth memory access, we investigate two quantization methods (binary and ternary) introduced in \cite{hubara} \cite{li} to quantize weights embedded in the network architecture. As the changes during gradient descent are small, it is important to maintain sufficient resolution otherwise no change is seen during the training process, therefore the real--valued gradients of weights are accumulated in real--valued variables. We also set the bias values to zero to achieve further efficiency in hardware realization while delivering an acceptable classification accuracy for the experimented biomedical case study. Such quantization methods can be considered as a form of regularization that can help the network to generalize.  In particular, the binary and ternary quantization are a variant of Dropout, in which weights are binarized/ternarized instead of randomly setting part of the activations to zero when computing the parameter gradients \cite{hubara}.

\par The quantization of weights in the forward path must be also reflected in the calculation of the gradient descent. Here, we use the version of the straight--through estimator introduced in \cite{hubara} that takes into account the saturation effect. Consider the sign and round functions for binary and ternary quantization respectively as follows:
\begin{equation}
\begin{cases}
q_b=\text{sign}(r)\\
q_t=\text{round}(r)
\end{cases}
\end{equation}
and assume that estimators $g_{q_b}$ and $g_{q_t}$ of the gradients $\frac{\delta C}{\delta q_b}$ and $\frac{\delta C}{\delta q_t}$ are derived. Then, the straight--through estimators of $\frac{\delta C}{\delta r}$ are:
\begin{equation}
\begin{cases}
g_{r_b}=g_{q_b}1_{\mid r\mid\leq 1}\\
g_{r_t}=g_{q_t}1_{\mid r\mid\leq 1}
\end{cases}
\end{equation}
\par This implies that the gradient is applied to the weights if their real values are between 1 and -1 otherwise the gradient is cancelled when $r$ is outside the range.

\begin{figure}[t]
\centering
\includegraphics[trim = 0.3in 0.2in 0.3in 0.2in, clip, width=4.0in]{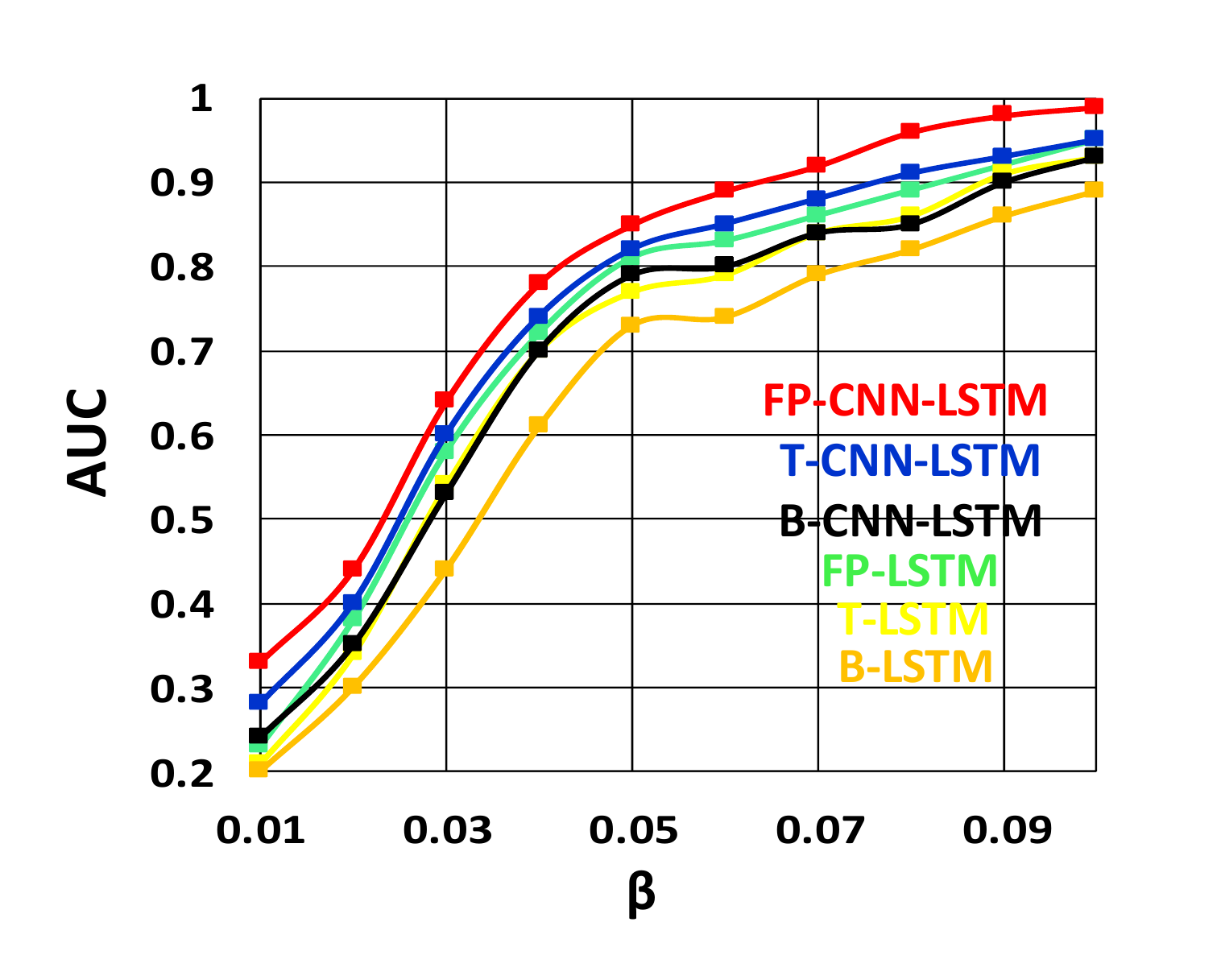}
\captionsetup{font=footnotesize}
\caption{Similarity index ($\beta$) sweeping for the LSTM network \cite{Hamid} and the proposed CNN--LSTM classifier with various precision.}
\label{fig:sin}
\end{figure}

\section{Hardware--oriented Simulations}
As mentioned before, the proposed classifier has two main parts: 1) feature extraction (CNN); 2) sequence learning (LSTM). Here, we use the sequence classifier introduced in \cite{Hamid}. The performance of the classifier in full, binary and ternary precision was measured in \cite{Hamid}. It was shown how the performance of the model is affected by sweeping various parameters in the design. Two chaotic nonlinear case studies were used as input time series signals. The same still applies here and in order to avoid repetition, we do not mention the experiments in detail again. Here, we investigate the impact of adding an extra feature extractor layer ro the classifier introduced in \cite{Hamid}. We use a simple case study as follows:
\begin{equation}
\Psi_j(t)=sin(t(\alpha+j\times\beta))
\end{equation}
where $j$ is the number of output classes which is 5, $\alpha$ is a constant value which is 3 and $\beta$ is the frequency difference between each channel ranging from 0.1 to 0.01. The reason why we choose such a separable case study is that the similarity index ($\beta$) between the channels can be linearly altered. The smaller $\beta$, the higher the similarity between the classes and consequently more difficulty to differentiate the output classes. The case studies utilized in \cite{Hamid} could not be used here as they are chaotic. As can be seen in Fig.~\ref{fig:sin}, by reducing $\beta$, the accuracy for all networks with different precisions drops. It is also shown that the due to the added feature extractor to the network the performance of the proposed CNN-LSTM architecture is always higher than the corresponding network \cite{Hamid} with various precisions. The performance of the proposed network is also compared to \cite{Hamid} for a inseparable dynamical system's time series with certain parameter sets extracted from Lorenz attractor \cite{lorenz} as the similarity index cannot be swept due to a chaotic nature of the dynamical system. Results shown in Table~\ref{table:chaotic} confirm that the proposed CNN--LSTM architecture achieves higher performance with various precisions compared to the classifier introduced in \cite{Hamid}.

\begin{table}[t]
\captionsetup{font=footnotesize}
\caption{Classification test error rates of the LSTM networks \cite{Hamid} with different weight resolutions/structure and the proposed CNN-LSTM structure trained on a inseparable dynamical system's time series. The CNN network has two layers with 20 and 50 neurons each respectively.}   
\centering          
\resizebox{\columnwidth}{!}{%
\begin{tabular}{c c c c c c}    
\hline\hline                        
  \textbf{Model} & \textbf{Learning~Rate}& \textbf{Hidden Neurons} & \textbf{Input~Window}& \textbf{Steps} &\textbf{AUC $\%$}\\ [0.5ex]  
\hline                      
\textbf{FP-LSTM} &0.05& 250 & 50& 30&90.32\\
\textbf{T-LSTM}&0.1& 350 & 50& 30 &88.37\\
\textbf{B-LSTM}& 0.1& 350 & 50 & 30& 83.23\\        
\textbf{FP-CNN-LSTM}& \textbf{0.05}& \textbf{250} & \textbf{50} & \textbf{30}& \textbf{93.43}\\        
\textbf{T-CNN-LSTM}& \textbf{0.1}& \textbf{350} & \textbf{50} & \textbf{30}& \textbf{91.65}\\        
\textbf{B-CNN-LSTM}& \textbf{0.1}& \textbf{350} & \textbf{50} & \textbf{30}& \textbf{88.91}\\ [1ex]        
\hline          
\end{tabular}
}
\label{table:chaotic}    
\end{table}

\section{Hardware Finite State Machine}
The proposed hardware classifier functions as a finite state machine that iterates through eight states and only one state is active at a time. This structure can also be implemented in a pipelined form with multiple active states and higher throughput at the expense of increased power and area consumption. The general functionality of each state is briefly described as follows:
\begin{algorithm}
    \SetKwInOut{Input}{Input}
    \SetKwInOut{Output}{Output}

    \underline{Hardware State Machine} $(x,y)$\;
    \Input{Time--series ($x$) with the length of $\omega_s$}
    \Output{Classified label ($y$)}
    \If{$state=1$}
      {
        calculate $\sum_{a=0}^{m-1}\bm W_{cnn(a)} x_{(i+a)}^{\ell - 1}$\;
        apply ReLU ($f(z)=max(0,z)$)\;
        \If{$l=2$}{
        set $state=2$\;}
        \Else
        {
        set $state=1$\;
        }
      }
      \ElseIf{$state=2$}
      {
        calculate $P_{(i)}=\sum_{c=1}^{f-1} \sum_{d=0}^{n-1} \bm{W}_{f(c)}\cdot\bm{r}_{(d)}$\;
        add $x$ to $P$\;
        set $state=3$\;}
            \ElseIf{$state=3$}
      {
      calculate $\bm W_f^T \bm {xx}$, $\bm W_i^T \bm {xx}$, $\bm W_o^T \bm {xx}$ and $\bm W_c^T \bm {xx}$\;
      set $state=4$\;}
            \ElseIf{$state=4$}
      {
      apply $\sigma(.)$ and $\tanh(.)$\;
      set $state=5$\;}
            \ElseIf{$state=5$}
      {
      calculate $\bm{h}^f_n \circ \bm{c}_n+\bm{h}_n^c \circ \bm h\bm_n^i$\;
      set $state=6$\;}
            \ElseIf{$state=6$}
      {
      apply $\tanh(.)$\;
      set $state=7$\;}
            \ElseIf{$state=7$}
      {
      calculate $\bm h^o_n \circ \tanh(\bm c_n)$\;
      set $state=8$\;}
            \ElseIf{$state=8$}
      {
      classify the output label ($\bm W^T \bm h_n +\bm b_y$)\;
      set $state=1$\;
      return $y$\;}

    \caption{Finite state machine algorithm for the proposed hardware classifier.}
\end{algorithm}

\par $\textbf{State~1:}$ After initialization at the start state, the first input transmission according to the defined $\omega_s$ is carried out and the system enters the first state where $\sum_{a=0}^{m-1}\bm W_{cnn} x_{(i+a)}^{\ell - 1}$ is calculated with 32 additions and multiplexing per clock cycle. A total of $(\omega_s-m+1)\times f\times\frac{m}{32}$ clock pulses are needed for each layer in this state to finish the calculations as we use zero--padding and stride is equal to one. In this state after completing each CNN layer, a ReLU function ($f(x)=max(0,x)$) is also applied which adds $(\omega_s-m+1)\times\frac{f}{32}$ operations overhead.
\par $\textbf{State~2:}$ Then, according to (3) the FC layer is implemented in this state using 32 MAC operations in parallel and the result is added to $x$ as seen in Fig.~\ref{fig:arch}.
\par $\textbf{State~3:}$ The output of the previous state is formed equal to $\omega_s$ and then $\bm W_f^T \bm {xx}$, $\bm W_i^T \bm {xx}$, $\bm W_o^T \bm {xx}$ and $\bm W_c^T \bm {xx}$ are independently calculated each with eight additions per clock cycle. A total of  $(N_h+\omega_s)\times \frac{N_h}{8}$ clock pulses are needed for this state to finish the calculations.
\par $\textbf{State~4:}$ At the end of the calculations, the system enters the second state. In this state, according to (2), nonlinear functions ($\sigma(.)$ and $\tanh(.)$) are applied to the previous values fetched from memories. After $N_h$ clock pulses, the system enters the next state.
\par $\textbf{State~5:}$ In this state, using two embedded multipliers, the value of variable $\bm c$ is calculated in $N_h$ clock pulses. The critical path of the proposed architecture is limited by this state which can be alleviated by using pipelined or serial multipliers at the expense of increased latency and hardware cost.
\par $\textbf{State~6:}$ As the variable $c$ is provided, the system enters this state where the $\tanh(.)$ function is applied to the previous values, taking $N_h$ clock pulses, then the system enters the next state.
\par $\textbf{State~7:}$ In this state, using one of the two embedded multipliers, the value of variable $h$ is calculated in $N_h$ clock pulses and the system enters the next state.
\par $\textbf{State~8:}$ Finally, if the number of scan times is equal to $s$, by calculating $\bm W_y^T \bm h$ in $N_h\times N_y$ clock pulses, the system determines the classified output and exits, otherwise enters State 1. This process is repeated for each window of the input signal(s) and managed by a master controller circuit, embedded into the system.

\section{Hardware Architecture}
\par In this section, hardware description of the proposed classifier is presented which is mainly inspired by the model introduced in \cite{Hamid}. The main goal of the design is to exploit the slow nature of physiological signals in particular heart activities for reducing hardware complexity and cost. In principle, in such systems, given that the classification rate is low, a few calculations per high speed clock (100 Mhz) are enough to handle the computational burden of the classifier. This architecture would also allow us to actively and efficiently reconfigure the system according to the user's specifications for the set number of neurons, window size, iterations and input/output classes.

\par Other design strategies such as implementing a large number of MAC units in hardware do not apply here as high throughput is not demanded. The architecture of the proposed system is shown in Fig.~\ref{fig:detail} in which the hardware modules (maximum 64 operations per clock cycle) are shared through a 96--bit bus. It should be stressed that since the accuracy of the ternary network is generally higher than its binary counterpart as shown in Fig.~\ref{fig:sin}, in the hardware implementation, the ternary quantization is used and two bits are allocated for storing each weight value. In the following, the architecture of each hardware module is explained in detail:

\par $\textbf{WBs~(Weight~Banks):}$ This block contains five sets of buffers to store the truncated 2--bit weights ($\bm{W}_f,~\bm{W}_i,~\bm{W}_o,~\bm{W}_c$, and $\bm{W}_{cnn}$), and two sets to store full precision weights for full--connection layers ($\bm{W}_{f}$ and $\bm{W}_y$). The $WBs$ module is able to read/write maximum 64 bits in each clock cycle. The utilised volume of each buffer is defined by the user which depends on the number of hidden neurons, $f$, $\omega_s$ and etc. However, the maximum volume of these buffers must be selected based on the available resources on the FPGA. The greater the volume size, the wider the range of flexibilities for the network/input size. The reading process of this block is controlled by the $MC$ unit and the block is only activated upon its use. It should be noted that, as the proposed architecture is implemented on a Xilinx FPGA in this work, the buffers are realised using block RAMs and the address of each reading operation is provided on the negative clock edge by the $MC$ module.

\begin{figure}[t]
\vspace{-10pt}
\centering
\includegraphics[trim = 0.2in 0.1in 0.2in 0.1in, clip, width=4.0in]{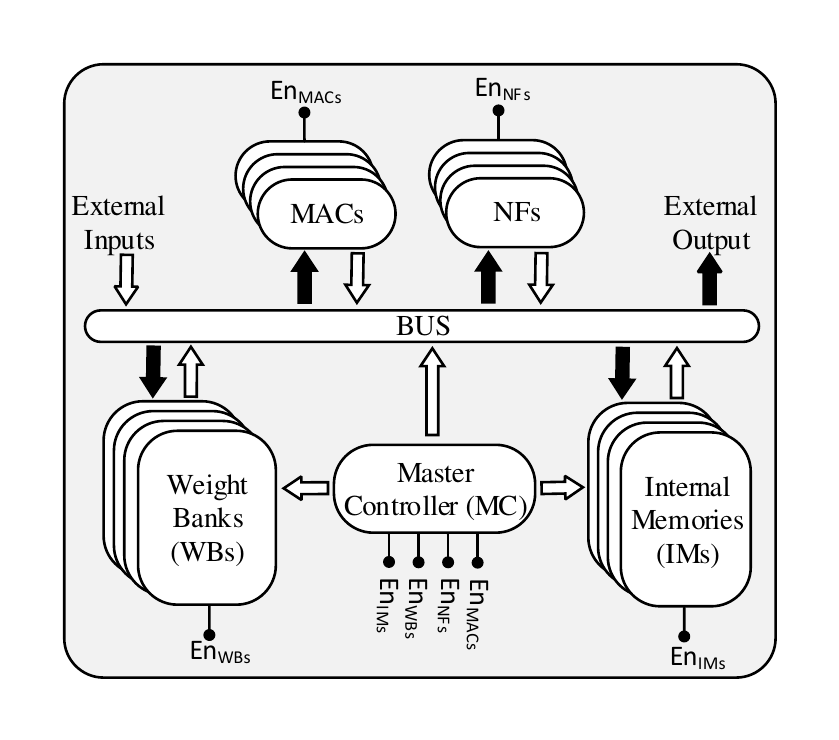}
\captionsetup{font=footnotesize}
\caption{The proposed architecture of the system comprising six blocks: MAC1s, NFs, MAC2s, WBs, IMs and MC. Black and white arrows represent output and input signals respectively in accordance with bus connections}
\vspace{-10pt}
\label{fig:detail}
\end{figure}

\par $\textbf{IMs~(Internal~Memories):}$ This block contains sets of buffers to store 12--bit values produced by the intermediate stages for both CNN and LSTM. The maximum and utilised volume of the buffers are again determined by the available on--chip memory and the user's specifications respectively. The writing and reading of this block is also controlled by the $MC$ unit and the reading address is provided on the negative clock edge as they are implemented using block RAMs. The maximum data bandwidth of this module is 48 bits and the module is active in almost all states.
\par $\textbf{MACs~(Multply--Acummulate):}$ This blocks contains 32 parallel MAC units with full precision. Each unit computes the product of 12--bit and 2--bit or 12--bit signed numbers and adds the product to an accumulator. The number of iterations that this unit needs to operate is defined by the user and assigned by the $MC$ block.

\par $\textbf{NFs~(Nonlinear~Functions)}:$ This block is responsible for the calculation of nonlinear functions ($\sigma(u)$ and $\tanh(u)$) and ReLU employed in the hardware state machine. The detailed architecture of this block is introduced in \cite{Hamid}.
\par $\textbf{MC~(Master~Controller)}:$ According to the defined parameters by the user, this block manages and controls all resources used in the architecture through a shared bus and controlling signals ($En_{MACs}$, $En_{NFs}$, $En_{IMs}$ and $En_{WBs}$). In other words, this block actively changes the state of the FSM by assigning proper tasks to the hardware modules and actively turning off the unused modules.

\begin{table}[t]
\captionsetup{font=footnotesize}
\caption{Performance comparison between the proposed system, implemented the Kintex--7 (XC7K325T) FPGA and other state of the art hardware.}   
\centering          
\resizebox{\columnwidth}{!}{%
\begin{tabular}{c c c c c c}    
\hline\hline                        
  & \textbf{2015\cite{sankaradas}} & \textbf{2016\cite{motamedi}} &\textbf{2016\cite{qiu}}&\textbf{2017\cite{yonekawa}}& \textbf{This Work}\\ [0.5ex]  
\hline                      
\textbf{Precision}& 32bits float&32bits float&16bits float&binary&12bits fixed\\
\textbf{Frequency}& 100 MHz & 100 MHz&150 MHz& 150 MHz&100 MHz\\
\textbf{FPGA Chip}& VX485T& VX485T& XC7Z045&ZU9EG&XC7K325T\\
\textbf{GOPs}& 61.6& 84.2& 187.8&460.8&6.3\\
\textbf{Slice}& 75123& 75924& 52458&47950&586\\
\textbf{GOPs/Slice}& 8.12E-04& 11.09E-04& 35.8E-04&96.1E-04&107.5E-4\\[1ex]        
\hline          
\end{tabular}
}
\label{table:hardware}    
\end{table}

\section{Hardware Results}
\par To verify the validity of the proposed hardware classifier, the architecture designed in the previous section is implemented on a Genesys 2 development system, which provides a high performance Kintex--7 (XC7K325T) FPGA surrounded by a comprehensive collection of peripheral components. The device utilization for the implementation of the proposed hardware is summarized in Table~\ref{table:hardware} along with other state of the art implementations. The focus of all other implementations is mainly on the hardware realization of CNNs; however, as the nature of computations in all deep learning algorithms is the same, for the sake of comparison the implementation results of those studies are included here. The results of hardware implementations show that the proposed classifier reaches 1.12$\times$ higher GOPs/Slice than similar state of the art FPGA--based accelerators. Obviously, less power consumption is also achieved as the number of FPGA slices used in the proposed system is lower than in other state of art hardware. Such a trade off constrains the GOPs factor, which is not critical for most slow biomedical applications. The required response time of the system must be seriously considered upon such modifications. For example, by adding layers to CNN or LSTM, the amount of calculations is increased, therefore, the number of parallel MAC processors in the \textbf{MACs} module must be increased to keep the response time of the systems constant.

\section{Heart--Related Case Studies}
\par To test the proposed generalized time series classifier we conduct our simulations through three datasets related to the heart diseases extracted from well-known UCR datasets \cite{UCRArchive} and  PhysioNet 2016 and 2017 challenges \cite{Physio2017}. UCR datasets recorded heart activities by use of electrocardiography (ECG)device. Mean and variance of UCR datasets are near to zero and unit respectively. ECG5000 dataset originates  from \cite{ECG5000}, the BIDMC congestive heart failure database, consisting of records of 15 subjects, with severe congestive heart failure (NYHA class 3-4). Records of each individual recorded in 20 hours, containing two ECG signals, sampled with rate of 250 Hz, with 12 bit resolution and over range of (-10--10) mV. ECG200 was formatted at \cite{Olszewski} including two datasets, normal heartbeat and a Myocardial Infarction, the dataset is subset of \cite{Greenwald}, which contains 35 half—hour records and sampled with rate of 125 Hz. In PhysioNet 2016 \cite{Physio2016}, heart sound recordings have been collected from several contributors around the world, gathered at either a clinical or nonclinical environment, from both healthy subjects and pathological patients. The Challenge training set consists of five databases (A through E) containing a total of 3,126 heart sound recordings, lasting from 5 seconds to just over 120 seconds. All recordings have been resampled to 2,000 Hz and have been provided as .wav format. Each recording contains only one PCG lead. PhysioNet 2017 challenge data sampled and stored as 300 Hz, 16-bit A/C conversion with bandwidth (0.5--40 Hz) and (-5--5 mV) dynamic range. It should be noted 70 percent of online provided dataset allocated  for training set and the rest for testing set.  All of the datasets extracted from one channel. However, our model simply is able to handle multivariate time series by adding another dimension to the convolution layers.

\par To evaluate our algorithms four experiments are performed. In table \ref{table:Network characterestics} learning parameters along with characteristics of each network for different datasets are represented. It should be noted no preprocessing was performed on the datasets, and the CNN network automatically extract the important features from the input signal. CNN has two layers including 10 and 30 filters with the size of respectively $1\times5$ and $1\times3$. As mentioned in Section III, the size of hidden neurons is chosen to be 350 for ternary precision experiments. Simulation were performed using both Python and MATLAB and results are demonstrated in table \ref{table:Accuracy} confirming this fact that full-precision version of the proposed model outperforms all presented state-of-art records. In addition, quantized models could achieve acceptable accuracy compared with full--precision implementation and even better accuracy on some benchmarks.

\begin{table}[t]
\captionsetup{font=footnotesize}
\caption{Chosen Parameters and Characteristics of Networks for Different Dataset.}   
\centering 
\resizebox{\columnwidth}{!}{%
  \begin{tabular}{c c c c c}
    \hline\hline
    \textbf{Datasets} & \textbf{Input--Window} & \textbf{Steps} & \textbf{Hidden Size}&\textbf{Learning Rate}\\
    \hline
    \textbf{ECG200}&20&4&250&.05 \\
    \textbf{ECG5000}&20&7&250&.05\\
    \textbf{PhysioNet 2016}&50&30&250&.05 \\
    \textbf{PhysioNet 2017}&50&30&250&.05\\
    \hline
\end{tabular}
}
\label{table:Network characterestics}    
\end{table}

\begin{table}[t]
\captionsetup{font=footnotesize}
\caption{Performance Comparison of FP--CNN--LSTM, T--CNN--LSTM, Hardware Results With Other Existing Scores.}   
\centering 
\resizebox{\columnwidth}{!}{%
  \begin{tabular}{c c c c c c}
    \hline\hline
    \textbf{Datasets} & \textbf{Existing SOTA} & \textbf{FP-LSTM} &\textbf{FP-CNN-LSTM} & \textbf{T--CNN--LSTM}&\textbf{Hardware}\\
    \hline
    \textbf{ECG200} & 0.92 \cite{Wang} & 0.88& 0.96 & 0.93& 0.93\\
    \textbf{ECG5000} & 0.948 \cite{Karim} & 0.85 &0.95 & 0.93&0.92\\
    \textbf{PhysioNet 2016}&0.86 \cite{Physio2016}& 0.87& 0.90&0.86&0.85\\
    \textbf{PhysioNet 2017} & 0.86 \cite{Physio2017} &  0.81& 0.87 & 0.84& 0.83 \\
    \hline
\end{tabular}
}
\label{table:Accuracy}    
\end{table}

\begin{table*}[t]
\captionsetup{font=footnotesize}
\caption{Memory and MAC estimations for all case studies with various architecture and weight precisions.}   
\centering 
\resizebox{\columnwidth}{!}{%
  \begin{tabular}{c c c c c c c c c}
    \hline\hline
    \textbf{Datasets}  & \multicolumn{4}{c}{ \textbf{Memory(Mb)}}   & \multicolumn{4}{c}{ \textbf{MAC Operations (M)}} \\
    \hline
     & \textbf{FP-LSTM} & \textbf{T-LSTM}&\textbf{FP-CNN-LSTM} & \textbf{T--CNN--LSTM}& \textbf{FP-LSTM} & \textbf{T-LSTM}&\textbf{FP-CNN-LSTM} & \textbf{T--CNN--LSTM}\\
    \hline
    \textbf{ECG200}  & 9.07&1.03& 11.27 & \textbf{1.84}& 0.27&0.52& 0.95 & \textbf{1.21}\\
    \textbf{ECG5000}  & 9.07&1.03 &11.27 & \textbf{1.84}& 0.27&0.52& 0.95 & \textbf{1.21}\\
    \textbf{PhysioNet 2016}& 10.06&1.12& 12.26&\textbf{1.93}&0.30 &0.56& 0.98 & \textbf{1.24}\\
    \textbf{PhysioNet 2017}  & 10.06 &1.12& 12.26 & \textbf{1.93}& 0.30&0.56& 0.98 & \textbf{1.24}\\
    \hline
\end{tabular}
}
\label{table:assesment}    
\end{table*}

\par Training loss traces for all case studies with various weight precisions are shown in Fig.\ref{fig:loss}. Results show that the loss function in the ternary network reaches to the required minimum value, albeit slower than the full precision networks in both all experiments. Note that, this would only create delay in the training phase which is not critical as the network is trained once for every application.

\begin{figure}[t]
\centering
\includegraphics[trim = 0.2in 0.1in 0.2in 0.1in, clip, width=5.0in]{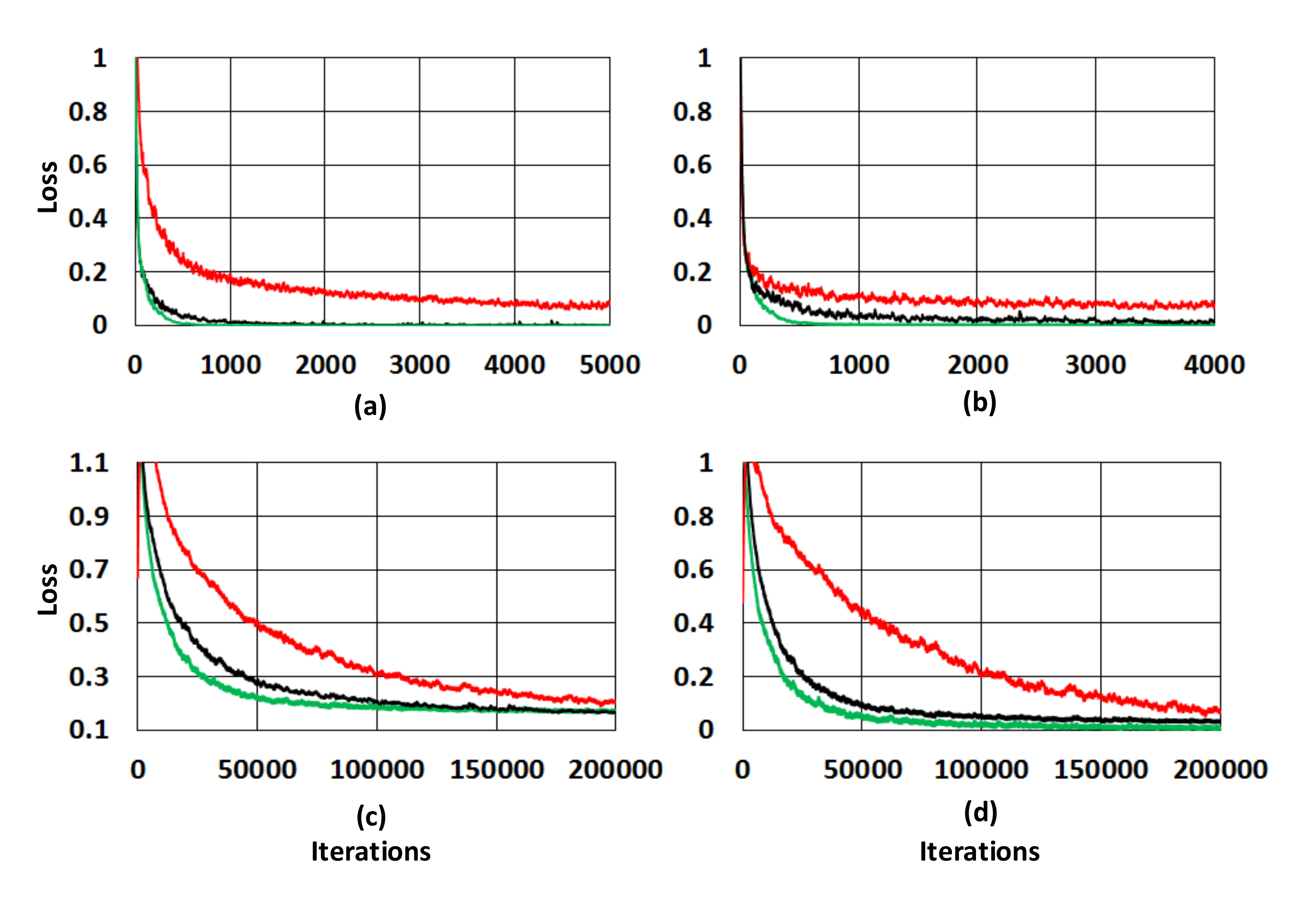}
\captionsetup{font=footnotesize}
\vspace{-15pt}
\caption{Training loss traces for various case studies (a) ECG200, (b) ECG5000, (c) PhysioNet 2016 and (d) PhysioNet 2017. FP-LSTM, FP-CNN-LSTM and T-CNN-LSTM are respectively shown with green, black and red traces. It is also evident that the ternary network converges to its final value slower that full precision networks.}
\vspace{-5pt}
\label{fig:loss}
\end{figure}

\section{Memory Assessment and MAC Operations}
\par \textbf{Memory}: There are three main sources of memory required for calculation of deep learning layers: 1) parameters including weights and biases; 2) intermediate data comes from output of each layer (e.g. features maps in CNN layers). It should be noticed that for saving memory bandwidth intermediate data is saved in on--chip memory and biases are neglected due to the minority of their sizes. Therefore, we just consider the memory required for storing weights and intermediate stages. The number of convolution weights per each layer can be estimated as follows:
\begin{eqnarray*}
\rm{CNN_{WS}} = I_d\times m\times f
\end{eqnarray*}
where $I_d$ the input depth. This value for LSTM networks equals to the following:
\begin{eqnarray*}
\rm{LSTM_{WS}}=4 \times(HN + \omega_s) \times HN
\end{eqnarray*}
where $HN$ is the number hidden neurons and $\omega_s$ is the input window size.
\par \textbf{MAC Operations}: In deep learning algorithms, MAC operation unit is normally quite dominant compared to other processing part, therefore, lower number of such units would save a significant area and latency in the design. Here, we first estimate the number of MAC operations required by the classifier, and then we will accordingly calculate the latency of the proposed hardware classifier based on the design specifications. Table \ref{table:assesment} illustrates memory and MAC estimations for for all case studies with various architecture and weight precisions. It should be stressed that the considered weight bit length for full precision (FP) is 32. As shown in the table, the required memory size for T-CNN-LSTM classifier for all case studies is lower than the FPGA Block RAM's capability implying that the model can be conveniently implemented on the FPGA. On the other hand, according to the number of required MAC operations and GOPs of the proposed hardware classifier (see Table \ref{table:hardware}), the response time is less than 0.2 mS per input window which is quite fast compared to the sampling frequency of the input heart signals.

\section{Discussion}
In this work, we have proposed an easily implementable hardware classifier capable of automatically extracting input features and accurately classifying them. Hardware resource sharing was used to achieve a flexible and efficient architecture where throughput of the system is traded off with hardware complexity and cost. It was also highlighted that such a compromise is only feasible in systems where the underlying time series has slow dynamics, as in the case of physiological systems. Otherwise, the faster the dynamics, the less resource sharing can be exploited. Moreover, four complex heart related case studies were investigated and results illustrated acceptable accuracy for all cases compared to state-of-the-art models. Hardware synthesis and physical implementation also confirmed that the proposed hardware achieved 1.12$\times$ higher GOPs/Slice than similar state of the art FPGA--based accelerators.
\chapter{Conclusion and Future Work}
\renewcommand{\baselinestretch}{\mystretch}
\label{chap:Future}

\par The applications of dynamical systems are highly diverse in science and engineering. Here, we focused on bioengineering applications where they are mainly used in large scale and generally categorised into two groups: (1) dynamical systems from biology (2) dynamical systems for biology. Computer--based simulations are not always suitable for interfacing with biological/physical systems where continuous monitoring with low power and area consumption might be required. Moreover, large scale and real time simulation of such systems with the use of computer--based software are quite time consuming and impractical. To tackle these issues, in this thesis a few novel hardware techniques for the both groups were proposed and their hardware results compared and validated by software simulations.

\par In particular, in \textbf{chapter 2} a fully reconfigurable synchronous cellular model, and its physical realisation on FPGA has been presented. Findings show that the proposed model properly demonstrates similar time domain and dynamical behaviour of the biological models and applies no serious limitation on the critical path of the system. This implies that a large number of intra/extracellular units can be realised on FPGA in real time. The proposed architecture was developed on a Kintex--7 (XC7K325T) FPGA. Hardware results showed that the proposed cellular model achieves a higher speed and lower area consumption compared to the regular digital implementations and previously published piecewise neuron models.. Moreover, it has also been demonstrated that the pipelined cellular network can accelerate the simulations 83 times compared to its CPU counterparts. The proposed cellular hardware is fully reconfigurable and only implemented once and reprogrammed for all cases with various sets of Hill functions. This implies that the proposed cellular model can achieve higher performance in terms of area and speed when modelling more complex dynamical systems. Therefore, the more complexity in targeted dynamical systems, the better the performance of the proposed cellular model. However, note that in order to obtain an optimum design in terms of area and accuracy, the cellular parameters must be properly tuned for a certain output dynamic range.

\par In \textbf{chapter 3}, it was demonstrated that slow nonlinear dynamical systems describing the behaviour of natural biological systems can be efficiently realised on digital platforms. The validity of this claim was confirmed by the emulation of intracellular calcium dynamics on a low power digital cytomimetic chip. Measured results showed that the proposed hardware model emulates the time domain behaviour of the biological model with an acceptable accuracy and applies no serious limitation on the critical path of the system. This implies that a large number of calcium units can be realised in real--time. Therefore, as a proof of concept, it has also been confirmed that the pipelined cellular network containing 16 calcium units can properly operate in real--time. A multi--dimensional comparison with the exact analog design realised in the same technology (AMS 0.35 $\mu m$) is offered. Results showed that when emulating slow biological dynamics in large--scale and real--time including A/D and D/A conversion, digital cytomimetic designs seem to carry advantages compared to their analog counterparts.

\par In \textbf{chapter 4}, a novel current--input current--output block and a systematic circuit synthesis method which allows for the direct mapping of nonlinear bilateral dynamical systems onto electrical circuits with considerably low power consumption or high speed and acceptable precision has been presented. The proposed systematic methodology has been applied successfully to four case studies. The applications of real--time simulation of nonlinear bilateral dynamical systems are diverse in science. However in biology and computational neuroscience they might have two major applications. First, the combination of neuronal and intra/extracellular dynamics of large--scale and dense biological systems can be efficiently simulated in real--time on specialised hardware with low power consumption and relatively small size. Emulating responses of a very large network of cells simultaneously, including aspects such as stochasticity and cell variability, is interesting since such a real--time simulation has a potential to accurately explain the functions of large cellular networks. The proposed circuit may help in laying the groundwork for the low--power and real--time simulations of large--scale biological networks such as small tissues or organs. Second, the proposed family of circuits may also be used in fast and efficient biosensors employed in drug testing platforms or, alternatively, be embedded in the robust and optimal control of biological systems (such as protheses). It was also shown that by sacrificing the power consumption the circuit is able to operate in strong--inversion region where a high speed nonlinear dynamics needs to be mimicked.

\par In \textbf{chapter 5}, as an application of \textit{dynamical systems for biology}, it was shown how complex deep learning algorithms can be modified for portable and embedded biomedical applications. By using hardware resource sharing, a flexible and efficient classifier can be delivered where throughput of the system is traded off with hardware complexity and cost. It was also shown that such a compromise is only feasible in systems where the underying time series has slow dynamics, such as physiological systems, otherwise, the faster the dynamics, the less resource sharing can be used. Moreover, for the purpose of performance evaluation, two case studies were investigated and results illustrated acceptable accuracy for both cases. Hardware synthesis and physical implementation also confirmed that the proposed hardware achieved 1.46$\times$ higher GOPs/Slice than similar state of the art FPGA--based accelerators. The proposed hardware architecture could lay the groundwork towards developing a generalized on--chip classifier for wearable biomedical applications.

\par In \textbf{chapter 6}, an efficient time-series classifier capable of automatically detecting effective features and classifying the input signals in real--time was proposed. A Convolutional Neural Network (CNN) was employed to extract input features and then a Long-Short-Term-Memory (LSTM) architecture with ternary weight precision classified the input signals according to the extracted features. Hardware implementation on a Xilinx FPGA confirmed that the proposed hardware can accurately classify multiple complex heart related time series data with low area and power consumption and outperform all previously presented state--of--the--art records.

\section{Future work}
\par In light of the above conclusions, the following recommendations are made for
future work:

\subsection{Fabrication in Smaller Technology}
The most important tradeoffs in the design of bioinspired digital circuits are power and area consumption. It is well understood that scaling fully digital designs in smaller technologies up to a certain point always leads to higher performance and lower area and power consumption. As mentioned above, the cellular model was fabricated in AMS 0.35 $\mu m$ technology which is quite large for digital design. While the results showed better performance compared to its analog counterpart, fabricating the design in smaller feature size would highlight the merits of the model even more.  

\subsection{Mixed Mode System-on-Chip (SoC)}
In this work, the proposed cellular models were tested using digital UART serial inputs and outputs. In the real-world applications where the model needs to be interfaced with biology, the system should be integrated with ADC module at the input and DAC module at the output. Such an architecture can be fabricated on a single chip through a mixed mode design in smaller technology. 

\subsection{Silicon Fabrication for NBDS Model}
The NBDS model introduced in chapter 4 was tested and verified by nominal and Monte Carlo simulated results with realistic process parameters from the AMS 0.35 $\mu m$ technology. However, as analog nonlinear circuits are generally prone to noise and mismatch, the model still needs to be fabricated and tested in silicon. The circuit could be realised even in smaller CMOS technology using the Cadence intelligent optimiser without sacrificing too much of the design corners.

\subsection{Silicon Fabrication of the Hardware Classifier}
While the proposed hardware classifier introduced in chapter 6 showed remarkable performance in classifying complex time--series, it should be still fabricated in silicon to measure its real power consumption. This would make the classifier one step closer to be used in real--world biomedical applications.

\frontmatter
\doublespace
\setlength{\tmplen}{\parskip}
\setlength{\parskip}{-1ex}
\renewcommand{\baselinestretch}{\mystretch}
\setlength{\parskip}{\tmplen}

\renewcommand{\baselinestretch}{1}

\bibliographystyle{IEEEtran}
\bibliography{IEEEabrv,references}

\fancyhead[RE]{\emph{References}}

\end{document}